\documentclass{article}

\PassOptionsToPackage{numbers, compress}{natbib}
 \usepackage[main, final]{neurips_2025}

\usepackage{tikz}
\usepackage{pgfplots}
\pgfplotsset{compat=1.18}
\pgfplotsset{
  log ticks with fixed point,
}
\usetikzlibrary{pgfplots.groupplots}
\usepackage[export]{adjustbox}
\usepackage[utf8]{inputenc} 
\usepackage[T1]{fontenc}    
\usepackage[hidelinks]{hyperref}       
\usepackage{url}            
\usepackage{booktabs}       
\usepackage{amsfonts}       
\usepackage{nicefrac}       
\usepackage{microtype}      
\usepackage{xcolor}         
\usepackage{mathtools}
\usepackage{amsthm}
\usepackage{bm}
\usepackage{subcaption}
\usepackage{wrapfig}

\usepackage[textsize=tiny]{todonotes}

\definecolor{tabblue}{RGB}{31, 119, 180}
\definecolor{taborange}{RGB}{255, 127, 14}
\definecolor{tabgreen}{RGB}{44, 160, 44}
\definecolor{tabred}{RGB}{214, 39, 40}
\definecolor{tabpurple}{RGB}{148, 103, 189}
\definecolor{tabbrown}{RGB}{140, 86, 75}
\definecolor{tabpink}{RGB}{227, 119, 194}
\definecolor{tabgray}{RGB}{127, 127, 127}
\definecolor{tabolive}{RGB}{188, 189, 34}
\definecolor{tabcyan}{RGB}{23, 190, 207}
\definecolor{ForestGreen}{RGB}{34, 139, 34}

\usepackage[acronym,nohypertypes={acronym,notation}]{glossaries}
\newacronym{gns}{GNS}{Graph Network Simulator}
\newacronym{mgn}{MGN}{MeshGraphNet}
\newacronym{mpc}{MPC}{Model Predictive Control}
\newacronym{mbrl}{MBRL}{Model-Based Reinforcement Learning}
\newacronym{gnn}{GNN}{Graph Neural Network}
\newacronym{sofa}{SOFA}{Simulation Open Framework Architecture}
\newacronym{mpn}{MPN}{Message Passing Network}
\newacronym{mlp}{MLP}{Multilayer Perceptron}
\newacronym{cnn}{CNN}{Convolutional Neural Network}
\newacronym{mse}{MSE}{Mean Squared Error}
\newacronym{iou}{IoU}{Intersection over Union}
\newacronym{ggns}{GGNS}{Grounding Graph Network Simulator}
\newacronym{ltsgns}{Meta-MGNO}{Meta Mesh Graph Neural Operator}
\newacronym{gmmnp}{GMM-NP}{Gaussian Mixture Model Neural Process}
\newacronym{prodmp}{ProDMP}{Probabilistic Dynamic Movement Primitive}
\newacronym{dmp}{DMP}{Dynamic Movement Primitive}
\newacronym{elbo}{ELBO}{Evidence Lower Bound}
\newacronym{cnp}{CNP}{Conditional Neural Process}
\newacronym{np}{NP}{Neural Process}
\newacronym{egno}{EGNO}{Equivariant Graph Neural Operator}

\newacronym{trngvi}{TRNG-VI}{Trust Region Natural Gradient Variational Inference}
\newacronym{gmm}{GMM}{Gaussian Mixture Model}
\newacronym{mp}{MP}{Movement Primitive}
\newacronym{fem}{FEM}{Finite Element Method}

\DeclareMathOperator{\enc}{enc}
\DeclareMathOperator{\dec}{dec}

\newcommand{\R}{\mathbb{R}}

\newcommand{\x}{\mathbf{x}}
\newcommand{\y}{\mathbf{y}}
\newcommand{\z}{\mathbf{z}}
\newcommand{\p}{\mathbf{p}}
\newcommand{\vsymb}{\mathbf{v}}
\newcommand{\h}{\mathbf{h}}
\newcommand{\m}{\mathbf{m}}
\newcommand{\e}{\mathbf{e}}

\newcommand{\Next}{N_{\text{ext}}}
\newcommand{\re}{\mathbf{r}}
\newcommand{\RR}{\mathbb{R}}
\newcommand{\Gr}{\mathcal{G}}

\usepackage[capitalize,noabbrev]{cleveref}

\newif\ifincludelist
\includelistfalse 

\title{MaNGO -- Adaptable Graph Network Simulators via Meta-Learning}

\author{\textbf{Philipp Dahlinger}\thanks{correspondence to \texttt{philipp.dahlinger@kit.edu}}~~
\textbf{Tai Hoang}~
\textbf{Denis Blessing}~
\textbf{Niklas Freymuth}~
\textbf{Gerhard Neumann}
\\
Autonomous Learning Robots\\
Karlsruhe Institute of Technology\\
Karlsruhe\\
}

\begin{document}

\maketitle

\begin{abstract}
    Accurately simulating physics is crucial across scientific domains, with applications spanning from robotics to materials science. While traditional mesh-based simulations are precise, they are often computationally expensive and require knowledge of physical parameters, such as material properties. In contrast, data-driven approaches like Graph Network Simulators (GNSs) offer faster inference but suffer from two key limitations: Firstly, they must be retrained from scratch for even minor variations in physical parameters, and secondly they require labor-intensive data collection for each new parameter setting. This is inefficient, as simulations with varying parameters often share a common underlying latent structure.
    In this work, we address these challenges by learning this shared structure through meta-learning, enabling fast adaptation to new physical parameters without retraining. 
    To this end, we propose a novel architecture that generates a latent representation by encoding graph trajectories using conditional neural processes (CNPs). To mitigate error accumulation over time, we combine CNPs with a novel neural operator architecture. 
    We validate our approach, Meta Neural Graph Operator (MaNGO), on several dynamics prediction tasks with varying material properties, demonstrating superior performance over existing GNS methods. Notably, MaNGO achieves accuracy on unseen material properties close to that of an oracle model.
\end{abstract}

\section{Introduction}
\begin{figure}[t]
    \centering
    \resizebox{\textwidth}{!}{%
        \input{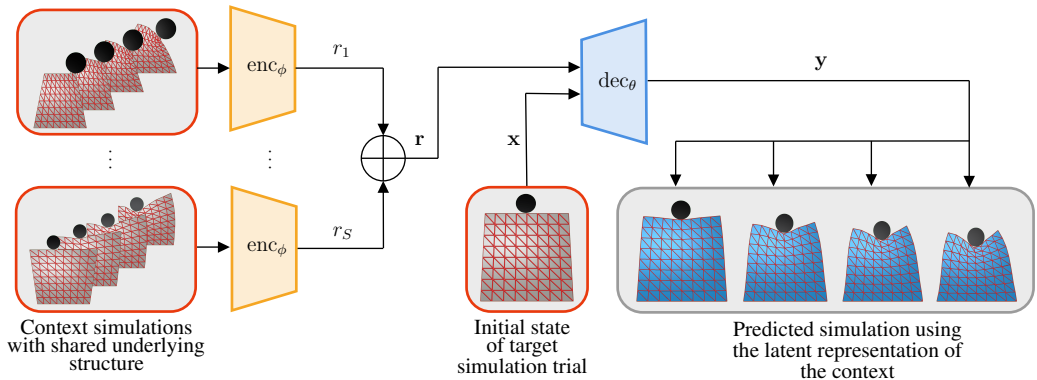}
    }
    \caption{
Our proposed Meta Neural Graph Operator (MaNGO) approach: The context set is aggregated to form a latent representation of material properties. Given an unseen initial state, the Graph Network Simulator (GNS) uses this latent representation to generate trials that follow the material laws of the context set, enabling accurate predictions for new conditions. The example \textcolor{blue}{prediction} aligns perfectly with the \textcolor{red}{ground truth data}.
\vspace{-0.5cm}}
\label{fig:meta_gns}
\end{figure}

The simulation of complex physical systems is of paramount importance in a wide variety of engineering disciplines, including structural mechanics~\citep{yazid2009state, zienkiewicz2005finite, stanova2015finite}, fluid dynamics~\citep{chung1978finite, zienkiewicz2013finite, connor2013finite}, and electromagnetism~\citep{jin2015finite, polycarpou2022introduction, reddy1994finite}.
In particular, simulating object deformations under external forces is essential for applications such as robotics~\citep{scheikl2022sim, caregiving2023deformable, linkerhagner2023grounding}.
Traditional mesh-based simulations are appealing for such problems due to the accuracy of the underlying finite element method~\citep{brenner2008mathematical, reddy2019introduction}.
However, these methods are typically slow and require precise knowledge of the simulation parameters, including material properties of objects.

In contrast, data-driven approaches for simulating complex systems have emerged as a promising alternative to traditional mesh-based simulators~\citep{guo2016convolutional, da2021deep, li2022machine}. %
Among them,~\glspl{gns} have recently become increasingly popular~\citep{battaglia2018relational, pfaff2020learning, allen2022graph, allen2023learning, linkerhagner2023grounding, egno2024}. 
\glspl{gns} encode the simulated system as a graph of interacting entities whose dynamics are predicted using~\glspl{gnn}~\citep{bronstein2021geometric}.
These models are often orders of magnitude faster than classical simulators~\citep{pfaff2020learning} while being fully differentiable, making them highly effective for downstream tasks such as inverse design problems~\citep{allen2022graph, xu2021end}. Moreover, these models do not require knowledge of simulation parameters as they directly learn from the training data. However, they must be retrained from
scratch for even minor variations in physical parameters, and require data collection for each new parameter setting that is often either costly and time-consuming or even impossible.

To overcome this, \citet{sanchezgonzalez2020learning} proposed to use the simulation parameter as conditional information when training the GNS, allowing for generalization to unseen simulation parameters. However, such an approach only works under two assumptions. Firstly, training data must be labeled with the corresponding simulation parameter, and secondly, the simulation parameter must be available at test time for the desired simulation. 
While the first assumption is mild and often satisfied in simulation settings, the second assumption requires solving the inverse problem of inferring the underlying physical parameters from observed system behavior \cite{isakov2006inverse}. Material estimation or system identification can be viewed as a specific instance of this problem \cite{Takahashi2019Video,ma2022risp,Feng2022Visual,Wang2015Deformation,qzhao2022graphpde,wang2024latent}. However, solving such inverse problems is challenging, as they are typically ill-posed and require explicit knowledge of the governing partial differential equation (PDE) \cite{Wang2015Deformation,Takahashi2019Video,ma2022risp}. 

To overcome this challenge, we investigate data-driven adaptation of GNS -- enabling fast and accurate simulations for unknown parameters using only a few simulation trials. Our work builds on the premise that training data from different simulations shares a common `latent' underlying structure. We aim to learn this structure via meta-learning using Conditional Neural Processes (CNPs). To that end, we propose a novel framework called Meta Neural Graph Operator (MaNGO) that builds on Message Passing Networks (MPNs) and neural operator methods to ensure efficient processing of spatiotemporal data. We validate our approach on several dynamics prediction tasks with varying material properties, demonstrating superior performance over existing GNS methods. Notably, our method achieves accuracy on unseen material properties close to that of an oracle model which has access to the simulation parameters at test time. To summarize, we identify our contribution as follows:
(i) we successfully use meta-learning with Conditional Neural Processes (CNPs) for graph network simulators allowing for fast and accurate adaptation to unseen physical parameters. (ii) we identify shortcomings of existing architectures for handling spatiotemporal data and propose a novel GNS architecture.   
(iii) we provide a set of new benchmark tasks suited for testing the adaptation capability of GNS\footnote{%
Code: \url{https://github.com/ALRhub/mango} \quad
Dataset: \url{https://zenodo.org/records/17287535}
}.

\section{Preliminaries} \label{sec: preliminaries}

\textbf{Graph and Message Passing Neural Networks.}
Graph Neural Networks (GNNs) are a class of neural networks designed to process graph-structured data by iteratively updating node representations through localized message passing. Here, a graph is a defined as \mbox{$\mathcal{G} = (\mathcal{V}, \mathcal{E}, \{\m_v^0\}_{v\in \mathcal V}, \{\m_e^0\}_{e\in \mathcal E})$} 
with nodes $\mathcal{V}$, edges $\mathcal{E}$, and associated vector-valued node and edge features $\m_v^0$ and $\m_e^0$. 
A~\gls{mpn}~\citep{sanchezgonzalez2020learning, pfaff2020learning}, a GNN architecture well-suited for graph-based simulations, consists of $K$ message passing steps, which iteratively update the node and edge features based on the graph topology.
Each such step is given as
\begin{equation}
    \m^{k+1}_{e} = f^{k}_{\mathcal{E}}(\m^{k}_e, \m_v^k, \m_w^k)\text{, with } e=(v, w) \text{,}\qquad 
    \m^{k+1}_{v} = f^{k}_{\mathcal{V}}(\m^{k}_{v}, \bigoplus_{e\in \mathcal{E}_v} \m^{k+1}_{e})\text{,} \label{eq:node_update}
\end{equation}
where \mbox{$\mathcal{E}_v \subset \mathcal{E}$} are the edges connected to $v$.
Further, $\bigoplus$ denotes a permutation-invariant aggregation operation such as the sum, the max, or the mean. 
The functions $f^m_{\mathcal V}$ and $f^m_{\mathcal E}$ are learned~\glspl{mlp}, usually with a residual connection.
The network's final output are the node-wise learned representations $\m_v^K $ that encode local information of the initial node and edge features.


\textbf{Neural Dynamics Prediction.} Our goal is to learn the dynamics of a multibody system, i.e., a trajectory of graphs $(\Gr^{(t)})_{t\in[0,T]}$ from the initial condition $\Gr^{(0)}$. 
We follow \citet{brandstetter2021message} and group existing approaches into two categories, \textit{neural operators} and \textit{autoregressive methods}. Neural operator methods treat the mapping from initial conditions to solutions at time $t$ as an input–output mapping learnable via supervised learning. Formally, a neural operator $F_{\text{NO}}$ predicts the graph at any $t\in[0,T]$ from the initial condition, that is,
$\Gr^{(t)} = F_{\text{NO}}\left(t,\Gr^{(0)}\right).
$
In contrast, autoregressive methods, learn to incrementally update the graph starting from $\Gr^{(0)}$:
\begin{equation*}
    \Gr^{(t + \Delta t)} = F_{\text{AR}}\left(\Delta t,\Gr^{(t)}\right).
\end{equation*}
Here, $F_{\text{AR}}$ is the temporal update and $\Delta t \in \R_{>0}$.

\textbf{Meta-Learning and Conditional Neural Processes.} To formalize the meta-learning problem using conditional neural processes (CNPs) \citep{garnelo_conditional_2018}, we consider a meta-dataset $\mathcal{D} = \mathcal{D}_{1:L}$ consisting of $L$, typically small, task datasets $\mathcal{D}_l = \{\x^l_i,\y^l_i\}_{i=1}^{S_l}$ of size $S_l$. Each task consists of inputs $\x^l_i \in \R^{d_{x}}$ and corresponding evaluations $\y^l_i \in \R^{d_{y}}$ of unknown functions $f_l$, that is, $\y^l_i = f_l(\x^l_i) + \epsilon_i$, where $\epsilon_s$ denotes (possibly heteroskedastic) noise. Meta-learning hinges on the idea that tasks share statistical structure, allowing for fast adaptation to a target function $f_*$ based on a small target dataset $\mathcal{D}_* = \{\x^*_i,\y^*_i \}_{i=1}^{S_*}$ of size $S_*$. 
To leverage this shared statistical structure, Conditional Neural Processes (CNPs) use the meta-dataset $\mathcal{D}$ to learn how to generate a latent representation $\re \in \R^{d_r}$ from a given set of $(\x,\y)$-pairs. At test time, this latent representation is generated from the target dataset $\mathcal{D}_*$ enabling generalization to unlabeled inputs $\x_*$ from the target task without requiring weight adaptation.
Formally, to generate the latent representation $\re$ for a set $\mathcal{S} = \{\x_i,\y_i\}_{i=1}^S$ with arbitrary size $S$, CNPs use a parameterized encoder with a permutation-invariant aggregation method,
\begin{equation}
\label{eq: aggregation}
     \re = \bigoplus_{i\in\{1,\dots,S\}} r_i \quad \text{with} 
    \quad r_i = \enc_{\phi}(\x_i,\y_i),
\end{equation}
with parameters $\phi$ and permutation-invariant aggregation method $\bigoplus$. The latent representation $\re$ is then used to predict the mean and variance of a Gaussian distribution over $\y$, given a new input $\x$,
\begin{equation}
    p_{\theta}(\y|\x,\mathcal{S}) = \mathcal{N}\left(\y|\dec_{\theta}^{\mu}(\x,\re), \ \dec_{\theta}^{\Sigma}(\x,\re)\right), \label{eq:gaussian_likelihood}
\end{equation}
using a parameterized decoder with parameters $\theta$. 
For training, the task datasets are further split into context sets $\mathcal{D}_l^c \subseteq \mathcal{D}_l$ to train the CNP on different dataset sizes $S_l^c$ which are used to minimize the negative task log-likelihood 
\begin{equation}
\label{eq: meta learning loss}
    \mathcal{L}_{l}(\phi, \theta) = -\mathbb E_{\mathcal{D}_l^c \subseteq \mathcal{D}_l}\bigg[  
 \sum_{i=1}^{S_l} \log p_{\theta}(\y^l_i | \x^l_i,  \mathcal{D}_l^c) \bigg]
\end{equation}
which implicitly depends on $\phi$ through the encoding of $\mathcal{D}_l^c$. Here, the expectation indicates randomly sampled subsets. Thus, meta-learning aims to maximize the overall task likelihood while conditioning the model on smaller subsets of the data.
The complete loss is obtained as $ \mathcal{L}(\phi,\theta) = \sum_l \mathcal{L}_{l}(\phi,\theta)$ and is optimized end-to-end using stochastic gradients with respect to $\phi$ and $\theta$.

\section{Adaptable Graph Network Simulators via Meta-Learning}
\label{sec:method}

Having established the foundations, we now explain how meta-learning with Conditional Neural Processes (CNPs) can be used to make graph network simulators adaptable to novel unseen physical parameters. To that end, we start by formalizing our setup and introducing the meta-dataset in \Cref{sec: setup}. Next, we propose an extension to meta-learning that improves training with additional information that is not available at test-time in \Cref{sec: improved training}. Lastly, we introduce a CNP architecture that is tailored to our setup. Specifically, we propose an encoder that generates a latent representation from spatiotemporal data in \Cref{sec: encoder}. Furthermore, we introduce a novel decoder that addresses shortcomings of existing methods as outlined in \Cref{sec: decoder}.
\begin{figure*}[t]
    \centering
    \resizebox{1.0\textwidth}{!}{%
        \input{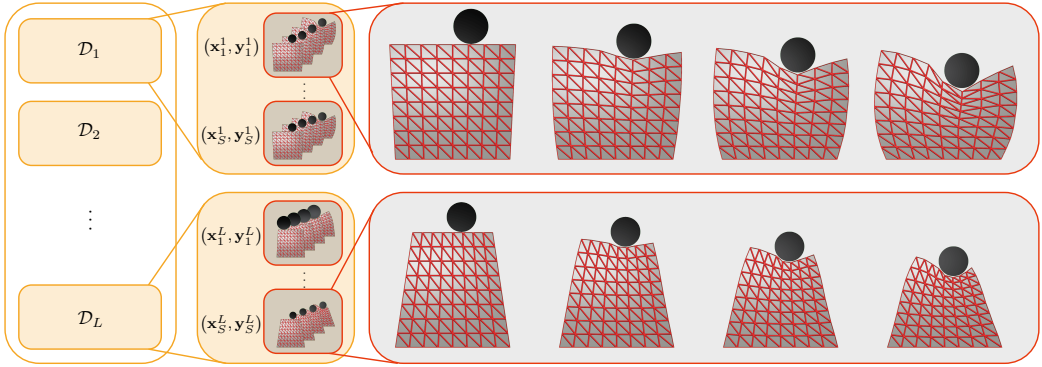}
    }
    \caption{Meta-dataset representation: This figure illustrates the structure of the meta-dataset, consisting of multiple task datasets. Each dataset includes simulations with shared material properties but varying starting conditions. Two example simulations, though starting similarly, produce vastly different results due to their distinct material properties.}
    \label{fig: meta dataset}
\end{figure*}
\vspace{-0.1cm}
\subsection{Problem Formalization and Setup} 
\vspace{-0.1cm} \label{sec: setup}
The aim is to learn a graph network simulator (GNS) that predicts a sequence of graphs $\mathcal{G}_{1:T}\coloneqq\{\mathcal{G}_t\}_{t=1}^T$ describing, e.g., mesh deformation over time without knowing some simulation or physical parameters, which we refer to as $\rho \in \RR^{d_\rho}$. For example, $\rho$ could be stiffness or compression properties of a deformable object in the simulation. To that end, we use meta-learning to extract latent representations from similar data to generalize to unseen physical parameters.
As outlined in \Cref{sec: preliminaries}, in order to perform meta-learning we require a meta dataset $\mathcal{D}$ containing $L$ task datasets $\mathcal{D}_l \in \mathcal{D}$ that in turn consist of multiple input-output pairs $(\x^l_i,\y^l_i) \in \mathcal{D}_l$. 
In our setting, each output is given by 
\begin{equation}
\label{eq: graph output}
    \y^l_i = (\p^l_{1:T}, \vsymb^l_{1:T}),
\end{equation}
where $\p^l_{1:T}, \vsymb^l_{1:T}\in \RR^{T \times N\times d}$ denote the positions and velocities of a sequence of $N$ nodes belonging to (possibly deformable) objects in $d$-dimensional space.
An input $\x^l_i$ in our setting is given by 
\begin{equation}
\label{eq: graph input}
    \x^l_i = (\p^l_0, \vsymb^l_0, \p^{l,\text{ext}}_{0:T}, \vsymb^{l,\text{ext}}_{0:T}, \h^l),
\end{equation}
where $\p^l_{0}, \vsymb^l_{0}\in \RR^{N\times d}$ are the initial node positions and velocities. Here $\p^{l,\text{ext}}_{0:T}, \vsymb^{l,\text{ext}}_{0:T}\in \RR^{(T+1) \times N^\text{ext}\times d}$ are positions and velocities of `external' objects, that is, objects that we do not wish to simulate, such as a rigid collider that interacts with the object of interest. 
Lastly, $\h^l \in \mathbb{R}^{(N + N^{\text{ext}}) \times d_{\h}}$ represents node features that remain constant over time, such as the node's type (deformable or collider) or whether a force is applied to the node. Initial condition $\x^l$ and simulation result $\y^l$ together result in a sequence of graphs $\mathcal{G}^{l}_{1:T}$ with $N + \Next$ nodes. In our work, we assume a fixed graph structure: the connectivity and the number of nodes in $\mathcal G^l_t$ is constant over time and task datasets $\mathcal D_l$. An illustration of a meta dataset $\mathcal{D}$ is shown in \Cref{fig: meta dataset}.

\vspace{-0.1cm}
\subsection{Incorporating Simulation Parameters into Meta-Training}
\vspace{-0.1cm}\label{sec: improved training}
For training, we could follow the procedure outlined in \Cref{sec: preliminaries} and optimize the loss defined in \Cref{eq: meta learning loss}. However, contrary to the setup discussed in \Cref{sec: preliminaries}, we additionally assume access to the simulation parameters for each training task, i.e., $\{\rho^l\}_{l=1}^L$ but not for target tasks $\mathcal{D}^*$ rendering them useless in the standard meta-learning formulation. Here, we slightly extend the meta training such that we obtain additional learning signals from $\rho^l$. To that end, we use an additional parameterized neural network $f^{\psi}: \R^{d_r} \rightarrow \R^{d_\rho}$ with parameters $\psi$ that aims to predict $\rho^l$ from a latent representation $\re^l$. Using the following definition for the joint likelihood between $\y$ and $\rho$, i.e.,
\begin{align*}
\label{eq: meta learning joint likelihood}
 p_{\theta, \psi}&(\y, \rho |\x,\mathcal{S}) = \mathcal{N} \left( 
\begin{bmatrix} \y\\ \rho \end{bmatrix} \bigg| \begin{bmatrix}  \dec_{\theta}^{\mu}(\x,\re) \\ f^{\psi}(\re)\end{bmatrix}, 
\begin{bmatrix}  \dec_{\theta}^{\Sigma}(\x,\re) & 0 \\ 0 & 1  \end{bmatrix}
\right) \nonumber,
\end{align*}
we obtain a novel per-task loss function as
\begin{equation*}
\label{eq: meta learning joint loss}
    \mathcal{L}_{l}(\phi, \theta, \psi) = -\mathbb E_{\mathcal{D}_l^c \subseteq \mathcal{D}_l}\bigg[  
 \sum_{s=1}^{S_l} \log p_{\theta, \psi}(\y^l, \rho^l|\x,\mathcal{D}_l^c) \bigg],
\end{equation*}
where the full loss again sums over the task losses, $\mathcal{L}(\phi, \theta, \psi) = \sum_l \mathcal{L}_{l}(\phi, \theta, \psi)$.
Intuitively, gradients with respect to the encoder parameters $\phi$ are informed by $\rho$ via $f^{\psi}$. As another positive side-effect, we obtain an estimate $\hat \rho = f^{\psi}(\re)$ which could be used for downstream tasks.

\vspace{-0.1cm}
\subsection{Spatiotemporal Encoder} 
\vspace{-0.1cm} \label{sec: encoder}
In \Cref{sec: preliminaries}, we treated the encoder and decoder of the CNP architecture as black boxes that are responsible for generating a latent representation and a predictive distribution over outputs, respectively. However, since our data consists of non-standard structures, specifically graphs with both spatial and temporal components, we introduce the following novel architectures.

Recall that the encoder of a CNP generates a latent representation $\re^l \in \R^{d_r}$ from a set of input-output pairs, typically a context set $\mathcal{D}_l^c \subseteq \mathcal{D}_l$. Then, for each $(\x_i^l, \y_i^l) \in \mathcal{D}_l^c$ we separately generate a latent representation, i.e.,  $r_i^l = \enc_{\phi}(\x_i^l,\y_i^l)$ which are then aggregated in a permutation-invariant fashion to obtain $\re^l$. As discussed in \Cref{sec: setup}, each input-output pair in our setting has a spatial and temporal component. We first aggregate over the temporal dimension and then over the spatial dimension. To that end, we combine the inputs $\x_i^l$ (see \Cref{eq: graph input}) and $\y_i^l$ (see \Cref{eq: graph output}) as 
\begin{equation*}
    \z_i^l = (\p^l_{0:T}, \vsymb^l_{0:T}, \p^{l,\text{ext}}_{0:T}, \vsymb^{l,\text{ext}}_{0:T}, \h^l_{0:T}),
\end{equation*}
where $\h^l$ is simply copied for each time step. We ensure translation invariance by subtracting the initial mean position. To remove the temporal component, we apply a 1D convolutional neural network, i.e., 
 $\hat{\z}_i^l =\text{CNN}^{\phi}_\text{1D}(\z^l_i). $
Lastly, to obtain a spatial-independent latent representation, we apply a deep set encoder \citep{zaheer_deep_2017} on the node level, that is, 
\begin{equation*}
    r^l_i = \enc_{\phi}(\x^l_i,\y^l_i) = f^{\phi}_{\text{outer}}\left(\frac{1}{N + \Next} 
 \sum_{n=1}^{N + \Next} f^{\phi}_{\text{inner}}(\hat{\z}^l_{i,n})\right),
\end{equation*}
where, $f^{\phi}_{\text{outer}}$ and $f^{\phi}_{\text{inner}}$ are neural networks. 
In this work, we choose deep sets as our spatial aggregation method, as they have demonstrated good performance in graph-level problems similar to ours \cite{cai2023deepset}. 
Finally, to aggregate a whole set of input-output pairs, we follow \Cref{eq: aggregation} and use a permutation invariant aggregation method. 
\begin{figure}[t]
    \centering
    \resizebox{1.0\textwidth}{!}{%
        \input{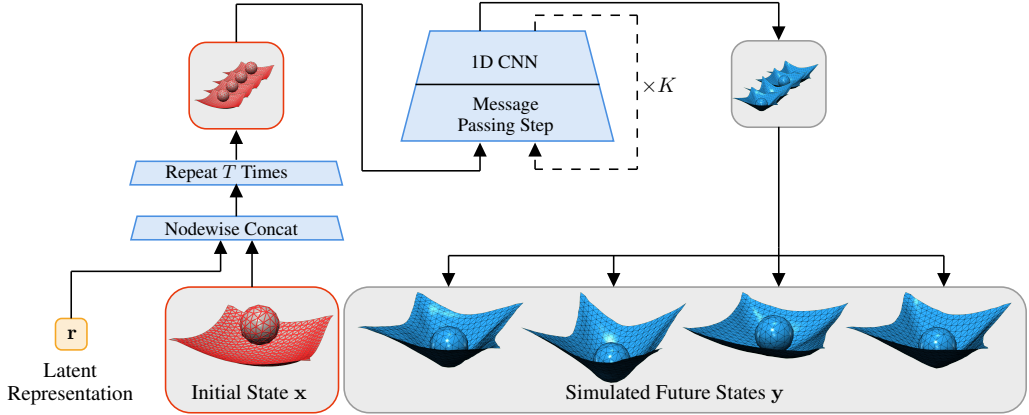}
    }
    \caption{MaNGO Decoder: Our simulator takes a latent representation and an initial state as input. The initial state is combined and iteratively processed to generate a trajectory of graphs. By alternating between a message-passing network for spatial processing and a 1D CNN for temporal processing, the simulator produces accurate dynamic simulations.}
    \label{fig:mgno_arch}
\end{figure}
\vspace{-0.1cm}
\subsection{Meta Neural Graph Operator Decoder}
\vspace{-0.1cm} \label{sec: decoder}

We propose a novel MaNGO decoder architecture that combines elements of {MeshGraphNet (MGN) \citep{pfaff2020learning}} and the {Equivariant Graph Neural Operator (EGNO) \citep{egno2024}}. EGNO predicts graph sequences using equivariant GNN layers and temporal convolutions in Fourier space. While beneficial in some settings, these equivariance constraints can limit performance when not required. We discuss this limitation in Appendix \ref{appx:egno_failed}, where we mathematically show that EGNO struggles on certain tasks.  
Alternatively, MGN models graph sequences autoregressively using {Message Passing Networks (MPNs)}, which are highly effective for graph-based simulations \citep{pfaff2020learning}. 
However, autoregressive prediction is not ideal when paired with the current meta-learning training setup, see Appendix \ref{app:mgn_cnp} for further discussions. 
To address these limitations, MaNGO retains the strengths of neural operator methods by replacing the equivariant GNN layers with {MPNs} and equivariant convolutions with a {1D CNN} (see Figure~\ref{fig:mgno_arch}).  
Formally, we use the MaNGO decoder to realize $\y  = \dec_{\theta}(\x,\re)$ (omitting the task index $l$ for readability). Thus, we define a single MaNGO block as a mapping 
\begin{equation*}
    (\m^{k}_{e,1:T}, \m_{v,1:T}^{k}) \mapsto (\m^{k+1}_{e,1:T}, \m_{v,1:T}^{k+1}),
\end{equation*}
following the notation in \Cref{sec: preliminaries}. The initial edge and note features $(\m^{0}_{e,1:T}, \m_{v,1:T}^{0})$ are extracted from $(\x,\re)$ by copying $\re$ to every node feature and replicating the initial graph $T$ times. Additionally, we include time embeddings for every time step $t$ and use relative positions as edge features. Further details on the feature creation are provided in Appendix \ref{app:mgno_features}. As a first step of the MaNGO block, we leverage the MPN for a spatial update, that is, we process the edge and node features according to \Cref{eq:node_update} for each timestep separately to obtain a new tuple $(\m^{k+1}_{e,1:T}, \tilde \m_{v,1:T}^{k+1})$. The temporal update is subsequently performed using a 1D residual convolutional layer, that is, 
\begin{equation*}
    \m_{v,1:T}^{k+1} = \text{Conv}^{\theta}_\text{1D}(\tilde{\m}_{v,1:T}^{k+1}) + \tilde{\m}_{v,1:T}^{k+1}.
\end{equation*}
After $K$ MaNGO blocks the final node features $\m_{v,1:T}^{K}$ are used for predicting the node positions for every time-step. Specifically, a displacement vector is computed using a parameterized network $f^{\theta}$, that is, $\mathbf{d}_{v, t}= f^{\theta}(\m_{v,t}^{K})$ to obtain node positions $\p_{v, t}$ as 
    $\p_{v, t} = \p_{v, 0} + \mathbf{d}_{v, t}. $

The resulting positions \(\p_{v, t}\) define the graph sequence \(\y\) over time. In this work, we do not predict input-dependent variances, and instead use a fixed $\dec_{\theta}^{\Sigma}(\x,\re) := 1$ to stabilize and simplify the training scheme.
If variances are required, for example, for uncertainty estimation, they can be easily predicted from the decoder as well.
By alternating between {spatial message passing} and {temporal convolution}, the MaNGO simulator efficiently models time-series graph data while avoiding the pitfalls of equivariance over-constraints and autoregressive prediction. 

\section{Related Work}
\glsreset{gns}
\textbf{Learning-based forward simulators.}
Using deep neural networks to learn physical simulations has become an emerging research direction in scientific machine learning \cite{pfaff2020learning, li2021fourier, thuerey2020deep}. Deep learning-based approaches have demonstrated success in applications such as fluid dynamics \cite{kochkov2021machine, prantl2022guaranteed}, aerodynamics \cite{bhatnagar2019prediction, pfaff2020learning}, and deformable object simulations \cite{linkerhagner2023grounding, yu2023learning}. 
A popular class of learned neural simulators are~\glspl{gns}~\citep{battaglia2016interaction, sanchezgonzalez2020learning}.
\glspl{gns} utilize~\glspl{mpn}, a special type of~\gls{gnn}~\citep{scarselli2009the, bronstein2021geometric} that representationally encompasses the function class of many classical solvers~\citep{brandstetter2021message}.
\glspl{gns} handle physical data by modeling arbitrary entities and their relations as a graph. Notably, all previously mentioned \glspl{gns} predict system dynamics iteratively from a given state, whereas we directly estimate entire trajectories, improving rollout stability and prediction speed. Related to our approach is the Equivariant Graph Neural Operator (EGNO)~\citep{egno2024}, which also predicts full trajectories using SE(3) equivariance to model 3D dynamics and capture spatial and temporal correlations.
In this work, we adopt the trajectory prediction framework of  EGNOs for our decoder (as shown in \cref{fig:mgno_arch}) but remove the equivariance constraint. We justify this choice in Appendix~\ref{appx:egno_failed}.

\begin{figure}[t]
    \makebox[\textwidth][c]{
    \begin{tikzpicture}
    \tikzstyle{every node}=[font=\small]
    \input{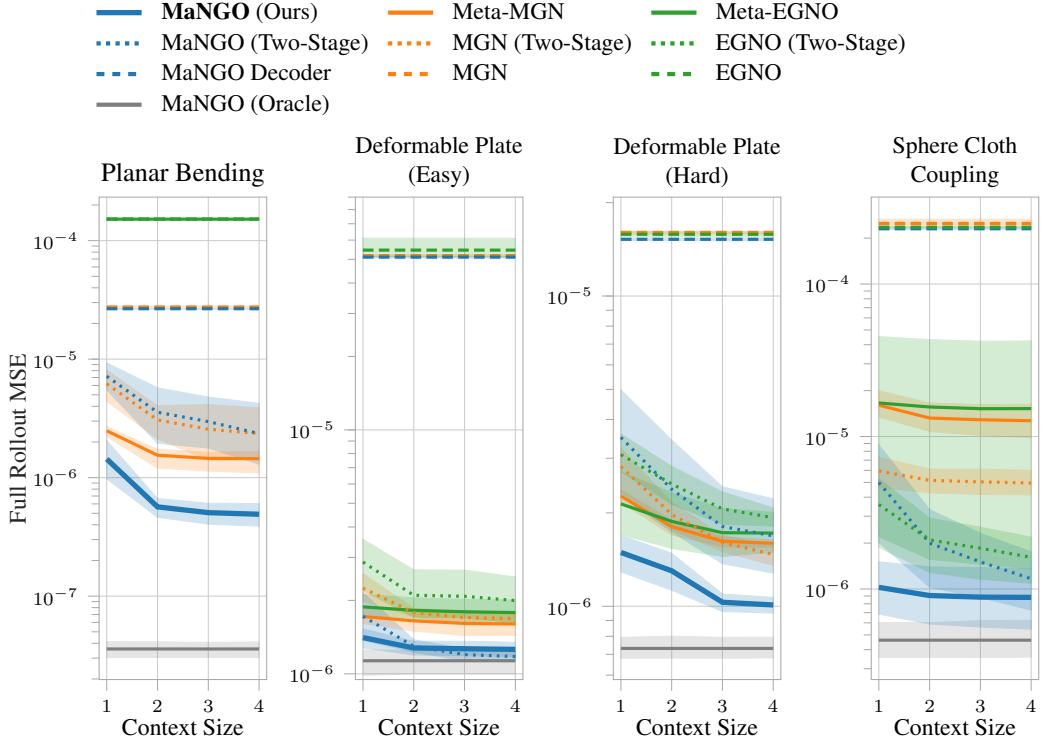}
    \begin{axis}[
        hide axis,
        xmin=0, xmax=1, 
        ymin=0, ymax=1,
        scale only axis, 
        width=0pt, height=0pt, 
        legend style={
            draw=none,
            legend cell align=left,
            legend columns=4,
            column sep=1ex,
            line width=1pt,
            at={(0.5,-0.1)}, 
            anchor=north, 
        },
        ]
        \addlegendimage{line legend, tabblue, line width=2pt} 
        \addlegendentry{\textbf{MaNGO} (Ours)}
        \addlegendimage{line legend, taborange, ultra thick} 
        \addlegendentry{Meta-MGN}
        \addlegendimage{line legend, tabgreen, ultra thick} 
        \addlegendentry{Meta-EGNO}
        \addlegendimage{empty legend}
        \addlegendentry{}
        
        \addlegendimage{line legend, tabblue, ultra thick, dotted}
        \addlegendentry{MaNGO (Two-Stage)}
        \addlegendimage{line legend, taborange, ultra thick, dotted}
        \addlegendentry{MGN (Two-Stage)}
        \addlegendimage{line legend, tabgreen, ultra thick, dotted}
        \addlegendentry{EGNO (Two-Stage)}
        \addlegendimage{empty legend}
        \addlegendentry{}
        \addlegendimage{line legend, tabblue, ultra thick, dashed} 
        \addlegendentry{MaNGO Decoder}
        \addlegendimage{line legend, taborange, ultra thick, dashed} 
        \addlegendentry{MGN}
        \addlegendimage{line legend, tabgreen, ultra thick, dashed} 
        \addlegendentry{EGNO}
        \addlegendimage{empty legend}
        \addlegendentry{}
        \addlegendimage{line legend, tabgray, ultra thick} 
        \addlegendentry{MaNGO (Oracle)}
    \end{axis}
\end{tikzpicture}
    }
    \centering
\begin{tikzpicture}
\pgfplotsset{every tick label/.append style={font=\scriptsize}} 
\pgfplotsset{every axis label/.append style={font=\small}} 

\definecolor{darkgray}{RGB}{169,169,169}
\definecolor{darkorange25512714}{RGB}{255,127,14}
\definecolor{darkslategray38}{RGB}{38,38,38}
\definecolor{forestgreen4416044}{RGB}{44,160,44}
\definecolor{gray127}{RGB}{127,127,127}
\definecolor{lightgray204}{RGB}{204,204,204}
\definecolor{steelblue31119180}{RGB}{31,119,180}

\begin{axis}[
xlabel style={yshift=0.15cm},
ylabel style={yshift=-0.15cm},
title style={align=center, yshift=-0.2cm},
x=0.675cm,
height=8cm,
axis line style={darkgray},
log basis y={10},
tick align=outside,
tick pos=left,
title={Planar Bending},
x grid style={lightgray204},
xlabel={Context Size},
yminorticks=true,
minor ytick={
0.00000002,
0.00000003, 
0.00000004,
0.00000005,
0.00000006,
0.00000007,
0.00000008,
0.00000009,
0.0000002,
0.0000003,
0.0000004,
0.0000005,
0.0000006, 
0.0000007,
0.0000008,
0.0000009,
0.000002, 
0.000003,
0.000004, 
0.000005,
0.000006, 
0.000007,
0.000008, 
0.000009,
0.00001,  
0.00002, 
0.00003,
0.00004, 
0.00005, 
0.00006,
0.00007,
0.00008, 
0.00009,
0.0002
},
xtick={1,2,3,4},
xmin=0.85, xmax=4.15,
xtick style={color=darkgray},
y grid style={lightgray204},
ymajorgrids,
xmajorgrids,
ymin=1.97793946615793e-08, ymax=0.000232925455323358,
ymode=log,
ytick pos=left,
ytick style={color=darkgray},
ytick={0.00000001, 0.0000001, 0.000001, 0.00001, 0.0001},
yticklabels={$10^{-8}$, $10^{-7}$, $10^{-6}$, $10^{-5}$, $10^{-4}$, },
ylabel=\textcolor{darkslategray38}{Full Rollout MSE},
]

\path [draw=darkorange25512714, fill=darkorange25512714, opacity=0.2]
(axis cs:1,2.79433337709634e-05)
--(axis cs:1,2.71947377768811e-05)
--(axis cs:2,2.71981653531839e-05)
--(axis cs:3,2.72075401881011e-05)
--(axis cs:4,2.72056622634409e-05)
--(axis cs:4,2.79592673905427e-05)
--(axis cs:4,2.79592673905427e-05)
--(axis cs:3,2.79745803709375e-05)
--(axis cs:2,2.79409357062832e-05)
--(axis cs:1,2.79433337709634e-05)
--cycle;

\path [draw=darkorange25512714, fill=darkorange25512714, opacity=0.2]
(axis cs:1,2.69782085524639e-06)
--(axis cs:1,2.21911900553096e-06)
--(axis cs:2,1.20685970159684e-06)
--(axis cs:3,1.13560525960565e-06)
--(axis cs:4,1.09704791384502e-06)
--(axis cs:4,1.66133174843708e-06)
--(axis cs:4,1.66133174843708e-06)
--(axis cs:3,1.63269505719654e-06)
--(axis cs:2,1.73409713397632e-06)
--(axis cs:1,2.69782085524639e-06)
--cycle;

\path [draw=steelblue31119180, fill=steelblue31119180, opacity=0.2]
(axis cs:1,2.67422514298232e-05)
--(axis cs:1,2.64315011008875e-05)
--(axis cs:2,2.64310619968455e-05)
--(axis cs:3,2.64315009189886e-05)
--(axis cs:4,2.64315171079943e-05)
--(axis cs:4,2.67773066298105e-05)
--(axis cs:4,2.67773066298105e-05)
--(axis cs:3,2.67777395492885e-05)
--(axis cs:2,2.6776862796396e-05)
--(axis cs:1,2.67422514298232e-05)
--cycle;

\path [draw=gray127, fill=gray127, opacity=0.2]
(axis cs:1,4.12679277417283e-08)
--(axis cs:1,3.02871654866976e-08)
--(axis cs:2,3.0396444472558e-08)
--(axis cs:3,3.03969461157294e-08)
--(axis cs:4,3.0397631434198e-08)
--(axis cs:4,4.10769096959029e-08)
--(axis cs:4,4.10769096959029e-08)
--(axis cs:3,4.0787790931418e-08)
--(axis cs:2,4.11013520107417e-08)
--(axis cs:1,4.12679277417283e-08)
--cycle;

\path [draw=steelblue31119180, fill=steelblue31119180, opacity=0.2]
(axis cs:1,2.05953010663507e-06)
--(axis cs:1,9.77179161054664e-07)
--(axis cs:2,4.6285100552268e-07)
--(axis cs:3,4.05029237526833e-07)
--(axis cs:4,3.89879829754136e-07)
--(axis cs:4,6.05255904275737e-07)
--(axis cs:4,6.05255904275737e-07)
--(axis cs:3,6.07708682309749e-07)
--(axis cs:2,6.65560673951404e-07)
--(axis cs:1,2.05953010663507e-06)
--cycle;

\path [draw=forestgreen4416044, fill=forestgreen4416044, opacity=0.2]
(axis cs:1,0.000152114746742882)
--(axis cs:1,0.0001520985388197)
--(axis cs:2,0.0001520985388197)
--(axis cs:3,0.0001520985388197)
--(axis cs:4,0.0001520985388197)
--(axis cs:4,0.000152114746742882)
--(axis cs:4,0.000152114746742882)
--(axis cs:3,0.000152114746742882)
--(axis cs:2,0.000152114746742882)
--(axis cs:1,0.000152114746742882)
--cycle;

\path [draw=forestgreen4416044, fill=forestgreen4416044, opacity=0.2]
(axis cs:1,0.000151916150643956)
--(axis cs:1,0.000151857809396461)
--(axis cs:2,0.000151802416075952)
--(axis cs:3,0.000151793414333952)
--(axis cs:4,0.000151794329067343)
--(axis cs:4,0.000151824165368453)
--(axis cs:4,0.000151824165368453)
--(axis cs:3,0.000151822969201021)
--(axis cs:2,0.000151824712520465)
--(axis cs:1,0.000151916150643956)
--cycle;

\path [draw=darkorange25512714, fill=darkorange25512714, opacity=0.2]
(axis cs:1,8.07417127361987e-06)
--(axis cs:1,4.41188667537062e-06)
--(axis cs:2,2.12070726774982e-06)
--(axis cs:3,1.40563099648716e-06)
--(axis cs:4,1.43030902108876e-06)
--(axis cs:4,3.87566437893838e-06)
--(axis cs:4,3.87566437893838e-06)
--(axis cs:3,4.13064692565968e-06)
--(axis cs:2,4.04199396939475e-06)
--(axis cs:1,8.07417127361987e-06)
--cycle;

\path [draw=steelblue31119180, fill=steelblue31119180, opacity=0.2]
(axis cs:1,9.3003149231663e-06)
--(axis cs:1,5.50785625819117e-06)
--(axis cs:2,1.94933816146659e-06)
--(axis cs:3,1.77678857653518e-06)
--(axis cs:4,1.29969203044311e-06)
--(axis cs:4,4.23282399424352e-06)
--(axis cs:4,4.23282399424352e-06)
--(axis cs:3,4.78755227959482e-06)
--(axis cs:2,5.72189076137874e-06)
--(axis cs:1,9.3003149231663e-06)
--cycle;

\path [draw=forestgreen4416044, fill=forestgreen4416044, opacity=0.2]
(axis cs:1,0.00015161550254561)
--(axis cs:1,0.000151588342851028)
--(axis cs:2,0.00015147702943068)
--(axis cs:3,0.000151468036347069)
--(axis cs:4,0.000151453414582647)
--(axis cs:4,0.000151488403207622)
--(axis cs:4,0.000151488403207622)
--(axis cs:3,0.00015151463157963)
--(axis cs:2,0.000151527466368861)
--(axis cs:1,0.00015161550254561)
--cycle;

\addplot [very thick, darkorange25512714, dash pattern=on 4pt off 2pt]
table {%
1 2.75561873422703e-05
2 2.75542814051732e-05
3 2.7554672487895e-05
4 2.75558722933056e-05
};
\addplot [very thick, darkorange25512714]
table {%
1 2.48522956098896e-06
2 1.54618844590004e-06
3 1.45347015632069e-06
4 1.445945292744e-06
};
\addplot [very thick, steelblue31119180, dash pattern=on 4pt off 2pt]
table {%
1 2.66248331172392e-05
2 2.66248393018031e-05
3 2.66248371190159e-05
4 2.66248422121862e-05
};
\addplot [very thick, gray127]
table {%
1 3.58196338368089e-08
2 3.58193592120415e-08
3 3.58196675875888e-08
4 3.58203656958267e-08
};
\addplot [line width=2pt, steelblue31119180]
table {%
1 1.43627767101862e-06
2 5.66459004858189e-07
3 5.06544205336468e-07
4 4.90820082177379e-07
};
\addplot [very thick, forestgreen4416044, dash pattern=on 4pt off 2pt]
table {%
1 0.000152106743189506
2 0.000152106743189506
3 0.000152106743189506
4 0.000152106743189506
};
\addplot [very thick, forestgreen4416044]
table {%
1 0.000151890330016613
2 0.000151813353295438
3 0.000151806868962012
4 0.000151807806105353
};
\addplot [very thick, darkorange25512714, dash pattern=on 1pt off 2pt]
table {%
1 6.19474212726345e-06
2 3.06858396470489e-06
3 2.56279556651862e-06
4 2.35844529470342e-06
};
\addplot [very thick, steelblue31119180, dash pattern=on 1pt off 2pt]
table {%
1 7.16930917405989e-06
2 3.5637345035866e-06
3 2.9705783845202e-06
4 2.35568472817249e-06
};
\addplot [very thick, forestgreen4416044, dash pattern=on 1pt off 2pt]
table {%
1 0.000151601922698319
2 0.000151502247899771
3 0.000151491188444197
4 0.000151470879791304
};
\end{axis}

\end{tikzpicture}%
\begin{tikzpicture}
\pgfplotsset{every tick label/.append style={font=\scriptsize}} 
\pgfplotsset{every axis label/.append style={font=\small}} 

\definecolor{darkgray}{RGB}{169,169,169}
\definecolor{darkorange25512714}{RGB}{255,127,14}
\definecolor{darkslategray38}{RGB}{38,38,38}
\definecolor{forestgreen4416044}{RGB}{44,160,44}
\definecolor{gray127}{RGB}{127,127,127}
\definecolor{lightgray204}{RGB}{204,204,204}
\definecolor{steelblue31119180}{RGB}{31,119,180}

\begin{axis}[
xlabel style={yshift=0.15cm},
ylabel style={yshift=-0.15cm},
title style={align=center, yshift=-0.2cm},
x=0.675cm,
height=8cm,
axis line style={darkgray},
log basis y={10},
tick align=outside,
tick pos=left,
title={\small{Deformable Plate} \\ \small{(Easy)}},
x grid style={lightgray204},
xlabel={Context Size},
yminorticks=true,
minor ytick={
0.00000002,
0.00000003, 
0.00000004,
0.00000005,
0.00000006,
0.00000007,
0.00000008,
0.00000009,
0.0000002,
0.0000003,
0.0000004,
0.0000005,
0.0000006, 
0.0000007,
0.0000008,
0.0000009,
0.000002, 
0.000003,
0.000004, 
0.000005,
0.000006, 
0.000007,
0.000008, 
0.000009,
0.00001,  
0.00002, 
0.00003,
0.00004, 
0.00005, 
0.00006,
0.00007,
0.00008, 
0.00009,
0.0002
},
xtick={1,2,3,4},
xmin=0.85, xmax=4.15,
xtick style={color=darkgray},
y grid style={lightgray204},
ymajorgrids,
xmajorgrids,
ymin=0.94741228683419e-06, ymax=0.00009,
ymode=log,
ytick pos=left,
ytick style={color=darkgray},
ytick={0.0000001, 0.000001, 0.00001, 0.0001},
yticklabels={$10^{-7}$, $10^{-6}$, $10^{-5}$, $10^{-4}$, },
]

\path [draw=darkorange25512714, fill=darkorange25512714, opacity=0.2]
(axis cs:1,5.22327070939355e-05)
--(axis cs:1,5.13468810822815e-05)
--(axis cs:2,5.13361061166506e-05)
--(axis cs:3,5.13631818830618e-05)
--(axis cs:4,5.1364424143685e-05)
--(axis cs:4,5.22394257131964e-05)
--(axis cs:4,5.22394257131964e-05)
--(axis cs:3,5.22273046954069e-05)
--(axis cs:2,5.22327136422973e-05)
--(axis cs:1,5.22327070939355e-05)
--cycle;

\path [draw=darkorange25512714, fill=darkorange25512714, opacity=0.2]
(axis cs:1,1.8694080381465e-06)
--(axis cs:1,1.60658814820636e-06)
--(axis cs:2,1.49793988839519e-06)
--(axis cs:3,1.4355596704263e-06)
--(axis cs:4,1.4362453157446e-06)
--(axis cs:4,1.77409679054108e-06)
--(axis cs:4,1.77409679054108e-06)
--(axis cs:3,1.79115311311762e-06)
--(axis cs:2,1.81472630629287e-06)
--(axis cs:1,1.8694080381465e-06)
--cycle;

\path [draw=steelblue31119180, fill=steelblue31119180, opacity=0.2]
(axis cs:1,5.14901847054716e-05)
--(axis cs:1,5.07158589971368e-05)
--(axis cs:2,5.07158812433772e-05)
--(axis cs:3,5.06979908095673e-05)
--(axis cs:4,5.06797623529565e-05)
--(axis cs:4,5.1452578190947e-05)
--(axis cs:4,5.1452578190947e-05)
--(axis cs:3,5.14490238856524e-05)
--(axis cs:2,5.14901876158547e-05)
--(axis cs:1,5.14901847054716e-05)
--cycle;

\path [draw=gray127, fill=gray127, opacity=0.2]
(axis cs:1,1.25154838315211e-06)
--(axis cs:1,9.8714750720319e-07)
--(axis cs:2,9.98750226699485e-07)
--(axis cs:3,9.98754615011421e-07)
--(axis cs:4,9.98754683223524e-07)
--(axis cs:4,1.25154147099238e-06)
--(axis cs:4,1.25154147099238e-06)
--(axis cs:3,1.25281119380816e-06)
--(axis cs:2,1.25279998428596e-06)
--(axis cs:1,1.25154838315211e-06)
--cycle;

\path [draw=steelblue31119180, fill=steelblue31119180, opacity=0.2]
(axis cs:1,1.52908387462958e-06)
--(axis cs:1,1.27810665162542e-06)
--(axis cs:2,1.19476019335707e-06)
--(axis cs:3,1.18766913601576e-06)
--(axis cs:4,1.17625613711425e-06)
--(axis cs:4,1.34696694033209e-06)
--(axis cs:4,1.34696694033209e-06)
--(axis cs:3,1.36030591875169e-06)
--(axis cs:2,1.357932319479e-06)
--(axis cs:1,1.52908387462958e-06)
--cycle;

\path [draw=forestgreen4416044, fill=forestgreen4416044, opacity=0.2]
(axis cs:1,6.10503439020249e-05)
--(axis cs:1,5.10667068738258e-05)
--(axis cs:2,5.10667068738258e-05)
--(axis cs:3,5.10667068738258e-05)
--(axis cs:4,5.10667068738258e-05)
--(axis cs:4,6.10503439020249e-05)
--(axis cs:4,6.10503439020249e-05)
--(axis cs:3,6.10503384450567e-05)
--(axis cs:2,6.10503393545514e-05)
--(axis cs:1,6.10503439020249e-05)
--cycle;

\path [draw=forestgreen4416044, fill=forestgreen4416044, opacity=0.2]
(axis cs:1,2.11589329524031e-06)
--(axis cs:1,1.75087734533008e-06)
--(axis cs:2,1.66767415521463e-06)
--(axis cs:3,1.64320584872257e-06)
--(axis cs:4,1.62486253429961e-06)
--(axis cs:4,2.00661892790777e-06)
--(axis cs:4,2.00661892790777e-06)
--(axis cs:3,2.01781347186625e-06)
--(axis cs:2,2.04923767910259e-06)
--(axis cs:1,2.11589329524031e-06)
--cycle;

\path [draw=darkorange25512714, fill=darkorange25512714, opacity=0.2]
(axis cs:1,2.57900474025519e-06)
--(axis cs:1,1.98174684555852e-06)
--(axis cs:2,1.71298126304009e-06)
--(axis cs:3,1.58621742230025e-06)
--(axis cs:4,1.59269366122317e-06)
--(axis cs:4,1.77409883690416e-06)
--(axis cs:4,1.77409883690416e-06)
--(axis cs:3,1.80664737854386e-06)
--(axis cs:2,1.85327830877213e-06)
--(axis cs:1,2.57900474025519e-06)
--cycle;

\path [draw=steelblue31119180, fill=steelblue31119180, opacity=0.2]
(axis cs:1,2.15726790884219e-06)
--(axis cs:1,1.37206463932671e-06)
--(axis cs:2,1.22659339467646e-06)
--(axis cs:3,1.131659632847e-06)
--(axis cs:4,1.13140959001612e-06)
--(axis cs:4,1.22248352454335e-06)
--(axis cs:4,1.22248352454335e-06)
--(axis cs:3,1.26583747714903e-06)
--(axis cs:2,1.37957561037183e-06)
--(axis cs:1,2.15726790884219e-06)
--cycle;

\path [draw=forestgreen4416044, fill=forestgreen4416044, opacity=0.2]
(axis cs:1,3.55811442176446e-06)
--(axis cs:1,2.19291121084098e-06)
--(axis cs:2,1.70976834255043e-06)
--(axis cs:3,1.69692165741253e-06)
--(axis cs:4,1.62502792022678e-06)
--(axis cs:4,2.49695534648708e-06)
--(axis cs:4,2.49695534648708e-06)
--(axis cs:3,2.66013748273508e-06)
--(axis cs:2,2.66875156285096e-06)
--(axis cs:1,3.55811442176446e-06)
--cycle;

\addplot [very thick, darkorange25512714, dash pattern=on 4pt off 2pt]
table {%
1 5.18228058353998e-05
2 5.1823539979523e-05
3 5.18237829965074e-05
4 5.18232423928566e-05
};
\addplot [very thick, darkorange25512714]
table {%
1 1.71850861079292e-06
2 1.64878908890387e-06
3 1.60946838150267e-06
4 1.6014035736589e-06
};
\addplot [very thick, steelblue31119180, dash pattern=on 4pt off 2pt]
table {%
1 5.1094083028147e-05
2 5.10940888489131e-05
3 5.10940888489131e-05
4 5.10940895765089e-05
};
\addplot [very thick, gray127]
table {%
1 1.1313413665448e-06
2 1.13133036165891e-06
3 1.13133679633393e-06
4 1.13133777404073e-06
};
\addplot [line width=2pt, steelblue31119180]
table {%
1 1.4080475011724e-06
2 1.27634625641804e-06
3 1.2672009461312e-06
4 1.25936028325668e-06
};
\addplot [very thick, forestgreen4416044, dash pattern=on 4pt off 2pt]
table {%
1 5.45769626114634e-05
2 5.45769617019687e-05
3 5.4576960792474e-05
4 5.45769626114634e-05
};
\addplot [very thick, forestgreen4416044]
table {%
1 1.88186280070113e-06
2 1.82064658815762e-06
3 1.796420718847e-06
4 1.78263687189428e-06
};
\addplot [very thick, darkorange25512714, dash pattern=on 1pt off 2pt]
table {%
1 2.24570767386467e-06
2 1.78301613686926e-06
3 1.70101632193109e-06
4 1.68339624906366e-06
};
\addplot [very thick, steelblue31119180, dash pattern=on 1pt off 2pt]
table {%
1 1.72148752426438e-06
2 1.29656766603148e-06
3 1.19874855499802e-06
4 1.17694655727973e-06
};
\addplot [very thick, forestgreen4416044, dash pattern=on 1pt off 2pt]
table {%
1 2.87273098820151e-06
2 2.09932463235418e-06
3 2.0795038722099e-06
4 1.99699681502352e-06
};
\end{axis}

\end{tikzpicture}%
\begin{tikzpicture}
\pgfplotsset{every tick label/.append style={font=\scriptsize}} 
\pgfplotsset{every axis label/.append style={font=\small}} 

\definecolor{darkgray}{RGB}{169,169,169}
\definecolor{darkorange25512714}{RGB}{255,127,14}
\definecolor{darkslategray38}{RGB}{38,38,38}
\definecolor{forestgreen4416044}{RGB}{44,160,44}
\definecolor{gray127}{RGB}{127,127,127}
\definecolor{lightgray204}{RGB}{204,204,204}
\definecolor{steelblue31119180}{RGB}{31,119,180}

\begin{axis}[
xlabel style={yshift=0.15cm},
ylabel style={yshift=-0.15cm},
title style={align=center, yshift=-0.2cm},
x=0.675cm,
height=8cm,
axis line style={darkgray},
log basis y={10},
tick align=outside,
tick pos=left,
title={\small{Deformable Plate} \\ \small{(Hard)}},
x grid style={lightgray204},
xlabel={Context Size},
yminorticks=true,
minor ytick={
0.00000002,
0.00000003, 
0.00000004,
0.00000005,
0.00000006,
0.00000007,
0.00000008,
0.00000009,
0.0000002,
0.0000003,
0.0000004,
0.0000005,
0.0000006, 
0.0000007,
0.0000008,
0.0000009,
0.000002, 
0.000003,
0.000004, 
0.000005,
0.000006, 
0.000007,
0.000008, 
0.000009,
0.00001,  
0.00002, 
0.00003,
0.00004, 
0.00005, 
0.00006,
0.00007,
0.00008, 
0.00009,
0.0002
},
xtick={1,2,3,4},
xmin=0.85, xmax=4.15,
xtick style={color=darkgray},
y grid style={lightgray204},
ymajorgrids,
xmajorgrids,
ymin=5.8045700380673e-07, ymax=2.0835657565936e-05,
ymode=log,
ytick pos=left,
ytick style={color=darkgray},
ytick={0.0000001, 0.000001, 0.00001, 0.0001},
yticklabels={$10^{-7}$, $10^{-6}$, $10^{-5}$, $10^{-4}$, },
]

\path [draw=darkorange25512714, fill=darkorange25512714, opacity=0.2]
(axis cs:1,1.61998992552981e-05)
--(axis cs:1,1.5841663844185e-05)
--(axis cs:2,1.58456885401392e-05)
--(axis cs:3,1.58460057718912e-05)
--(axis cs:4,1.58464838023065e-05)
--(axis cs:4,1.62003088917118e-05)
--(axis cs:4,1.62003088917118e-05)
--(axis cs:3,1.62006330356235e-05)
--(axis cs:2,1.61989442858612e-05)
--(axis cs:1,1.61998992552981e-05)
--cycle;

\path [draw=darkorange25512714, fill=darkorange25512714, opacity=0.2]
(axis cs:1,2.37252806982724e-06)
--(axis cs:1,2.16297944461985e-06)
--(axis cs:2,1.69871228877128e-06)
--(axis cs:3,1.55553441970824e-06)
--(axis cs:4,1.55193821967714e-06)
--(axis cs:4,1.63362902355857e-06)
--(axis cs:4,1.63362902355857e-06)
--(axis cs:3,1.67363552918687e-06)
--(axis cs:2,1.89014274383226e-06)
--(axis cs:1,2.37252806982724e-06)
--cycle;

\path [draw=steelblue31119180, fill=steelblue31119180, opacity=0.2]
(axis cs:1,1.53653905726969e-05)
--(axis cs:1,1.50947898691811e-05)
--(axis cs:2,1.50942729305825e-05)
--(axis cs:3,1.5077056013979e-05)
--(axis cs:4,1.50942882100935e-05)
--(axis cs:4,1.53659195348155e-05)
--(axis cs:4,1.53659195348155e-05)
--(axis cs:3,1.53653994129854e-05)
--(axis cs:2,1.5364847422461e-05)
--(axis cs:1,1.53653905726969e-05)
--cycle;

\path [draw=gray127, fill=gray127, opacity=0.2]
(axis cs:1,7.94611162291403e-07)
--(axis cs:1,6.80167477185023e-07)
--(axis cs:2,6.81033486671367e-07)
--(axis cs:3,6.80169137012854e-07)
--(axis cs:4,6.81719814110693e-07)
--(axis cs:4,7.94609638887778e-07)
--(axis cs:4,7.94609638887778e-07)
--(axis cs:3,7.94610161847231e-07)
--(axis cs:2,8.00542568413221e-07)
--(axis cs:1,7.94611162291403e-07)
--cycle;

\path [draw=steelblue31119180, fill=steelblue31119180, opacity=0.2]
(axis cs:1,1.69180580087414e-06)
--(axis cs:1,1.29036261569127e-06)
--(axis cs:2,1.12666918994364e-06)
--(axis cs:3,9.61776549956994e-07)
--(axis cs:4,9.50767775975692e-07)
--(axis cs:4,1.06810741726804e-06)
--(axis cs:4,1.06810741726804e-06)
--(axis cs:3,1.09566065020772e-06)
--(axis cs:2,1.48230266233895e-06)
--(axis cs:1,1.69180580087414e-06)
--cycle;

\path [draw=forestgreen4416044, fill=forestgreen4416044, opacity=0.2]
(axis cs:1,1.61081799160456e-05)
--(axis cs:1,1.55157158587826e-05)
--(axis cs:2,1.55250600073487e-05)
--(axis cs:3,1.55157213157509e-05)
--(axis cs:4,1.55156962136971e-05)
--(axis cs:4,1.60920306370826e-05)
--(axis cs:4,1.60920306370826e-05)
--(axis cs:3,1.61097392265219e-05)
--(axis cs:2,1.61099917022511e-05)
--(axis cs:1,1.61081799160456e-05)
--cycle;

\path [draw=forestgreen4416044, fill=forestgreen4416044, opacity=0.2]
(axis cs:1,2.71203407464782e-06)
--(axis cs:1,1.70307118878554e-06)
--(axis cs:2,1.53560808939801e-06)
--(axis cs:3,1.44807952437986e-06)
--(axis cs:4,1.49460204283969e-06)
--(axis cs:4,2.00753501587769e-06)
--(axis cs:4,2.00753501587769e-06)
--(axis cs:3,2.06602467187622e-06)
--(axis cs:2,2.32250601015949e-06)
--(axis cs:1,2.71203407464782e-06)
--cycle;

\path [draw=darkorange25512714, fill=darkorange25512714, opacity=0.2]
(axis cs:1,3.23821254823997e-06)
--(axis cs:1,2.33658470278897e-06)
--(axis cs:2,1.76758651377895e-06)
--(axis cs:3,1.49917459566495e-06)
--(axis cs:4,1.35482928271813e-06)
--(axis cs:4,1.56116227572056e-06)
--(axis cs:4,1.56116227572056e-06)
--(axis cs:3,1.69528616424941e-06)
--(axis cs:2,2.16058206945036e-06)
--(axis cs:1,3.23821254823997e-06)
--cycle;

\path [draw=steelblue31119180, fill=steelblue31119180, opacity=0.2]
(axis cs:1,4.98428985338251e-06)
--(axis cs:1,2.26546151679941e-06)
--(axis cs:2,1.73743721916253e-06)
--(axis cs:3,1.3710496028807e-06)
--(axis cs:4,1.27935131786217e-06)
--(axis cs:4,2.2199069690032e-06)
--(axis cs:4,2.2199069690032e-06)
--(axis cs:3,2.42668013470393e-06)
--(axis cs:2,3.43485485245765e-06)
--(axis cs:1,4.98428985338251e-06)
--cycle;

\path [draw=forestgreen4416044, fill=forestgreen4416044, opacity=0.2]
(axis cs:1,3.59596430143938e-06)
--(axis cs:1,2.7229395982431e-06)
--(axis cs:2,2.13179032471089e-06)
--(axis cs:3,1.83777731876944e-06)
--(axis cs:4,1.81510404217988e-06)
--(axis cs:4,2.07236149663004e-06)
--(axis cs:4,2.07236149663004e-06)
--(axis cs:3,2.33847765684914e-06)
--(axis cs:2,2.82473570223374e-06)
--(axis cs:1,3.59596430143938e-06)
--cycle;

\addplot [very thick, darkorange25512714, dash pattern=on 4pt off 2pt]
table {%
1 1.60291496285936e-05
2 1.60285100719193e-05
3 1.60299179697176e-05
4 1.60296494868817e-05
};
\addplot [very thick, darkorange25512714]
table {%
1 2.26427127927309e-06
2 1.80666138476226e-06
3 1.62045453180326e-06
4 1.59600140250404e-06
};
\addplot [very thick, steelblue31119180, dash pattern=on 4pt off 2pt]
table {%
1 1.52351700307918e-05
2 1.52351623910363e-05
3 1.52351705764886e-05
4 1.52351758515579e-05
};
\addplot [very thick, gray127]
table {%
1 7.31373097551113e-07
2 7.31372381324036e-07
3 7.31371653728274e-07
4 7.31372313111933e-07
};
\addplot [line width=2pt, steelblue31119180]
table {%
1 1.4910842082827e-06
2 1.30109697238368e-06
3 1.02998503734852e-06
4 1.00975116765767e-06
};
\addplot [very thick, forestgreen4416044, dash pattern=on 4pt off 2pt]
table {%
1 1.58127480972325e-05
2 1.58128637849586e-05
3 1.58127302711364e-05
4 1.58127786562545e-05
};
\addplot [very thick, forestgreen4416044]
table {%
1 2.13807509226172e-06
2 1.87837235898769e-06
3 1.72654767993663e-06
4 1.72150114394753e-06
};
\addplot [very thick, darkorange25512714, dash pattern=on 1pt off 2pt]
table {%
1 2.82132914435351e-06
2 1.97798451608833e-06
3 1.6072558310043e-06
4 1.46738104831456e-06
};
\addplot [very thick, steelblue31119180, dash pattern=on 1pt off 2pt]
table {%
1 3.50282221006637e-06
2 2.38380114296888e-06
3 1.80835977516836e-06
4 1.68624728758004e-06
};
\addplot [very thick, forestgreen4416044, dash pattern=on 1pt off 2pt]
table {%
1 3.08086333689062e-06
2 2.47826301347231e-06
3 2.05984233048184e-06
4 1.93035856455026e-06
};
\end{axis}

\end{tikzpicture}%
\begin{tikzpicture}
\pgfplotsset{every tick label/.append style={font=\scriptsize}} 
\pgfplotsset{every axis label/.append style={font=\small}}

\definecolor{darkgray}{RGB}{169,169,169}
\definecolor{darkorange25512714}{RGB}{255,127,14}
\definecolor{darkslategray38}{RGB}{38,38,38}
\definecolor{forestgreen4416044}{RGB}{44,160,44}
\definecolor{gray127}{RGB}{127,127,127}
\definecolor{lightgray204}{RGB}{204,204,204}
\definecolor{steelblue31119180}{RGB}{31,119,180}

\begin{axis}[
xlabel style={yshift=0.15cm},
ylabel style={yshift=-0.15cm},
title style={align=center, yshift=-0.2cm},
x=0.675cm,
height=8cm,
axis line style={darkgray},
log basis y={10},
tick align=outside,
tick pos=left,
title={\small{Sphere Cloth} \\ \small{Coupling}},
x grid style={lightgray204},
xlabel={Context Size},
yminorticks=true,
minor ytick={
0.00000002,
0.00000003, 
0.00000004,
0.00000005,
0.00000006,
0.00000007,
0.00000008,
0.00000009,
0.0000002,
0.0000003,
0.0000004,
0.0000005,
0.0000006, 
0.0000007,
0.0000008,
0.0000009,
0.000002, 
0.000003,
0.000004, 
0.000005,
0.000006, 
0.000007,
0.000008, 
0.000009,
0.00001,  
0.00002, 
0.00003,
0.00004, 
0.00005, 
0.00006,
0.00007,
0.00008, 
0.00009,
0.0002
},
xtick={1,2,3,4},
xmin=0.85, xmax=4.15,
xtick style={color=darkgray},
y grid style={lightgray204},
ymajorgrids,
xmajorgrids,
ymin=2.54741228683419e-07, ymax=0.000371669015013751,
ymode=log,
ytick pos=left,
ytick style={color=darkgray},
ytick={0.0000001, 0.000001, 0.00001, 0.0001},
yticklabels={$10^{-7}$, $10^{-6}$, $10^{-5}$, $10^{-4}$, },
]

\path [draw=darkorange25512714, fill=darkorange25512714, opacity=0.2]
(axis cs:1,0.000266834322537761)
--(axis cs:1,0.00023741445329506)
--(axis cs:2,0.00023730483953841)
--(axis cs:3,0.000237448999541812)
--(axis cs:4,0.000237397351884283)
--(axis cs:4,0.000266892013823963)
--(axis cs:4,0.000266892013823963)
--(axis cs:3,0.000266829435713589)
--(axis cs:2,0.000266864159493707)
--(axis cs:1,0.000266834322537761)
--cycle;

\path [draw=darkorange25512714, fill=darkorange25512714, opacity=0.2]
(axis cs:1,1.99593317574909e-05)
--(axis cs:1,1.34206902657752e-05)
--(axis cs:2,1.08034444565419e-05)
--(axis cs:3,1.01735138741788e-05)
--(axis cs:4,9.82726368874864e-06)
--(axis cs:4,1.63123884703964e-05)
--(axis cs:4,1.63123884703964e-05)
--(axis cs:3,1.62179711514909e-05)
--(axis cs:2,1.65879917858547e-05)
--(axis cs:1,1.99593317574909e-05)
--cycle;

\path [draw=steelblue31119180, fill=steelblue31119180, opacity=0.2]
(axis cs:1,0.000233529687684495)
--(axis cs:1,0.000227532393182628)
--(axis cs:2,0.000227670636377297)
--(axis cs:3,0.00022745122842025)
--(axis cs:4,0.00022756096732337)
--(axis cs:4,0.000233527607633732)
--(axis cs:4,0.000233527607633732)
--(axis cs:3,0.000233608766575344)
--(axis cs:2,0.000233608763664961)
--(axis cs:1,0.000233529687684495)
--cycle;

\path [draw=gray127, fill=gray127, opacity=0.2]
(axis cs:1,6.02016461925814e-07)
--(axis cs:1,3.56154203018377e-07)
--(axis cs:2,3.56154197334035e-07)
--(axis cs:3,3.54748050312992e-07)
--(axis cs:4,3.57596132971594e-07)
--(axis cs:4,6.23642591790485e-07)
--(axis cs:4,6.23642591790485e-07)
--(axis cs:3,6.19353585307181e-07)
--(axis cs:2,6.02017610162875e-07)
--(axis cs:1,6.02016461925814e-07)
--cycle;

\path [draw=steelblue31119180, fill=steelblue31119180, opacity=0.2]
(axis cs:1,1.50739363931507e-06)
--(axis cs:1,6.83883388319373e-07)
--(axis cs:2,5.89286818382106e-07)
--(axis cs:3,5.63330831937492e-07)
--(axis cs:4,5.45648333627469e-07)
--(axis cs:4,1.35409551376142e-06)
--(axis cs:4,1.35409551376142e-06)
--(axis cs:3,1.38144195034329e-06)
--(axis cs:2,1.39615200396292e-06)
--(axis cs:1,1.50739363931507e-06)
--cycle;

\path [draw=forestgreen4416044, fill=forestgreen4416044, opacity=0.2]
(axis cs:1,0.000238249872927554)
--(axis cs:1,0.000233389572458691)
--(axis cs:2,0.000232989684445783)
--(axis cs:3,0.000232986159317079)
--(axis cs:4,0.000232989678625017)
--(axis cs:4,0.000238391104576294)
--(axis cs:4,0.000238391104576294)
--(axis cs:3,0.000238249872927554)
--(axis cs:2,0.000238249870017171)
--(axis cs:1,0.000238249872927554)
--cycle;

\path [draw=forestgreen4416044, fill=forestgreen4416044, opacity=0.2]
(axis cs:1,4.52554256185067e-05)
--(axis cs:1,2.20703952891199e-06)
--(axis cs:2,1.56470923684537e-06)
--(axis cs:3,1.45303121712459e-06)
--(axis cs:4,1.35893938590925e-06)
--(axis cs:4,4.23187290721216e-05)
--(axis cs:4,4.23187290721216e-05)
--(axis cs:3,4.21480341969982e-05)
--(axis cs:2,4.31863585390602e-05)
--(axis cs:1,4.52554256185067e-05)
--cycle;

\path [draw=darkorange25512714, fill=darkorange25512714, opacity=0.2]
(axis cs:1,7.31861682652379e-06)
--(axis cs:1,4.6436761294899e-06)
--(axis cs:2,4.27077866334002e-06)
--(axis cs:3,4.16969555772084e-06)
--(axis cs:4,4.14492687923484e-06)
--(axis cs:4,5.99889381192042e-06)
--(axis cs:4,5.99889381192042e-06)
--(axis cs:3,6.10418683208991e-06)
--(axis cs:2,6.10588249401189e-06)
--(axis cs:1,7.31861682652379e-06)
--cycle;

\path [draw=steelblue31119180, fill=steelblue31119180, opacity=0.2]
(axis cs:1,8.81600990396692e-06)
--(axis cs:1,2.01579777012739e-06)
--(axis cs:2,1.02748805375086e-06)
--(axis cs:3,8.77398110787908e-07)
--(axis cs:4,7.27246913356794e-07)
--(axis cs:4,1.75057425622072e-06)
--(axis cs:4,1.75057425622072e-06)
--(axis cs:3,2.31525654044162e-06)
--(axis cs:2,3.31791185317343e-06)
--(axis cs:1,8.81600990396692e-06)
--cycle;

\path [draw=forestgreen4416044, fill=forestgreen4416044, opacity=0.2]
(axis cs:1,5.27795828020317e-06)
--(axis cs:1,1.87936143447587e-06)
--(axis cs:2,1.28507139152134e-06)
--(axis cs:3,1.15994168936595e-06)
--(axis cs:4,1.08867430981263e-06)
--(axis cs:4,2.18725566014655e-06)
--(axis cs:4,2.18725566014655e-06)
--(axis cs:3,2.54185749781755e-06)
--(axis cs:2,2.91638741600764e-06)
--(axis cs:1,5.27795828020317e-06)
--cycle;

\addplot [very thick, darkorange25512714, dash pattern=on 4pt off 2pt]
table {%
1 0.000249824969796464
2 0.000249805880594067
3 0.000249825048376806
4 0.000249799250741489
};
\addplot [very thick, darkorange25512714]
table {%
1 1.60461988343741e-05
2 1.32186802147771e-05
3 1.28806133943726e-05
4 1.26951237689354e-05
};
\addplot [very thick, steelblue31119180, dash pattern=on 4pt off 2pt]
table {%
1 0.000230399493011646
2 0.000230399475549348
3 0.000230399466818199
4 0.000230399478459731
};
\addplot [very thick, gray127]
table {%
1 4.61645589666659e-07
2 4.61645970517566e-07
3 4.61645765881258e-07
4 4.61645419136403e-07
};
\addplot [line width=2pt, steelblue31119180]
table {%
1 1.02496456975132e-06
2 9.05943176121582e-07
3 8.84228842323864e-07
4 8.80752793364081e-07
};
\addplot [very thick, forestgreen4416044, dash pattern=on 4pt off 2pt]
table {%
1 0.000236006203340366
2 0.000236006200429983
3 0.000236006200429983
4 0.000236006203340366
};
\addplot [very thick, forestgreen4416044]
table {%
1 1.65942093985905e-05
2 1.56216560753819e-05
3 1.52216343565215e-05
4 1.5243597118797e-05
};
\addplot [very thick, darkorange25512714, dash pattern=on 1pt off 2pt]
table {%
1 5.93824643146945e-06
2 5.15704155077401e-06
3 5.04235590597091e-06
4 4.95309727739368e-06
};
\addplot [very thick, steelblue31119180, dash pattern=on 1pt off 2pt]
table {%
1 4.99552165820205e-06
2 1.9982504568361e-06
3 1.50725106777827e-06
4 1.16728816692557e-06
};
\addplot [very thick, forestgreen4416044, dash pattern=on 1pt off 2pt]
table {%
1 3.57865985733952e-06
2 2.10072940376449e-06
3 1.85089959359175e-06
4 1.6135350051627e-06
};
\end{axis}

\end{tikzpicture}%
    \caption{Performance comparison of our proposed methods and baselines across four datasets.
We report the mean and 95\% confidence interval of the \emph{Full Rollout MSE} over five random seeds.
The $x$-axis indicates the number of context samples used by the meta-learning approaches.
We compare (i) meta-learning methods that employ a latent simulation parameter representation, (ii) a two-stage setup that explicitly predicts simulation parameters, and (iii) a baseline model that ignores context and performs direct simulation.
The MaNGO (Oracle) variant, which has privileged access to the ground-truth simulation parameters, serves as an upper performance bound.
Overall, MaNGO consistently outperforms both non–meta-learning and alternative meta-learning approaches, and achieves results close to the oracle.
    \vspace{-0.7cm}}
    \label{fig:main_eval}
\end{figure}

\textbf{Material estimation.}
Determining physical parameters from observational data is a challenging and ill-posed problem \cite{isakov2006inverse}. Machine learning methods have shown success in inferring material properties from videos \cite{Takahashi2019Video,ma2022risp,Feng2022Visual} and point clouds \cite{Wang2015Deformation}, but they rely on knowledge of the underlying PDE. 
Recently, approaches utilizing pre-trained Graph Network Simulators (GNS) to infer material parameters gained popularity due to their computational efficiency and differentiability \cite{qzhao2022graphpde, wang2024latent}. By back-propagating through the learned simulator, these methods estimate latent material codes directly from observations. While the approach in \citet{qzhao2022graphpde} and ours share this goal, only we aggregate a context set of simulation trials to extract the underlying structure. Furthermore, our method does not require any backward pass and model weight updates, resulting in a faster adaptation.

\textbf{Meta-learning and Neural Processes.}
Learning models that can quickly adapt to small context datasets at test time is often referred to as Meta-Learning. This emerging paradigm has found wide applications in fields such as language models \cite{brown2020language} and robotics \cite{vosylius2024instantpolicyincontextimitation,di2024keypoint}, where it is commonly referred to as In-Context Learning. Meta-Learning is typically categorized into two approaches: optimization-based methods with few-shot examples and context-aggregation methods in a zero-shot fashion. A prominent example of the former is Model-Agnostic Meta-Learning (MAML)~\citep{finn_model-agnostic_2017,grant_recasting_2018,finn_probabilistic_2018,kim_bayesian_2018}, which employs gradient-based updates to adapt the model to new tasks using few examples. In the latter, Neural Processes (NPs) \cite{garnelo_conditional_2018,Garnelo2018NeuralP,kim_attentive_2019,gordon_meta-learning_2019,louizos_functional_2019,Volpp2021BayesianCA,volpp2023} aggregate latent features from a variable-sized context set to produce a latent task description that can be used directly during inference.  

In this paper, we adopt Conditional Neural Processes (CNPs)~\citep{garnelo_conditional_2018} as the core mechanism for aggregating context from different simulation trials to produce a latent description representing the simulation parameters. A \gls{gns} is then trained to condition on this latent descriptor, enabling it to adapt dynamically to new materials. To the best of our knowledge, this is the first time such a framework has been proposed, combining meta-learning with graph network simulators to quickly adapt to unknown material properties given only few context-examples without requiring retraining or knowledge of the underlying PDE.

\vspace{-0.2cm}
\section{Experiments}
\vspace{-0.2cm}
\label{sec:experiments}

\begin{figure}[t]
    \begin{tikzpicture}
    \tikzstyle{every node}=[font=\small]
    \input{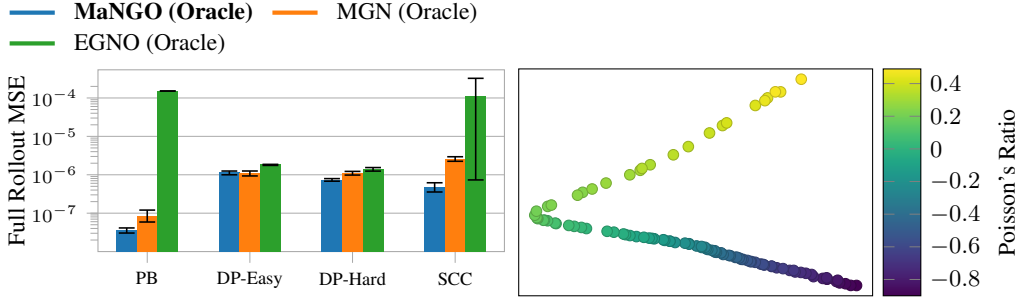}
    \begin{axis}[
        hide axis,
        xmin=0, xmax=1, 
        ymin=0, ymax=1,
        scale only axis, 
        width=0pt, height=0pt, 
        legend style={
            draw=none,
            legend cell align=left,
            legend columns=2,
            column sep=1ex,
            line width=1pt,
            at={(0.5,-0.1)}, 
            anchor=north, 
        },
        ]
        \addlegendimage{line legend, tabblue, ultra thick} 
        \addlegendentry{\textbf{MaNGO (Oracle)}}
        \addlegendimage{line legend, taborange, ultra thick} 
        \addlegendentry{MGN (Oracle)}
        \addlegendimage{line legend, tabgreen, ultra thick} 
        \addlegendentry{EGNO (Oracle)}
    \end{axis}
\end{tikzpicture}\\
    \begin{tikzpicture}
\pgfplotsset{every tick label/.append style={font=\scriptsize}} 
\pgfplotsset{every axis label/.append style={font=\small}} 

\definecolor{darkgray}{RGB}{169,169,169}
\definecolor{darkgray176}{RGB}{176,176,176}
\definecolor{darkorange25512714}{RGB}{255,127,14}
\definecolor{forestgreen4416044}{RGB}{44,160,44}
\definecolor{lightgray204}{RGB}{204,204,204}
\definecolor{steelblue31119180}{RGB}{31,119,180}


\begin{axis}[
height=4cm,
width=0.5\textwidth,
axis line style={darkgray},
legend cell align={left},
legend style={
  fill opacity=0.8,
  draw opacity=1,
  text opacity=1,
  at={(0.03,0.97)},
  anchor=north west,
  draw=lightgray204
},
minor ytick={
0.00000002,
0.00000003, 
0.00000004,
0.00000005,
0.00000006,
0.00000007,
0.00000008,
0.00000009,
0.0000001,
0.0000002,
0.0000003,
0.0000004,
0.0000005,
0.0000006, 
0.0000007,
0.0000008,
0.0000009,
0.000002, 
0.000003,
0.000004, 
0.000005,
0.000006, 
0.000007,
0.000008, 
0.000009,
0.00002, 
0.00003,
0.00004, 
0.00005, 
0.00006,
0.00007,
0.00008, 
0.00009,
0.0002
},
tick align=outside,
tick pos=left,
x grid style={darkgray176},
xmin=-0.48, xmax=3.48,
xtick style={color=darkgray},
xtick={0,1,2,3},
xticklabels={PB, DP-Easy,DP-Hard,SCC},
ymajorgrids,
y grid style={darkgray176},
ylabel={Full Rollout MSE},
y dir=normal,
ymin=0.00000001, ymax=0.00054082278580172,
ylabel style={yshift=-0.15cm},
ymode=log,
ytick style={color=darkgray},
ytick={0.0000001, 0.000001, 0.00001, 0.0001},
yticklabels={$10^{-7}$, $10^{-6}$, $10^{-5}$, $10^{-4}$, } 
]
\draw[draw=none,fill=steelblue31119180] (axis cs:-0.3,1e-12) rectangle (axis cs:-0.1,3.58191023508425e-08);
\draw[draw=none,fill=steelblue31119180] (axis cs:0.7,1e-12) rectangle (axis cs:0.9,1.13132830392715e-06);
\draw[draw=none,fill=steelblue31119180] (axis cs:1.7,1e-12) rectangle (axis cs:1.9,7.31367777007108e-07);
\draw[draw=none,fill=steelblue31119180] (axis cs:2.7,1e-12) rectangle (axis cs:2.9,4.61645100813257e-07);

\draw[draw=none,fill=darkorange25512714] (axis cs:-0.1,1e-12) rectangle (axis cs:0.1,8.28291177867868e-08);
\draw[draw=none,fill=darkorange25512714] (axis cs:0.9,1e-12) rectangle (axis cs:1.1,1.09302914097498e-06);
\draw[draw=none,fill=darkorange25512714] (axis cs:1.9,1e-12) rectangle (axis cs:2.1,1.10352482352027e-06);
\draw[draw=none,fill=darkorange25512714] (axis cs:2.9,1e-12) rectangle (axis cs:3.1,2.58124214269628e-06);

\draw[draw=none,fill=forestgreen4416044] (axis cs:0.1,1e-12) rectangle (axis cs:0.3,0.00015158363385126);
\draw[draw=none,fill=forestgreen4416044] (axis cs:1.1,1e-12) rectangle (axis cs:1.3,1.82571602636017e-06);
\draw[draw=none,fill=forestgreen4416044] (axis cs:2.1,1e-12) rectangle (axis cs:2.3,1.42075873554859e-06);
\draw[draw=none,fill=forestgreen4416044] (axis cs:3.1,1e-12) rectangle (axis cs:3.3,0.000108705760794692);

\path [draw=black, semithick]
(axis cs:-0.2,3.03961513736795e-08)
--(axis cs:-0.2,4.12672633842703e-08);

\path [draw=black, semithick]
(axis cs:0.8,9.98750238068169e-07)
--(axis cs:0.8,1.25152871532919e-06);

\path [draw=black, semithick]
(axis cs:1.8,6.80163680044643e-07)
--(axis cs:1.8,7.94609604781726e-07);

\path [draw=black, semithick]
(axis cs:2.8,3.56154203018377e-07)
--(axis cs:2.8,6.19422881698028e-07);

\addplot [semithick, black, mark=-, mark size=3, mark options={solid}, only marks]
table {%
-0.2 3.03961513736795e-08
0.8 9.98750238068169e-07
1.8 6.80163680044643e-07
2.8 3.56154203018377e-07
};

\addplot [semithick, black, mark=-, mark size=3, mark options={solid}, only marks]
table {%
-0.2 4.12672633842703e-08
0.8 1.25152871532919e-06
1.8 7.94609604781726e-07
2.8 6.19422881698028e-07
};

\path [draw=black, semithick]
(axis cs:0,5.8877759556708e-08)
--(axis cs:0,1.203692221452e-07);

\path [draw=black, semithick]
(axis cs:1,9.32439604639512e-07)
--(axis cs:1,1.25361850678019e-06);

\path [draw=black, semithick]
(axis cs:2,9.82421170192538e-07)
--(axis cs:2,1.22462847684801e-06);

\path [draw=black, semithick]
(axis cs:3,2.24528866965557e-06)
--(axis cs:3,2.94834671876743e-06);

\addplot [semithick, black, mark=-, mark size=3, mark options={solid}, only marks]
table {%
0 5.8877759556708e-08
1 9.32439604639512e-07
2 9.82421170192538e-07
3 2.24528866965557e-06
};

\addplot [semithick, black, mark=-, mark size=3, mark options={solid}, only marks]
table {%
0 1.203692221452e-07
1 1.25361850678019e-06
2 1.22462847684801e-06
3 2.94834671876743e-06
};

\path [draw=black, semithick]
(axis cs:0.2,0.000151561078382656)
--(axis cs:0.2,0.000151603220729157);

\path [draw=black, semithick]
(axis cs:1.2,1.75241007127624e-06)
--(axis cs:1.2,1.86839383786719e-06);

\path [draw=black, semithick]
(axis cs:2.2,1.24559653613687e-06)
--(axis cs:2.2,1.54152019149478e-06);

\path [draw=black, semithick]
(axis cs:3.2,7.34354728137987e-07)
--(axis cs:3.2,0.000324593129334971);

\addplot [semithick, black, mark=-, mark size=3, mark options={solid}, only marks]
table {%
0.2 0.000151561078382656
1.2 1.75241007127624e-06
2.2 1.24559653613687e-06
3.2 7.34354728137987e-07
};

\addplot [semithick, black, mark=-, mark size=3, mark options={solid}, only marks]
table {%
0.2 0.000151603220729157
1.2 1.86839383786719e-06
2.2 1.54152019149478e-06
3.2 0.000324593129334971
};

\end{axis}

\end{tikzpicture}
\begin{tikzpicture}

\definecolor{darkgray176}{RGB}{176,176,176}

\begin{axis}[
colorbar,
colorbar style={
    font=\small,
    ylabel={Poisson's Ratio},
    yticklabel style={font=\small},
    at={(1.03,0.5)}, 
    anchor=west,
    ytick={-0.8,-0.6,-0.4,-0.2, 0.0,0.2, 0.4}
},
height=4.6cm,
width=0.45\textwidth,
colormap/viridis,
point meta max=0.490000009536743,
point meta min=-0.899999976158142,
tick align=outside,
tick pos=left,
x grid style={darkgray176},
xmin=1.1938727915287, xmax=4.65765692591667,
xtick style={color=black},
y grid style={darkgray176},
ymin=-6.52396874427795, ymax=16.1255150318146,
ytick style={color=black},
xtick=\empty,   
ytick=\empty,   
]
\addplot [colormap/viridis, only marks, scatter, scatter src=explicit]
table [x=x, y=y, meta=colordata]{%
x  y  colordata
4.5002121925354 -5.49444675445557 -0.8999999761581421
4.44368982315063 -5.48879528045654 -0.8859596252441406
4.39245176315308 -5.3196234703064 -0.8719192147254944
4.3348126411438 -5.23547649383545 -0.8578788042068481
4.20999050140381 -4.86789989471436 -0.8438383936882019
4.30291223526001 -5.09689044952393 -0.8297979831695557
4.31608486175537 -4.97504377365112 -0.8157575726509094
4.21514415740967 -4.83798933029175 -0.8017171621322632
4.16975975036621 -4.79019737243652 -0.7876767516136169
4.20757389068604 -4.72850036621094 -0.7736363410949707
4.07900762557983 -4.61186981201172 -0.7595959305763245
4.06074237823486 -4.39983510971069 -0.745555579662323
4.00487995147705 -4.2795991897583 -0.7315151691436768
3.95506453514099 -4.19195461273193 -0.7174747586250305
3.89936900138855 -4.1444845199585 -0.7034343481063843
3.84641098976135 -3.99131679534912 -0.689393937587738
3.88546252250671 -3.99144268035889 -0.6753535270690918
3.71905946731567 -3.72754120826721 -0.6613131165504456
3.78087711334229 -3.81150197982788 -0.6472727060317993
3.66939187049866 -3.59827375411987 -0.6332322955131531
3.6690833568573 -3.55882692337036 -0.6191919445991516
3.59780192375183 -3.4534866809845 -0.6051515340805054
3.56331944465637 -3.3586893081665 -0.5911111235618591
3.58687424659729 -3.3991003036499 -0.5770707130432129
3.50884222984314 -3.18222379684448 -0.5630303025245667
3.47118806838989 -3.10946369171143 -0.5489898920059204
3.4612238407135 -3.11251211166382 -0.5349494814872742
3.43274450302124 -2.99565267562866 -0.5209090709686279
3.38938903808594 -2.89411401748657 -0.5068686604499817
3.37236285209656 -2.88323879241943 -0.49282827973365784
3.35146450996399 -2.75868844985962 -0.4787878692150116
3.31444597244263 -2.69331097602844 -0.46474748849868774
3.32757306098938 -2.67219066619873 -0.4507070779800415
3.26466584205627 -2.52736854553223 -0.43666666746139526
3.25052237510681 -2.48921060562134 -0.422626256942749
3.19532179832458 -2.39831686019897 -0.4085858464241028
3.17493391036987 -2.23784780502319 -0.39454546570777893
3.12905049324036 -2.18388438224792 -0.3805050551891327
3.11428833007812 -2.15022706985474 -0.36646464467048645
3.11296987533569 -2.02717518806458 -0.3524242341518402
3.08454036712646 -1.99023616313934 -0.33838382363319397
3.04458212852478 -1.89804995059967 -0.3243434429168701
3.02912616729736 -1.8275271654129 -0.3103030323982239
3.01797652244568 -1.6867663860321 -0.29626262187957764
2.98367428779602 -1.69686996936798 -0.2822222113609314
2.93479990959167 -1.54953169822693 -0.26818183064460754
2.9829797744751 -1.60211253166199 -0.2541414201259613
2.86175894737244 -1.32704496383667 -0.24010100960731506
2.86168909072876 -1.37107491493225 -0.22606059908866882
2.82138228416443 -1.27127301692963 -0.21202020347118378
2.72510838508606 -1.20382106304169 -0.19797979295253754
2.67412042617798 -1.0763601064682 -0.1839393973350525
2.67369294166565 -1.07801151275635 -0.16989898681640625
2.62313795089722 -0.993236124515533 -0.1558585911989212
2.60870146751404 -0.931271731853485 -0.14181818068027496
2.52085185050964 -0.833170354366302 -0.12777778506278992
2.49833226203918 -0.794303417205811 -0.11373737454414368
2.4703004360199 -0.765838265419006 -0.09969697147607803
2.34614968299866 -0.613447904586792 -0.08565656840801239
2.41880869865417 -0.686249673366547 -0.07161616533994675
2.28257966041565 -0.510964393615723 -0.057575758546590805
2.23929905891418 -0.41972079873085 -0.04353535175323486
2.21897602081299 -0.356471002101898 -0.02949494868516922
2.12552714347839 -0.17357063293457 -0.015454545617103577
2.0592885017395 -0.0863124430179596 -0.001414141384884715
1.85667455196381 0.190050303936005 0.01262626238167286
1.80756390094757 0.299360454082489 0.02666666731238365
1.73784637451172 0.406549870967865 0.040707070380449295
1.68389081954956 0.499588906764984 0.05474747344851494
1.55969858169556 0.684962391853333 0.06878788024187088
1.55056202411652 0.72787207365036 0.08282828330993652
1.47645843029022 0.96714061498642 0.09686868637800217
1.44638526439667 1.02447617053986 0.11090908944606781
1.43582594394684 1.01403069496155 0.12494949251413345
1.38084888458252 1.17656350135803 0.1389898955821991
1.3760883808136 1.22991454601288 0.15303030610084534
1.37136507034302 1.27923953533173 0.16707070171833038
1.35131752490997 1.52061069011688 0.18111111223697662
1.36725997924805 1.92006707191467 0.19515150785446167
1.47750806808472 2.39934778213501 0.2091919183731079
1.51343476772308 2.53279304504395 0.22323232889175415
1.81981980800629 3.80418300628662 0.2372727245092392
1.76994812488556 3.51767373085022 0.25131312012672424
1.91241312026978 4.13760375976562 0.2653535306453705
2.04878568649292 4.72773313522339 0.2793939411640167
2.28126263618469 5.55387592315674 0.29343435168266296
2.3830668926239 6.00299501419067 0.3074747622013092
2.48071599006653 6.55628776550293 0.32151514291763306
2.40082335472107 6.18898582458496 0.3355555534362793
2.70739674568176 7.55043315887451 0.34959596395492554
2.8541100025177 8.36861419677734 0.3636363744735718
3.06122612953186 9.7293872833252 0.37767675518989563
3.18856191635132 10.4723062515259 0.39171716570854187
3.22916960716248 10.736349105835 0.4057575762271881
3.50983691215515 12.4765968322754 0.41979798674583435
3.62950682640076 13.2309408187866 0.4338383972644806
3.60361170768738 12.9575338363647 0.44787877798080444
3.75718665122986 13.8326549530029 0.4619191884994507
3.70575070381165 13.8368759155273 0.4759595990180969
3.95926332473755 15.0959930419922 0.49000000953674316
};
\end{axis}

\end{tikzpicture}
    \vspace{-0.2cm}
    \caption{\textbf{Left:} Comparison of different GNS decoders with oracle information. MaNGO outperforms both MGN and EGNO, with the performance gap being more visible in the \emph{Sphere Cloth Coupling} task due to its highly complex underlying dynamics. Additionally, EGNO fails to learn in the \emph{Planar Bending} task, a phenomenon further analyzed in Appendix~\ref{appx:egno_failed}.
    \textbf{Right:}
    Visualization of the $2$D latent space for Deformable Plate (Easy), with points color-coded by Poisson’s ratio. The structured separation (at Poisson value $0$) shows that $\re$ effectively captures the underlying material properties.}
    \vspace{-0.25cm}
    \label{fig:add_results}
\end{figure}

We validate MaNGO on four different simulation datasets derived from three distinct simulation platforms. All tasks are normalized to $[0, 1]^3$.  
The first dataset is a $2$D \textit{Deformable Plate} (DP) task~\citep{linkerhagner2023grounding}, which has two variants: \textit{DP-easy} and \textit{DP-hard}. In \textit{DP-easy}, the material property of interest is Poisson's ratio~\citep{lim2015auxetic}. To increase complexity, \textit{DP-hard} additionally varies Young's modulus and the initial velocity across trials within the same task.
Next, we introduce a novel \textit{Planar Bending} (PB) dataset, which simulates the bending of a $2$D sheet subjected to two external forces perpendicular to the sheet. 
In this task, Young's modulus is the property of interest.
The final dataset is a new \textit{Sphere Cloth Coupling} (SCC) task, inspired by the Physion++ dataset~\cite{tung2023physion++}, which involves a coupling system consisting of a sphere and a cloth. In this task, spheres of varying sizes are dropped onto the cloth. Their density is varied to influence the cloth's deformation upon contact.  

The length of each simulation trial varies across datasets: $52$ time steps for DP, $50$ for PB, and $100$ time steps for SCC to account for the increased complexity of simulating elastic behaviors. 
Further details about preprocessing steps and ground-truth simulators can be found in Appendix~\ref{app:datasets}.

\subsection{Training setup and baselines} \label{sec:baselines} 
In this section, we present all methods evaluated in our experiments. Each model variant, whether meta-learning or non-meta-learning, has approximately three million trainable parameters, ensuring a fair comparison across architectures.

\textbf{Meta-learning.} We first describe the setup used for our meta-learning framework. Each task dataset $\mathcal D_l$ contains $16$ different simulation trials, where each trial shares the same material properties $\rho$ (e.g. Poisson's ratio or Young's modulus) but varies in initial conditions (e.g., collider position and size, or force position).  
During training, we randomly select a subset of \( S_l \) of a size between $1$ to $8$ to serve as the context dataset. This context is then used to predict a random trial\footnote{We empirically find that predicting a random trial per batch instead of all trials improves performance.} using different decoder methods. We compare our proposed MaNGO decoder against the EGNO architecture and a step-based MGN simulator.  
For testing, we split $\mathcal{D_*}$ into two distinct subsets. The first subset is used as the context dataset, while the second subset provides the initial conditions for predicting the full trajectory. We evaluate various context sizes to assess adaptation capability.

\textbf{Two-Stage Training Setup.}
Since we assume access to the simulation parameters during training, we compare the meta-learning methods against a two-stage training scheme. In the first stage, we train an encoder to predict the simulation parameters given a simulation trial. In the second stage, we freeze the encoder and train a decoder that uses the predicted simulation parameters from the encoder as an additional input. Therefore, this baseline predicts an explicit representation rather than a latent representation of the simulation parameters. We refer to this approach as MaNGO/MGN/EGNO (Two-Stage).

\textbf{Vanilla GNS.} We compare our method against non-meta-learning baselines, including EGNO, and MGN, and our proposed MaNGO decoder without context information. To assess the upper bound, we additionally train an oracle method using the MaNGO decoder which has access to the simulation parameters during training \textit{and test time}. We refer to it as MaNGO-Oracle. For these baselines, the full training set containing all $16$ simulation trials is provided. 
Additionally, for the MGN baseline, we follow the approach of~\citet{pfaff2020learning}, adding noise during training to mitigate error accumulation and stabilize rollouts at inference. We tune the input noise level for each task to maximize MGN's performance. The full experimental protocol, along with computational budget details, is available in \cref{app:exp_protocol}.

\subsection{Results}
\textbf{Main evaluation.} We compare the mean-squared error (MSE) between the ground truth and the predicted simulation averaged over all time steps. Overall, Figure~\ref{fig:main_eval} shows that meta-learning approaches consistently outperform non-meta-learning baselines, achieving a relatively low MSE close to the oracle method, MaNGO-Oracle. The two-stage training scheme is only competitive in the simplest environment, Deformable Plate (Easy), and across all environments it requires a larger context size to achieve its best performance. In general, within the meta-learning setting, performance improves with larger context sizes. Furthermore, we emphasize that strong performance is achieved with as few as 2 to 3 context samples (even under the high-complexity DP-hard dataset), highlighting the ability of our meta-learning approach to adapt quickly with minimal context data. All non-meta-learning baselines perform similarly. Interestingly, EGNO struggled to learn in the Planar Bending task, a phenomenon we further analyze in Appendix~\ref{appx:egno_failed}. Qualitative visualizations of all methods and tasks are provided in \cref{app:vis}.

\begin{wrapfigure}{r}{0.45\textwidth}
    \centering
    \includegraphics[width=0.45\textwidth]{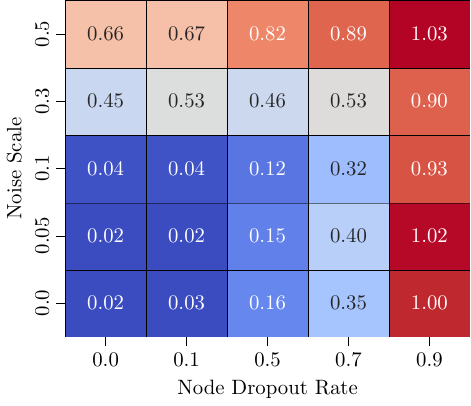}
    \vspace{-0.4cm}
    \caption{Robustness analysis of MaNGO on the Deformable Plate task under varying levels of Gaussian noise and node dropout, a setup mimicking real-world conditions. The normalized MSE is reported. Our method remains stable with up to $10\%$ noise (relative to the width of the mesh) and $10\%$ node dropout.}
     \vspace{-0.5cm} 
    \label{fig:stability_matrix}
\end{wrapfigure}

To directly 
compare our proposed MaNGO decoder with existing architectures, we evaluate it in the oracle setting, where material properties are known. As shown in Figure~\ref{fig:add_results} on the left side, MaNGO is outperforming other baselines,
confirming the effectiveness of our approach. We suspect that for MGN, auto-regressive prediction suffers from accumulated errors, 

an issue particularly evident in the Sphere Cloth Coupling dataset, where highly nonlinear contact dynamics \cite{karl2017dvbf, tung2023physion++} lead to MSE an order of magnitude higher than MaNGO.
 As for EGNO, while equivariance reduces the amount of required training data, it can also be overly restrictive, as real-world physics is not strictly E(3)-equivariant \cite{wang2024discovering,linden2024learning}. Various factors, such as friction and gravity, can break this symmetry, leading to suboptimal generalization.

\textbf{Latent visualization.}
To understand how the learned latent representation correlates with simulation parameters $\rho$, we visualize the 2D latent space with $\dim(\re) = 2$ for the Deformable Plate task, color-coded by Poisson’s ratio. Figure~\ref{fig:add_results} (right) shows a strong correlation between $\re$ and Poisson’s ratio, with two linear trends emerging: one from $0$ to $0.49$ and the other from $-0.9$ to $0$. This separation reflects the underlying material behavior — plates with a positive Poisson’s ratio expand on contact, while those with a negative Poisson’s ratio contract. The learned representation captures this distinction, indicating that the model encodes meaningful physical properties.

\textbf{Robustness under sparse and noisy observations.} 
Inspired by the setup in \cite{qzhao2022graphpde}, we evaluate the robustness of MaNGO on the Deformable Plate task by introducing Gaussian noise into the observational context data and reducing the number of observed nodes at test time. To this end, we also introduce a small noise level of $0.05$ for the context split during training to enhance the robustness of the encoder. We report the normalized MSE in the range $[0,1]$, where the minimum is set by MaNGO-Oracle and the maximum by the non-meta-learning MaNGO approach (as shown in Figure~\ref{fig:main_eval}).  
As shown in Figure~\ref{fig:stability_matrix}, the lower-left region of the matrix—corresponding to a 10\% noise level and 10\% node dropout—demonstrates near-optimal performance. Even with 50\% of nodes unobserved, the performance drop remains around 15\%, highlighting the model’s ability to handle sparse and noisy observations. These results confirm that our encoder is robust under such conditions, reflecting real-world scenarios.

\textbf{Runtime Efficiency and Memory Consumption} A key advantage of MaNGO is its ability to predict entire trajectories in a single forward pass, enabling efficient batched inference over time. This design leads to substantial improvements in inference speed compared to traditional autoregressive next-step simulators.
On the Sphere-Cloth Coupling benchmark, the CNP encoder requires approximately 6 ms to compute the latent representation, and the MaNGO decoder simulates the full trajectory in only 13 ms. In contrast, the MGN next-step simulator takes about 500 ms, as it predicts one step at a time, making MaNGO over an order of magnitude faster than the already efficient autoregressive neural simulator.

However, the batched inference and training scheme comes with increased memory requirements. GPU memory consumption scales approximately linearly with the number of predicted time steps, since the model retains intermediate activations across the temporal dimension. In our experiments, we observe that
\[
\text{Memory(MaNGO)} \approx \text{Memory(MGN)} \times \text{Number of predicted steps.}
\]

\section{Conclusion}
\label{sec:conclusion}
In this work, we explored data-driven adaptation of graph network simulators (GNS) via meta-learning. Specifically, we investigated the setting where simulation parameters are unknown at test time which would require retraining or labor-intensive data collection for existing methods. In a series of experiments, we demonstrated the potential of meta-learning for GNS, where our approach achieves accuracy on unseen material properties comparable to that of an oracle model. We view this work as a first stepping-stone towards the next generation of data-driven simulators, that are fast, differentiable, and capable of adapting to a wide variety of simulation settings. A discussion of the broader impact of this work is provided in \cref{app:broader_impact}.

\paragraph{Limitations and Future Work}
One limitation of our current approach is its focus on a single data modality at test time. In practical scenarios, however, test-time data may come in diverse formats, such as point clouds captured by cameras. Our method does not yet support such modalities, which constrains its applicability in real-world settings. Addressing this limitation would require the development of new architectures capable of handling heterogeneous data.

A further limitation is that MaNGO assumes a fixed graph topology and requires substantial GPU memory when predicting full trajectories in a single forward pass. These constraints limit its scalability and applicability to systems with dynamic connectivity. A promising direction for future work is to adopt a hybrid decoding strategy that predicts shorter temporal segments while dynamically updating the graph structure after each segment. Such an approach could reduce memory demands and enable the modeling of systems with evolving topologies, bridging the gap between efficiency and flexibility.

\section*{Acknowledgements}
This work is part of the DFG AI Resarch Unit 5339 regarding the combination of physics-based simulation with AI-based methodologies for the fast maturation of manufacturing processes. The financial support by German Research Foundation (DFG, Deutsche Forschungsgemeinschaft) is gratefully acknowledged. 
The authors acknowledge support by the state of Baden-Württemberg through bwHPC, as well as the HoreKa supercomputer funded by the Ministry of Science, Research and the Arts Baden-Württemberg and by the German Federal Ministry of Education and Research. We thank Philipp Becker for suggesting the title acronym used in this work.

\newpage
\bibliographystyle{unsrtnat}
\bibliography{mgno}

\begin{thebibliography}{67}
\providecommand{\natexlab}[1]{#1}
\providecommand{\url}[1]{\texttt{#1}}
\expandafter\ifx\csname urlstyle\endcsname\relax
  \providecommand{\doi}[1]{doi: #1}\else
  \providecommand{\doi}{doi: \begingroup \urlstyle{rm}\Url}\fi

\bibitem[Yazid et~al.(2009)Yazid, Abdelkader, and Abdelmadjid]{yazid2009state}
Abdelaziz Yazid, Nabbou Abdelkader, and Hamouine Abdelmadjid.
\newblock A state-of-the-art review of the x-fem for computational fracture mechanics.
\newblock \emph{Applied Mathematical Modelling}, 33\penalty0 (12):\penalty0 4269--4282, 2009.

\bibitem[Zienkiewicz and Taylor(2005)]{zienkiewicz2005finite}
Olek~C Zienkiewicz and Robert~Leroy Taylor.
\newblock \emph{The finite element method for solid and structural mechanics}.
\newblock Elsevier, 2005.

\bibitem[Stanova et~al.(2015)Stanova, Fedorko, Kmet, Molnar, and Fabian]{stanova2015finite}
Eva Stanova, Gabriel Fedorko, Stanislav Kmet, Vieroslav Molnar, and Michal Fabian.
\newblock Finite element analysis of spiral strands with different shapes subjected to axial loads.
\newblock \emph{Advances in engineering software}, 83:\penalty0 45--58, 2015.

\bibitem[Chung(1978)]{chung1978finite}
TJ~Chung.
\newblock Finite element analysis in fluid dynamics.
\newblock \emph{NASA STI/Recon Technical Report A}, 78:\penalty0 44102, 1978.

\bibitem[Zienkiewicz et~al.(2013)Zienkiewicz, Taylor, and Nithiarasu]{zienkiewicz2013finite}
Olek~C Zienkiewicz, Robert~Leroy Taylor, and Perumal Nithiarasu.
\newblock \emph{The finite element method for fluid dynamics}.
\newblock Butterworth-Heinemann, 2013.

\bibitem[Connor and Brebbia(2013)]{connor2013finite}
Jerome~J Connor and Carlos~Alberto Brebbia.
\newblock \emph{Finite element techniques for fluid flow}.
\newblock Newnes, 2013.

\bibitem[Jin(2015)]{jin2015finite}
Jian-Ming Jin.
\newblock \emph{The finite element method in electromagnetics}.
\newblock John Wiley \& Sons, 2015.

\bibitem[Polycarpou(2022)]{polycarpou2022introduction}
Anastasis~C Polycarpou.
\newblock \emph{Introduction to the finite element method in electromagnetics}.
\newblock Springer Nature, 2022.

\bibitem[Reddy(1994)]{reddy1994finite}
CJ~Reddy.
\newblock \emph{Finite element method for eigenvalue problems in electromagnetics}, volume 3485.
\newblock NASA, Langley Research Center, 1994.

\bibitem[Scheikl et~al.(2022)Scheikl, Tagliabue, Gyenes, Wagner, Dall'Alba, Fiorini, and Mathis-Ullrich]{scheikl2022sim}
Paul~Maria Scheikl, Eleonora Tagliabue, Bal{\'a}zs Gyenes, Martin Wagner, Diego Dall'Alba, Paolo Fiorini, and Franziska Mathis-Ullrich.
\newblock Sim-to-real transfer for visual reinforcement learning of deformable object manipulation for robot-assisted surgery.
\newblock \emph{IEEE Robotics and Automation Letters}, 8\penalty0 (2):\penalty0 560--567, 2022.

\bibitem[Wang and Zhu(2023)]{caregiving2023deformable}
Liman Wang and Jihong Zhu.
\newblock Deformable object manipulation in caregiving scenarios: A review.
\newblock \emph{Machines}, 11\penalty0 (11):\penalty0 1013, 2023.

\bibitem[Linkerh{\"a}gner et~al.(2023)Linkerh{\"a}gner, Freymuth, Scheikl, Mathis-Ullrich, and Neumann]{linkerhagner2023grounding}
Jonas Linkerh{\"a}gner, Niklas Freymuth, Paul~Maria Scheikl, Franziska Mathis-Ullrich, and Gerhard Neumann.
\newblock Grounding graph network simulators using physical sensor observations.
\newblock In \emph{The Eleventh International Conference on Learning Representations}, 2023.

\bibitem[Brenner and Scott(2008)]{brenner2008mathematical}
Susanne~C Brenner and L~Ridgway Scott.
\newblock \emph{The mathematical theory of finite element methods}, volume~3.
\newblock Springer, 2008.

\bibitem[Reddy(2019)]{reddy2019introduction}
Junuthula~Narasimha Reddy.
\newblock \emph{Introduction to the finite element method}.
\newblock McGraw-Hill Education, 2019.

\bibitem[Guo et~al.(2016)Guo, Li, and Iorio]{guo2016convolutional}
Xiaoxiao Guo, Wei Li, and Francesco Iorio.
\newblock Convolutional neural networks for steady flow approximation.
\newblock In \emph{Proceedings of the 22nd ACM SIGKDD International Conference on Knowledge Discovery and Data Mining}, KDD '16, page 481–490, New York, NY, USA, 2016. Association for Computing Machinery.
\newblock ISBN 9781450342322.
\newblock \doi{10.1145/2939672.2939738}.
\newblock URL \url{https://doi.org/10.1145/2939672.2939738}.

\bibitem[Da~Wang et~al.(2021)Da~Wang, Blunt, Armstrong, and Mostaghimi]{da2021deep}
Ying Da~Wang, Martin~J Blunt, Ryan~T Armstrong, and Peyman Mostaghimi.
\newblock Deep learning in pore scale imaging and modeling.
\newblock \emph{Earth-Science Reviews}, 215:\penalty0 103555, 2021.

\bibitem[Li et~al.(2022)Li, Du, and Martins]{li2022machine}
Jichao Li, Xiaosong Du, and Joaquim~RRA Martins.
\newblock Machine learning in aerodynamic shape optimization.
\newblock \emph{Progress in Aerospace Sciences}, 134:\penalty0 100849, 2022.

\bibitem[Battaglia et~al.(2018)Battaglia, Hamrick, Bapst, Sanchez{-}Gonzalez, Zambaldi, Malinowski, Tacchetti, Raposo, Santoro, Faulkner, G{\"{u}}l{\c{c}}ehre, Song, Ballard, Gilmer, Dahl, Vaswani, Allen, Nash, Langston, Dyer, Heess, Wierstra, Kohli, Botvinick, Vinyals, Li, and Pascanu]{battaglia2018relational}
Peter~W. Battaglia, Jessica~B. Hamrick, Victor Bapst, Alvaro Sanchez{-}Gonzalez, Vin{\'{\i}}cius~Flores Zambaldi, Mateusz Malinowski, Andrea Tacchetti, David Raposo, Adam Santoro, Ryan Faulkner, {\c{C}}aglar G{\"{u}}l{\c{c}}ehre, H.~Francis Song, Andrew~J. Ballard, Justin Gilmer, George~E. Dahl, Ashish Vaswani, Kelsey~R. Allen, Charles Nash, Victoria Langston, Chris Dyer, Nicolas Heess, Daan Wierstra, Pushmeet Kohli, Matthew Botvinick, Oriol Vinyals, Yujia Li, and Razvan Pascanu.
\newblock Relational inductive biases, deep learning, and graph networks.
\newblock \emph{CoRR}, abs/1806.01261, 2018.
\newblock URL \url{http://arxiv.org/abs/1806.01261}.

\bibitem[Pfaff et~al.(2021)Pfaff, Fortunato, Sanchez-Gonzalez, and Battaglia]{pfaff2020learning}
Tobias Pfaff, Meire Fortunato, Alvaro Sanchez-Gonzalez, and Peter~W. Battaglia.
\newblock Learning mesh-based simulation with graph networks.
\newblock In \emph{International Conference on Learning Representations}, 2021.
\newblock URL \url{https://arxiv.org/abs/2010.03409}.

\bibitem[Allen et~al.(2022)Allen, Guevara, Rubanova, Stachenfeld, Sanchez-Gonzalez, Battaglia, and Pfaff]{allen2022graph}
Kelsey~R Allen, Tatiana~Lopez Guevara, Yulia Rubanova, Kimberly Stachenfeld, Alvaro Sanchez-Gonzalez, Peter Battaglia, and Tobias Pfaff.
\newblock Graph network simulators can learn discontinuous, rigid contact dynamics.
\newblock \emph{Conference on Robot Learning (CoRL).}, 2022.

\bibitem[Allen et~al.(2023)Allen, Rubanova, Lopez-Guevara, Whitney, Sanchez-Gonzalez, Battaglia, and Pfaff]{allen2023learning}
Kelsey~R Allen, Yulia Rubanova, Tatiana Lopez-Guevara, William~F Whitney, Alvaro Sanchez-Gonzalez, Peter Battaglia, and Tobias Pfaff.
\newblock Learning rigid dynamics with face interaction graph networks.
\newblock In \emph{The Eleventh International Conference on Learning Representations}, 2023.

\bibitem[Xu et~al.(2024)Xu, Han, Lou, Kossaifi, Ramanathan, Azizzadenesheli, Leskovec, Ermon, and Anandkumar]{egno2024}
Minkai Xu, Jiaqi Han, Aaron Lou, Jean Kossaifi, Arvind Ramanathan, Kamyar Azizzadenesheli, Jure Leskovec, Stefano Ermon, and Anima Anandkumar.
\newblock Equivariant graph neural operator for modeling 3d dynamics.
\newblock In \emph{Forty-first International Conference on Machine Learning, {ICML} 2024, Vienna, Austria, July 21-27, 2024}. OpenReview.net, 2024.
\newblock URL \url{https://openreview.net/forum?id=dccRCYmL5x}.

\bibitem[Bronstein et~al.(2021)Bronstein, Bruna, Cohen, and Velickovic]{bronstein2021geometric}
Michael~M. Bronstein, Joan Bruna, Taco Cohen, and Petar Velickovic.
\newblock Geometric deep learning: Grids, groups, graphs, geodesics, and gauges.
\newblock \emph{CoRR}, abs/2104.13478, 2021.
\newblock URL \url{https://arxiv.org/abs/2104.13478}.

\bibitem[Xu et~al.(2021)Xu, Chen, Zlokapa, Foshey, Matusik, Sueda, and Agrawal]{xu2021end}
Jie Xu, Tao Chen, Lara Zlokapa, Michael Foshey, Wojciech Matusik, Shinjiro Sueda, and Pulkit Agrawal.
\newblock An end-to-end differentiable framework for contact-aware robot design.
\newblock In \emph{Robotics: Science \& Systems}, 2021.

\bibitem[Sanchez-Gonzalez et~al.(2020)Sanchez-Gonzalez, Godwin, Pfaff, Ying, Leskovec, and Battaglia]{sanchezgonzalez2020learning}
Alvaro Sanchez-Gonzalez, Jonathan Godwin, Tobias Pfaff, Rex Ying, Jure Leskovec, and Peter Battaglia.
\newblock Learning to simulate complex physics with graph networks.
\newblock In \emph{Proceedings of the 37th International Conference on Machine Learning}, pages 8459--8468. PMLR, 2020.

\bibitem[Isakov(2006)]{isakov2006inverse}
Victor Isakov.
\newblock \emph{Inverse problems for partial differential equations}, volume 127.
\newblock Springer, 2006.

\bibitem[Takahashi and Lin(2019)]{Takahashi2019Video}
Tetsuya Takahashi and Ming~C. Lin.
\newblock Video-guided real-to-virtual parameter transfer for viscous fluids.
\newblock \emph{ACM Trans. Graph.}, 38\penalty0 (6), November 2019.
\newblock ISSN 0730-0301.
\newblock \doi{10.1145/3355089.3356551}.
\newblock URL \url{https://doi.org/10.1145/3355089.3356551}.

\bibitem[Ma et~al.(2022)Ma, Du, Tenenbaum, Matusik, and Gan]{ma2022risp}
Pingchuan Ma, Tao Du, Joshua~B. Tenenbaum, Wojciech Matusik, and Chuang Gan.
\newblock {RISP}: Rendering-invariant state predictor with differentiable simulation and rendering for cross-domain parameter estimation.
\newblock In \emph{International Conference on Learning Representations}, 2022.
\newblock URL \url{https://openreview.net/forum?id=uSE03demja}.

\bibitem[Feng et~al.(2022)Feng, Ogren, Daraio, and Bouman]{Feng2022Visual}
Berthy~T. Feng, Alexander~C. Ogren, Chiara Daraio, and Katherine~L. Bouman.
\newblock Visual vibration tomography: Estimating interior material properties from monocular video.
\newblock In \emph{2022 IEEE/CVF Conference on Computer Vision and Pattern Recognition (CVPR)}, pages 16210--16219, 2022.
\newblock \doi{10.1109/CVPR52688.2022.01575}.

\bibitem[Wang et~al.(2015)Wang, Wu, Yin, Liu, and Huang]{Wang2015Deformation}
Bin Wang, Longhua Wu, Kangkang Yin, Libin Liu, and Hui Huang.
\newblock Deformation capture and modeling of soft objects.
\newblock \emph{ACM Transactions on Graphics(Proc. of SIGGRAPH 2015)}, 34\penalty0 (4):\penalty0 94:1--94:12, 2015.

\bibitem[Zhao et~al.(2022)Zhao, Lindell, and Wetzstein]{qzhao2022graphpde}
Qingqing Zhao, David~B. Lindell, and Gordon Wetzstein.
\newblock Learning to solve pde-constrained inverse problems with graph networks.
\newblock 2022.

\bibitem[Wang and Wang(2024)]{wang2024latent}
Tian Wang and Chuang Wang.
\newblock Latent neural operator for solving forward and inverse {PDE} problems.
\newblock In \emph{The Thirty-eighth Annual Conference on Neural Information Processing Systems}, 2024.
\newblock URL \url{https://openreview.net/forum?id=VLw8ZyKfcm}.

\bibitem[Brandstetter et~al.(2022)Brandstetter, Worrall, and Welling]{brandstetter2021message}
Johannes Brandstetter, Daniel~E Worrall, and Max Welling.
\newblock Message passing neural pde solvers.
\newblock In \emph{International Conference on Learning Representations}, 2022.

\bibitem[Garnelo et~al.(2018{\natexlab{a}})Garnelo, Rosenbaum, Maddison, Ramalho, Saxton, Shanahan, Teh, Rezende, and Eslami]{garnelo_conditional_2018}
Marta Garnelo, Dan Rosenbaum, Christopher Maddison, Tiago Ramalho, David Saxton, Murray Shanahan, Yee~Whye Teh, Danilo Rezende, and S.~M.~Ali Eslami.
\newblock Conditional neural processes.
\newblock \emph{{International} {Conference} on {Machine} {Learning}}, 2018{\natexlab{a}}.

\bibitem[Zaheer et~al.(2017)Zaheer, Kottur, Ravanbakhsh, Póczos, Salakhutdinov, and Smola]{zaheer_deep_2017}
Manzil Zaheer, Satwik Kottur, Siamak Ravanbakhsh, Barnabás Póczos, Ruslan Salakhutdinov, and Alexander~J. Smola.
\newblock Deep sets.
\newblock \emph{{Advances in Neural Information Processing Systems}}, 2017.

\bibitem[Cai et~al.(2023)Cai, Hy, Yu, and Wang]{cai2023deepset}
Chen Cai, Truong~Son Hy, Rose Yu, and Yusu Wang.
\newblock On the connection between mpnn and graph transformer.
\newblock In \emph{International Conference on Machine Learning}, pages 3408--3430. PMLR, 2023.

\bibitem[Li et~al.(2021)Li, Kovachki, Azizzadenesheli, liu, Bhattacharya, Stuart, and Anandkumar]{li2021fourier}
Zongyi Li, Nikola~Borislavov Kovachki, Kamyar Azizzadenesheli, Burigede liu, Kaushik Bhattacharya, Andrew Stuart, and Anima Anandkumar.
\newblock Fourier neural operator for parametric partial differential equations.
\newblock In \emph{International Conference on Learning Representations}, 2021.
\newblock URL \url{https://openreview.net/forum?id=c8P9NQVtmnO}.

\bibitem[Thuerey et~al.(2020)Thuerey, Wei{\ss}enow, Prantl, and Hu]{thuerey2020deep}
Nils Thuerey, Konstantin Wei{\ss}enow, Lukas Prantl, and Xiangyu Hu.
\newblock Deep learning methods for reynolds-averaged navier--stokes simulations of airfoil flows.
\newblock \emph{AIAA Journal}, 58\penalty0 (1):\penalty0 25--36, 2020.

\bibitem[Kochkov et~al.(2021)Kochkov, Smith, Alieva, Wang, Brenner, and Hoyer]{kochkov2021machine}
Dmitrii Kochkov, Jamie~A Smith, Ayya Alieva, Qing Wang, Michael~P Brenner, and Stephan Hoyer.
\newblock Machine learning--accelerated computational fluid dynamics.
\newblock \emph{Proceedings of the National Academy of Sciences}, 118\penalty0 (21):\penalty0 e2101784118, 2021.

\bibitem[Prantl et~al.(2022)Prantl, Ummenhofer, Koltun, and Thuerey]{prantl2022guaranteed}
Lukas Prantl, Benjamin Ummenhofer, Vladlen Koltun, and Nils Thuerey.
\newblock Guaranteed conservation of momentum for learning particle-based fluid dynamics.
\newblock In Alice~H. Oh, Alekh Agarwal, Danielle Belgrave, and Kyunghyun Cho, editors, \emph{Advances in Neural Information Processing Systems}, 2022.
\newblock URL \url{https://openreview.net/forum?id=6niwHlzh10U}.

\bibitem[Bhatnagar et~al.(2019)Bhatnagar, Afshar, Pan, Duraisamy, and Kaushik]{bhatnagar2019prediction}
Saakaar Bhatnagar, Yaser Afshar, Shaowu Pan, Karthik Duraisamy, and Shailendra Kaushik.
\newblock Prediction of aerodynamic flow fields using convolutional neural networks.
\newblock \emph{Computational Mechanics}, 64\penalty0 (2):\penalty0 525--545, jun 2019.
\newblock \doi{10.1007/s00466-019-01740-0}.
\newblock URL \url{https://doi.org/10.1007\%2Fs00466-019-01740-0}.

\bibitem[Yu et~al.(2023)Yu, Choi, Cho, Lee, Kim, Chang, Woo, Kim, Lee, Yang, et~al.]{yu2023learning}
Youn-Yeol Yu, Jeongwhan Choi, Woojin Cho, Kookjin Lee, Nayong Kim, Kiseok Chang, Chang-Seung Woo, Ilho Kim, Seok-Woo Lee, Joon-Young Yang, et~al.
\newblock Learning flexible body collision dynamics with hierarchical contact mesh transformer.
\newblock \emph{arXiv preprint arXiv:2312.12467}, 2023.

\bibitem[Battaglia et~al.(2016)Battaglia, Pascanu, Lai, Jimenez~Rezende, and kavukcuoglu]{battaglia2016interaction}
Peter Battaglia, Razvan Pascanu, Matthew Lai, Danilo Jimenez~Rezende, and koray kavukcuoglu.
\newblock Interaction networks for learning about objects, relations and physics.
\newblock In D.~Lee, M.~Sugiyama, U.~Luxburg, I.~Guyon, and R.~Garnett, editors, \emph{Advances in Neural Information Processing Systems}, volume~29. Curran Associates, Inc., 2016.

\bibitem[Scarselli et~al.(2009)Scarselli, Gori, Tsoi, Hagenbuchner, and Monfardini]{scarselli2009the}
Franco Scarselli, Marco Gori, Ah~Chung Tsoi, Markus Hagenbuchner, and Gabriele Monfardini.
\newblock The graph neural network model.
\newblock \emph{IEEE Transactions on Neural Networks}, 20\penalty0 (1):\penalty0 61--80, 2009.
\newblock \doi{10.1109/TNN.2008.2005605}.

\bibitem[Brown et~al.(2020)Brown, Mann, Ryder, Subbiah, Kaplan, Dhariwal, Neelakantan, Shyam, Sastry, Askell, Agarwal, Herbert-Voss, Krueger, Henighan, Child, Ramesh, Ziegler, Wu, Winter, Hesse, Chen, Sigler, Litwin, Gray, Chess, Clark, Berner, McCandlish, Radford, Sutskever, and Amodei]{brown2020language}
Tom Brown, Benjamin Mann, Nick Ryder, Melanie Subbiah, Jared~D Kaplan, Prafulla Dhariwal, Arvind Neelakantan, Pranav Shyam, Girish Sastry, Amanda Askell, Sandhini Agarwal, Ariel Herbert-Voss, Gretchen Krueger, Tom Henighan, Rewon Child, Aditya Ramesh, Daniel Ziegler, Jeffrey Wu, Clemens Winter, Chris Hesse, Mark Chen, Eric Sigler, Mateusz Litwin, Scott Gray, Benjamin Chess, Jack Clark, Christopher Berner, Sam McCandlish, Alec Radford, Ilya Sutskever, and Dario Amodei.
\newblock Language models are few-shot learners.
\newblock In H.~Larochelle, M.~Ranzato, R.~Hadsell, M.F. Balcan, and H.~Lin, editors, \emph{Advances in Neural Information Processing Systems}, volume~33, pages 1877--1901. Curran Associates, Inc., 2020.

\bibitem[Vosylius and Johns(2024)]{vosylius2024instantpolicyincontextimitation}
Vitalis Vosylius and Edward Johns.
\newblock Instant policy: In-context imitation learning via graph diffusion, 2024.
\newblock URL \url{https://arxiv.org/abs/2411.12633}.

\bibitem[Di~Palo and Johns(2024)]{di2024keypoint}
Norman Di~Palo and Edward Johns.
\newblock Keypoint action tokens enable in-context imitation learning in robotics.
\newblock \emph{arXiv preprint arXiv:2403.19578}, 2024.

\bibitem[Finn et~al.(2017)Finn, Abbeel, and Levine]{finn_model-agnostic_2017}
Chelsea Finn, Pieter Abbeel, and Sergey Levine.
\newblock Model-agnostic meta-learning for fast adaptation of deep networks.
\newblock \emph{International {Conference} on {Machine} {Learning}}, 2017.

\bibitem[Grant et~al.(2018)Grant, Finn, Levine, Darrell, and Griffiths]{grant_recasting_2018}
Erin Grant, Chelsea Finn, Sergey Levine, Trevor Darrell, and Thomas~L. Griffiths.
\newblock Recasting {Gradient}-{Based} {Meta}-{Learning} as {Hierarchical} {Bayes}.
\newblock \emph{International {Conference} on {Learning} {Representations}}, 2018.

\bibitem[Finn et~al.(2018)Finn, Xu, and Levine]{finn_probabilistic_2018}
Chelsea Finn, Kelvin Xu, and Sergey Levine.
\newblock Probabilistic {Model}-{Agnostic} {Meta}-{Learning}.
\newblock \emph{{Advances in Neural Information Processing Systems}}, 2018.

\bibitem[Kim et~al.(2018)Kim, Yoon, Dia, Kim, Bengio, and Ahn]{kim_bayesian_2018}
Taesup Kim, Jaesik Yoon, Ousmane Dia, Sungwoong Kim, Yoshua Bengio, and Sungjin Ahn.
\newblock Bayesian {Model}-{Agnostic} {Meta}-{Learning}.
\newblock \emph{Advances in Neural Information Processing Systems}, 2018.

\bibitem[Garnelo et~al.(2018{\natexlab{b}})Garnelo, Schwarz, Rosenbaum, Viola, Rezende, Eslami, and Teh]{Garnelo2018NeuralP}
Marta Garnelo, Jonathan Schwarz, Dan Rosenbaum, Fabio Viola, Danilo~Jimenez Rezende, S.~M.~Ali Eslami, and Yee~Whye Teh.
\newblock Neural processes.
\newblock \emph{ICML Workshop on Theoretical Foundations and Applications of Deep Generative Models}, 2018{\natexlab{b}}.

\bibitem[Kim et~al.(2019)Kim, Mnih, Schwarz, Garnelo, Eslami, Rosenbaum, Vinyals, and Teh]{kim_attentive_2019}
Hyunjik Kim, Andriy Mnih, Jonathan Schwarz, Marta Garnelo, Ali Eslami, Dan Rosenbaum, Oriol Vinyals, and Yee~Whye Teh.
\newblock Attentive neural processes.
\newblock \emph{International Conference on Learning Representations}, 2019.

\bibitem[Gordon et~al.(2019)Gordon, Bronskill, Bauer, Nowozin, and Turner]{gordon_meta-learning_2019}
Jonathan Gordon, John Bronskill, Matthias Bauer, Sebastian Nowozin, and Richard~E. Turner.
\newblock Meta-{Learning} {Probabilistic} {Inference} for {Prediction}.
\newblock \emph{International {Conference} on {Learning} {Representations}}, 2019.

\bibitem[Louizos et~al.(2019)Louizos, Shi, Schutte, and Welling]{louizos_functional_2019}
Christos Louizos, Xiahan Shi, Klamer Schutte, and M.~Welling.
\newblock The functional neural process.
\newblock \emph{Advances in Neural Information Processing Systems}, 2019.

\bibitem[Volpp et~al.(2021)Volpp, Fl{\"u}renbrock, Gro{\ss}berger, Daniel, and Neumann]{Volpp2021BayesianCA}
Michael Volpp, Fabian Fl{\"u}renbrock, Lukas Gro{\ss}berger, Christian Daniel, and Gerhard Neumann.
\newblock Bayesian context aggregation for neural processes.
\newblock \emph{International Conference on Learning Representations}, 2021.

\bibitem[Volpp et~al.(2023)Volpp, Dahlinger, Becker, Daniel, and Neumann]{volpp2023}
Michael Volpp, Philipp Dahlinger, Philipp Becker, Christian Daniel, and Gerhard Neumann.
\newblock Accurate bayesian meta-learning by accurate task posterior inference.
\newblock In \emph{The Eleventh International Conference on Learning Representations, {ICLR} 2023, Kigali, Rwanda, May 1-5, 2023}. OpenReview.net, 2023.
\newblock URL \url{https://openreview.net/pdf?id=sb-IkS8DQw2}.

\bibitem[Lim(2015)]{lim2015auxetic}
Teik-Cheng Lim.
\newblock \emph{Auxetic Materials and Structures}.
\newblock Springer Singapore, 01 2015.
\newblock ISBN 978-981-287-274-6.
\newblock \doi{10.1007/978-981-287-275-3}.
\newblock URL \url{https://doi.org/10.1007/978-981-287-275-3}.

\bibitem[Tung et~al.(2023)Tung, Ding, Chen, Bear, Gan, Tenenbaum, Yamins, Fan, and Smith]{tung2023physion++}
Fish Tung, Mingyu Ding, Zhenfang Chen, Daniel~M. Bear, Chuang Gan, Joshua~B. Tenenbaum, Daniel L.~K. Yamins, Judith Fan, and Kevin~A. Smith.
\newblock Physion++: Evaluating physical scene understanding that requires online inference of different physical properties.
\newblock \emph{arXiv}, 2023.

\bibitem[Karl et~al.(2017)Karl, Soelch, Bayer, and van~der Smagt]{karl2017dvbf}
Maximilian Karl, Maximilian Soelch, Justin Bayer, and Patrick van~der Smagt.
\newblock Deep variational bayes filters: Unsupervised learning of state space models from raw data.
\newblock In \emph{International Conference on Learning Representations}, 2017.
\newblock URL \url{https://openreview.net/forum?id=HyTqHL5xg}.

\bibitem[Wang et~al.(2024)Wang, Hofgard, Gao, Walters, and Smidt]{wang2024discovering}
Rui Wang, Elyssa Hofgard, Han Gao, Robin Walters, and Tess Smidt.
\newblock Discovering symmetry breaking in physical systems with relaxed group convolution.
\newblock In \emph{Forty-first International Conference on Machine Learning}, 2024.
\newblock URL \url{https://openreview.net/forum?id=59oXyDTLJv}.

\bibitem[der Linden et~al.(2024)der Linden, Castellanos, Vadgama, Kuipers, and Bekkers]{linden2024learning}
Putri A~Van der Linden, Alejandro~Garc{\'\i}a Castellanos, Sharvaree Vadgama, Thijs~P. Kuipers, and Erik~J Bekkers.
\newblock Learning symmetries via weight-sharing with doubly stochastic tensors.
\newblock In \emph{The Thirty-eighth Annual Conference on Neural Information Processing Systems}, 2024.
\newblock URL \url{https://openreview.net/forum?id=44WWOW4GPF}.

\bibitem[Satorras et~al.(2021)Satorras, Hoogeboom, and Welling]{egnn}
V{\i}ctor~Garcia Satorras, Emiel Hoogeboom, and Max Welling.
\newblock E (n) equivariant graph neural networks.
\newblock In \emph{International conference on machine learning}, pages 9323--9332. PMLR, 2021.

\bibitem[Faure et~al.(2012)Faure, Duriez, Delingette, Allard, Gilles, Marchesseau, Talbot, Courtecuisse, Bousquet, Peterlik, and Cotin]{faure2012sofa}
Fran{\c c}ois Faure, Christian Duriez, Herv{\'e} Delingette, J{\'e}r{\'e}mie Allard, Benjamin Gilles, St{\'e}phanie Marchesseau, Hugo Talbot, Hadrien Courtecuisse, Guillaume Bousquet, Igor Peterlik, and St{\'e}phane Cotin.
\newblock {SOFA: A Multi-Model Framework for Interactive Physical Simulation}.
\newblock In Yohan Payan, editor, \emph{{Soft Tissue Biomechanical Modeling for Computer Assisted Surgery}}, volume~11 of \emph{Studies in Mechanobiology, Tissue Engineering and Biomaterials}, pages 283--321. {Springer}, June 2012.
\newblock \doi{10.1007/8415\_2012\_125}.
\newblock URL \url{https://hal.inria.fr/hal-00681539}.

\bibitem[NVIDIA(2022{\natexlab{a}})]{isaac-sim}
NVIDIA.
\newblock Isaac sim, 2022{\natexlab{a}}.
\newblock URL \url{https://developer.nvidia.com/isaac-sim}.

\bibitem[NVIDIA(2022{\natexlab{b}})]{physx}
NVIDIA.
\newblock Nvidia physx sdk, 2022{\natexlab{b}}.
\newblock URL \url{https://developer.nvidia.com/physx-sdk}.

\bibitem[Loshchilov and Hutter(2019)]{adamw}
Ilya Loshchilov and Frank Hutter.
\newblock Decoupled weight decay regularization.
\newblock In \emph{7th International Conference on Learning Representations, {ICLR} 2019, New Orleans, LA, USA, May 6-9, 2019}. OpenReview.net, 2019.
\newblock URL \url{https://openreview.net/forum?id=Bkg6RiCqY7}.

\end{thebibliography}

\newpage
\appendix

\section{Braoder Impact Statement} \label{app:broader_impact}
Our proposed Graph Network Simulator, with its ability to adapt to new material properties through a small context set, offers significant advancements across fields that rely on computational modeling and simulation. By reducing the need for extensive simulation data and recalibration, it can lower computational costs while maintaining high accuracy. This adaptability could benefit industries ranging from materials science to robotics, enabling more efficient design and testing of novel materials.

However, this flexibility in simulating a wide range of material properties could also be misused in contexts where precise material behavior is critical, such as in the development of advanced weaponry or other high-risk technologies. While the primary intent is to advance scientific and engineering applications, ethical considerations must be taken into account to prevent unintended harmful uses.

\section{Limitations of the Equivariant Graph Neural Network}
\label{appx:egno_failed}

In this section, we analyze the limitations of the Equivariant Graph Neural Network (EGNN) \citep{egnn}, which serves as the GNN backbone of the Equivariant Graph Neural Operator (EGNO) baseline \citep{egno2024}. During our experiments, we observed that EGNO fails to predict any deformation in the Planar Bending task.

In this task, all neural operators receive an initial input configuration \( \p_0 \), representing a completely flat plane as the positions of the mesh. Consequently, all points in this initial graph lie within a plane \( E \) embedded in three-dimensional space. In this appendix, we demonstrate that the spatial output of any trained EGNN will also remain confined to a plane. This implies that an EGNN cannot solve the Planar Bending task, as the required deformed state does not lie within a single plane.

We establish this result by first proving that an EGNN maps all points in the specific plane  
\[
\bm Z_{=0} = \{(x,y,z)\in \mathbb{R}^3 \mid z = 0\}
\]
back onto the same plane \( \bm Z_{=0} \). To see this, consider any node \( v \in \mathcal{G} \) with an initial position \( \p^0_v = (x, y, 0) \in \bm Z_{=0} \). Applying a single layer of EGNN to update the node positions, we obtain  
\[
\p_v^1 = \p_v^0 + C \sum_{w \in \mathcal{N}(v)} (\p_v^0 - \p_w^0) \varphi_x(\m_{vw}).
\]
Examining the \( z \)-coordinate, we note that the message term \( \varphi_x(\m_{vw}) \) is scaled by \( (\p_v^0 - \p_w^0) \). Since all \( \p^0_v \) lie in \( \bm Z_{=0} \), their \( z \)-coordinates are zero, meaning \( (\p_v^0 - \p_w^0) \) has no \( z \)-component. Consequently, the updated \( z \)-coordinate in \( \p_v^1 \) remains zero, ensuring that \( \p_v^1 \in \bm Z_{=0} \). By induction, this property holds for all subsequent layers, proving that EGNN maps \( \bm Z_{=0} \) onto itself.

Next, we extend this result to any arbitrary plane \( E \) in three-dimensional space. Since \( E \) is a plane, there exists a transformation \( T \in \text{SE}(3) \) such that  
\[
T(E) = \bm Z_{=0}.
\]
By the equivariance property of EGNN, there exists another transformation \( S \in \text{SE}(3) \) satisfying  
\[
\text{EGNN}(\underbrace{T(\mathcal{G})}_{\subset \bm Z_{=0}}) = S (\text{EGNN}(\mathcal{G})).
\]
From our previous result, the left-hand side remains confined to \( \bm Z_{=0} \). Thus, applying \( S^{-1} \) yields  
\[
\text{EGNN}(\mathcal{G}) \subset S^{-1} \bm Z_{=0},
\]
which is another plane in three-dimensional space. This establishes the key result: an EGNN always maps planar inputs to planar outputs, rendering it incapable of solving the Planar Bending task.

A similar argument extends to the full EGNO architecture since its equivariant temporal convolution layer in Fourier space is also equivariant and it preserves the property of mapping \( \bm Z_{=0} \) onto itself. Thus, EGNO, like EGNN, is inherently unable to capture the required deformations in the Planar Bending task.

\section{Feature creation for the MaNGO decoder} \label{app:mgno_features}
This appendix provides a detailed explanation of the input processing for the MaNGO simulator. First, similar to the Spatiotemporal encoder, we create a sequence of graphs $\mathcal G_t$ over time. However, unlike the encoder, we repeat the initial positions and velocities of all deformable nodes across time, as the future positions—our target variable—are unknown and need to be estimated. 
In $\mathcal G_t$, we define the node features for a node $v$ as the tensor $[\re, \h_v, \vsymb_t, \text{TE}(t)]$ consisting of the latent system identification $\re$, the node features $\h_v$, the velocity $\vsymb_t$ and a time embedding of the time step $t$. Note that the position is not part of the node features. Instead, following \cite{pfaff2020learning}, we compute for each edge $e$ the relative position $\p^\text{rel}_{e,t}$ to ensure translation invariance. The complete edge features are given as $[\e_e, \p^\text{rel}_{e,t}, \text{TE}(t)]$ consisting of the (time-independent) edge features, the relative position and a time step embedding.

\section{Datasets and Preprocessing information} 
\label{app:datasets}

In this section, we provide detailed information about the datasets used in our experiments. Table~\ref{tab:datasets} summarizes the key characteristics of each dataset, including the dataset splits, simulation length, and the number of nodes used for prediction.  

For brevity, we use the following abbreviations throughout the paper:  
\begin{itemize}
    \item \textbf{PB}: \textit{Planar Bending}.
    \item \textbf{DP}: \textit{Deformable Plate}, with two variants: \textit{DP-easy} and \textit{DP-hard}.
    \item \textbf{SCC}: \textit{Sphere Cloth Coupling}.
\end{itemize}  

Table~\ref{tab:training_setup} further details the training setup for each dataset, specifying the material properties considered and the variations in initial conditions. These variations influence the dynamics of each task, ensuring diverse training scenarios that test the generalization capabilities of our method.

\begin{table}[h]
\centering
\caption{Dataset descriptions}
\label{tab:datasets}
\begin{adjustbox}{max width=\textwidth}
\begin{tabular}{lcccc}
\toprule
    Name & Train/Val/Test Splits & Number of Steps & Number of Nodes for Prediction \\ 
\midrule
PB  & $460$/$50$/$50$ & $50$ & $225$ \\
DP-easy  & $600$/$100$/$100$ & $52$ & $81$ \\
DP-hard  & $600$/$100$/$100$ & $52$ & $81$ \\
SCC      & $600$/$100$/$100$ & $100$ & $400$ (cloth) + $98$ (sphere) \\
\bottomrule
\end{tabular}
\end{adjustbox}
\end{table}

\begin{table}[h]
\centering
\caption{Training setup for each dataset}
\label{tab:training_setup}
\begin{adjustbox}{max width=\textwidth}
\begin{tabular}{lcc}
\toprule
    Name & Material Properties & Initial Condition Variations \\ 
\midrule
PB  & Young's modulus & Two forces: $(x,y)$ position, $\{-1,1\}$ direction, constant magnitude \\
DP-easy  & Poisson’s ratio & Collider’s $x$ position, size, constant initial velocity \\
DP-hard  & Young's modulus, Poisson’s ratio & Collider’s $x$ position, size, \textbf{varied} initial velocity \\
SCC      & Sphere's density & Sphere’s size, same initial position \\
\bottomrule
\end{tabular}
\end{adjustbox}
\end{table}

\textbf{Planar Bending.}
We uniformly select Young's modulus from $10$ to $500$, from a very deformable to an almost stiff sheet. 
The boundary nodes of the sheet are kept in place. 

\textbf{Deformable Plate.}
The original task was introduced in \cite{linkerhagner2023grounding}, generated using \gls{sofa}~\citep{faure2012sofa}. We extended to meta-learning setting by sampling Poisson's ratios between $-0.9$ to $0.49$, under different trapezoidal meshes. We further increase the difficulty of this dataset by also randomizing the Young's modulus within a range from $500$ to $10000$ using Log-Uniform distribution. 

\textbf{Sphere Cloth Coupling.}  
Each trajectory in this dataset is generated by selecting a sphere radius from the range $[0.2, 0.8]$. The material property of interest is the sphere’s density, which varies between $[2.0, 100.0]$. The cloth is initialized in a stable state, remaining consistent across all tasks and trials. This dataset is created using NVIDIA Isaac Sim~\citep{isaac-sim}, which leverages PhysX 5.0~\citep{physx} to simulate position-based dynamics (PBD) particle interactions.

For meta-learning setup, we assume a set \( S_i \) containing $16$ different simulation trials, where each trial shares the same material properties (e.g., density, Poisson's ratio, Young's modulus) but varies in initial conditions (e.g., sphere size, force positions, and magnitudes).  

\section{MGN Decoder for Conditional Neural Processes} 
\label{app:mgn_cnp}
In step-based prediction tasks, batches are typically shuffled to include different simulations and time steps. However, when using a Conditional Neural Process (CNP), only data from the same task can be used for each batch, which can impact performance. During hyper parameter optimization, we tested a modified version of the MGN where we reduced batch shuffling to mimic the CNP approach. This modification resulted in poorer performance compared to the standard MGN. This difference in performance may partially explain why Meta-MGN underperforms relative to MaNGO.

\section{Experimental Protocol} 
\label{app:exp_protocol}

In order to promote reproducibility, we provide details of our experimental methodology. 
Table~\ref{tab:training_setup} presents the hyperparameters used in our experiments. For a comprehensive description of the creation of all datasets, please refer to Appendix \ref{app:datasets}.

The training took place on an NVIDIA A100 GPU, with each method given the same computation budget of 48 hours. Consequently, the number of epochs varied, as the batching differed significantly between the meta-learning methods and the step-based \gls{mgn}. In total, generating the results presented in this paper required approximately 8,500 GPU hours.

We conducted a multi-staged grid-based hyperparameter search for the learning rate, input noise, and other hyperparameters as residual connections and layer norms. We did not use the test data for this, but tuned all hyperparameters on a separate validation split. This split was also used to determine the best epoch checkpoint to mitigate any overfitting effects. Hyperparameter tuning required an additional computational budget of approximately 6,000 GPU hours.

For \gls{mgn}, we included velocity features of the current step.

\begin{table}[t]
\centering
\caption{\textbf{Left:} Training setup for each dataset. \textbf{Right:} Noise-scale per task for Auto-regressive methods}
\label{tab:appx_training_setup}
\begin{minipage}[t]{0.6\textwidth}
\begin{tabular}{lc}
\toprule
    Parameter & Value  \\ 
\midrule
Node feature dimension & $128$ \\
Latent representation dimension  & $128$ \\
Decoder hidden dimension & $128$ \\
Message passing blocks & $15$ \\
\gls{gnn} Aggregation function & Mean \\
\gls{gnn} Activation function & Leaky ReLU \\
Learning rate (Auto-regressive methods) & $5.0 \times 10^{-4}$ \\
Learning rate & $1.0 \times 10^{-4}$ \\
Optimizer & AdamW \cite{adamw} \\
Min Context Size (Training) & 1 \\
Max Context Size (Training) & 8 \\
MaNGO-CNN Decoder Kernel size & 7 \\
CNN-Deepset Encoder Kernel size & 3 \\
Latent representation aggregation & Maximum \\
\bottomrule
\end{tabular}
\end{minipage}\hfill
\begin{minipage}[t]{0.3\textwidth}
\begin{tabular}{lc}
\toprule
    Task & Value  \\ 
\midrule
PB & $5.0 \times 10^{-4}$ \\
DP-easy & $7.0 \times 10^{-4}$ \\
DP-hard & $7.0 \times 10^{-4}$ \\
SCC & $1.0 \times 10^{-3}$ \\
\bottomrule
\end{tabular}
\end{minipage}
\end{table}

\section{Visualizations} \label{app:vis}
In this section, we present qualitative results for all tasks and methods discussed in the main paper.
\begin{figure*}[ht!]
    \centering
    \begin{minipage}{0.119\textwidth}
            \centering
            \includegraphics[width=\textwidth]{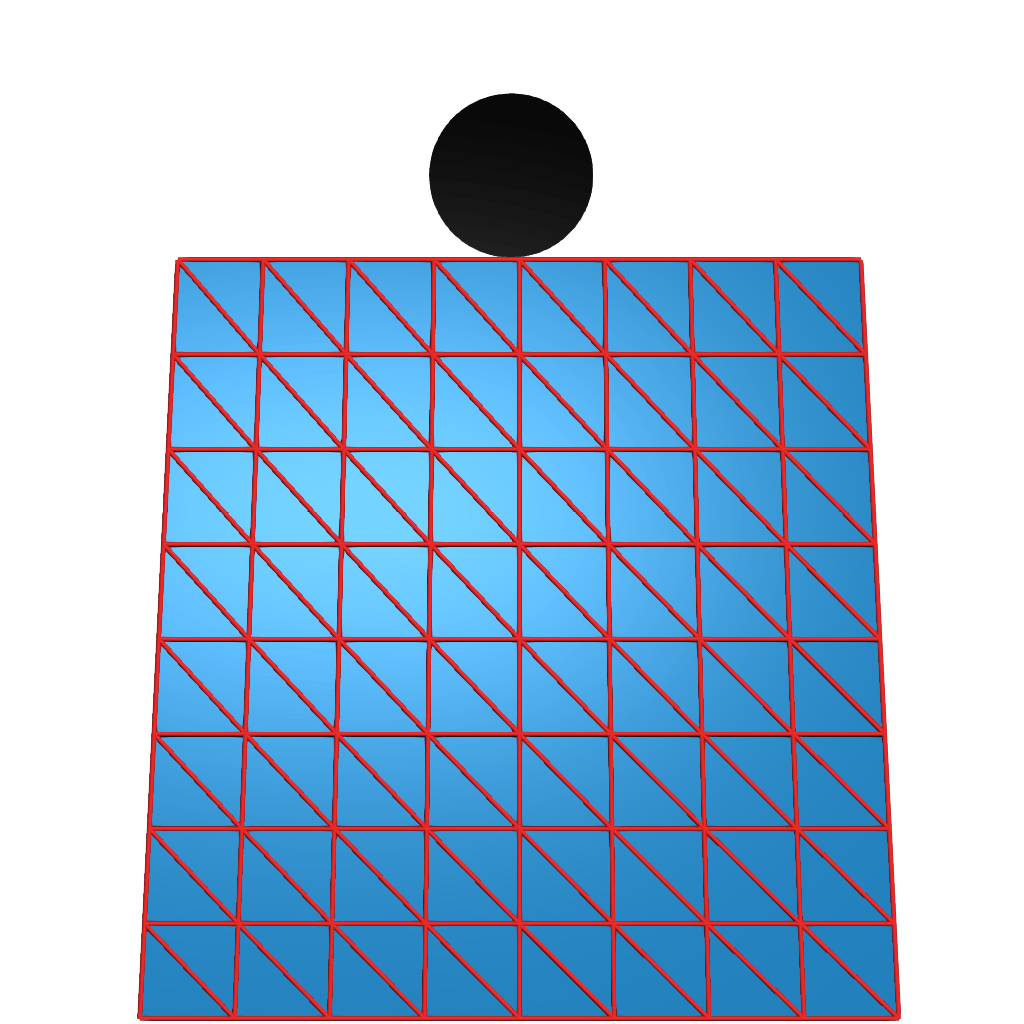}
    \end{minipage}
    \begin{minipage}{0.119\textwidth}
            \centering
            \includegraphics[width=\textwidth]{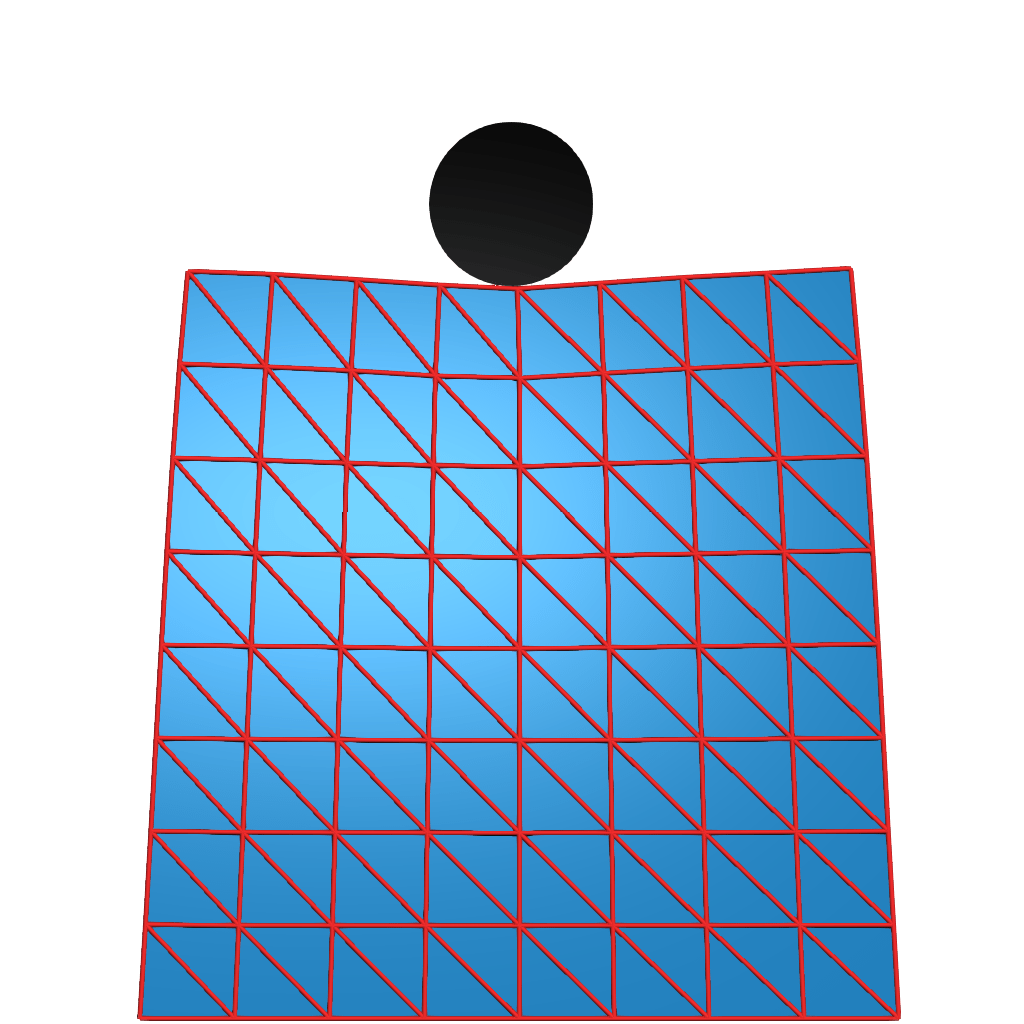}
    \end{minipage}
    \begin{minipage}{0.119\textwidth}
            \centering
            \includegraphics[width=\textwidth]{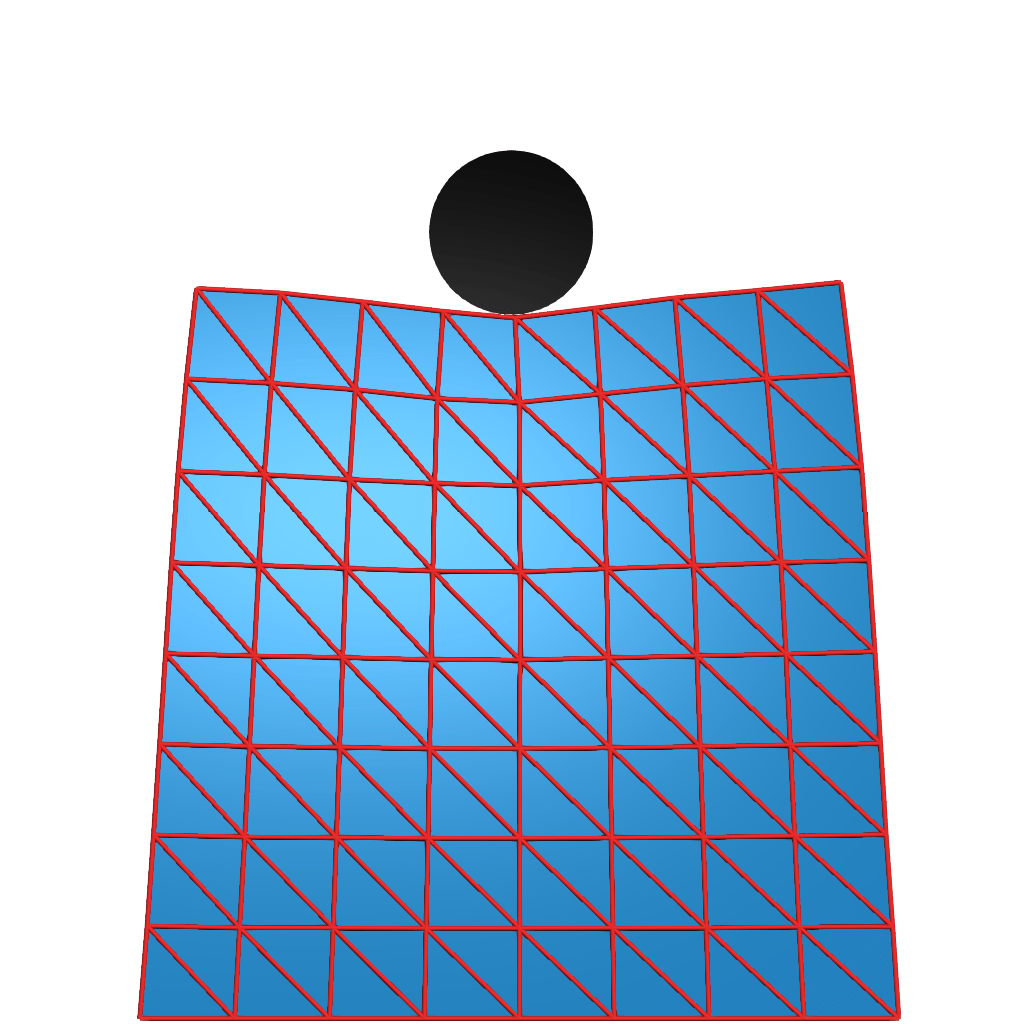}
    \end{minipage}
    \begin{minipage}{0.119\textwidth}
            \centering
            \includegraphics[width=\textwidth]{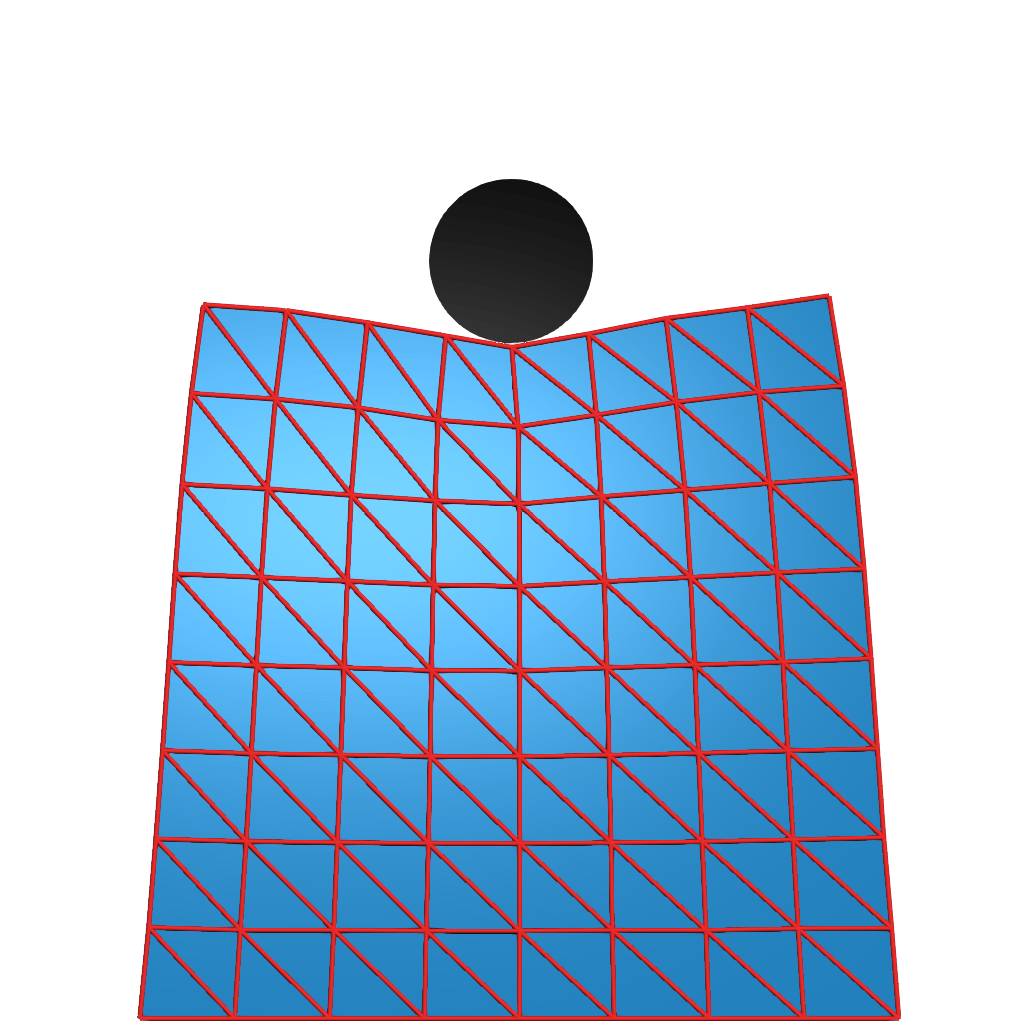}
    \end{minipage}
    \begin{minipage}{0.119\textwidth}
            \centering
            \includegraphics[width=\textwidth]{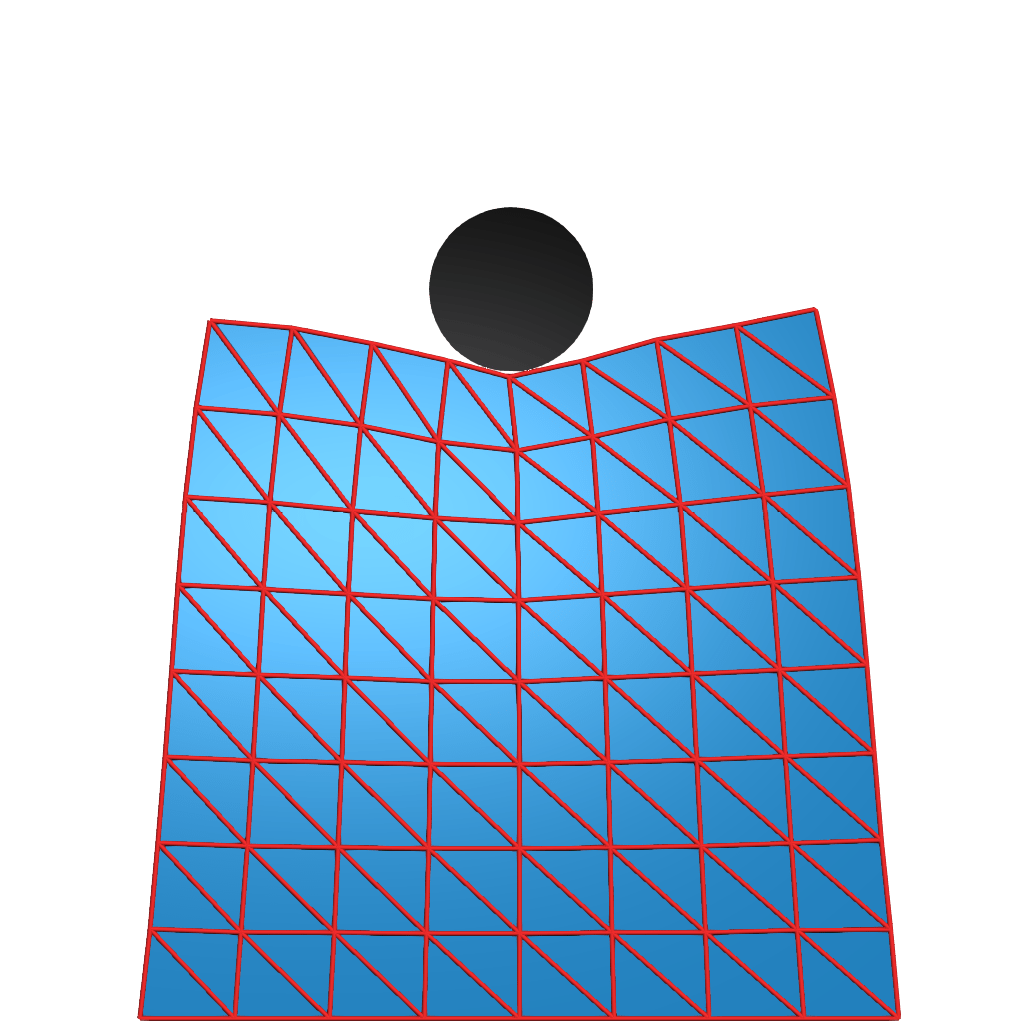}
    \end{minipage}
    \begin{minipage}{0.119\textwidth}
            \centering
            \includegraphics[width=\textwidth]{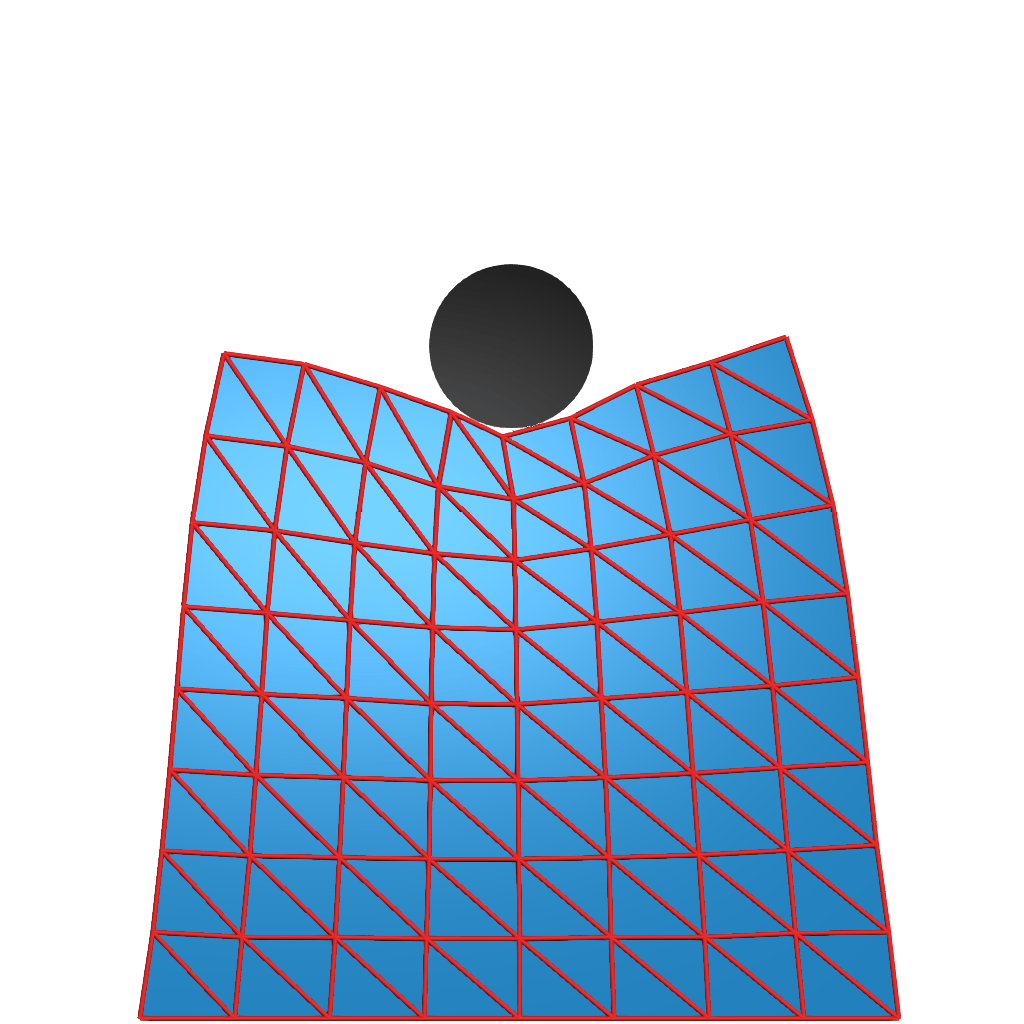}
    \end{minipage}
    \begin{minipage}{0.119\textwidth}
            \centering
            \includegraphics[width=\textwidth]{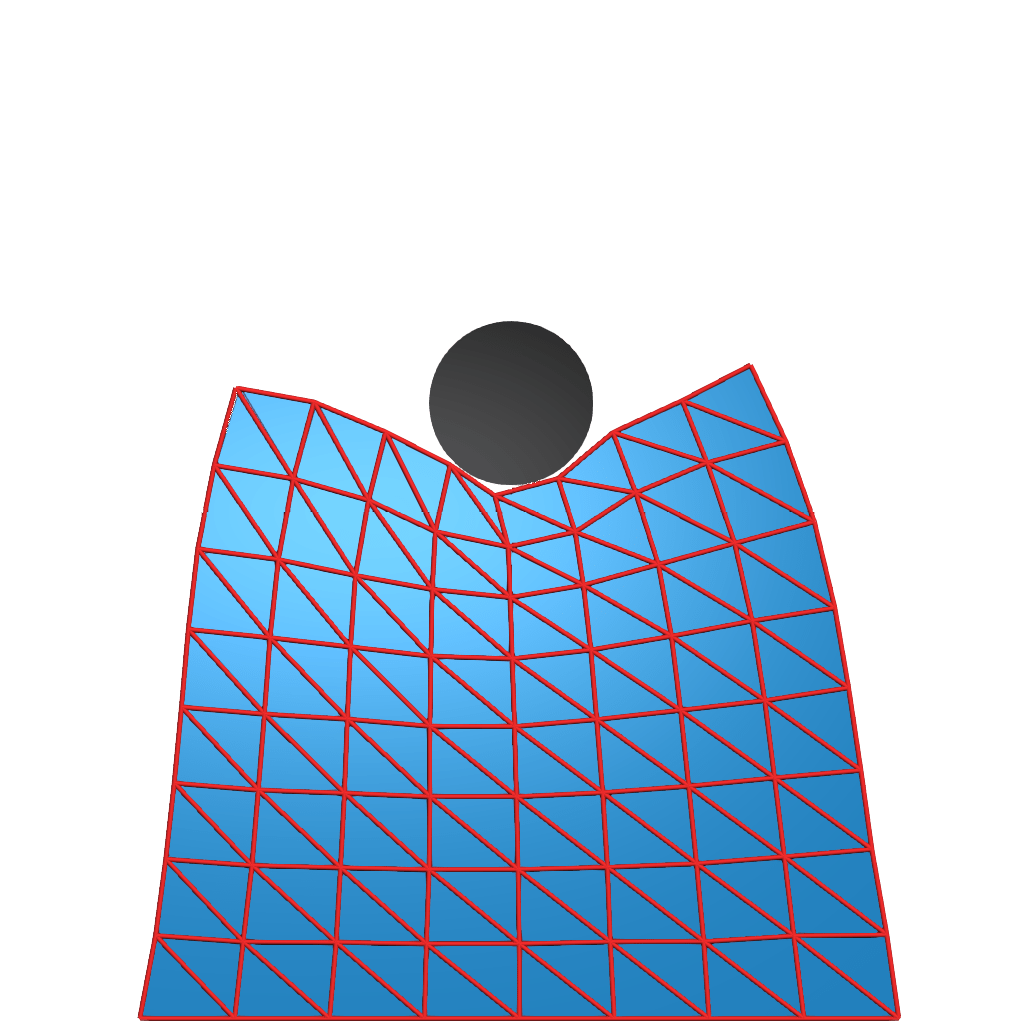}
    \end{minipage}
    \begin{minipage}{0.119\textwidth}
            \centering
            \includegraphics[width=\textwidth]{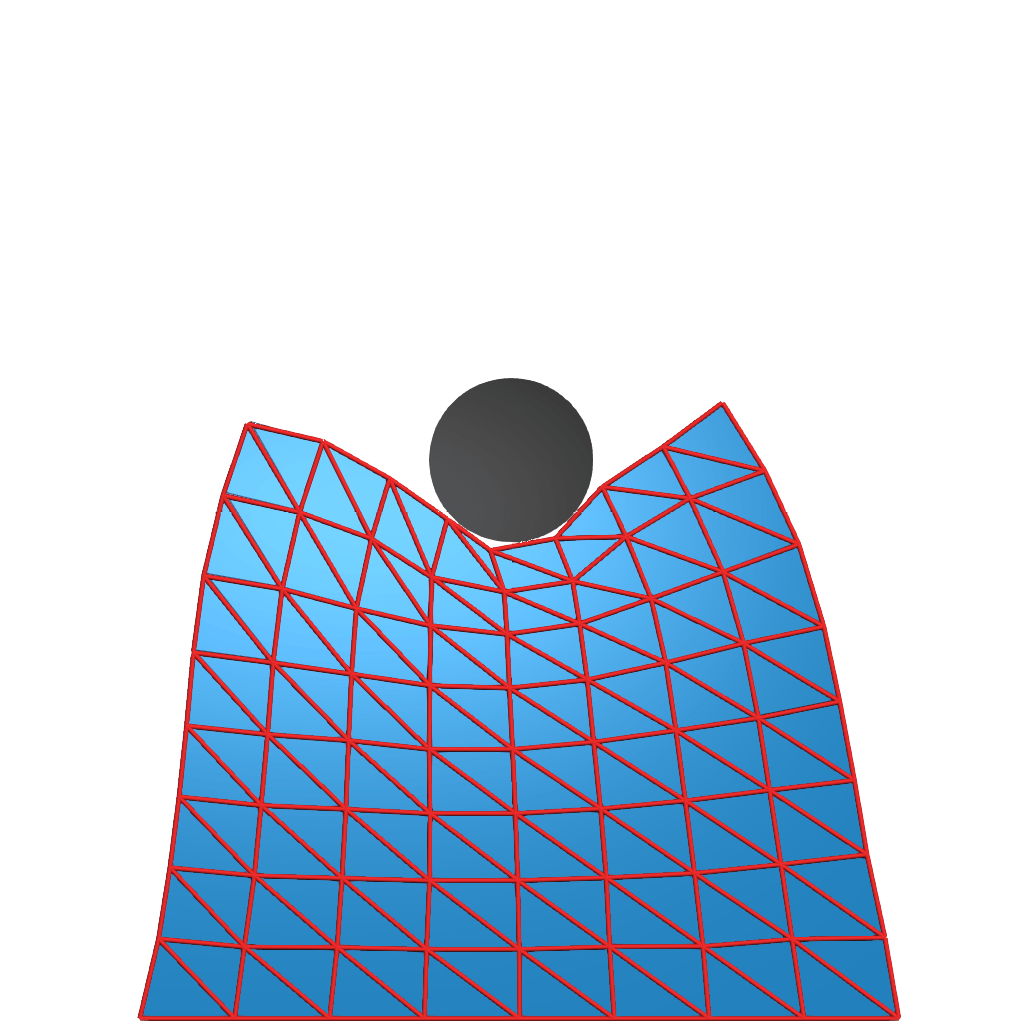}
    \end{minipage}

    \begin{minipage}{0.119\textwidth}
            \centering
            \includegraphics[width=\textwidth]{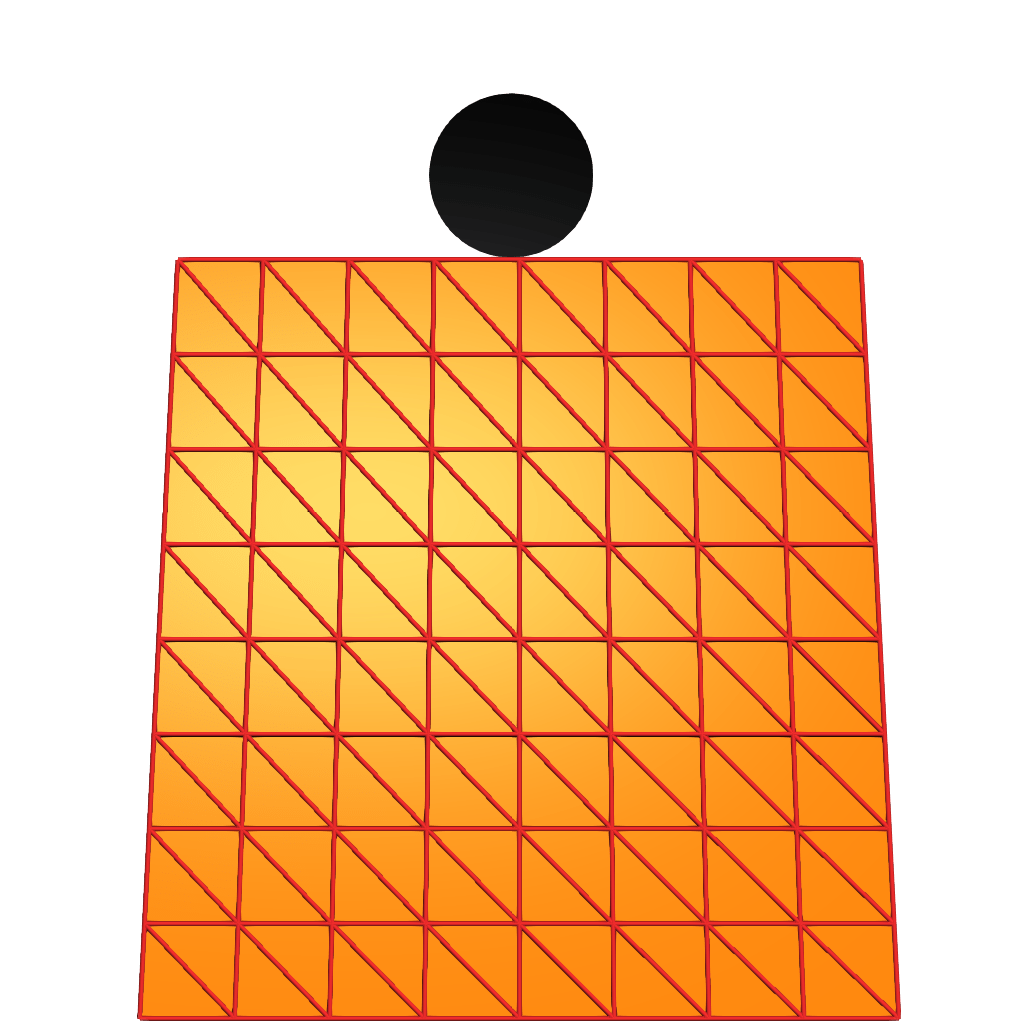}
    \end{minipage}
    \begin{minipage}{0.119\textwidth}
            \centering
            \includegraphics[width=\textwidth]{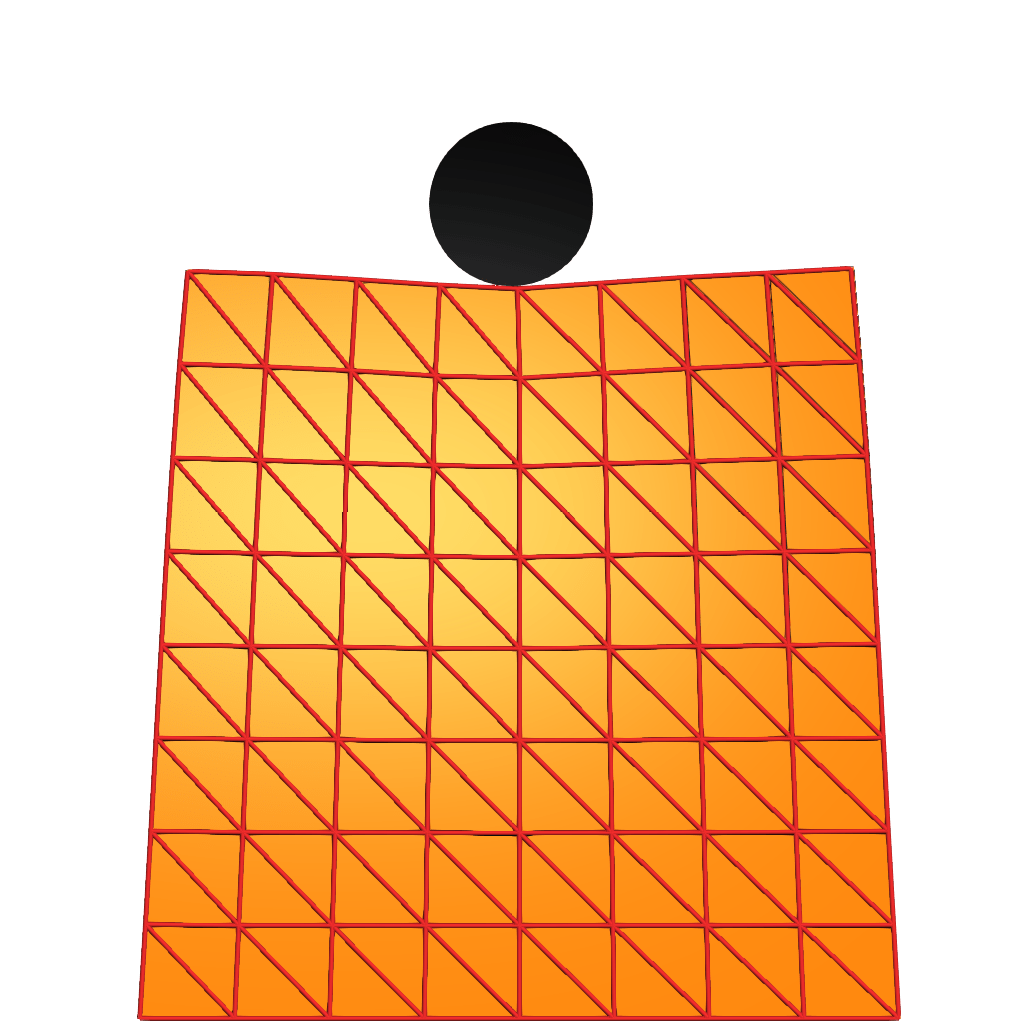}
    \end{minipage}
    \begin{minipage}{0.119\textwidth}
            \centering
            \includegraphics[width=\textwidth]{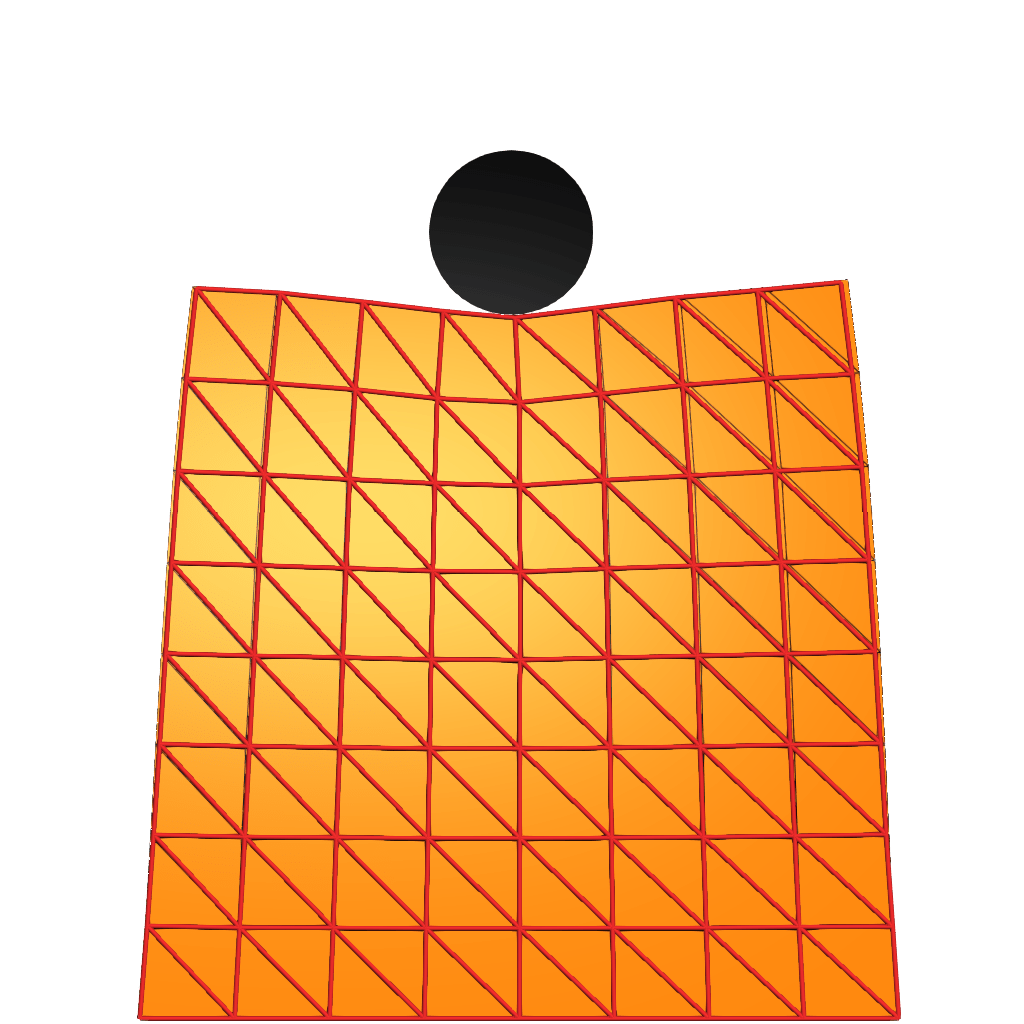}
    \end{minipage}
    \begin{minipage}{0.119\textwidth}
            \centering
            \includegraphics[width=\textwidth]{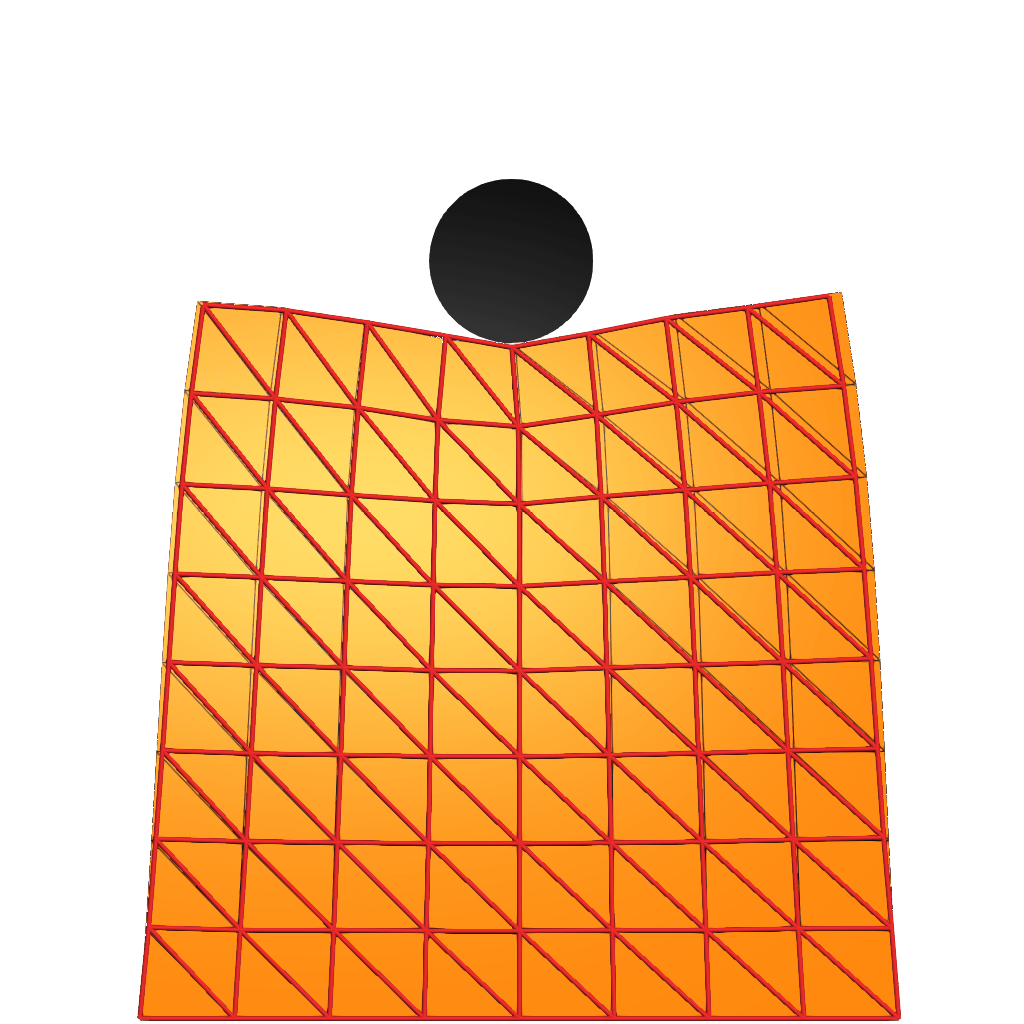}
    \end{minipage}
    \begin{minipage}{0.119\textwidth}
            \centering
            \includegraphics[width=\textwidth]{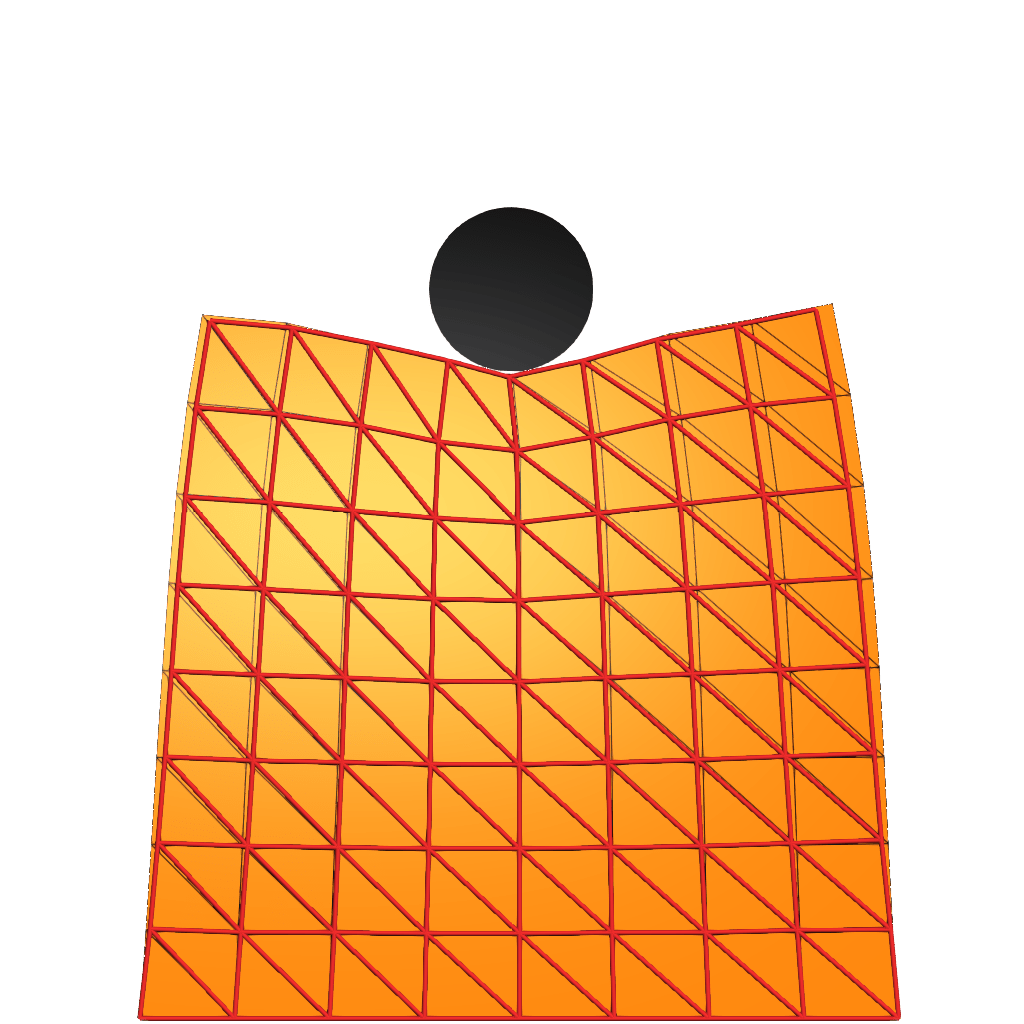}
    \end{minipage}
    \begin{minipage}{0.119\textwidth}
            \centering
            \includegraphics[width=\textwidth]{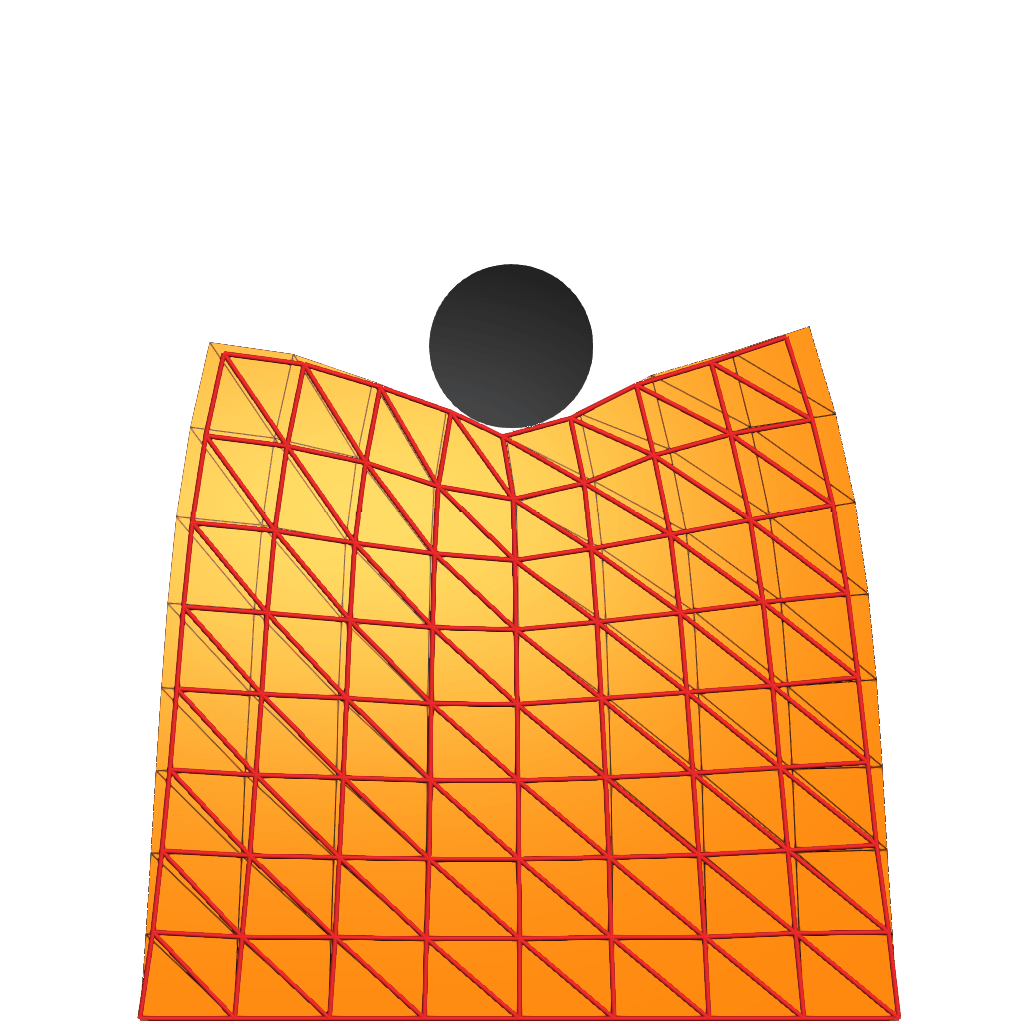}
    \end{minipage}
    \begin{minipage}{0.119\textwidth}
            \centering
            \includegraphics[width=\textwidth]{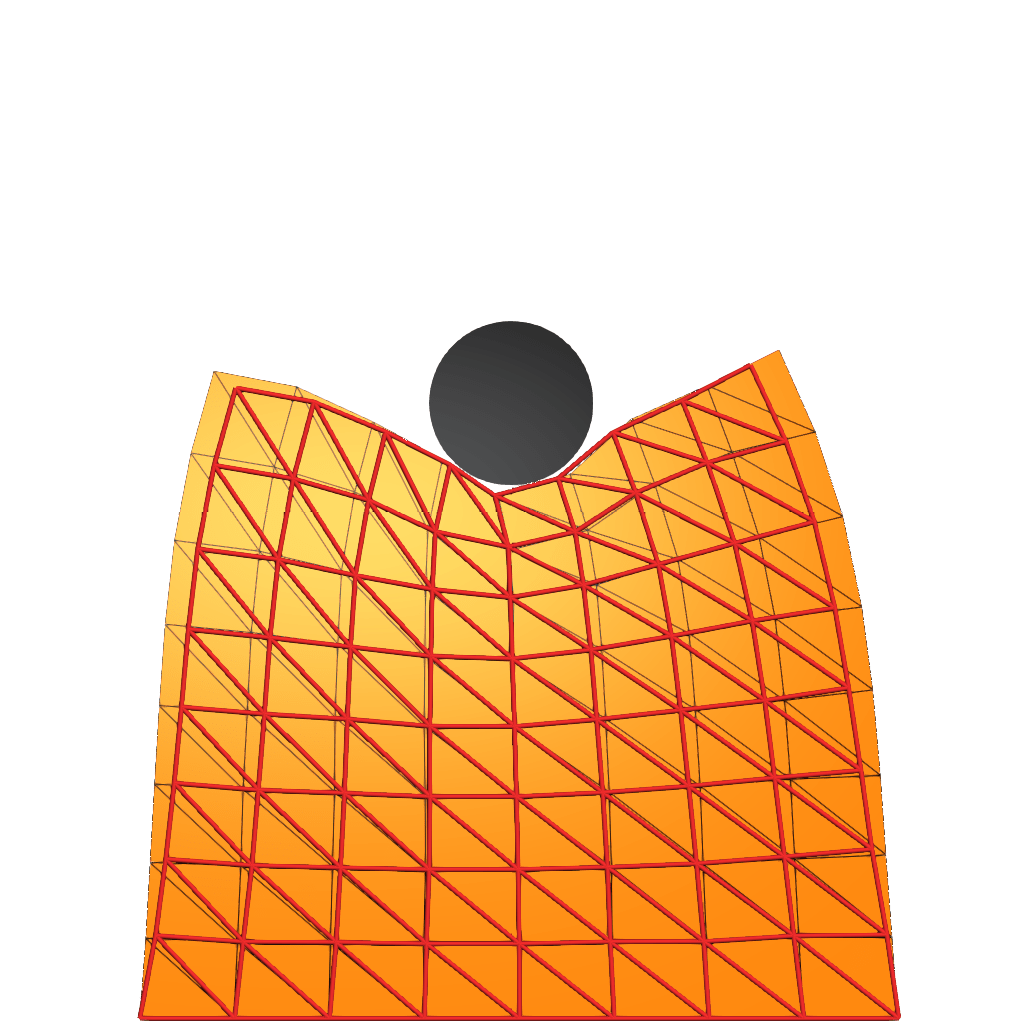}
    \end{minipage}
    \begin{minipage}{0.119\textwidth}
            \centering
            \includegraphics[width=\textwidth]{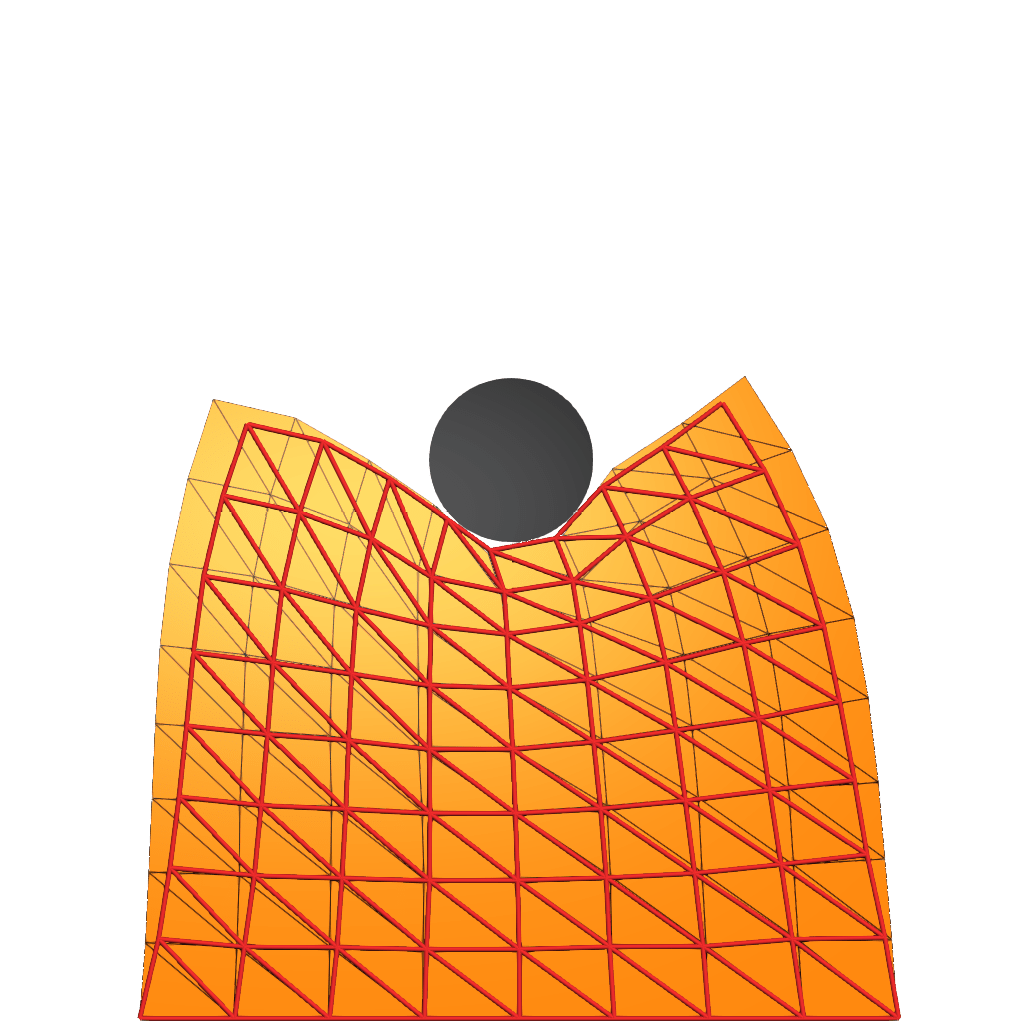}
    \end{minipage}

    \begin{minipage}{0.119\textwidth}
            \centering
            \includegraphics[width=\textwidth]{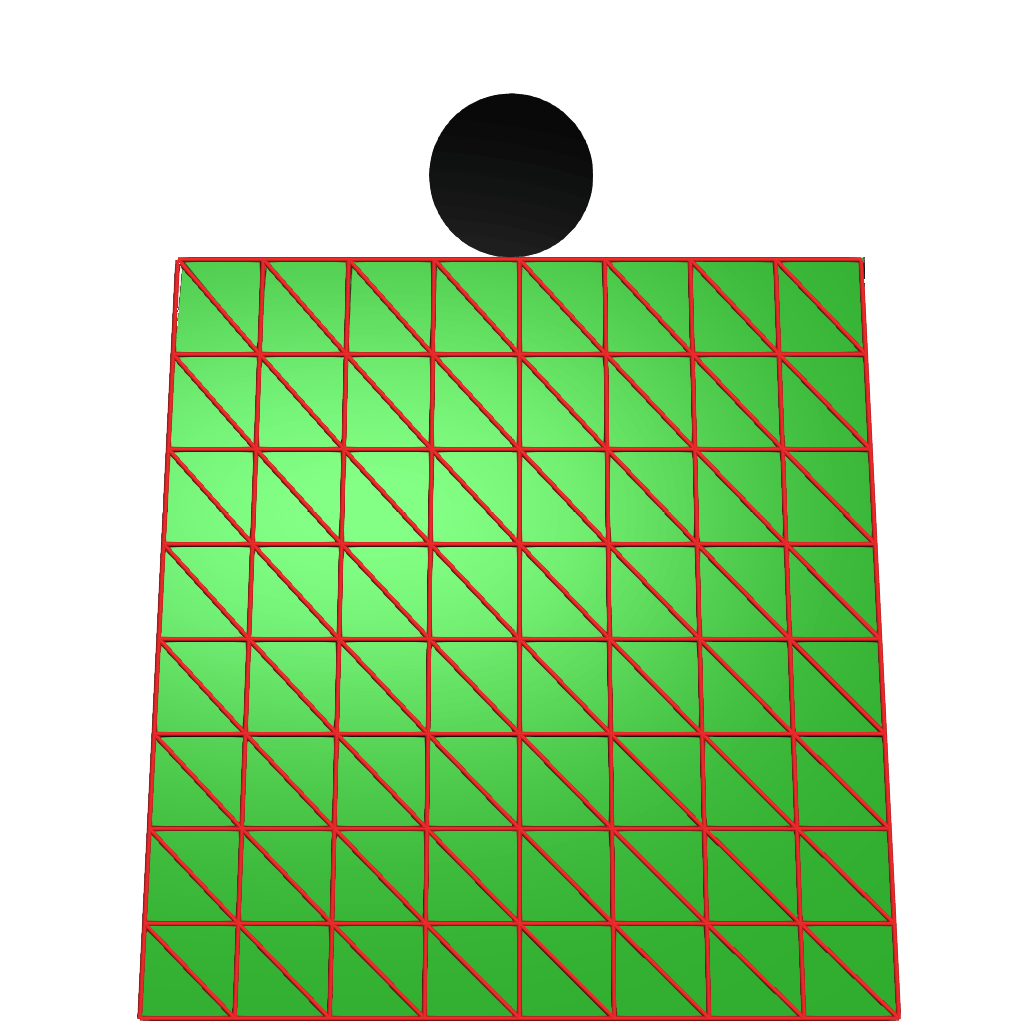}
    \end{minipage}
    \begin{minipage}{0.119\textwidth}
            \centering
            \includegraphics[width=\textwidth]{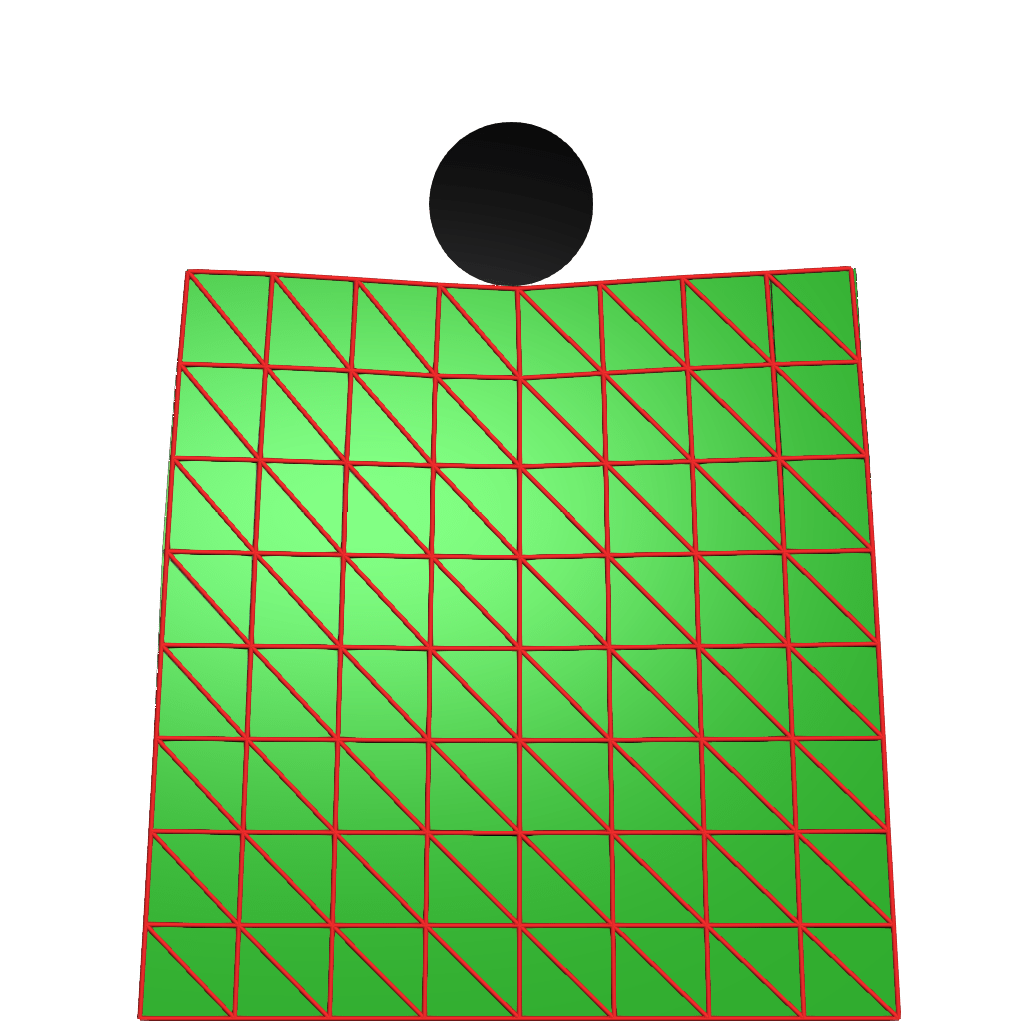}
    \end{minipage}
    \begin{minipage}{0.119\textwidth}
            \centering
            \includegraphics[width=\textwidth]{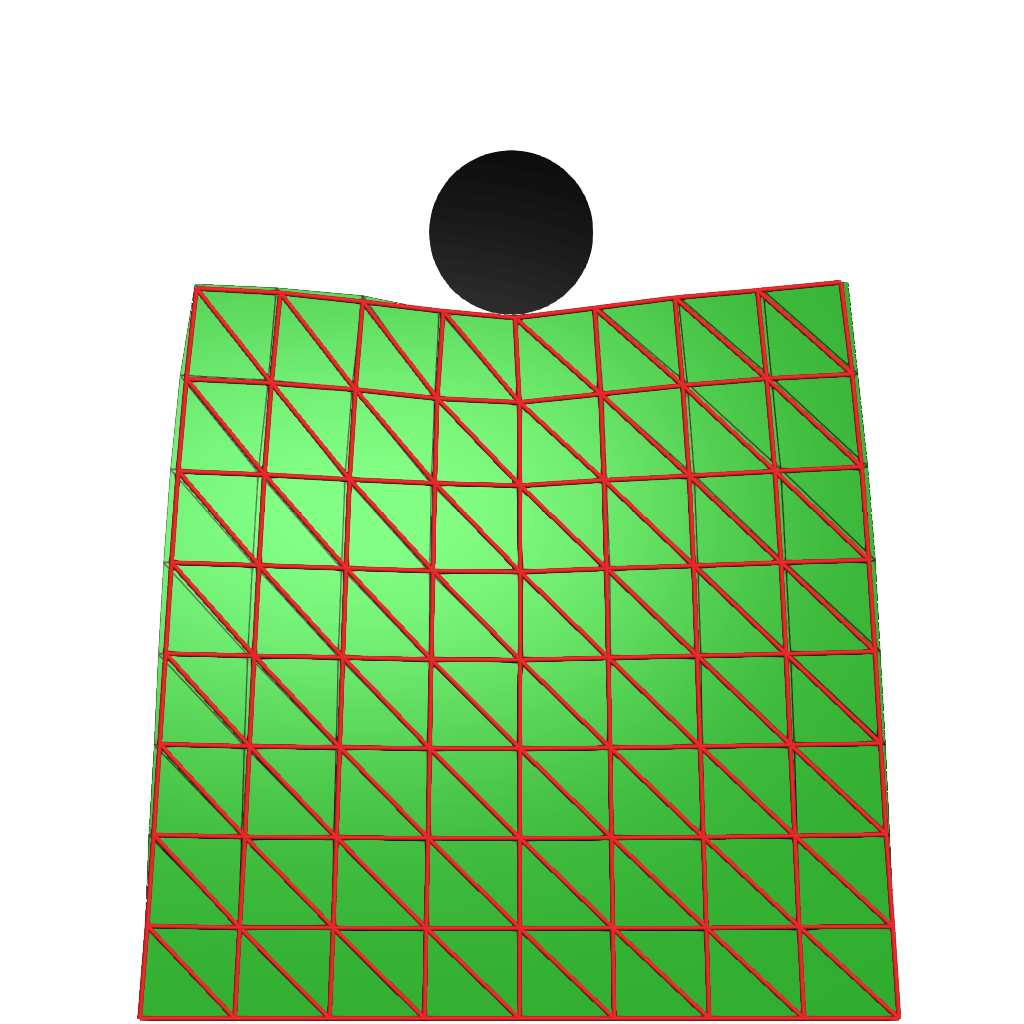}
    \end{minipage}
    \begin{minipage}{0.119\textwidth}
            \centering
            \includegraphics[width=\textwidth]{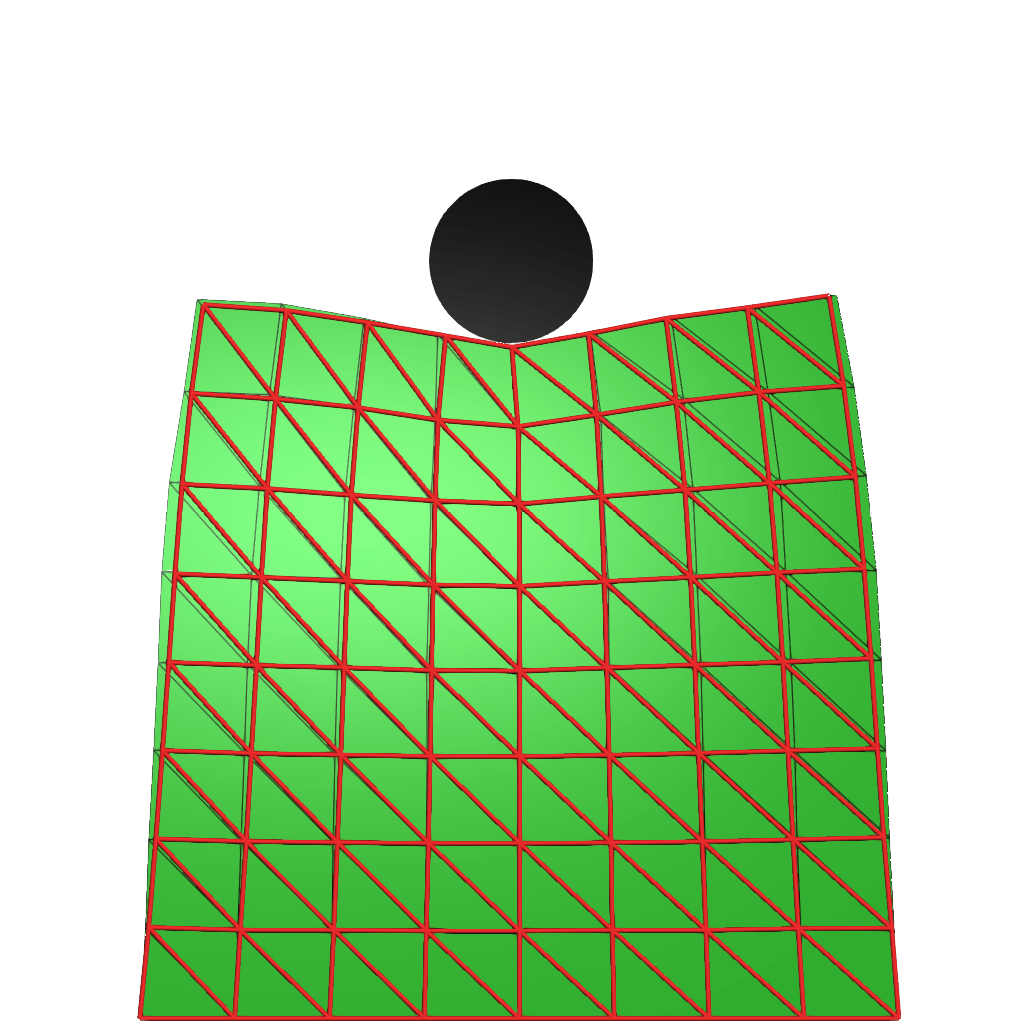}
    \end{minipage}
    \begin{minipage}{0.119\textwidth}
            \centering
            \includegraphics[width=\textwidth]{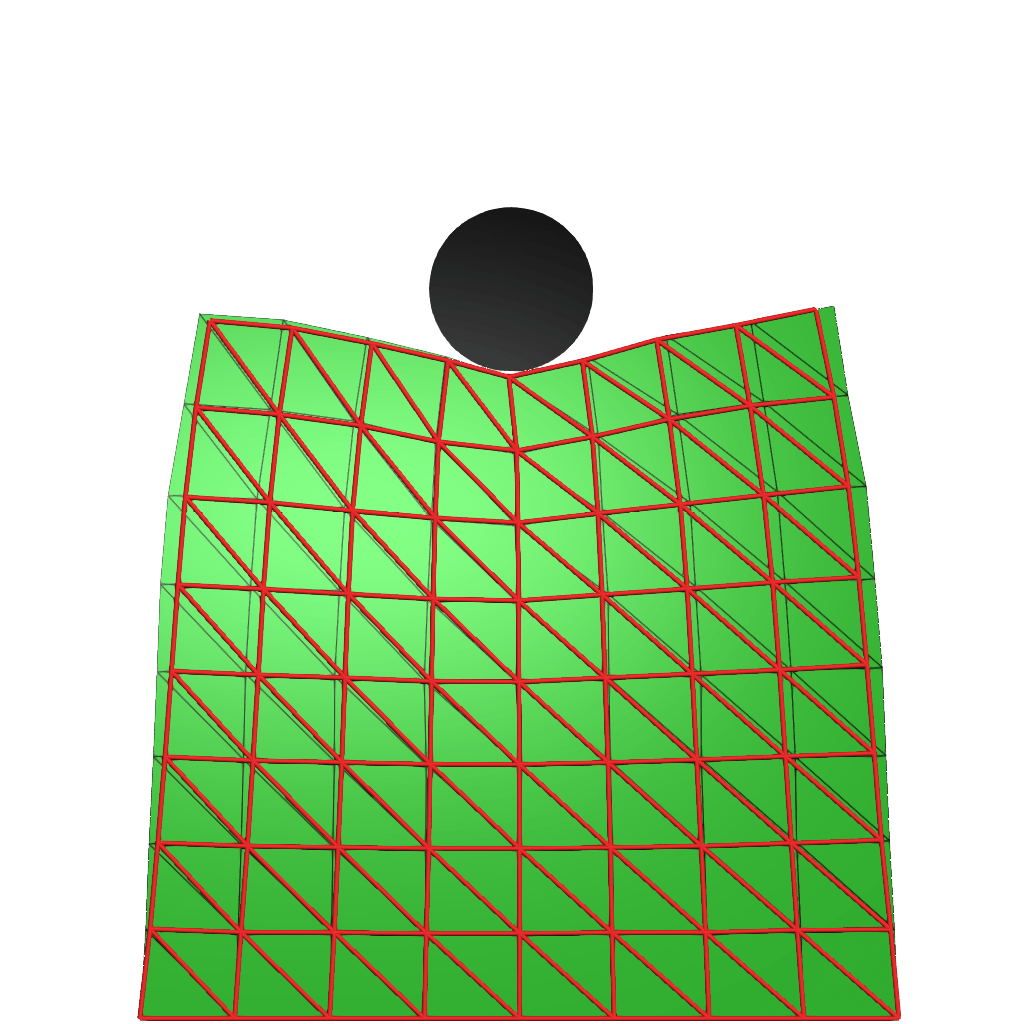}
    \end{minipage}
    \begin{minipage}{0.119\textwidth}
            \centering
            \includegraphics[width=\textwidth]{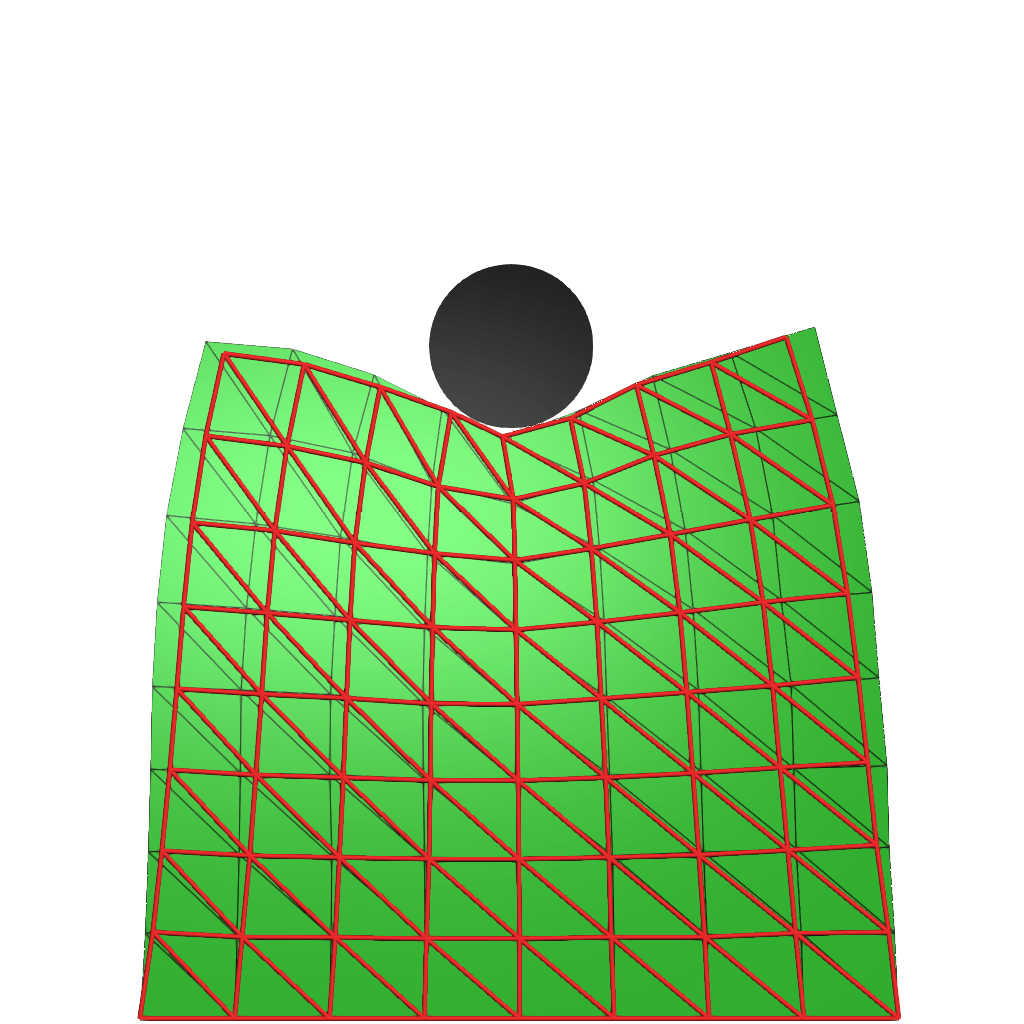}
    \end{minipage}
    \begin{minipage}{0.119\textwidth}
            \centering
            \includegraphics[width=\textwidth]{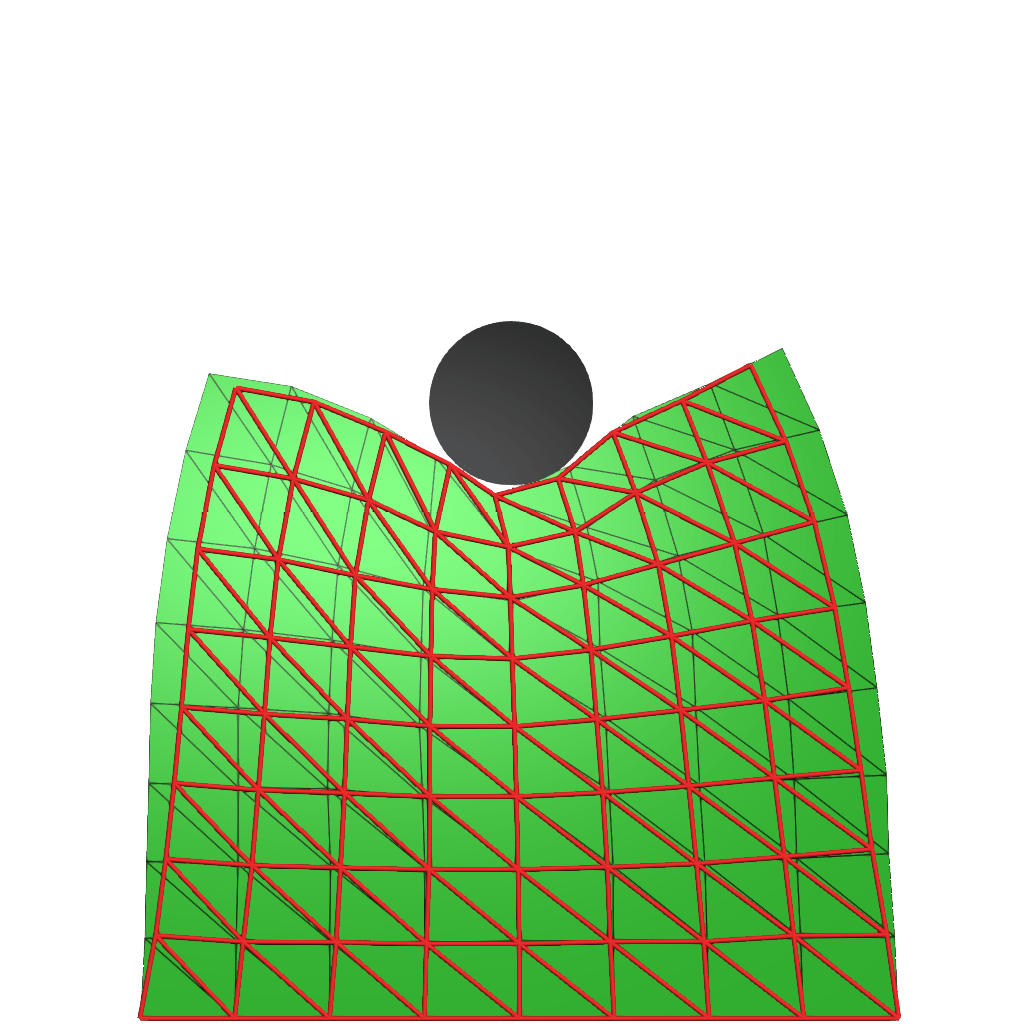}
    \end{minipage}
    \begin{minipage}{0.119\textwidth}
            \centering
            \includegraphics[width=\textwidth]{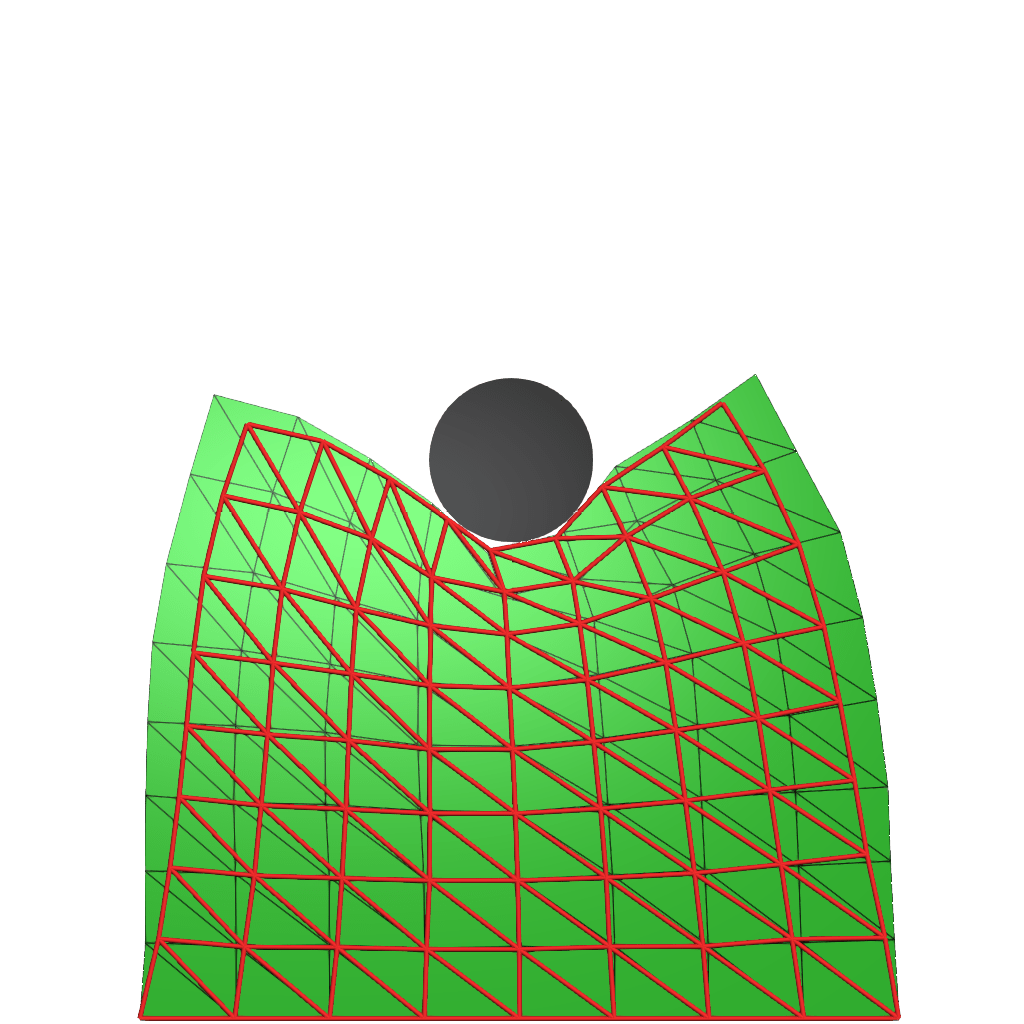}            
    \end{minipage}

    \vspace{1.0em} 
    \noindent\hrulefill 

    \vspace{-1.0em}
    \noindent\hrulefill 
    \vspace{0.5em} 

    \begin{minipage}{0.119\textwidth}
            \centering
            \includegraphics[width=\textwidth]{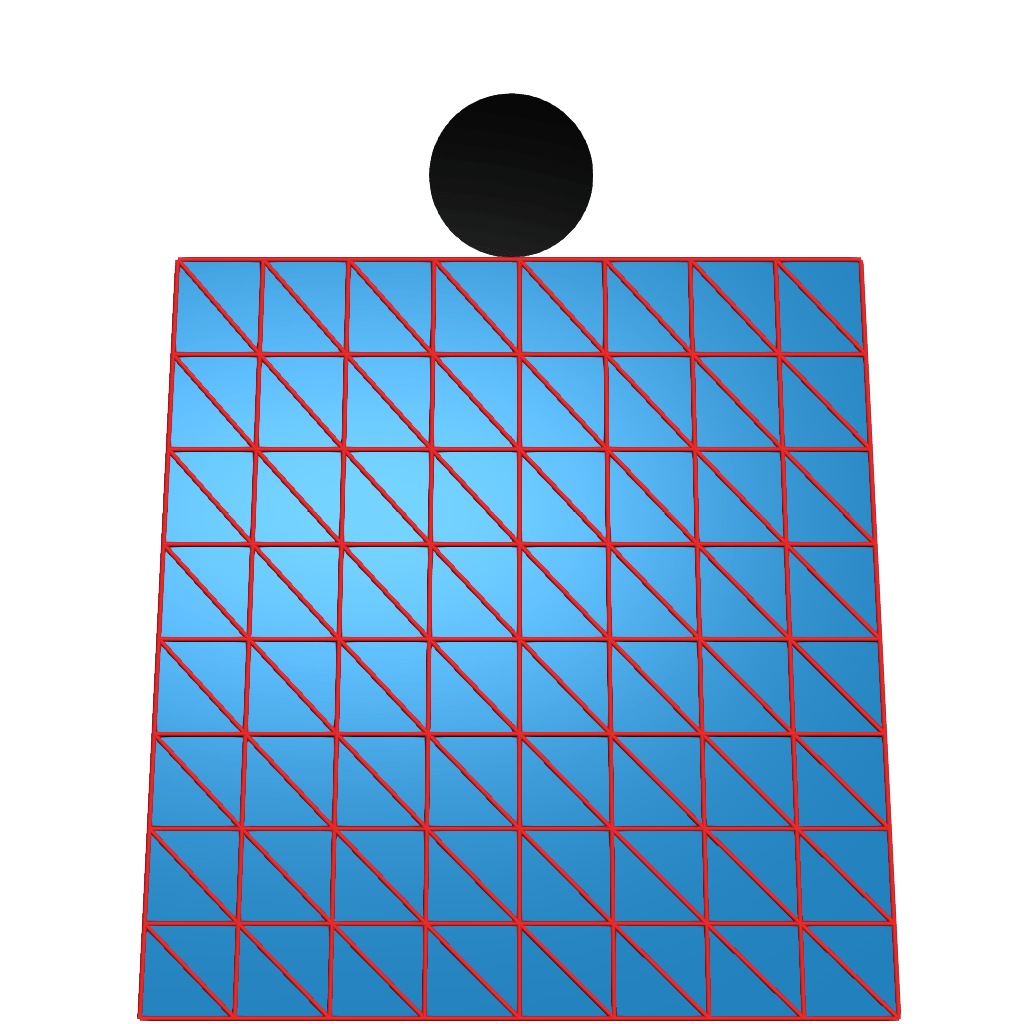}
            \subcaption*{$t=1$}
    \end{minipage}
    \begin{minipage}{0.119\textwidth}
            \centering
            \includegraphics[width=\textwidth]{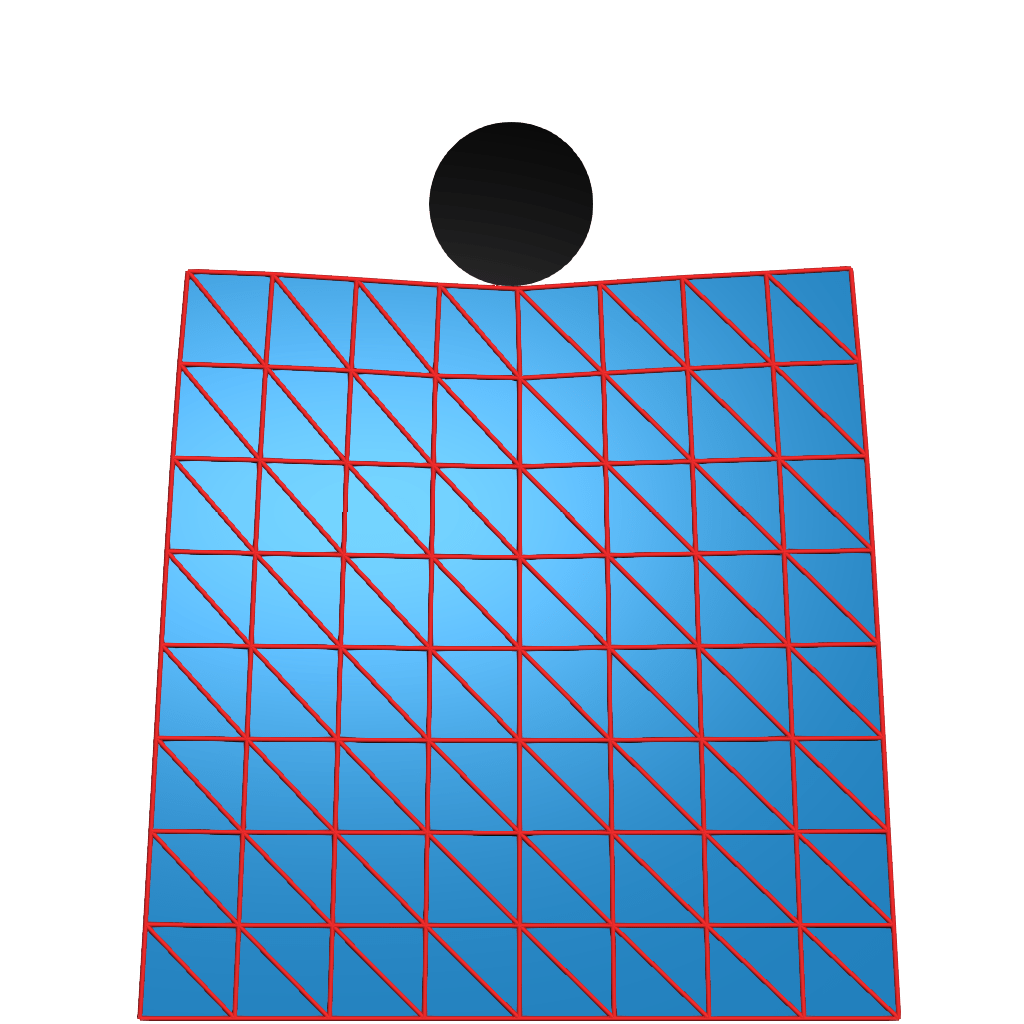}
            \subcaption*{$t=6$}
    \end{minipage}
    \begin{minipage}{0.119\textwidth}
            \centering
            \includegraphics[width=\textwidth]{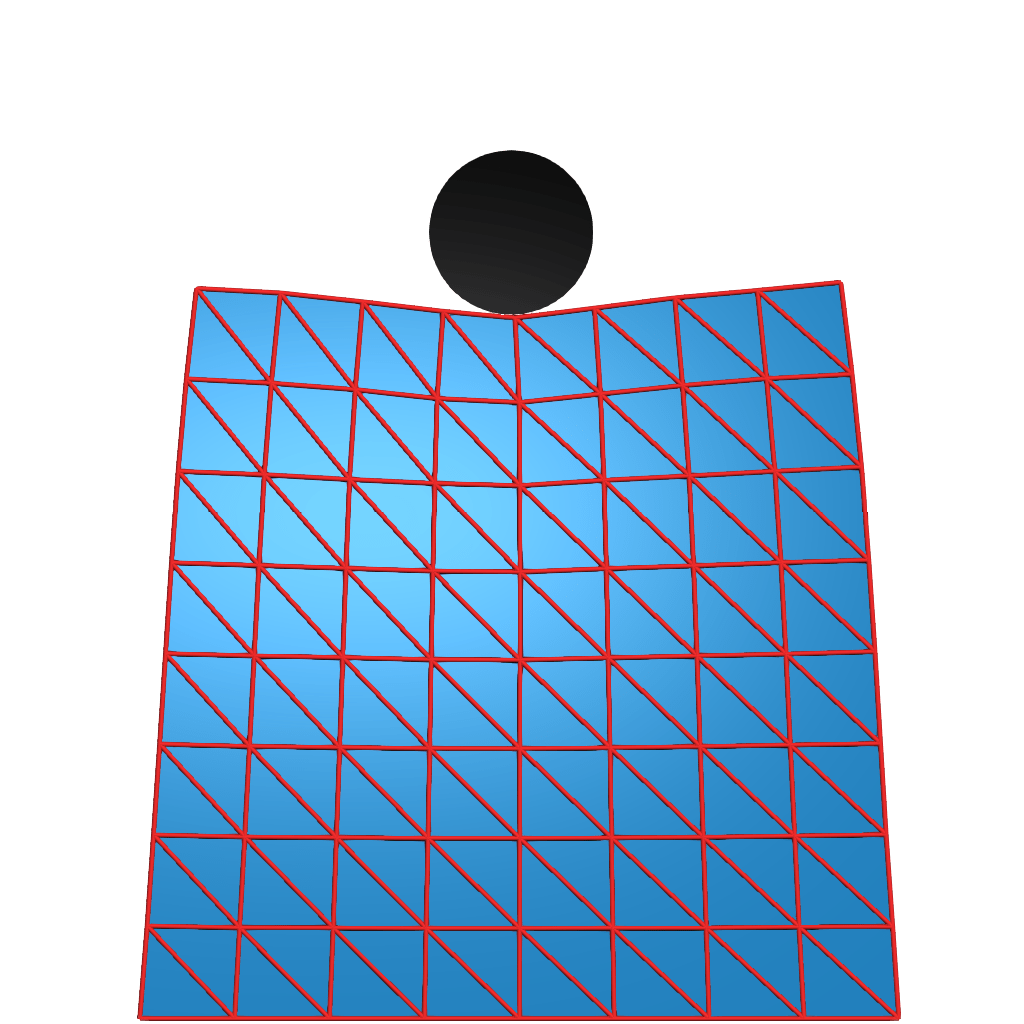}
            \subcaption*{$t=11$}
    \end{minipage}
    \begin{minipage}{0.119\textwidth}
            \centering
            \includegraphics[width=\textwidth]{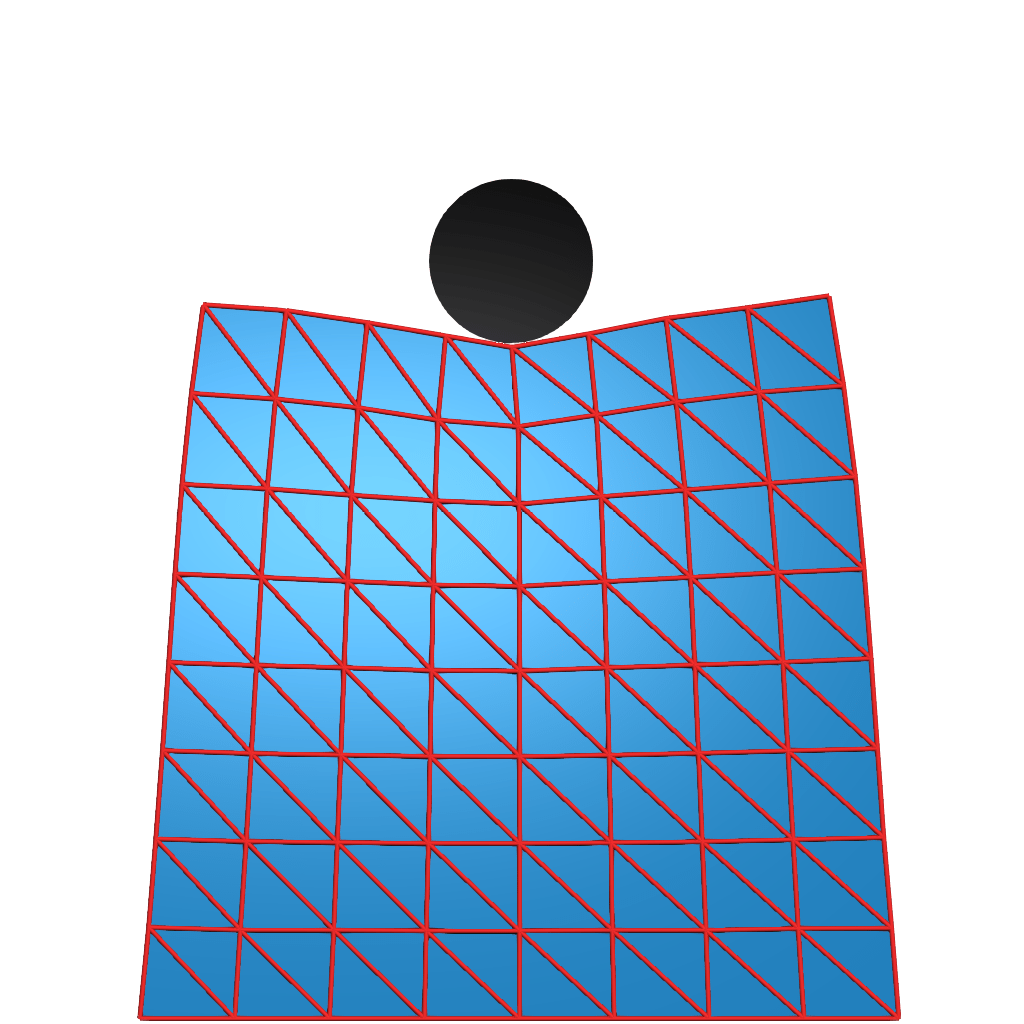}
            \subcaption*{$t=16$}
    \end{minipage}
    \begin{minipage}{0.119\textwidth}
            \centering
            \includegraphics[width=\textwidth]{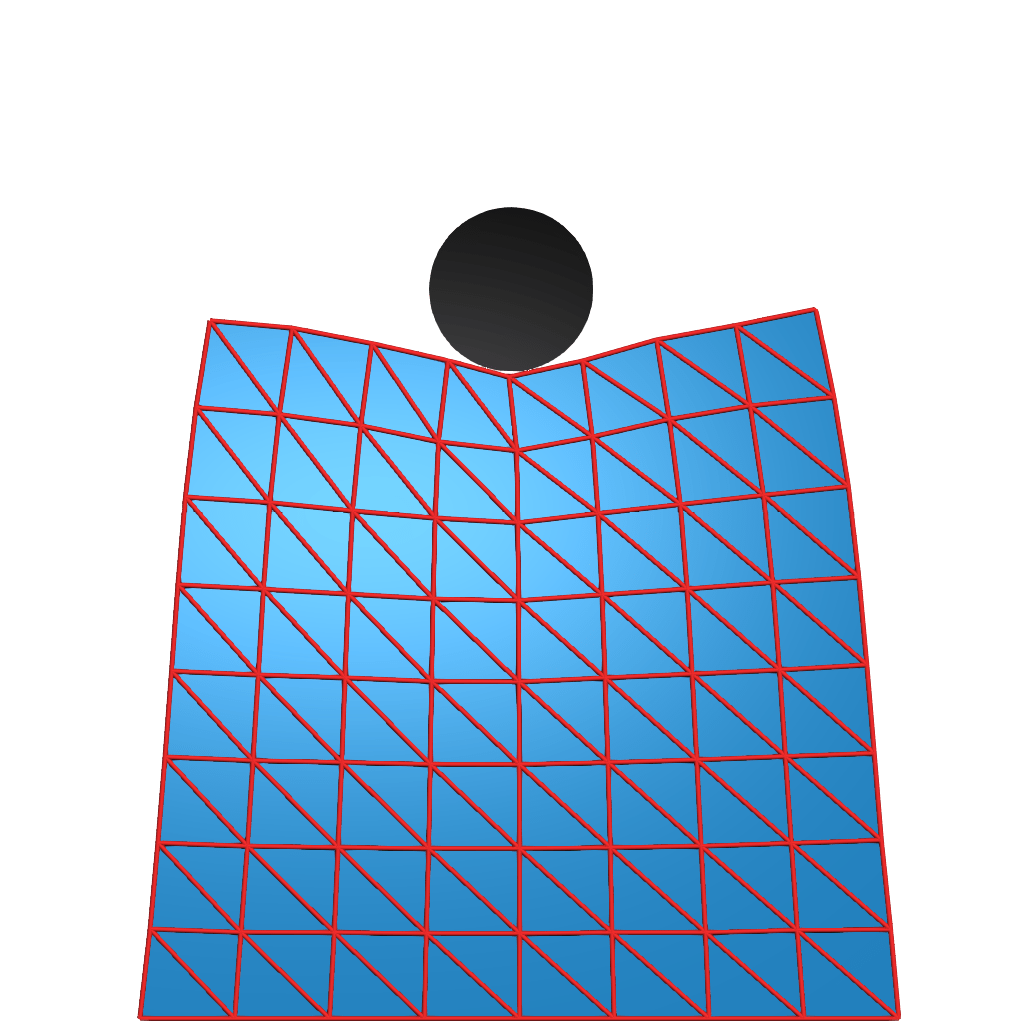}
            \subcaption*{$t=21$}
    \end{minipage}
    \begin{minipage}{0.119\textwidth}
            \centering
            \includegraphics[width=\textwidth]{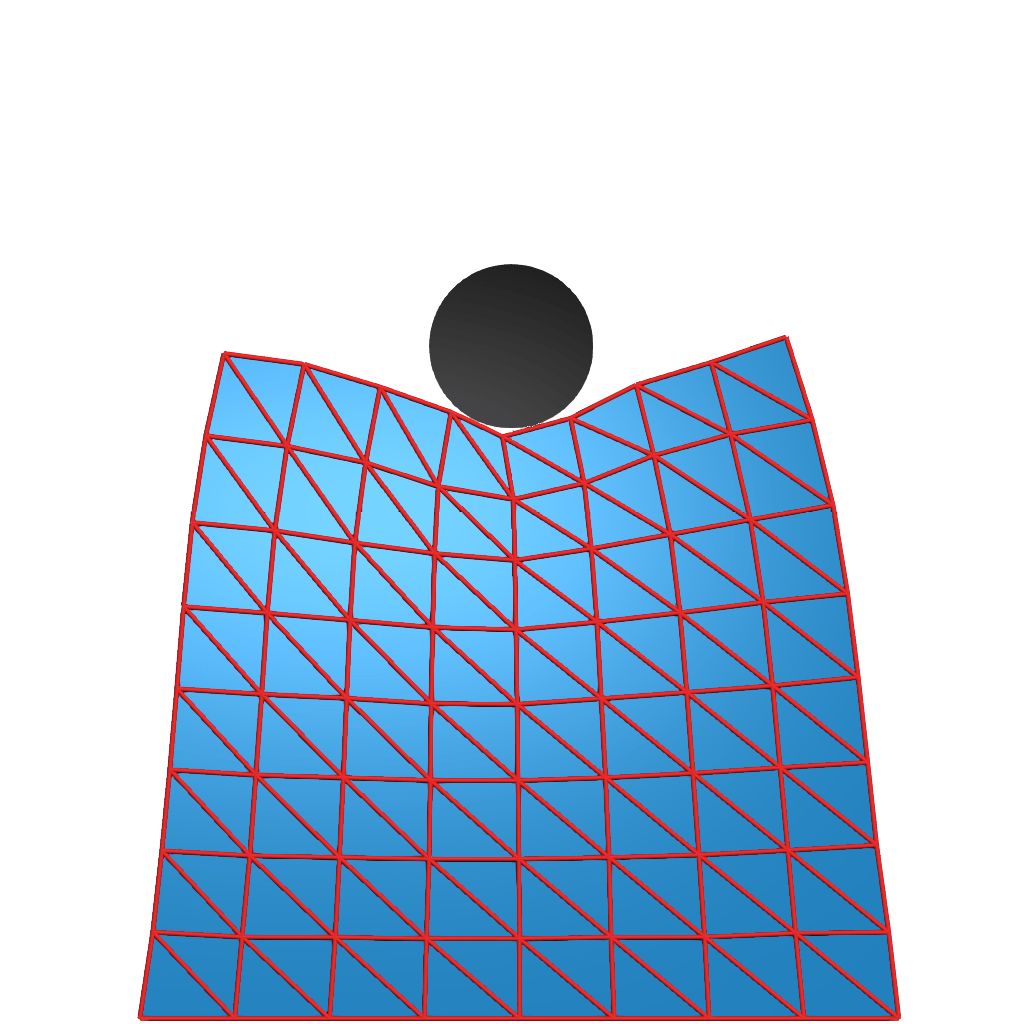}
            \subcaption*{$t=31$}
    \end{minipage}
    \begin{minipage}{0.119\textwidth}
            \centering
            \includegraphics[width=\textwidth]{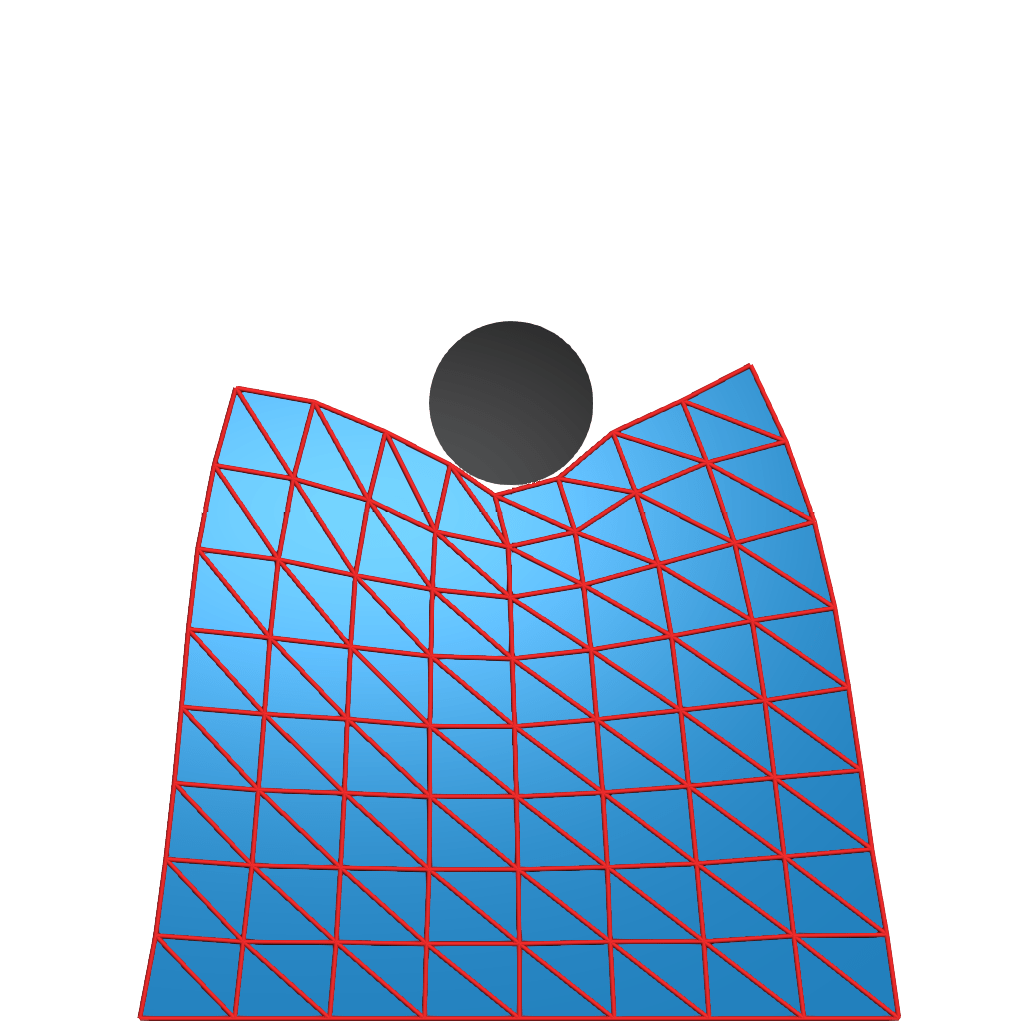}
            \subcaption*{$t=41$}
    \end{minipage}
    \begin{minipage}{0.119\textwidth}
            \centering
            \includegraphics[width=\textwidth]{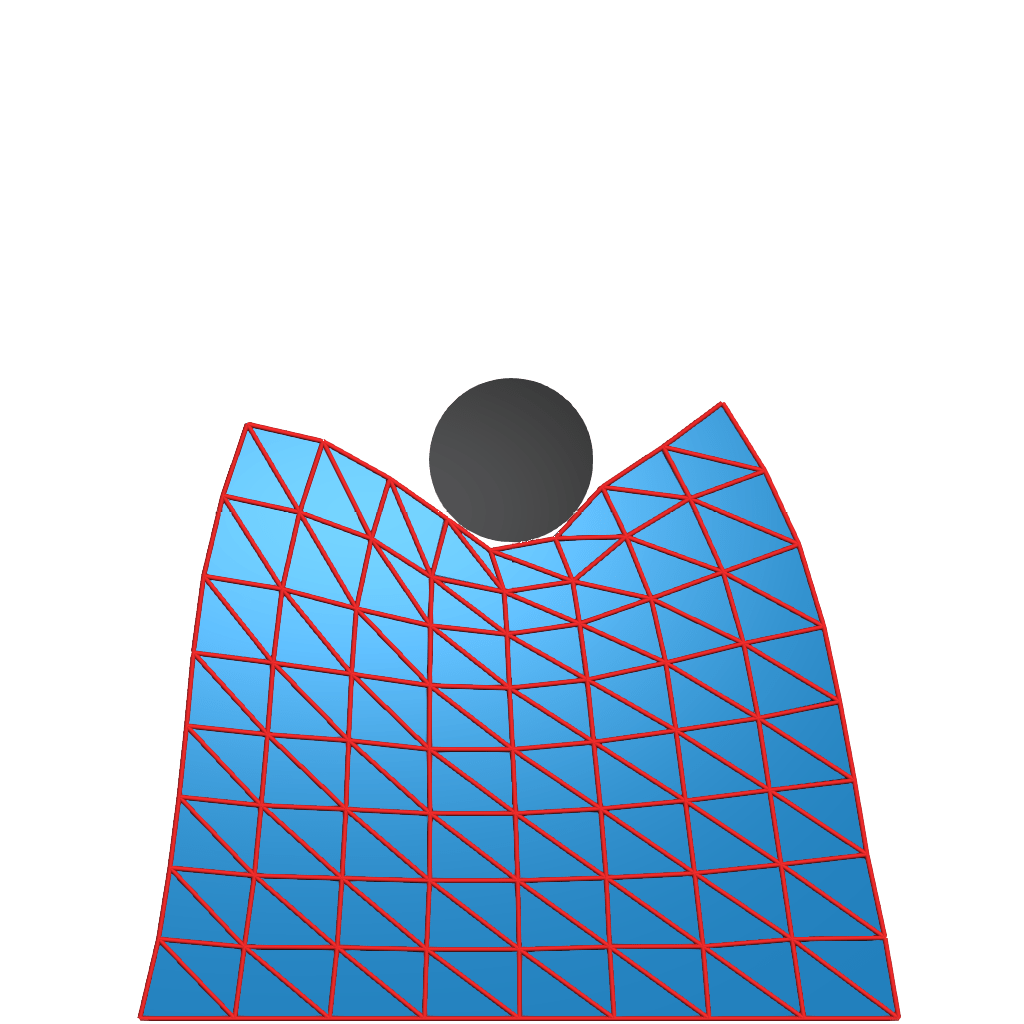}
            \subcaption*{$t=51$}
    \end{minipage}

    \vspace{0.01\textwidth}%

    \caption{
    Simulation over time of an exemplary test trajectory from the \textbf{DP-easy} task. The figure compares predictions from \textcolor{tabblue}{\textbf{MaNGO}}, \textcolor{taborange}{MGN}, and \textcolor{ForestGreen}{EGNO}. The last row, \textcolor{tabblue}{MaNGO-Oracle}, is separated by a horizontal line and represents predictions using oracle information. The \textbf{context set size} is set to $4$. All visualizations show the colored \textbf{predicted mesh}, with a \textbf{\textcolor{red}{wireframe}} representing the ground-truth simulation. \textcolor{tabblue}{\textbf{MaNGO}} accurately predicts the correct material properties, leading to a highly accurate simulation.
    }
    \label{fig:appendix_dp_easy}
\end{figure*}
\begin{figure*}[ht!]
    \centering
    \begin{minipage}{0.119\textwidth}
            \centering
            \includegraphics[width=\textwidth]{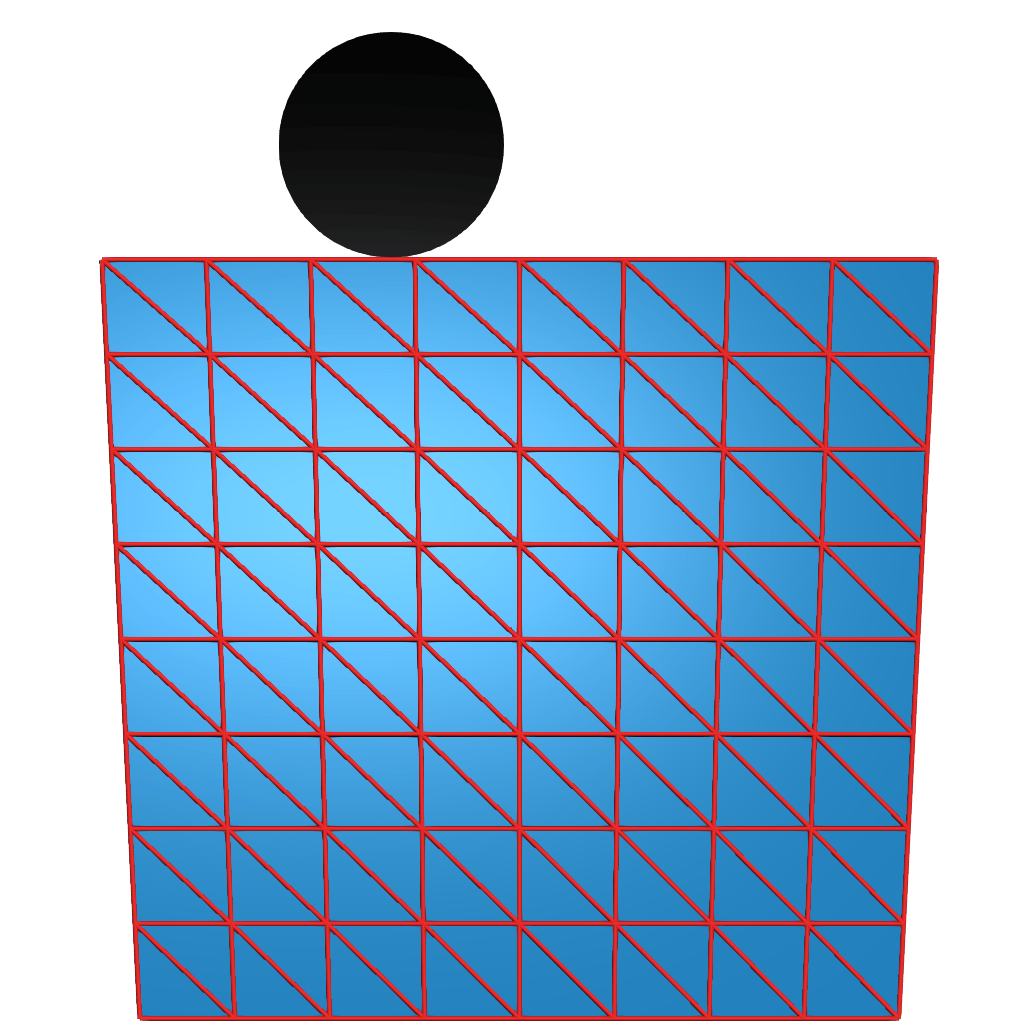}
    \end{minipage}
    \begin{minipage}{0.119\textwidth}
            \centering
            \includegraphics[width=\textwidth]{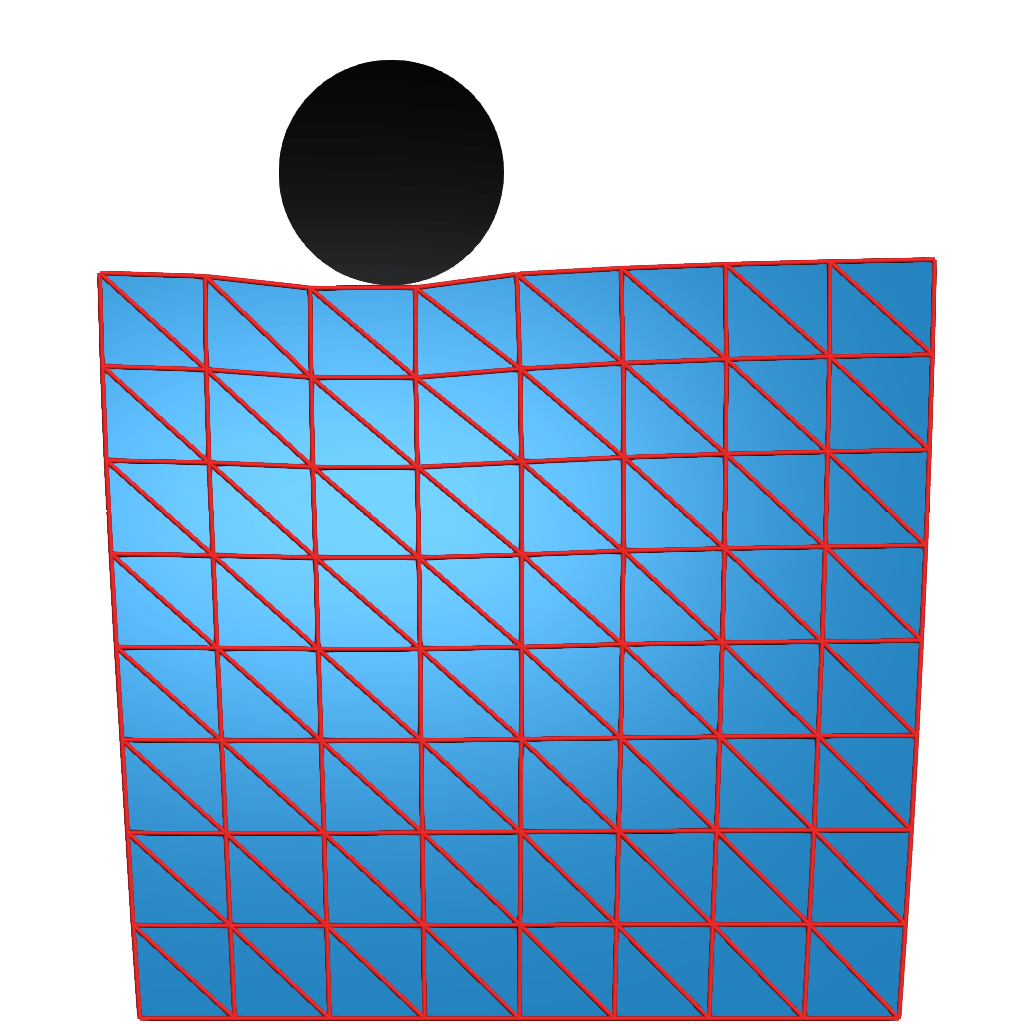}
    \end{minipage}
    \begin{minipage}{0.119\textwidth}
            \centering
            \includegraphics[width=\textwidth]{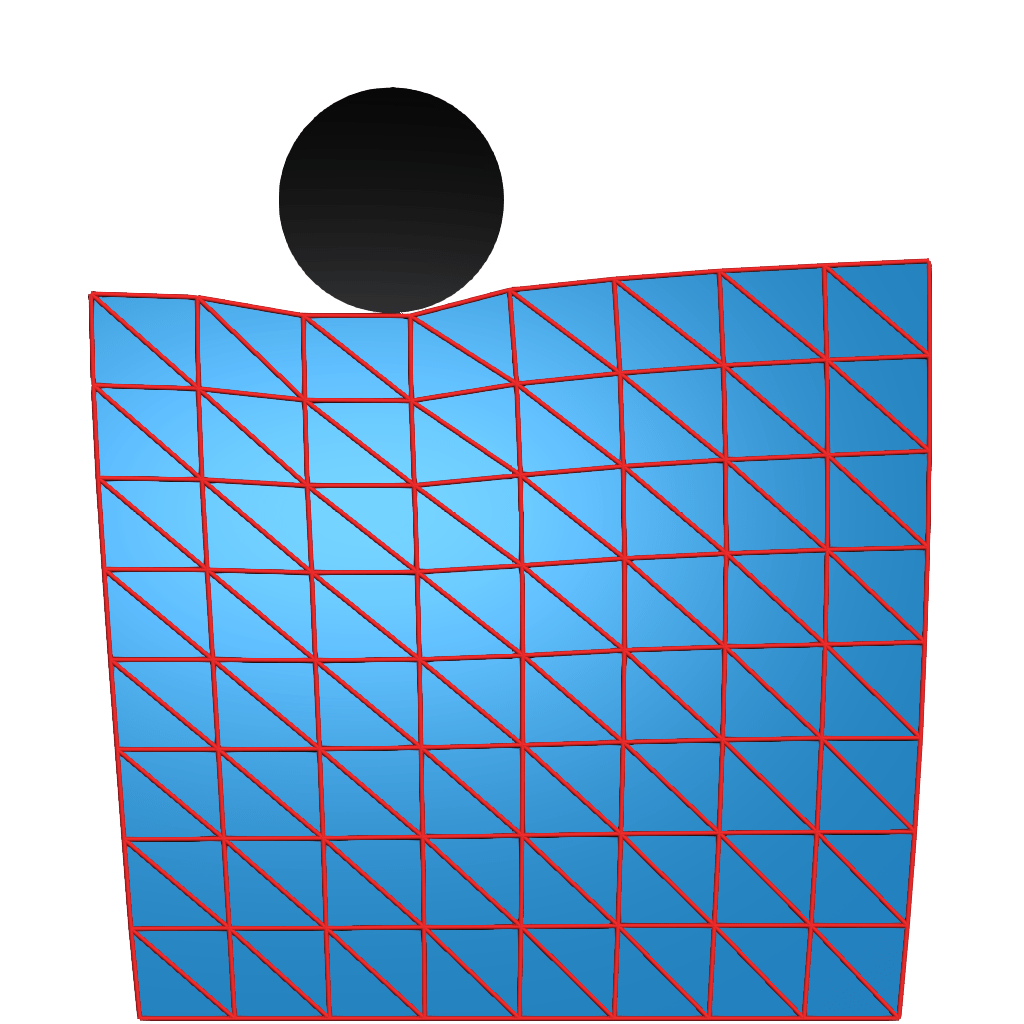}
    \end{minipage}
    \begin{minipage}{0.119\textwidth}
            \centering
            \includegraphics[width=\textwidth]{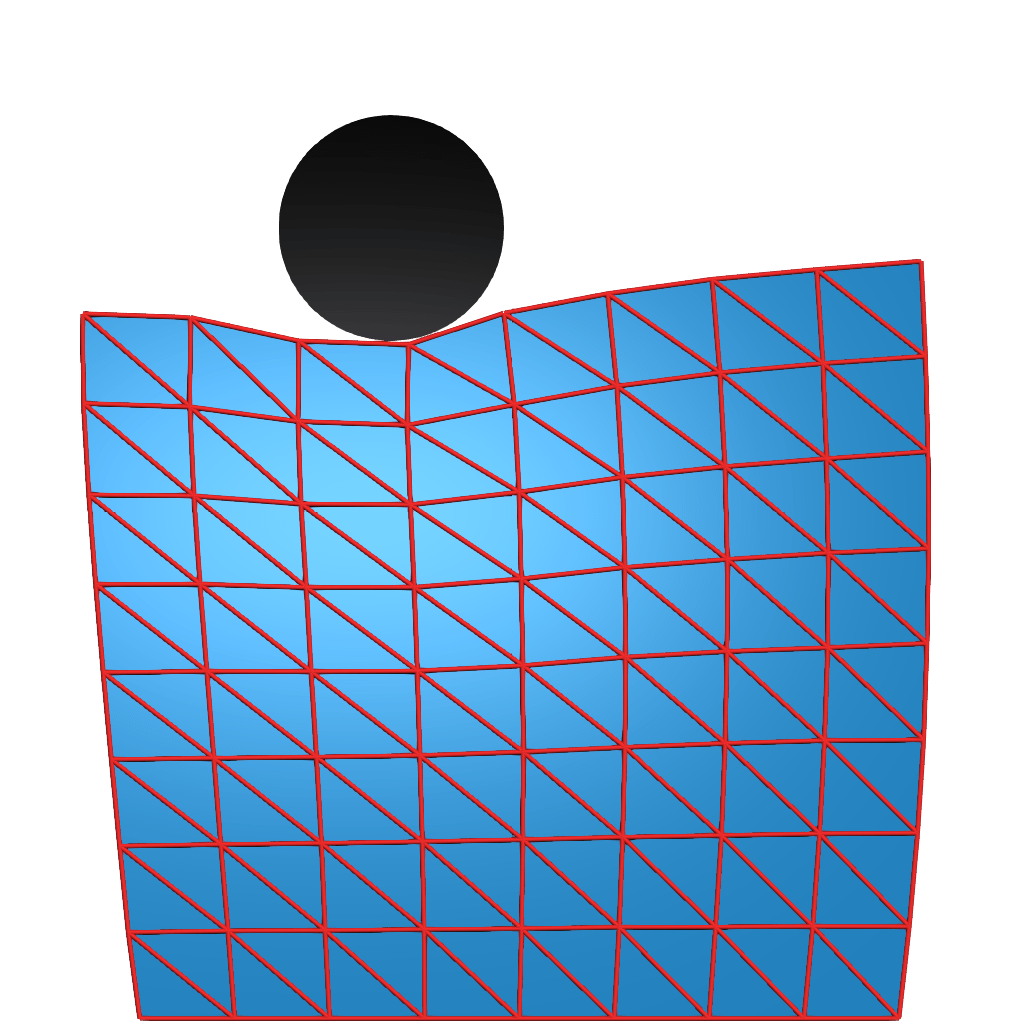}
    \end{minipage}
    \begin{minipage}{0.119\textwidth}
            \centering
            \includegraphics[width=\textwidth]{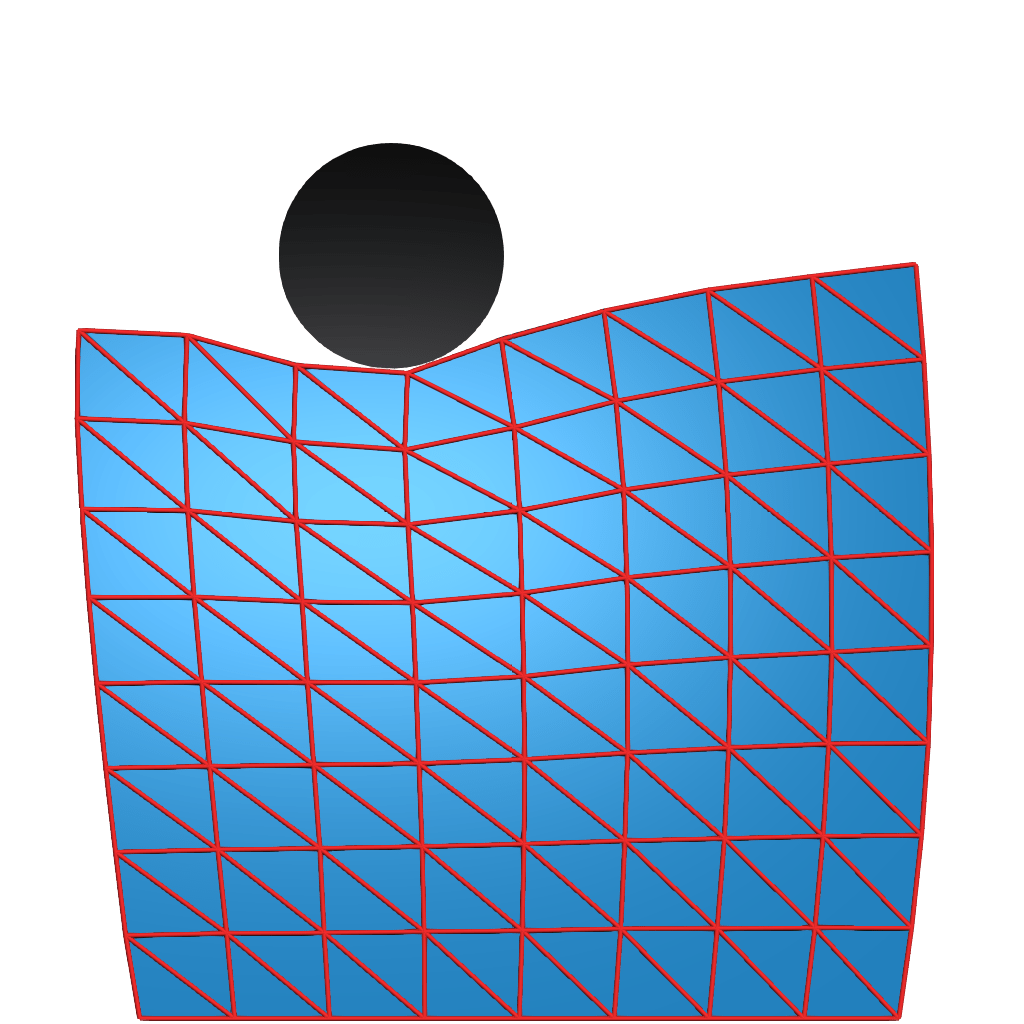}
    \end{minipage}
    \begin{minipage}{0.119\textwidth}
            \centering
            \includegraphics[width=\textwidth]{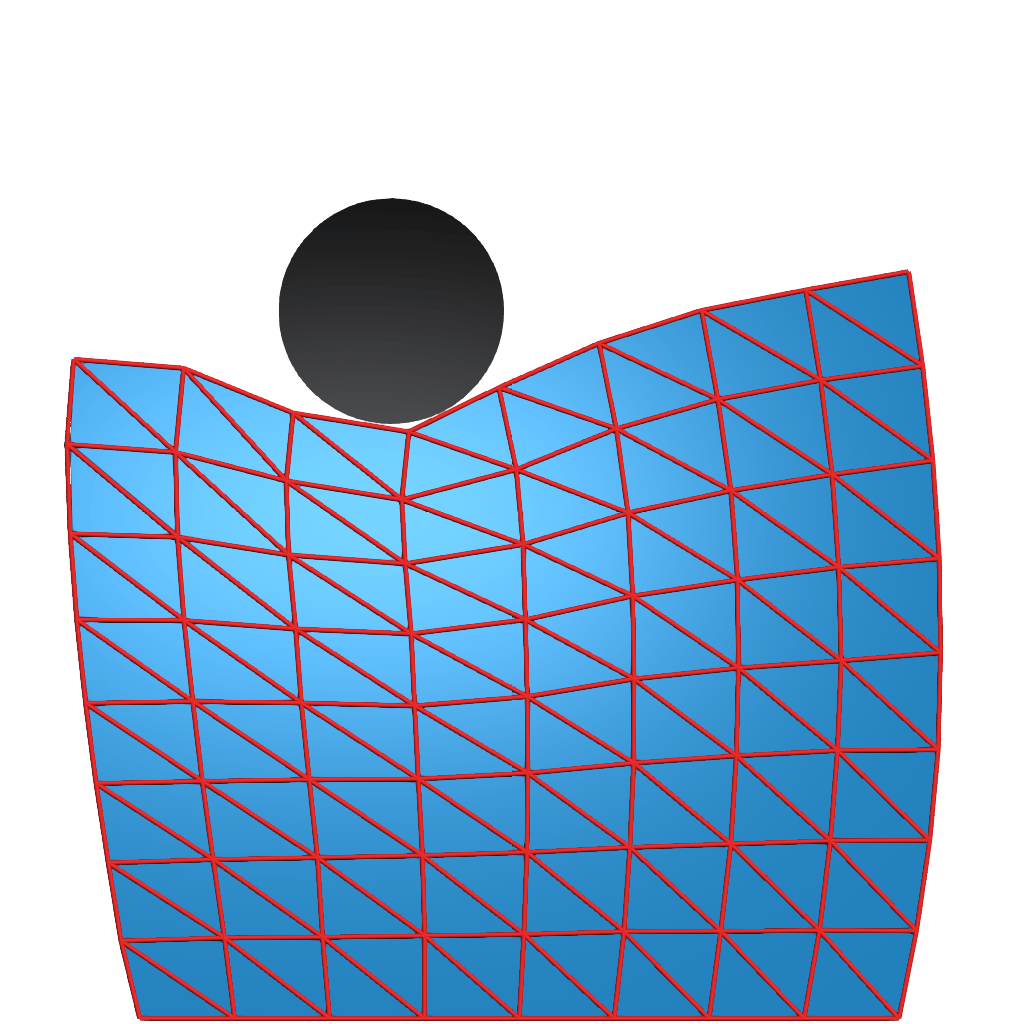}
    \end{minipage}
    \begin{minipage}{0.119\textwidth}
            \centering
            \includegraphics[width=\textwidth]{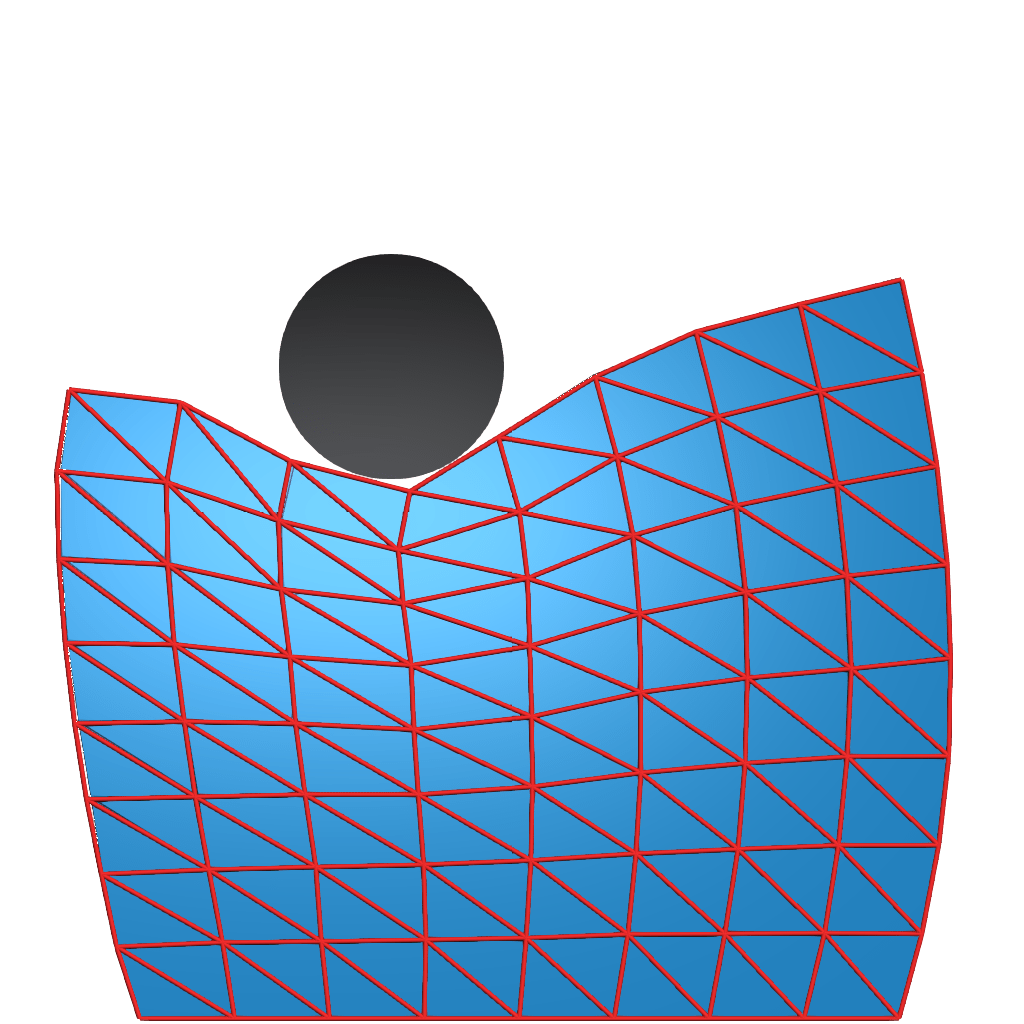}
    \end{minipage}
    \begin{minipage}{0.119\textwidth}
            \centering
            \includegraphics[width=\textwidth]{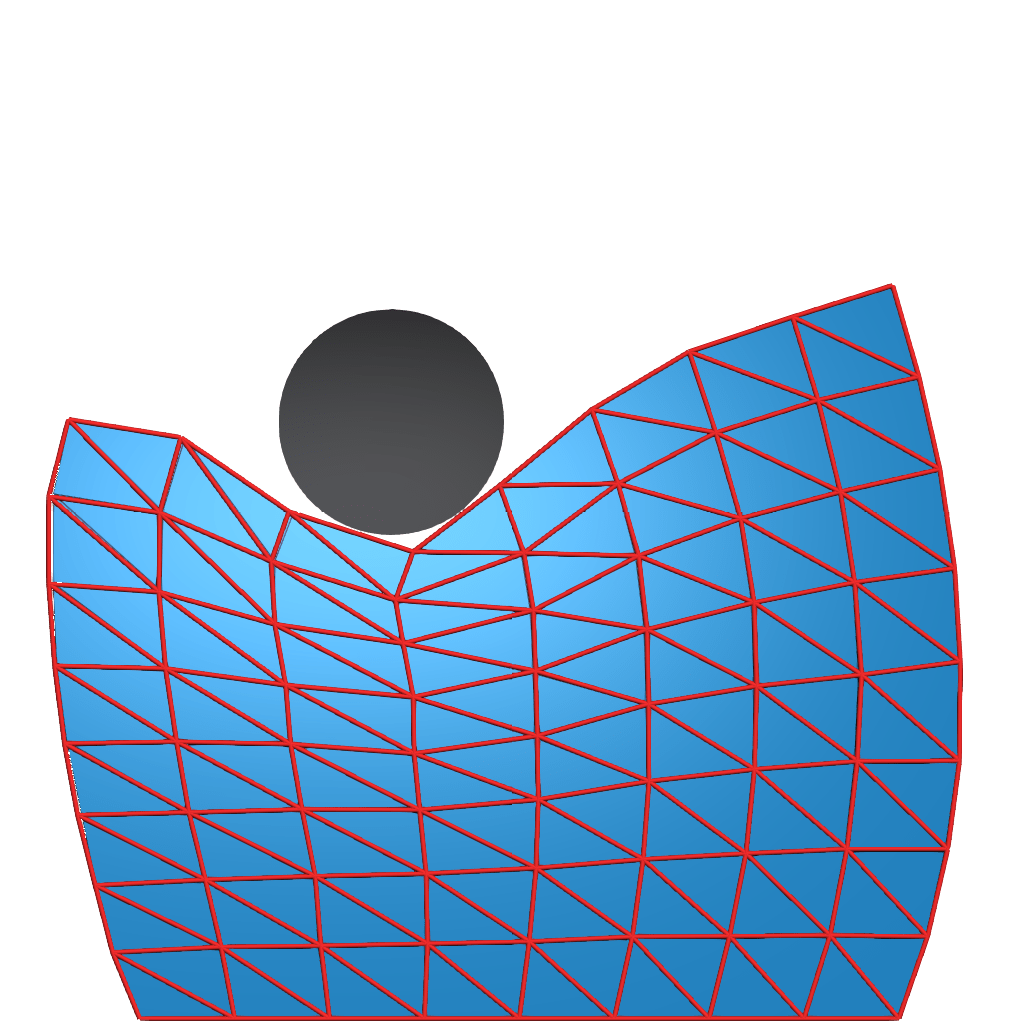}
    \end{minipage}

    \begin{minipage}{0.119\textwidth}
            \centering
            \includegraphics[width=\textwidth]{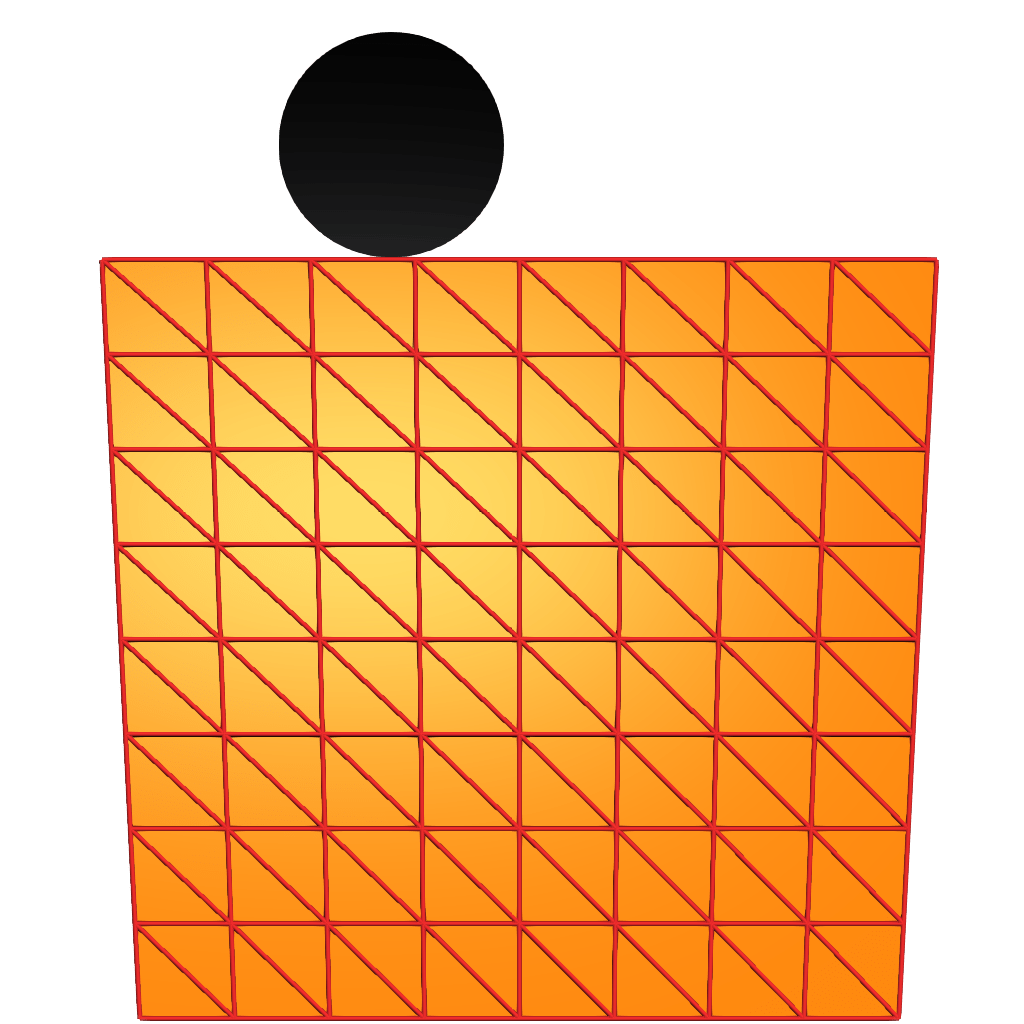}
    \end{minipage}
    \begin{minipage}{0.119\textwidth}
            \centering
            \includegraphics[width=\textwidth]{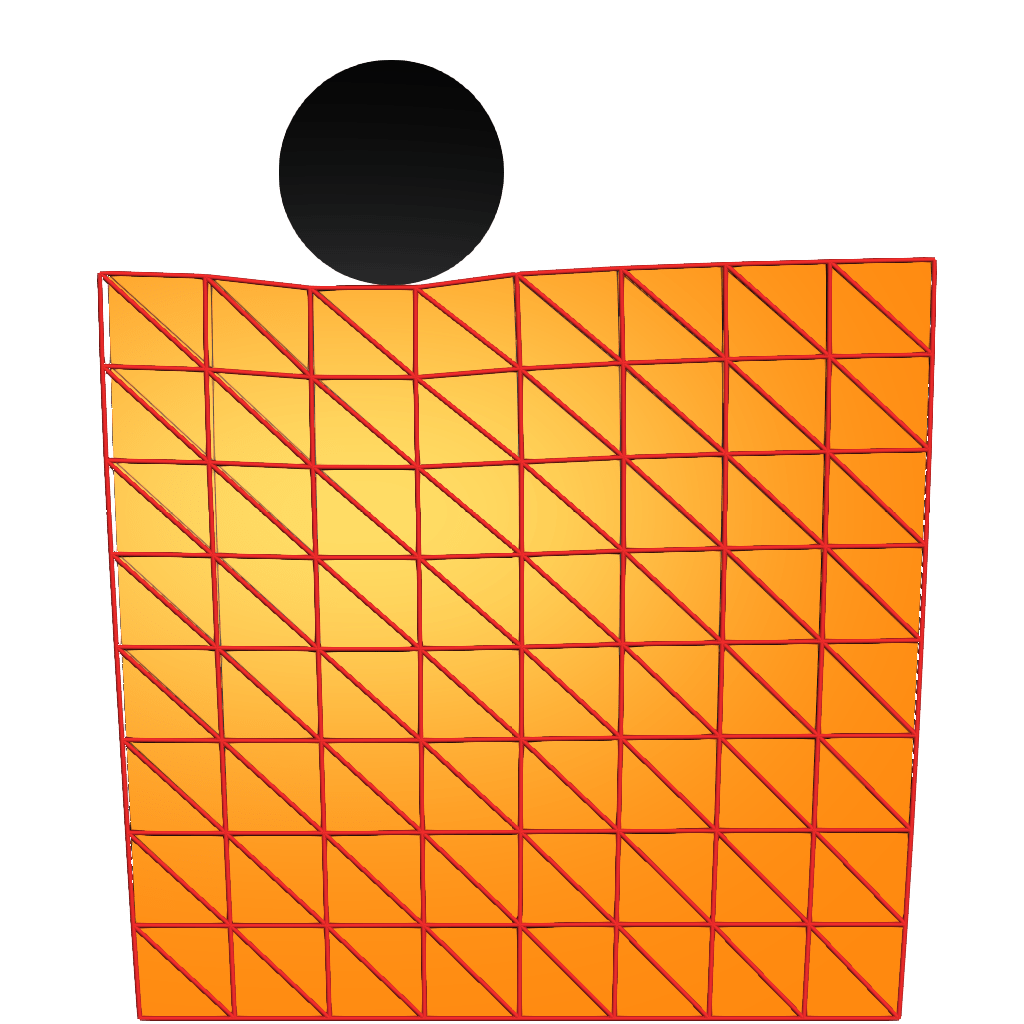}
    \end{minipage}
    \begin{minipage}{0.119\textwidth}
            \centering
            \includegraphics[width=\textwidth]{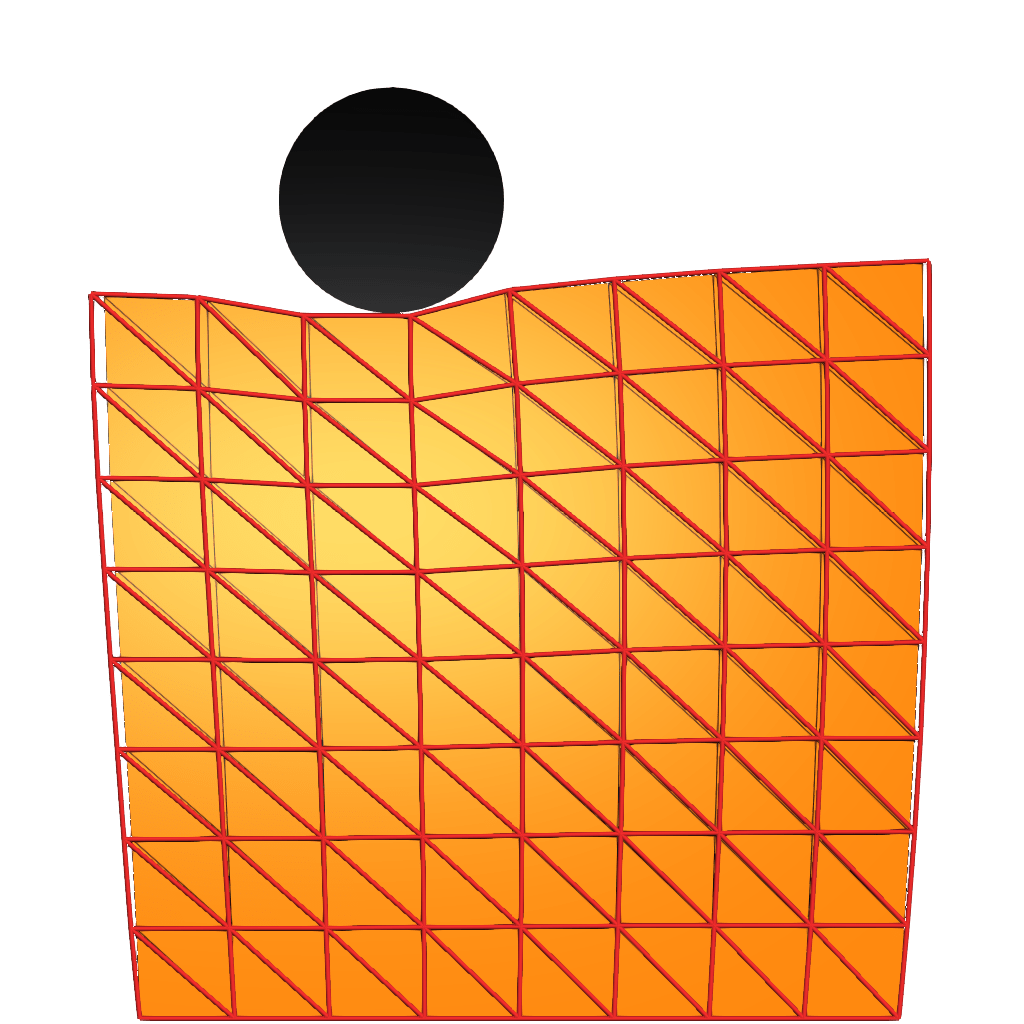}
    \end{minipage}
    \begin{minipage}{0.119\textwidth}
            \centering
            \includegraphics[width=\textwidth]{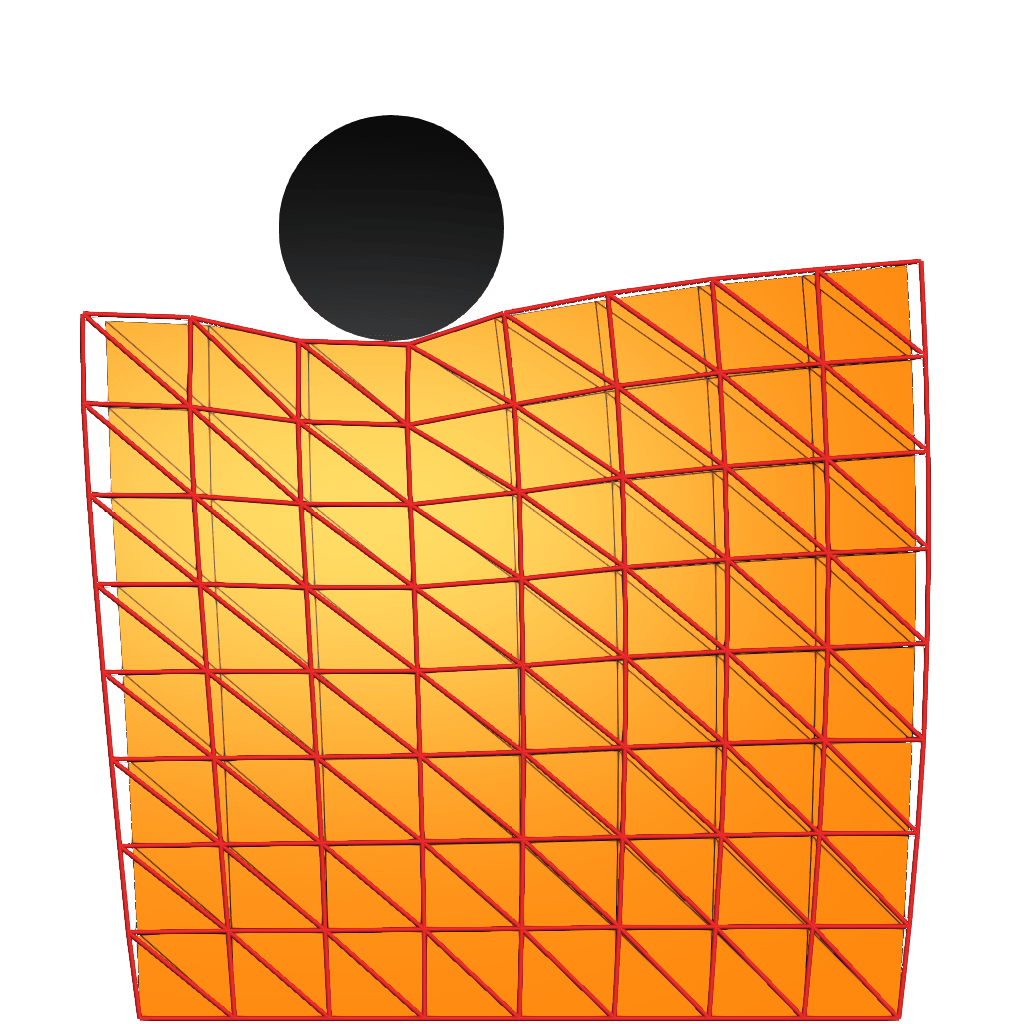}
    \end{minipage}
    \begin{minipage}{0.119\textwidth}
            \centering
            \includegraphics[width=\textwidth]{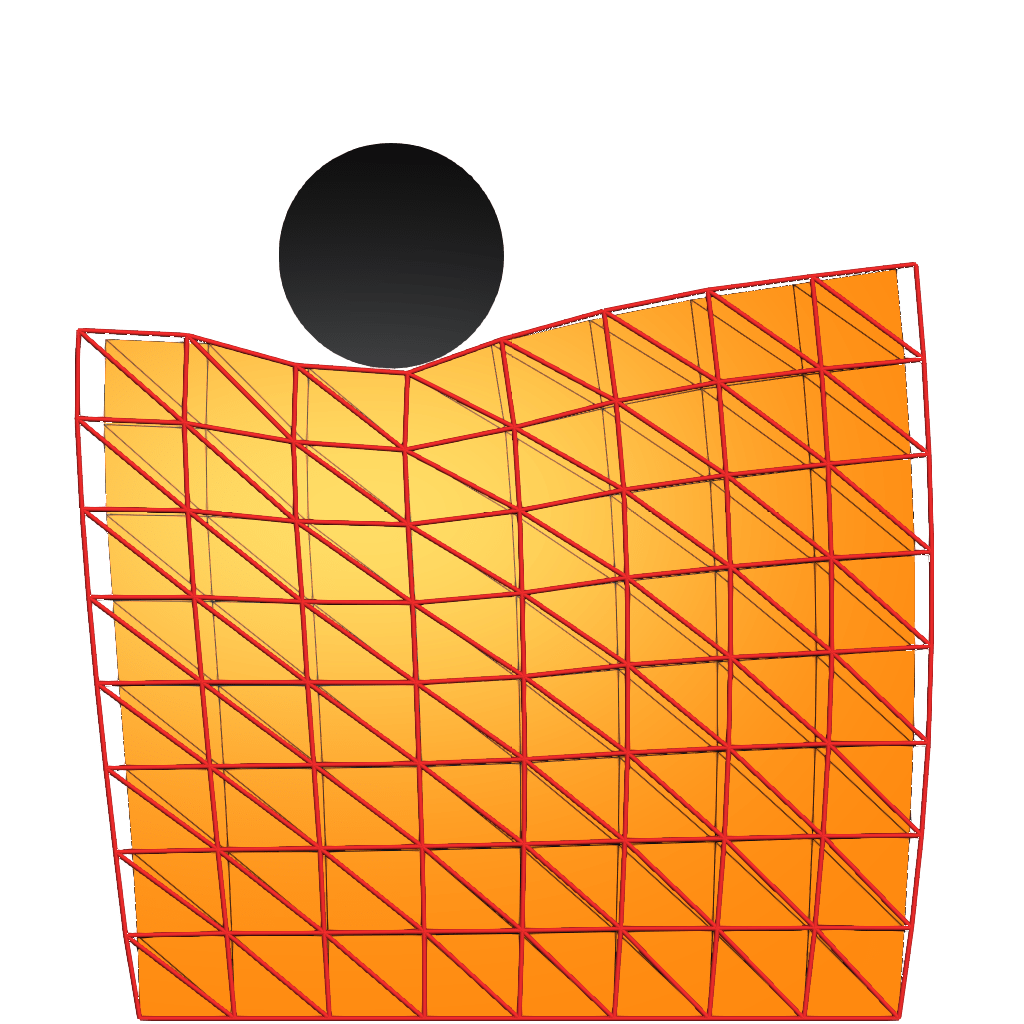}
    \end{minipage}
    \begin{minipage}{0.119\textwidth}
            \centering
            \includegraphics[width=\textwidth]{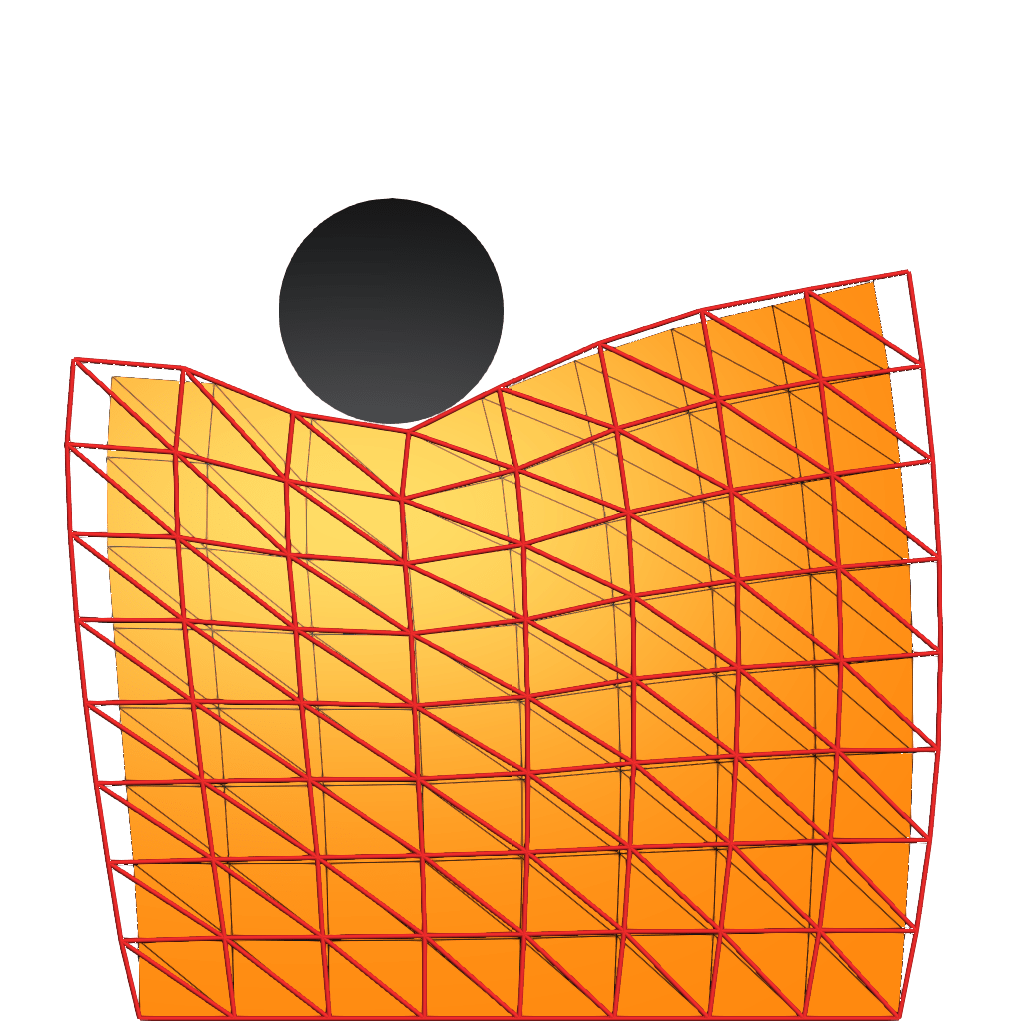}
    \end{minipage}
    \begin{minipage}{0.119\textwidth}
            \centering
            \includegraphics[width=\textwidth]{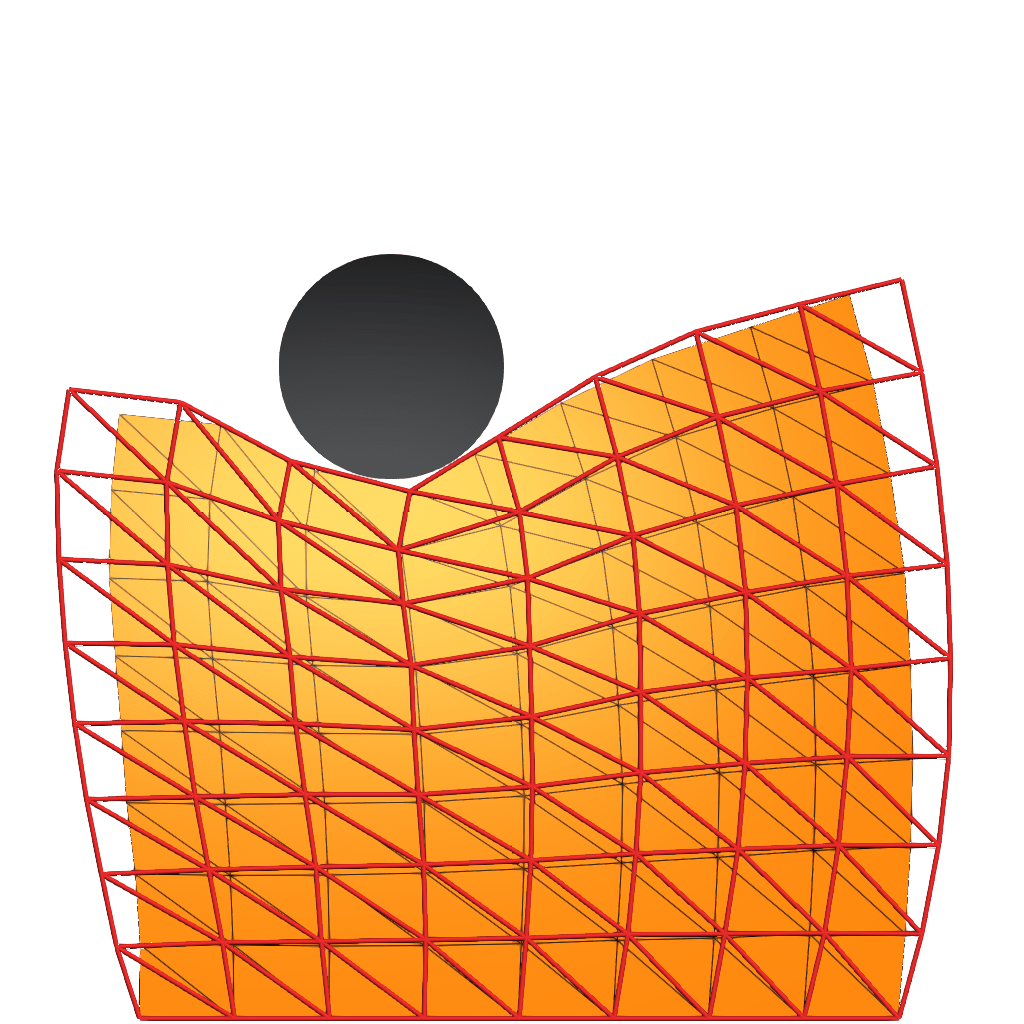}
    \end{minipage}
    \begin{minipage}{0.119\textwidth}
            \centering
            \includegraphics[width=\textwidth]{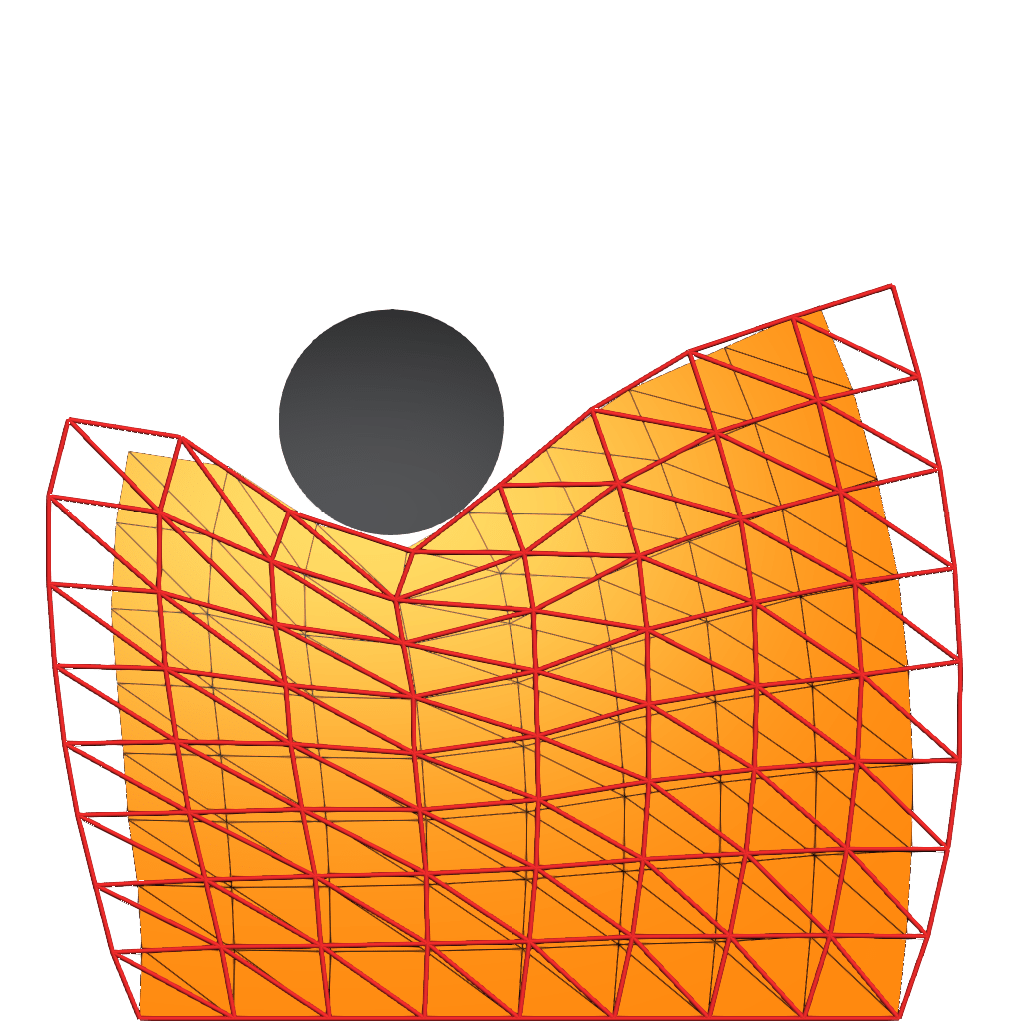}
    \end{minipage}

    \begin{minipage}{0.119\textwidth}
            \centering
            \includegraphics[width=\textwidth]{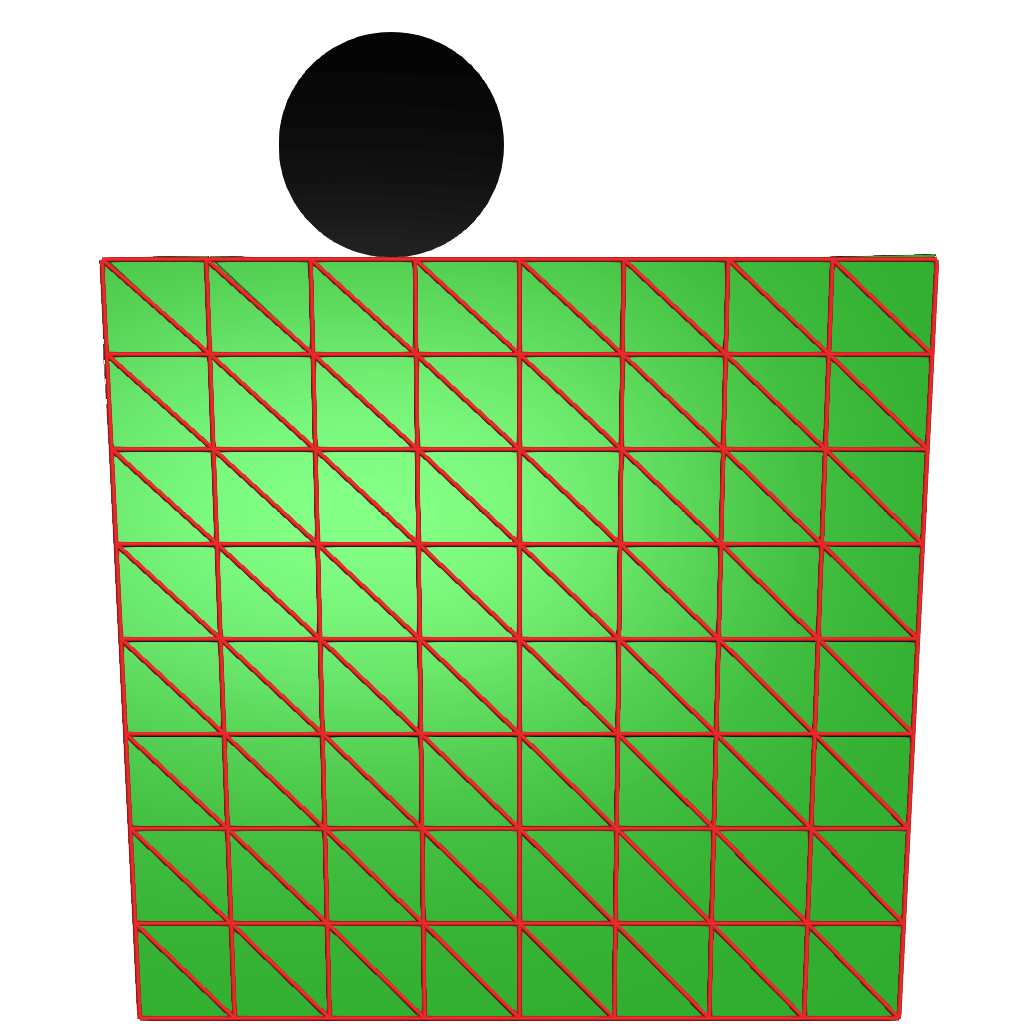}
    \end{minipage}
    \begin{minipage}{0.119\textwidth}
            \centering
            \includegraphics[width=\textwidth]{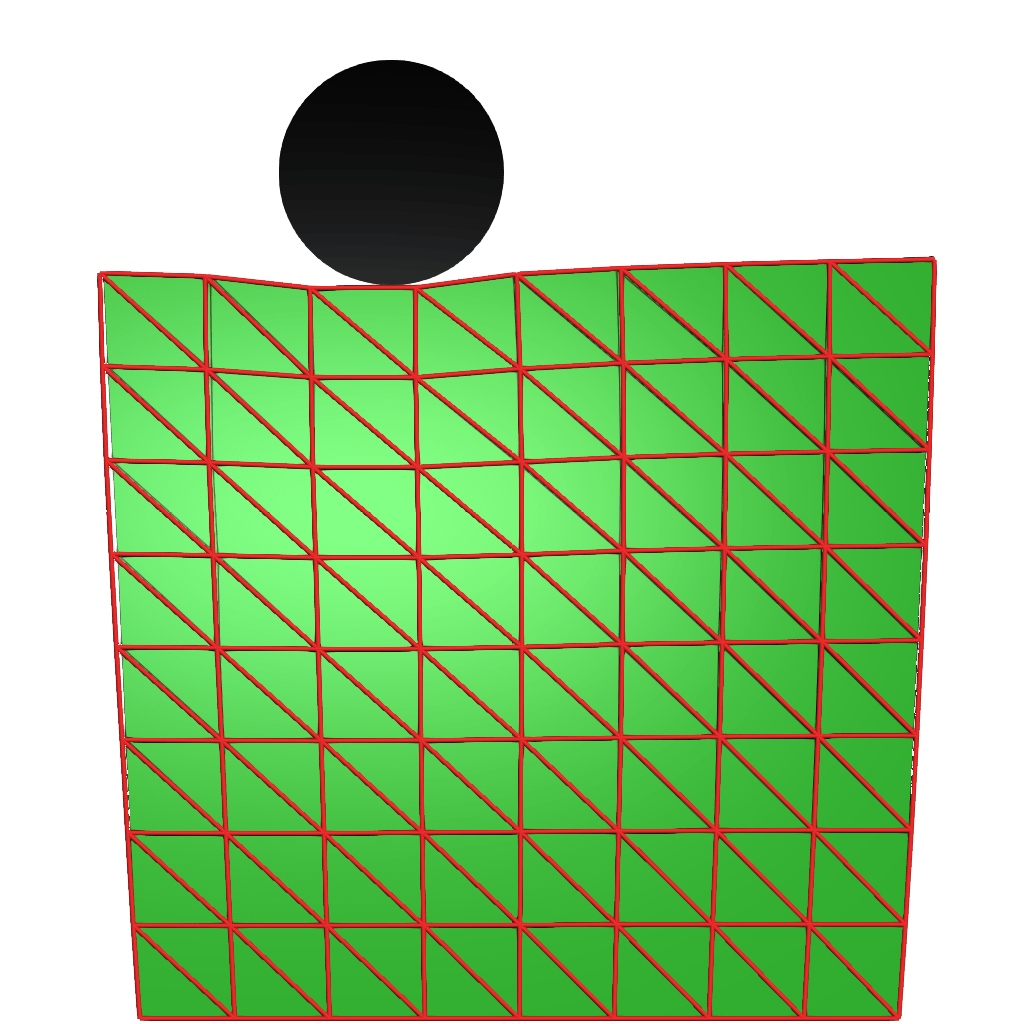}
    \end{minipage}
    \begin{minipage}{0.119\textwidth}
            \centering
            \includegraphics[width=\textwidth]{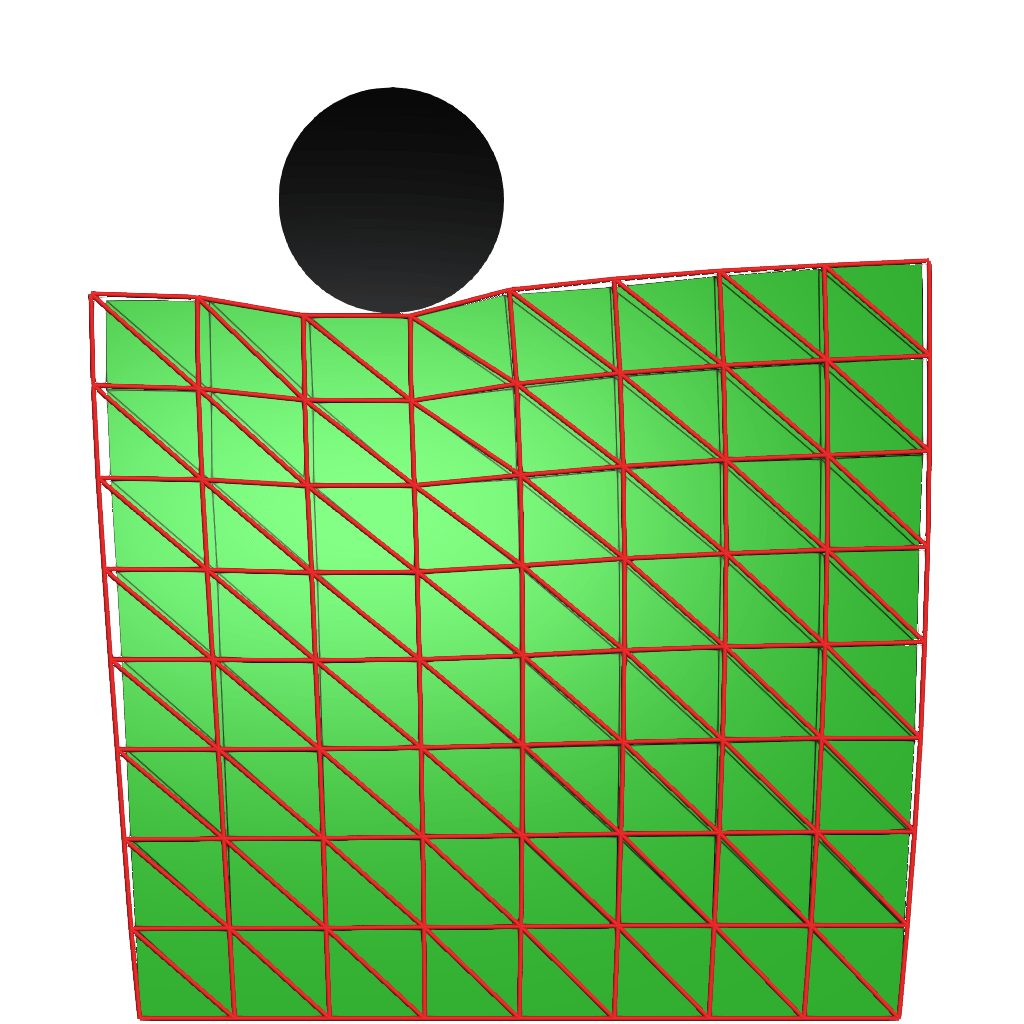}
    \end{minipage}
    \begin{minipage}{0.119\textwidth}
            \centering
            \includegraphics[width=\textwidth]{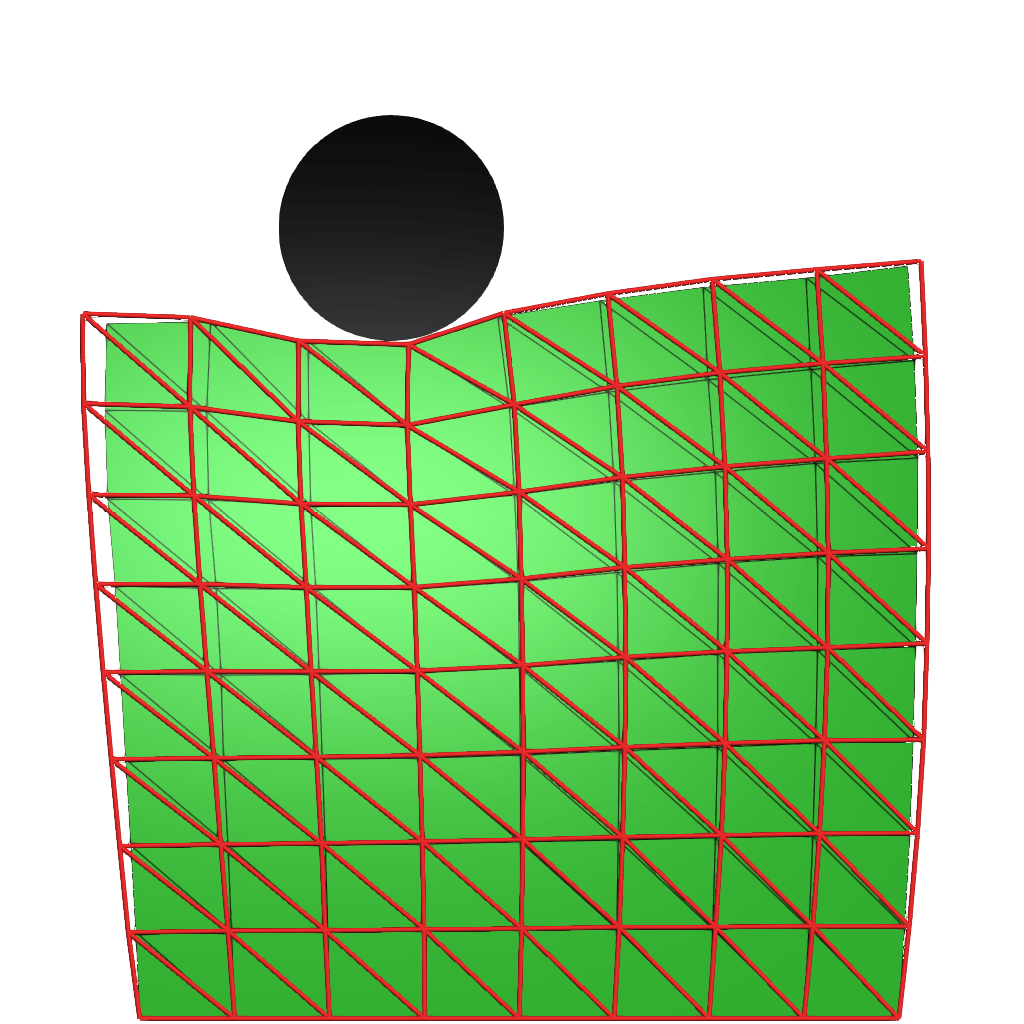}
    \end{minipage}
    \begin{minipage}{0.119\textwidth}
            \centering
            \includegraphics[width=\textwidth]{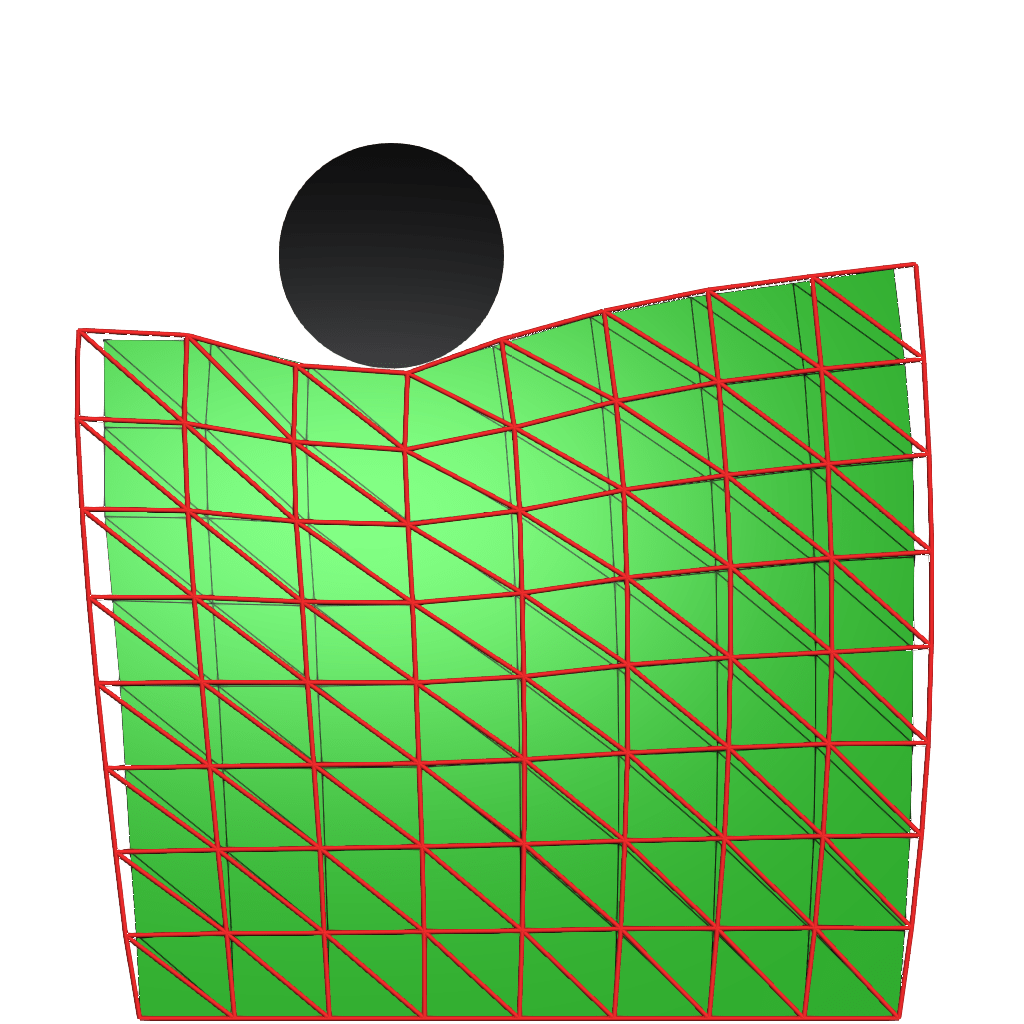}
    \end{minipage}
    \begin{minipage}{0.119\textwidth}
            \centering
            \includegraphics[width=\textwidth]{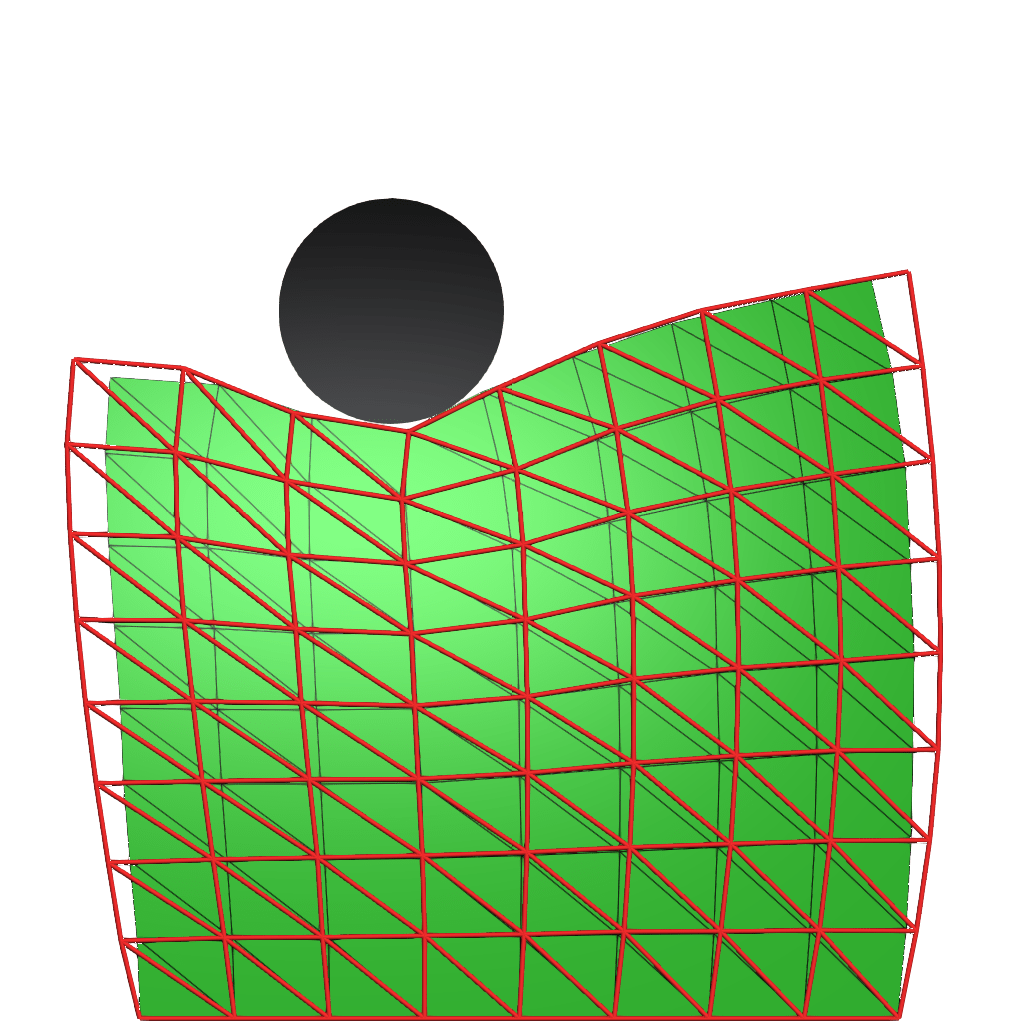}
    \end{minipage}
    \begin{minipage}{0.119\textwidth}
            \centering
            \includegraphics[width=\textwidth]{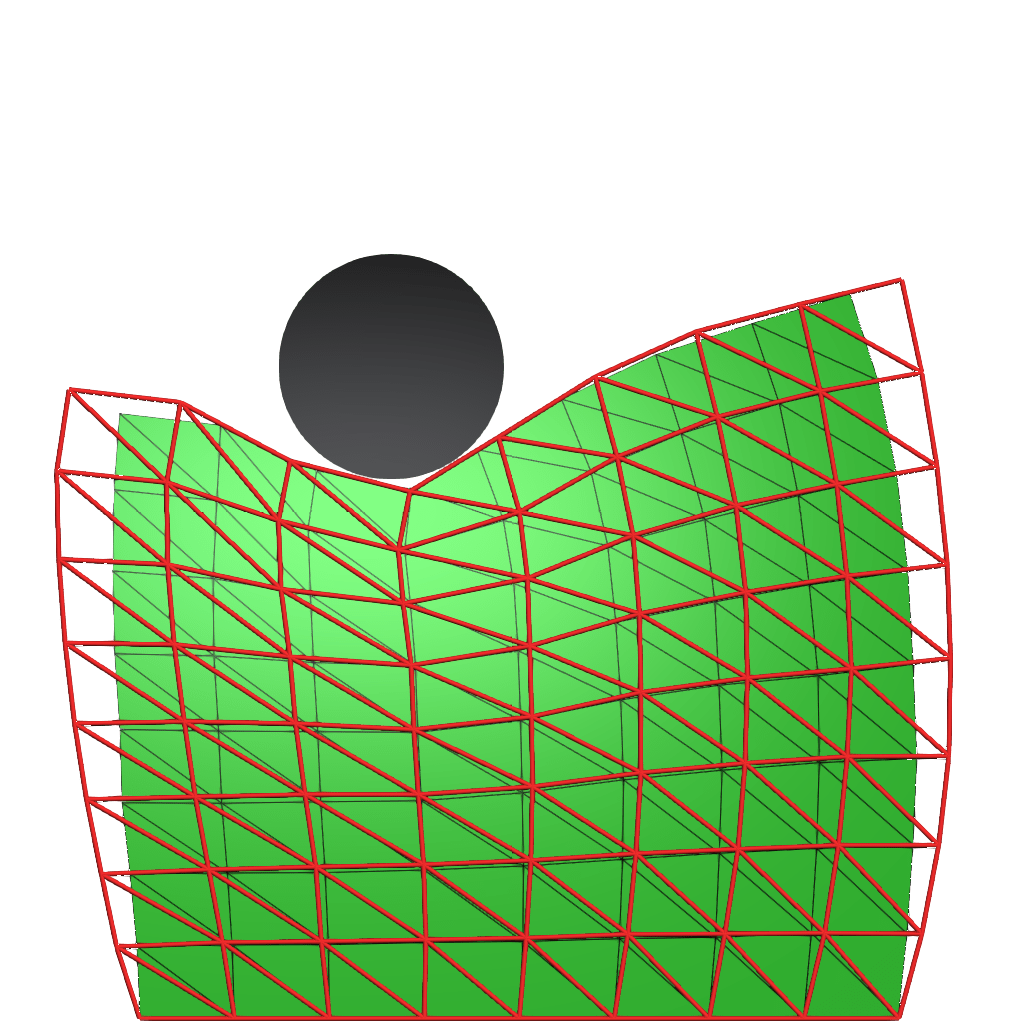}
    \end{minipage}
    \begin{minipage}{0.119\textwidth}
            \centering
            \includegraphics[width=\textwidth]{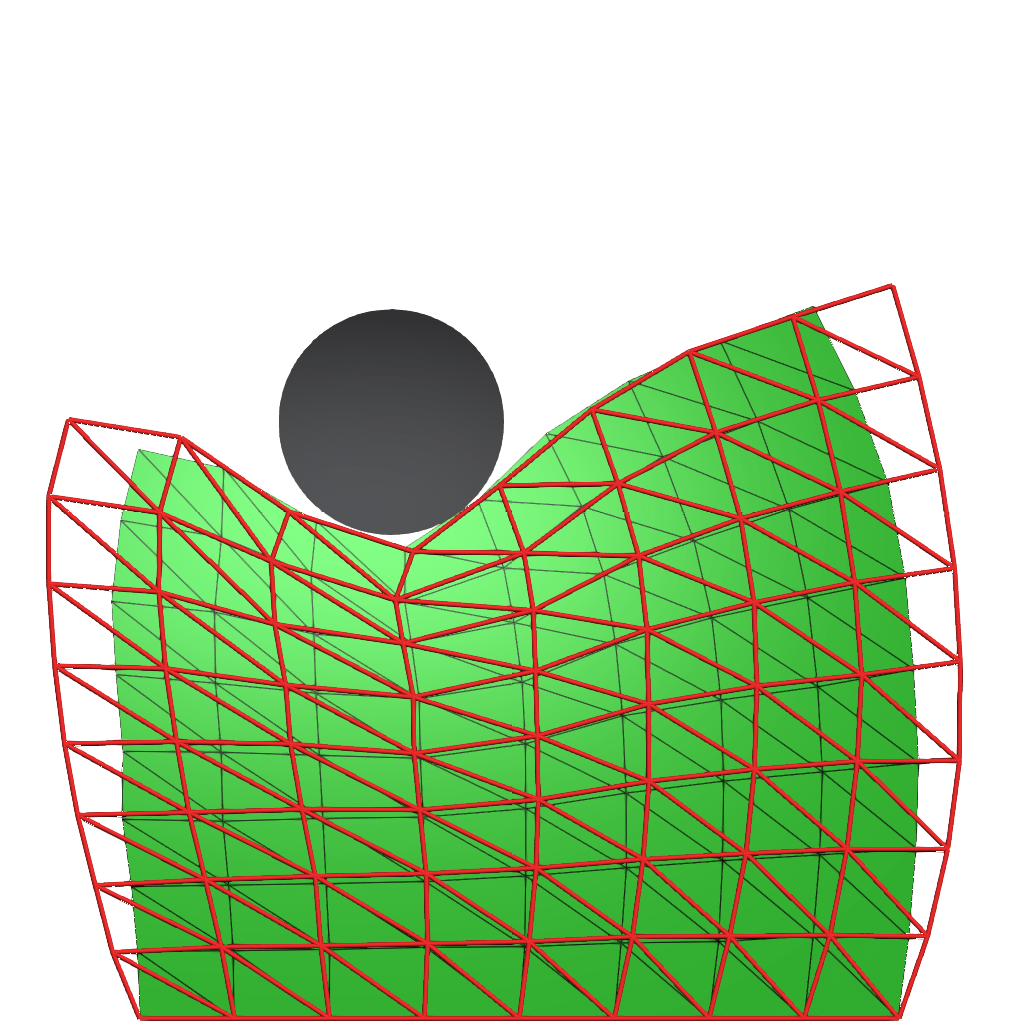}            
    \end{minipage}

    \vspace{1.0em} 
    \noindent\hrulefill 

    \vspace{-1.0em}
    \noindent\hrulefill 
    \vspace{0.5em} 

    \begin{minipage}{0.119\textwidth}
            \centering
            \includegraphics[width=\textwidth]{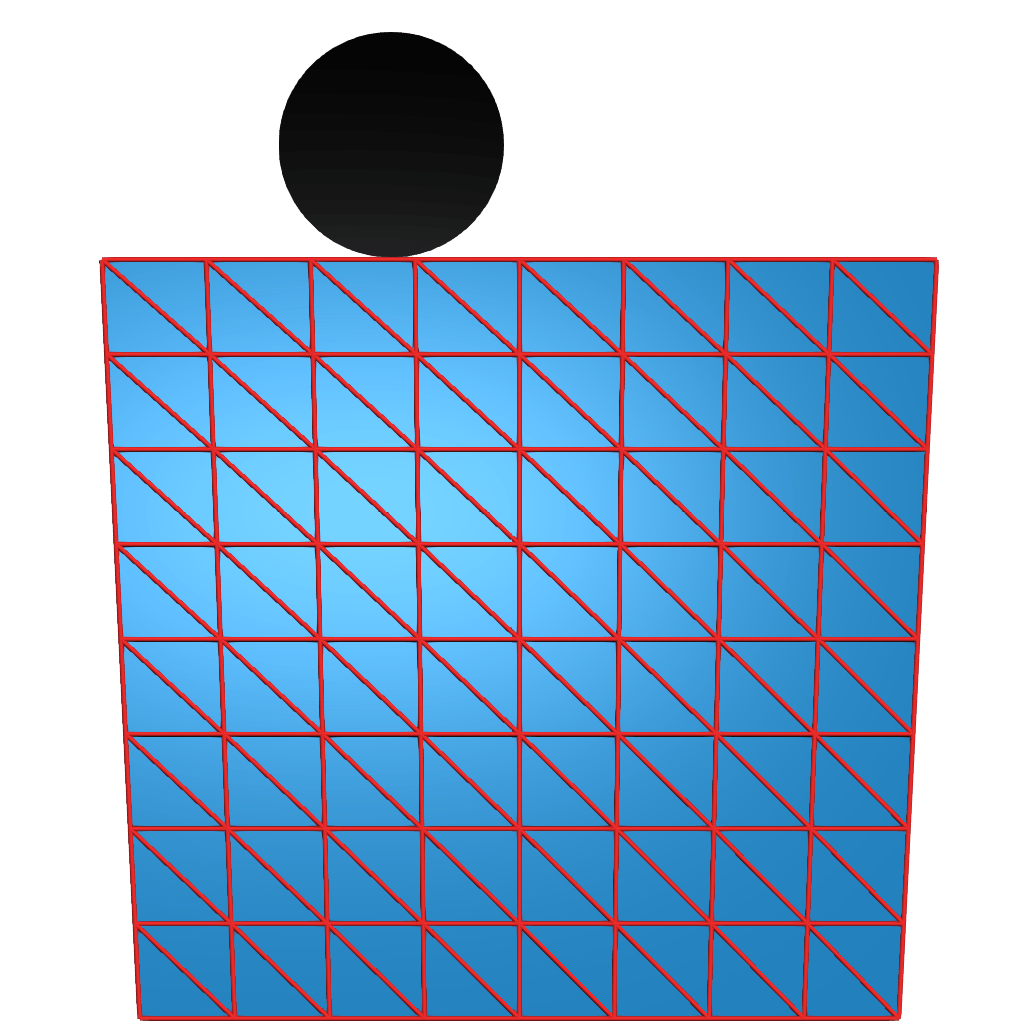}
            \subcaption*{$t=1$}
    \end{minipage}
    \begin{minipage}{0.119\textwidth}
            \centering
            \includegraphics[width=\textwidth]{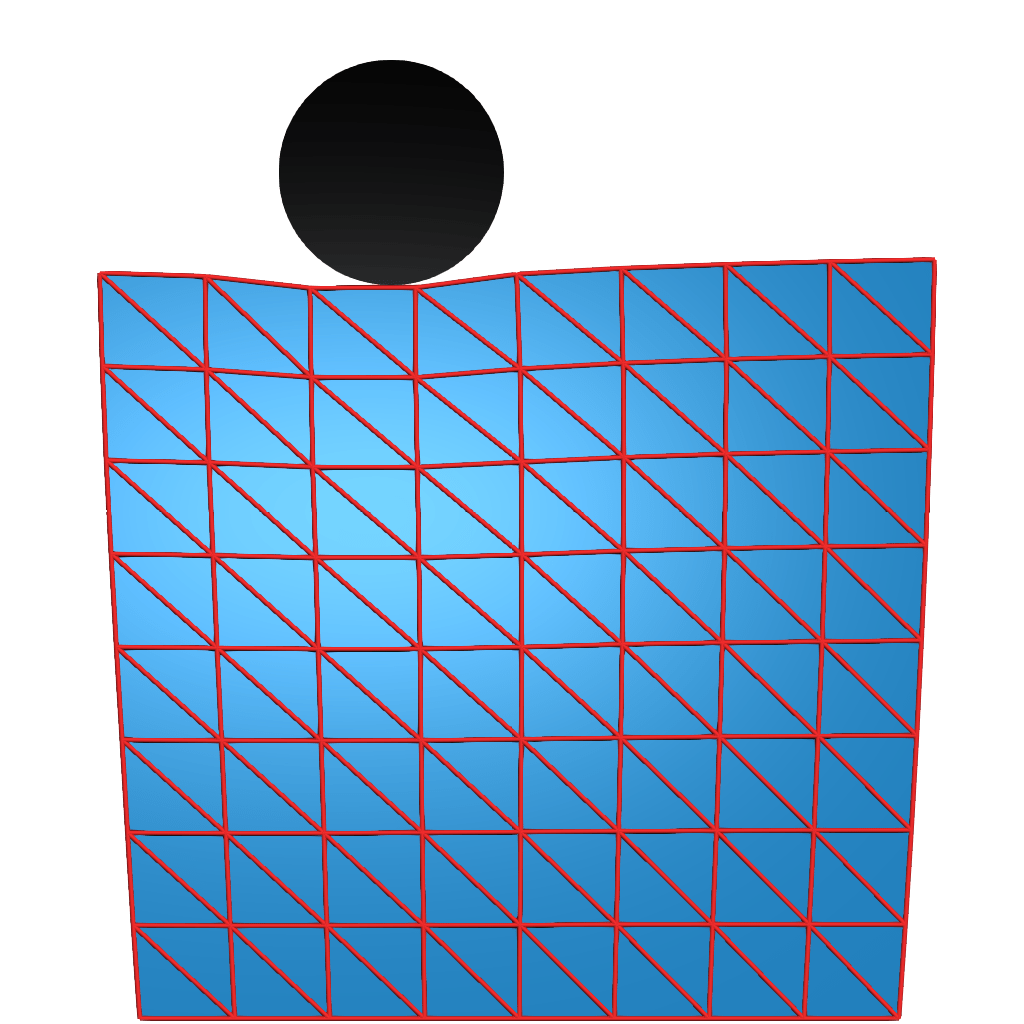}
            \subcaption*{$t=6$}
    \end{minipage}
    \begin{minipage}{0.119\textwidth}
            \centering
            \includegraphics[width=\textwidth]{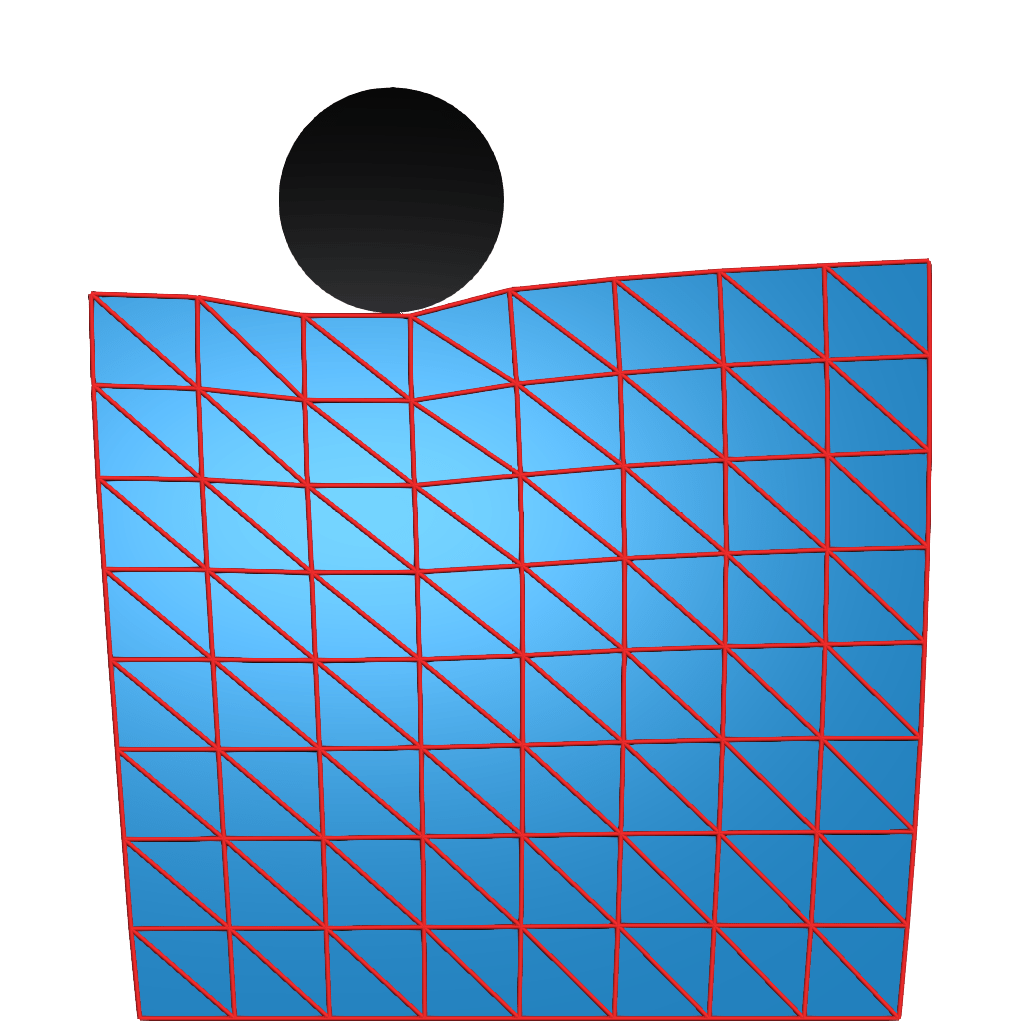}
            \subcaption*{$t=11$}
    \end{minipage}
    \begin{minipage}{0.119\textwidth}
            \centering
            \includegraphics[width=\textwidth]{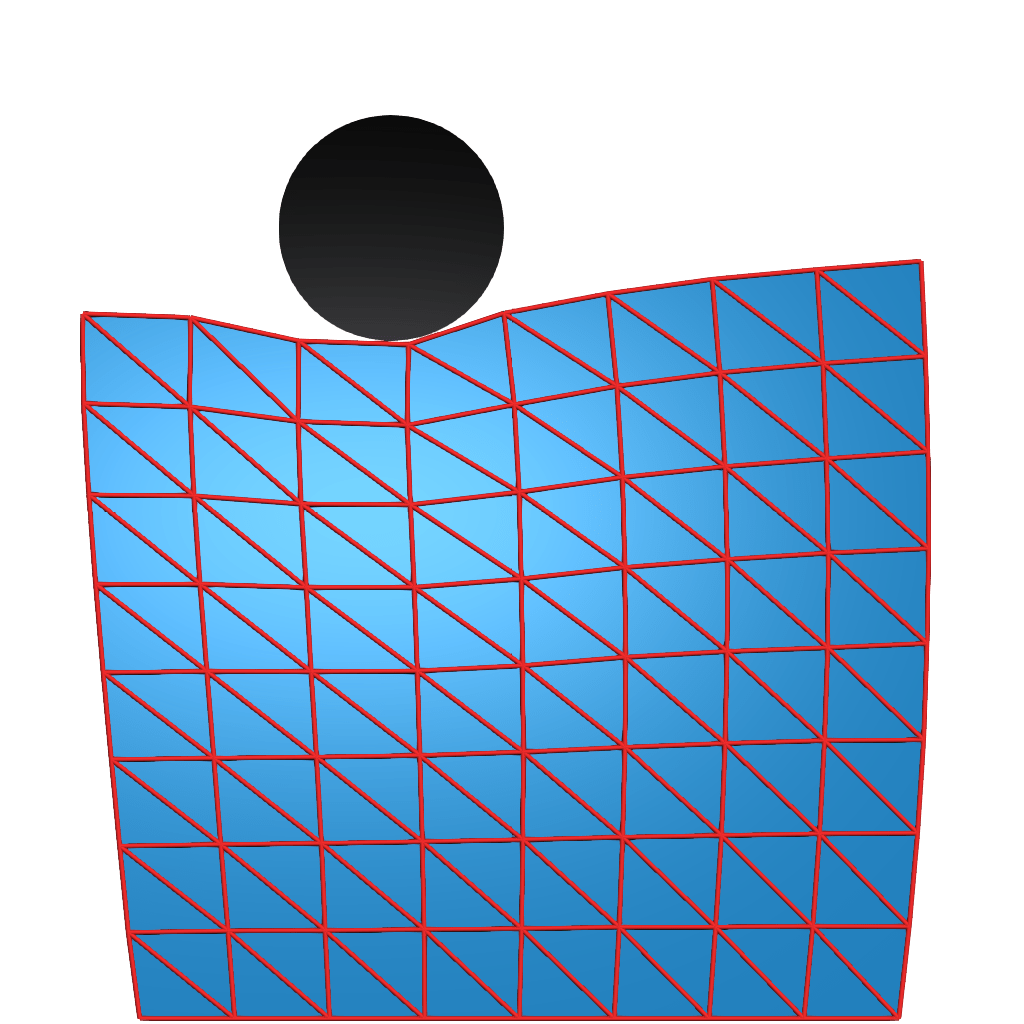}
            \subcaption*{$t=16$}
    \end{minipage}
    \begin{minipage}{0.119\textwidth}
            \centering
            \includegraphics[width=\textwidth]{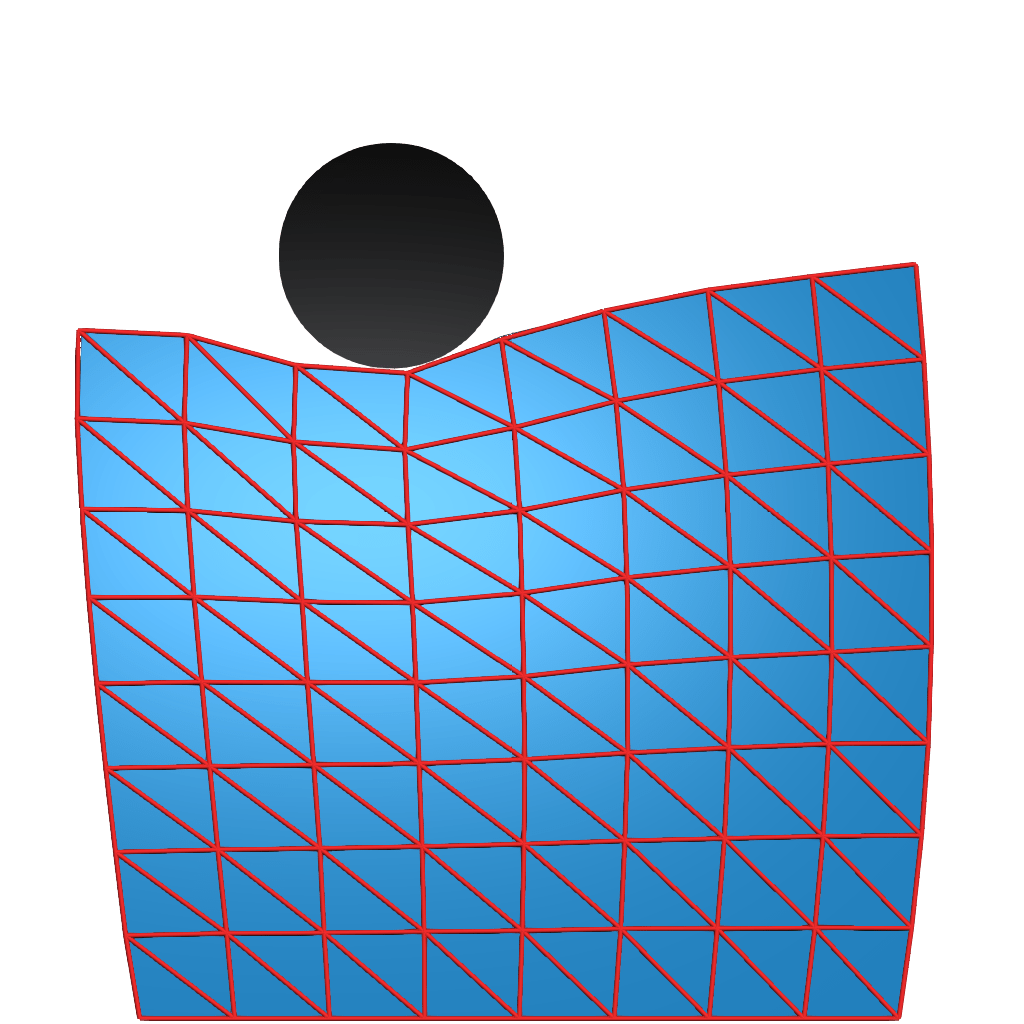}
            \subcaption*{$t=21$}
    \end{minipage}
    \begin{minipage}{0.119\textwidth}
            \centering
            \includegraphics[width=\textwidth]{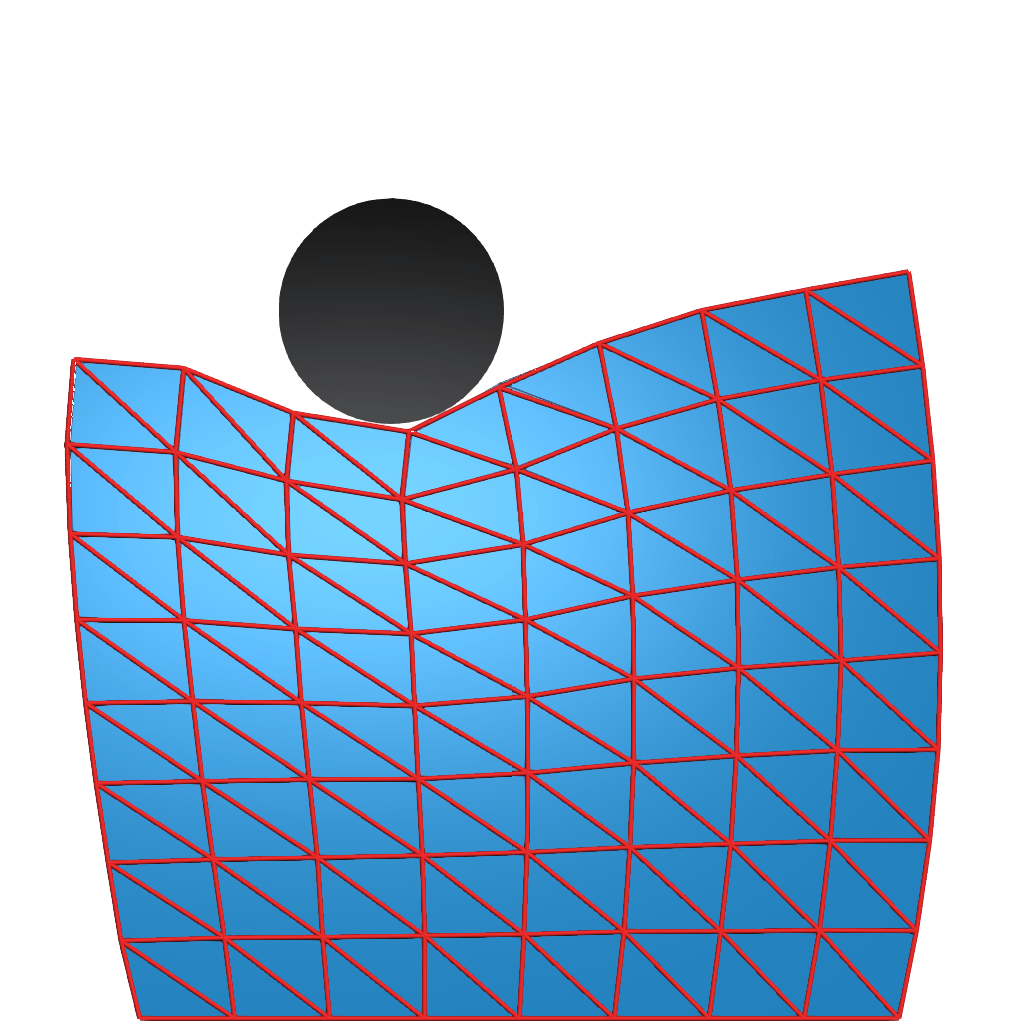}
            \subcaption*{$t=31$}
    \end{minipage}
    \begin{minipage}{0.119\textwidth}
            \centering
            \includegraphics[width=\textwidth]{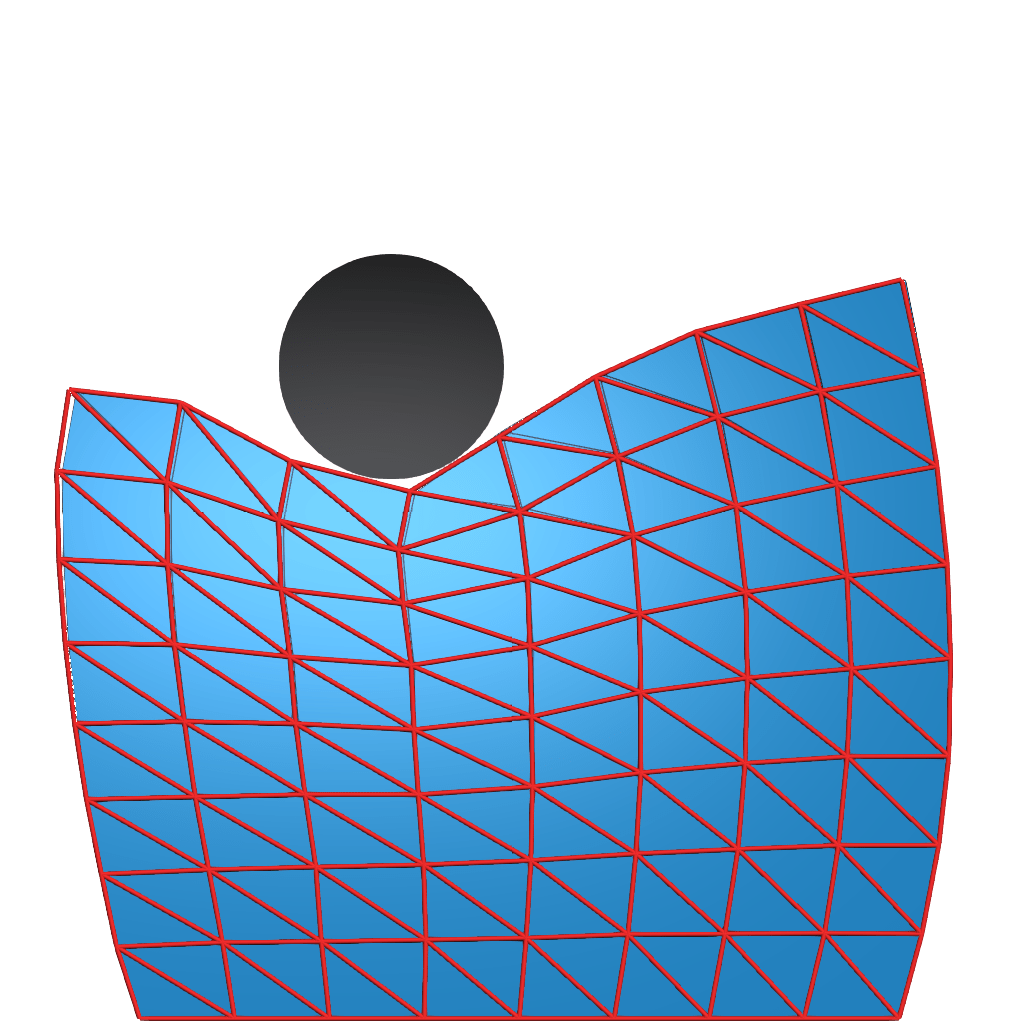}
            \subcaption*{$t=41$}
    \end{minipage}
    \begin{minipage}{0.119\textwidth}
            \centering
            \includegraphics[width=\textwidth]{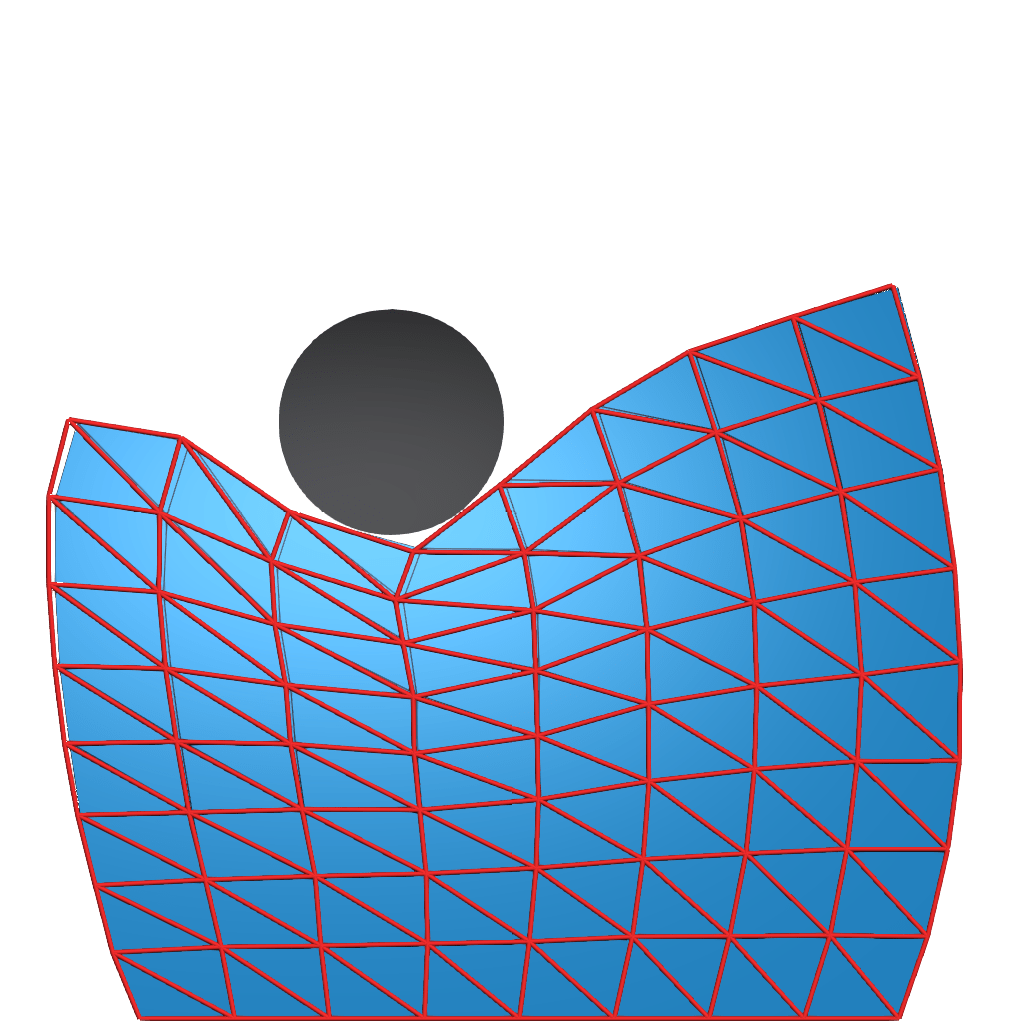}
            \subcaption*{$t=51$}
    \end{minipage}

    \vspace{0.01\textwidth}%

    \caption{
    Simulation over time of an exemplary test trajectory from the \textbf{DP-hard} task. The figure compares predictions from \textcolor{tabblue}{\textbf{MaNGO}}, \textcolor{taborange}{MGN}, and \textcolor{ForestGreen}{EGNO}. The last row, \textcolor{tabblue}{MaNGO-Oracle}, is separated by a horizontal line and represents predictions using oracle information. The \textbf{context set size} is set to $4$. All visualizations show the colored \textbf{predicted mesh}, with a \textbf{\textcolor{red}{wireframe}} representing the ground-truth simulation. \textcolor{tabblue}{\textbf{MaNGO}} accurately predicts the correct material properties, leading to a highly accurate simulation.
    }
    \label{fig:appendix_dp_hard}
\end{figure*}
\begin{figure*}[ht!]
    \centering
    \begin{minipage}{0.119\textwidth}
            \centering
            \includegraphics[width=\textwidth]{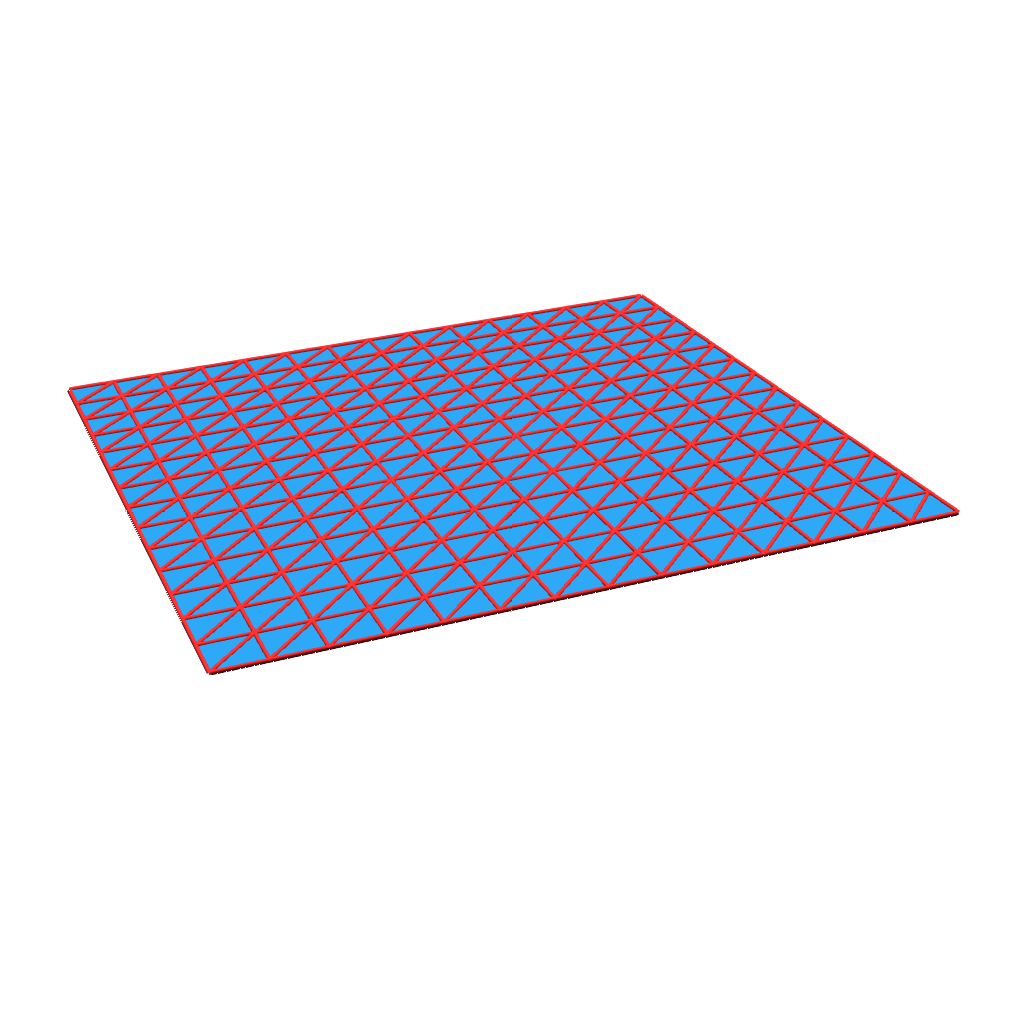}
    \end{minipage}
    \begin{minipage}{0.119\textwidth}
            \centering
            \includegraphics[width=\textwidth]{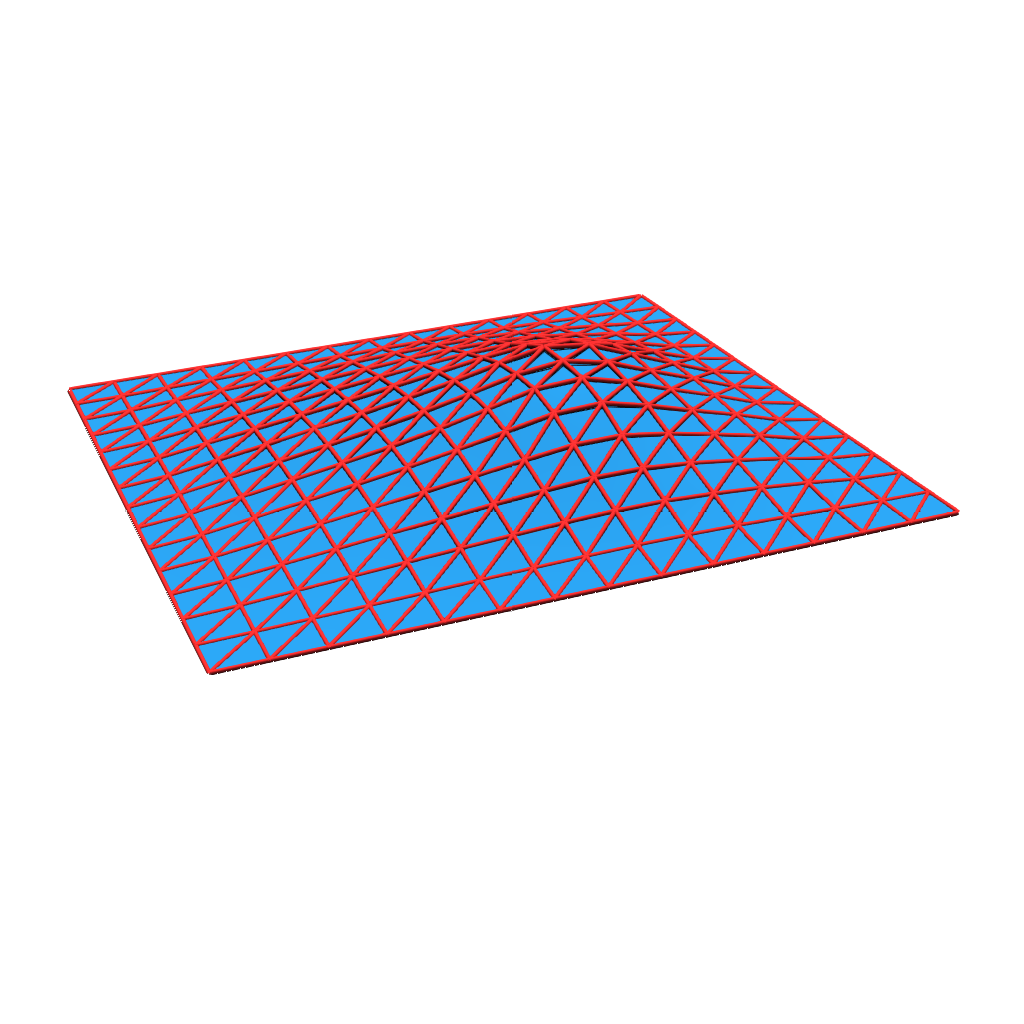}
    \end{minipage}
    \begin{minipage}{0.119\textwidth}
            \centering
            \includegraphics[width=\textwidth]{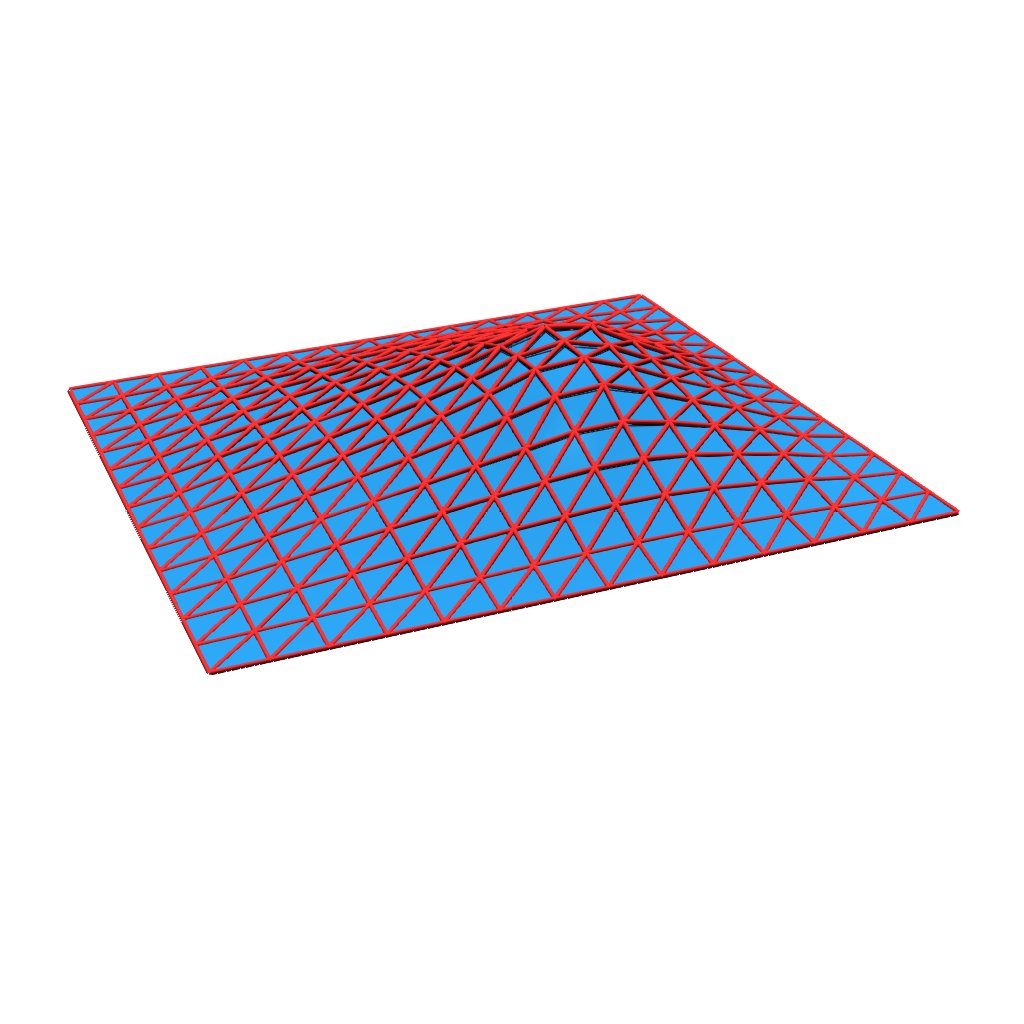}
    \end{minipage}
    \begin{minipage}{0.119\textwidth}
            \centering
            \includegraphics[width=\textwidth]{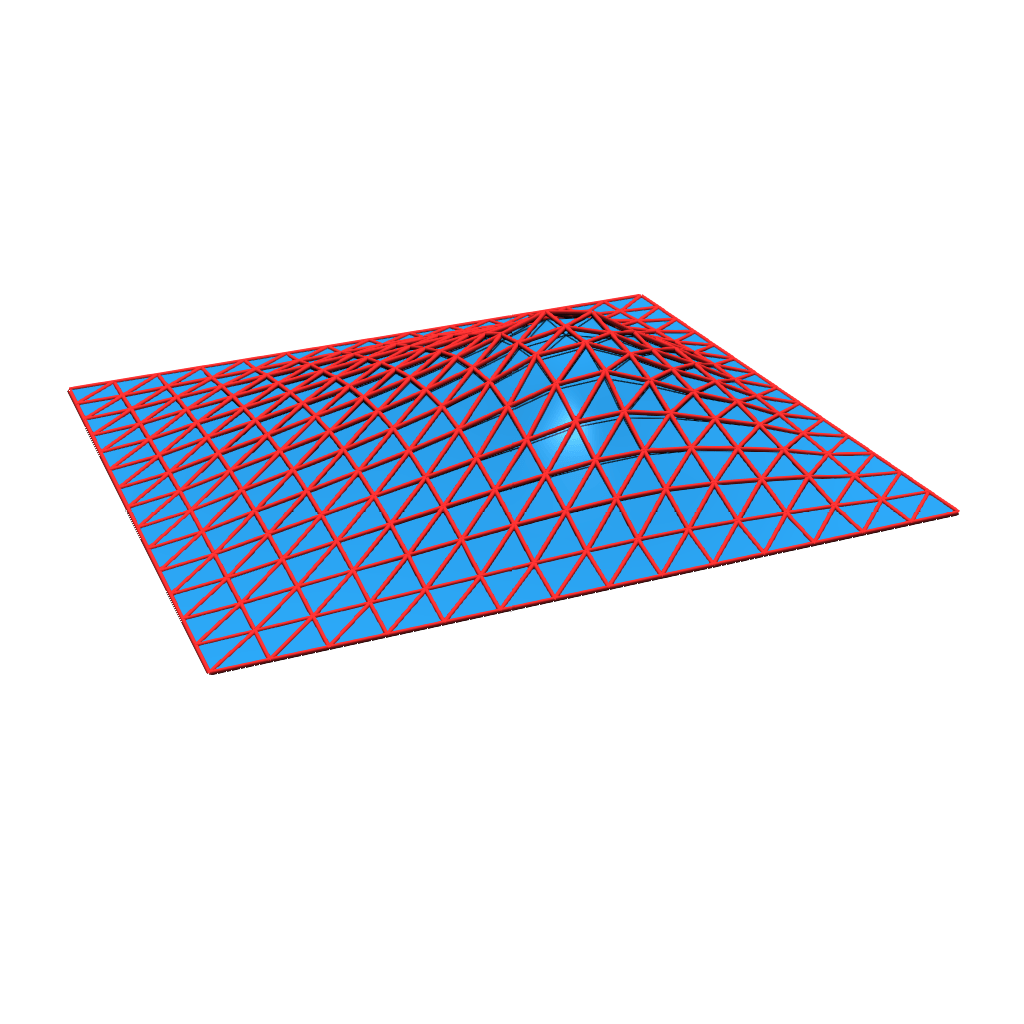}
    \end{minipage}
    \begin{minipage}{0.119\textwidth}
            \centering
            \includegraphics[width=\textwidth]{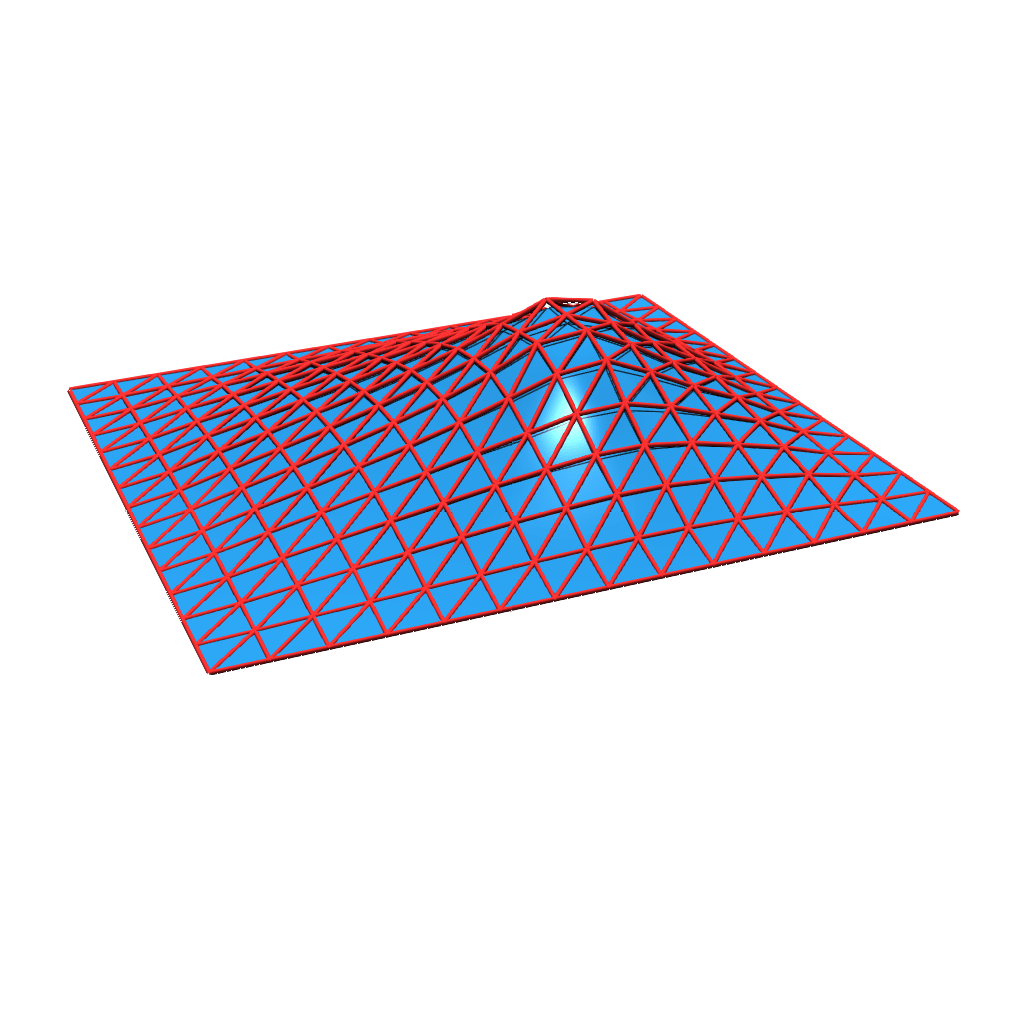}
    \end{minipage}
    \begin{minipage}{0.119\textwidth}
            \centering
            \includegraphics[width=\textwidth]{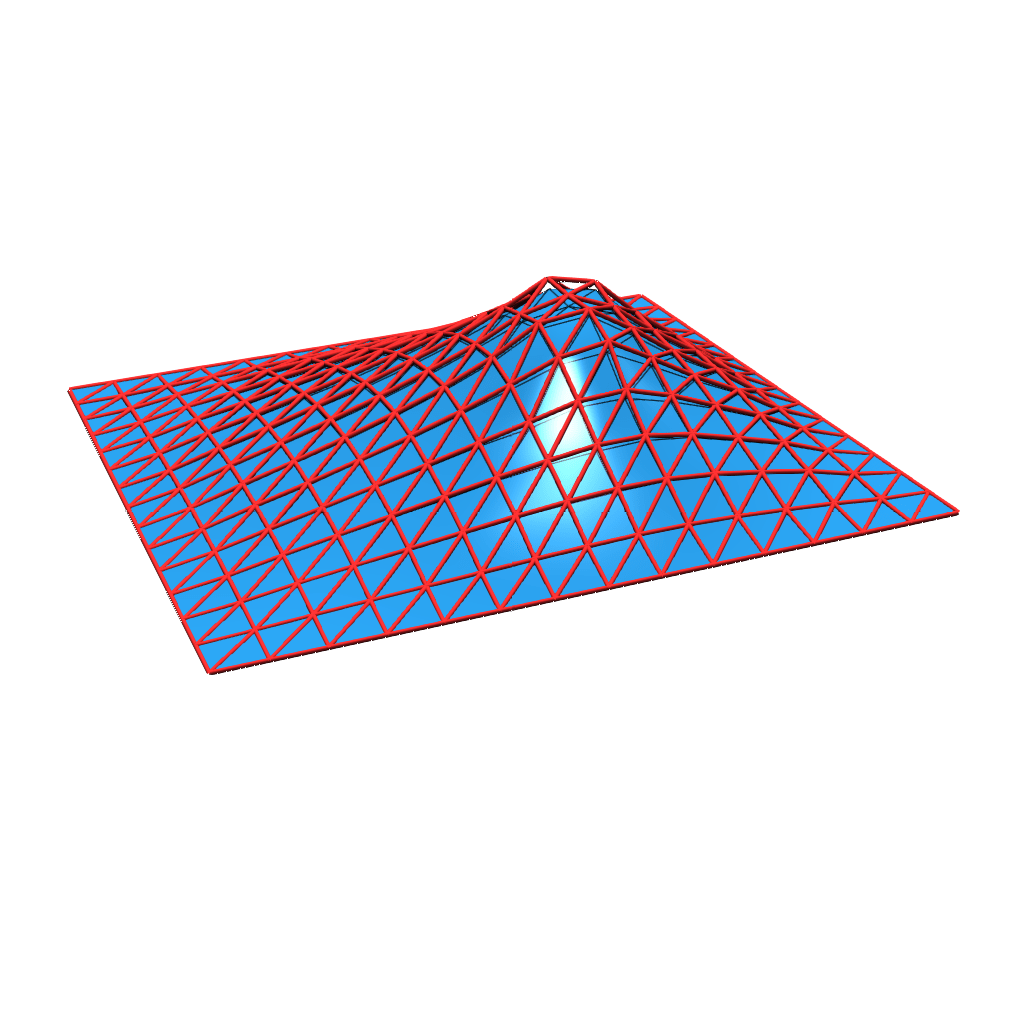}
    \end{minipage}
    \begin{minipage}{0.119\textwidth}
            \centering
            \includegraphics[width=\textwidth]{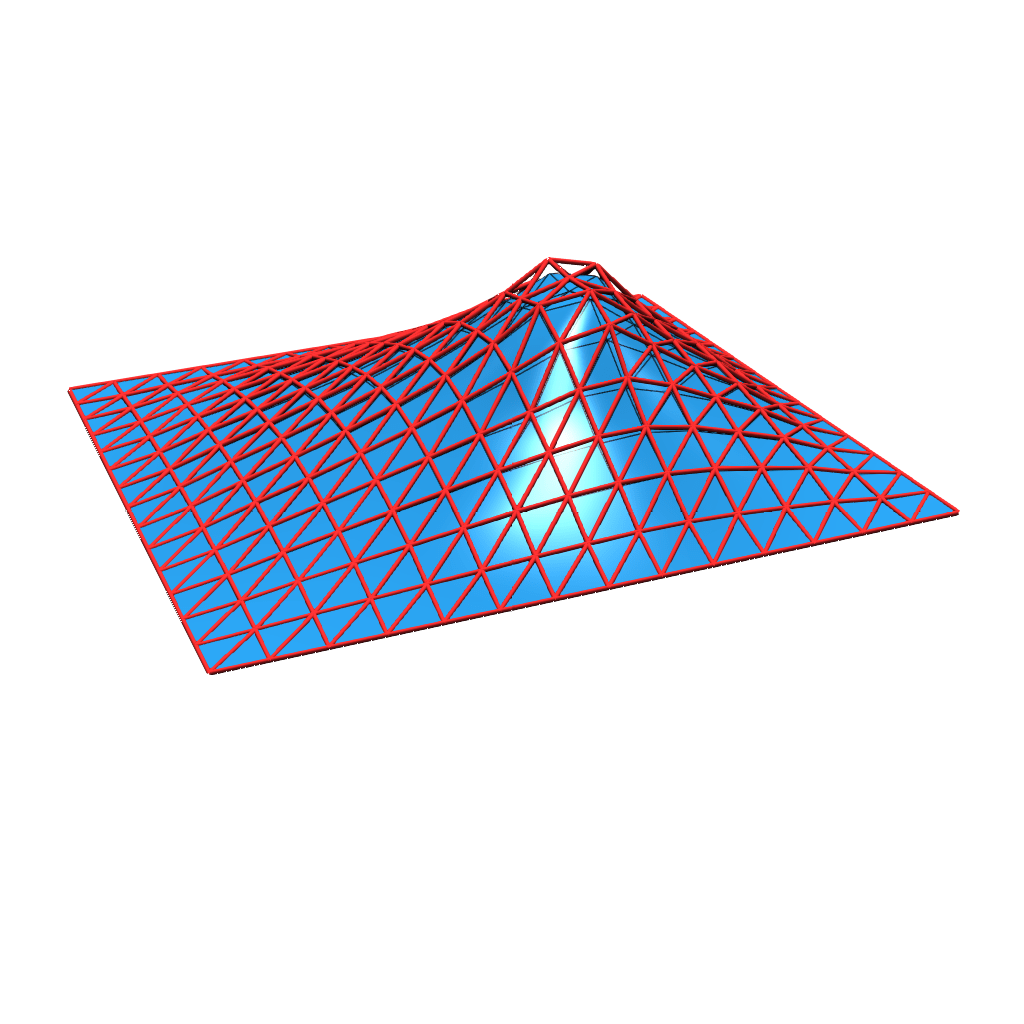}
    \end{minipage}
    \begin{minipage}{0.119\textwidth}
            \centering
            \includegraphics[width=\textwidth]{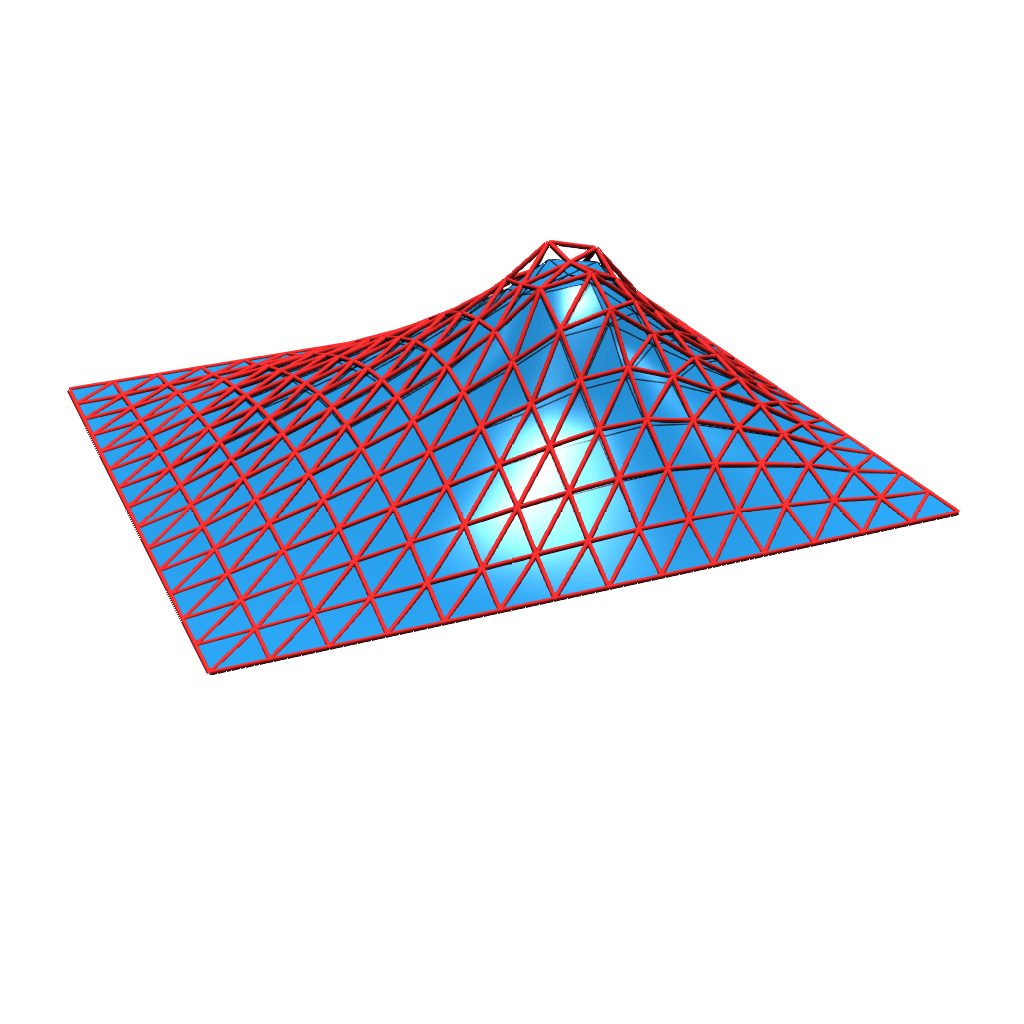}
    \end{minipage}

    \begin{minipage}{0.119\textwidth}
            \centering
            \includegraphics[width=\textwidth]{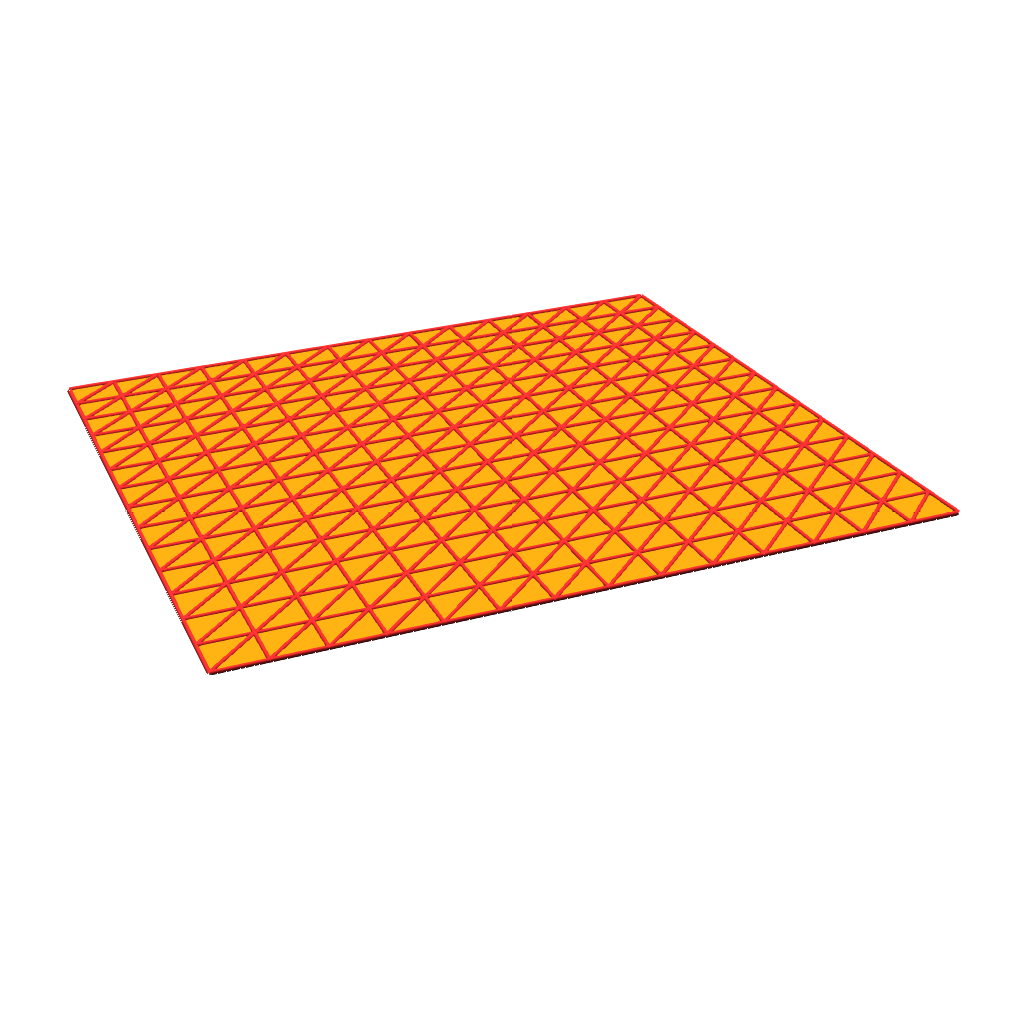}
    \end{minipage}
    \begin{minipage}{0.119\textwidth}
            \centering
            \includegraphics[width=\textwidth]{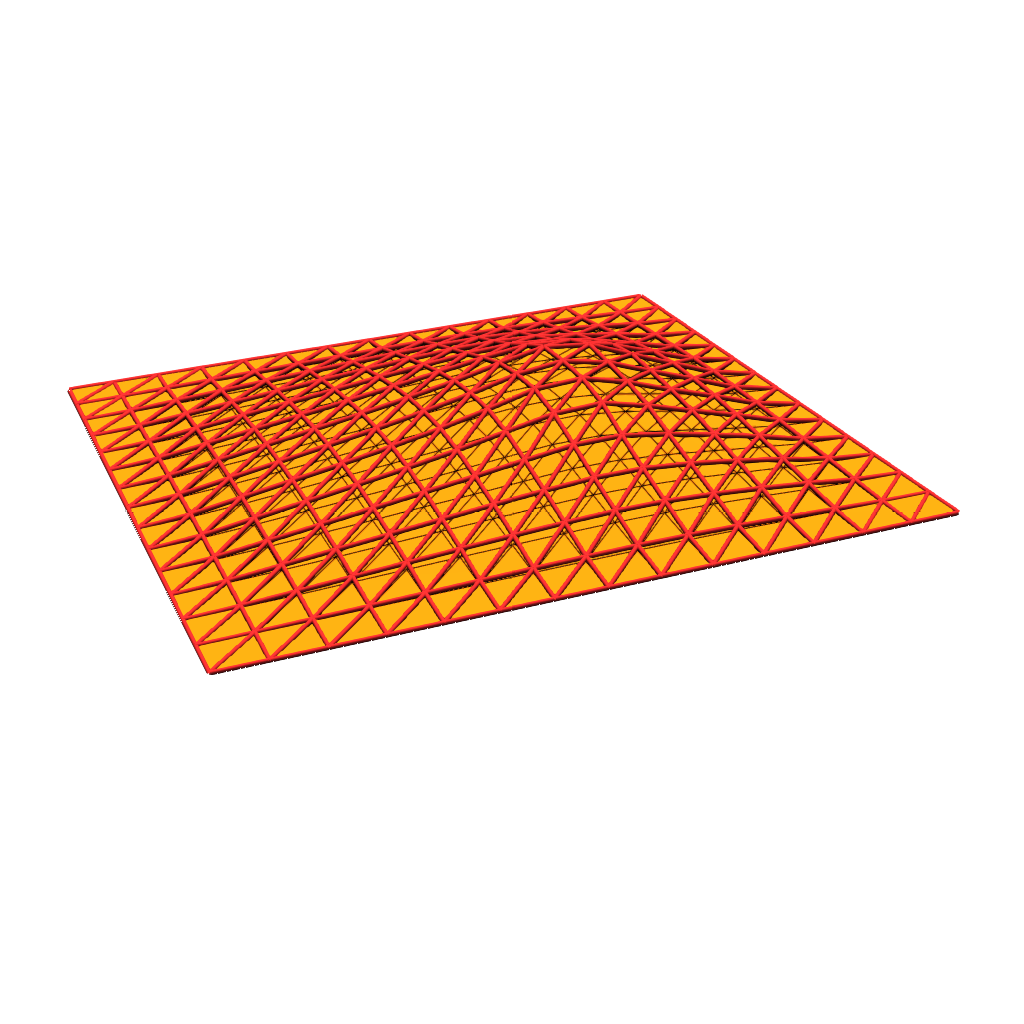}
    \end{minipage}
    \begin{minipage}{0.119\textwidth}
            \centering
            \includegraphics[width=\textwidth]{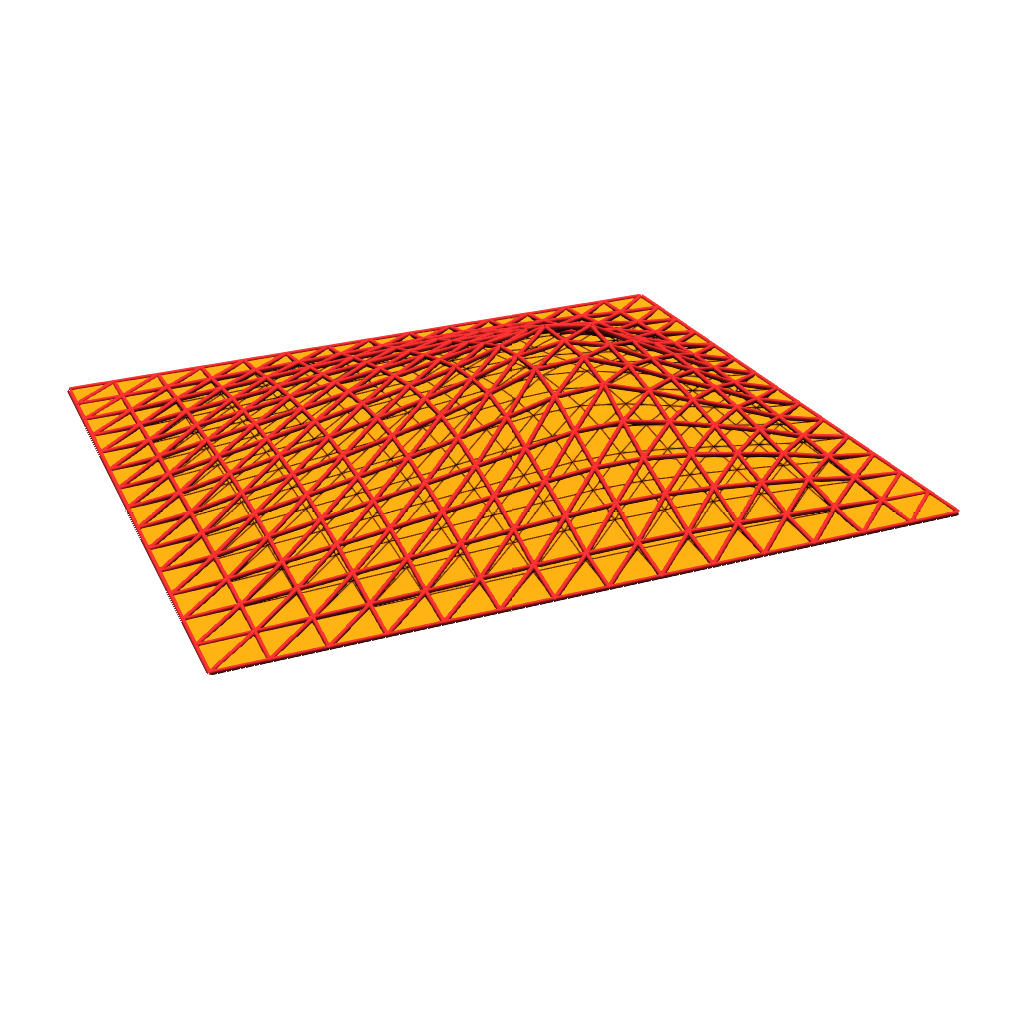}
    \end{minipage}
    \begin{minipage}{0.119\textwidth}
            \centering
            \includegraphics[width=\textwidth]{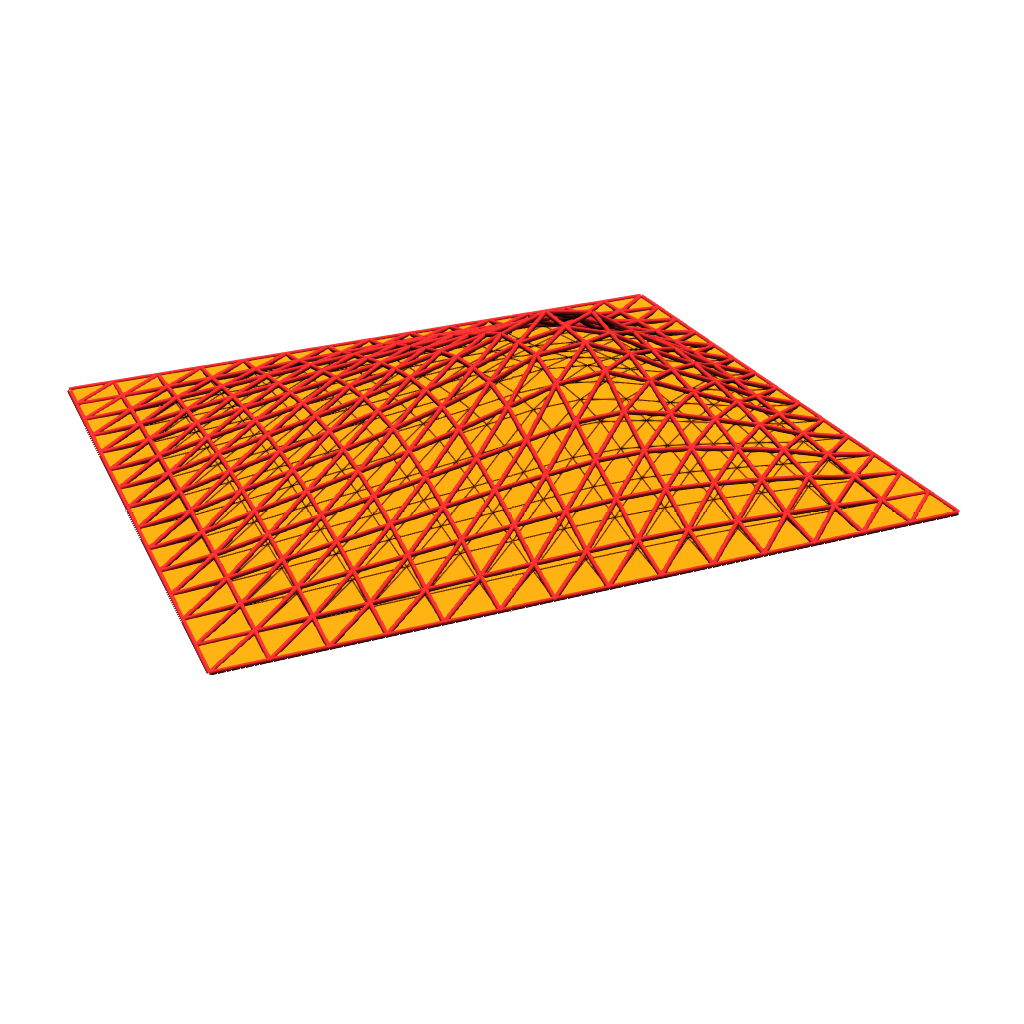}
    \end{minipage}
    \begin{minipage}{0.119\textwidth}
            \centering
            \includegraphics[width=\textwidth]{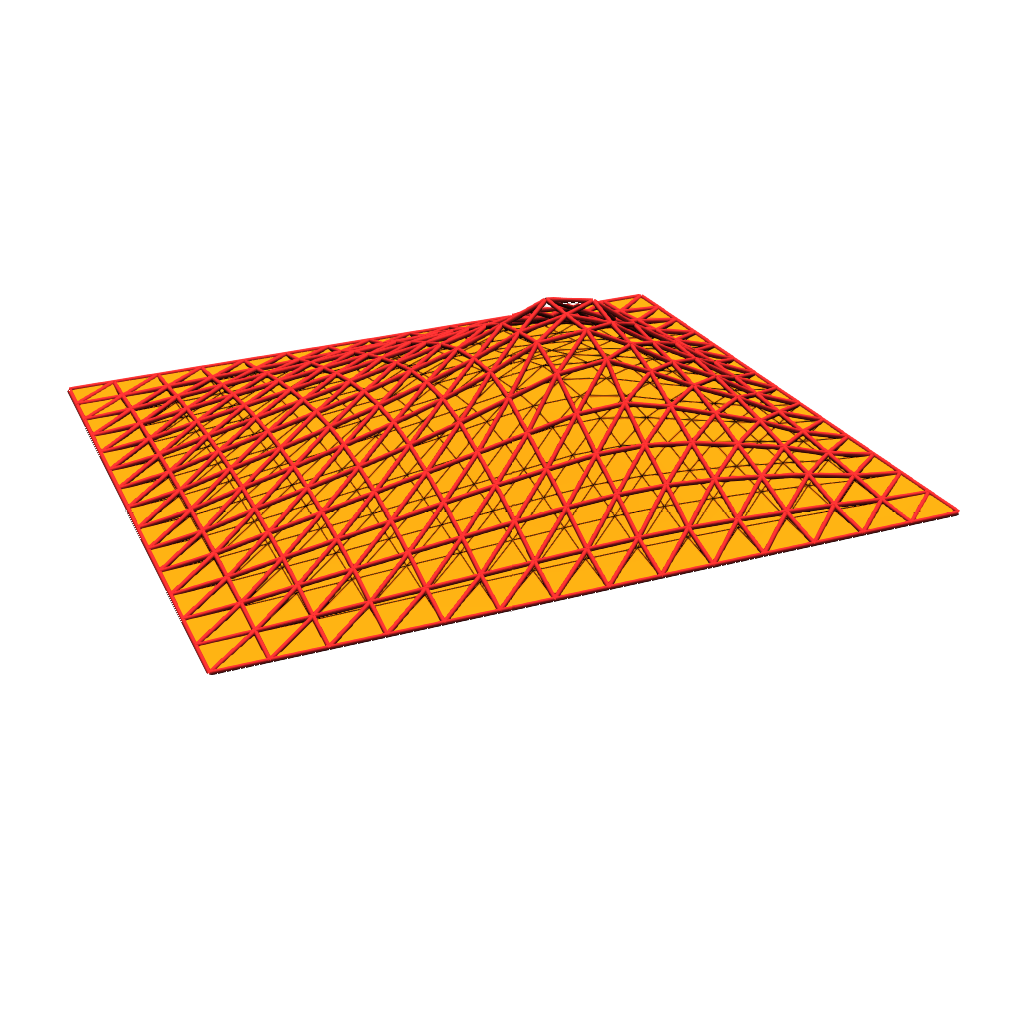}
    \end{minipage}
    \begin{minipage}{0.119\textwidth}
            \centering
            \includegraphics[width=\textwidth]{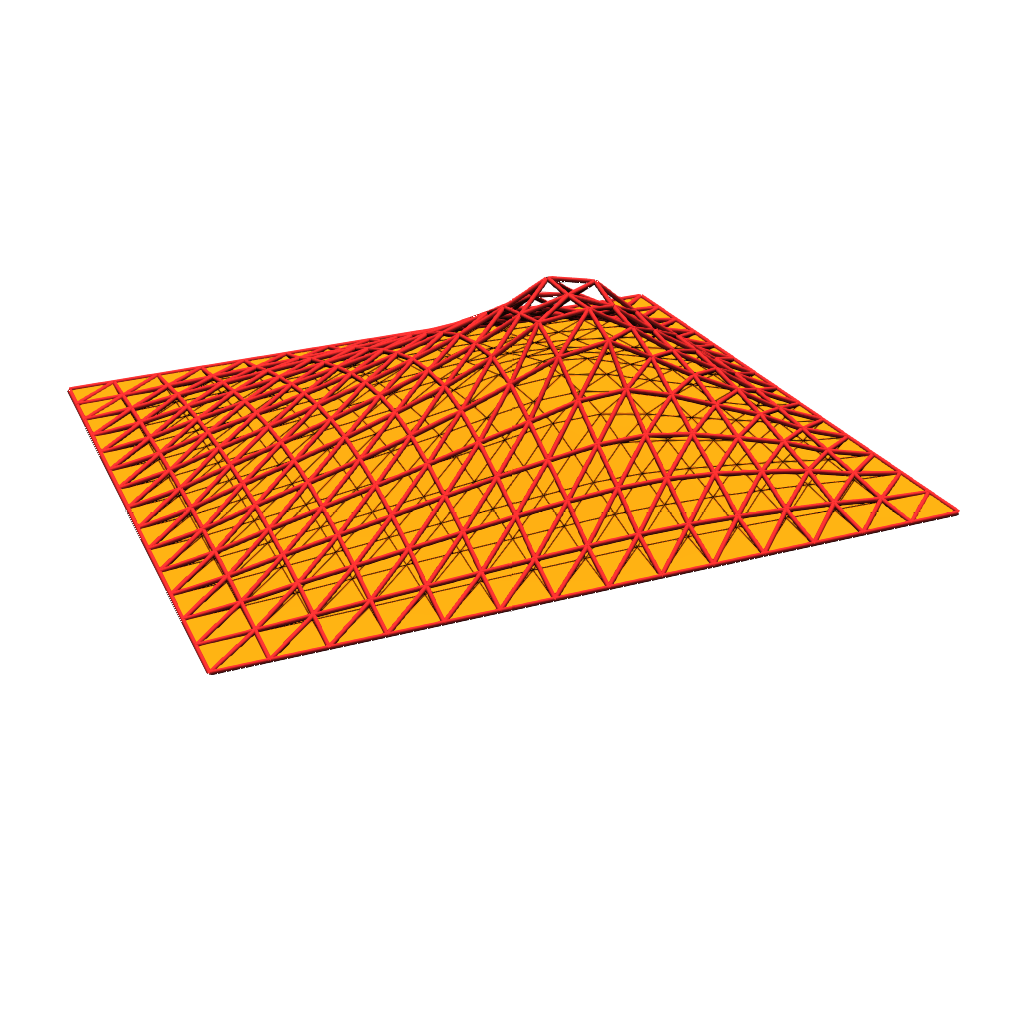}
    \end{minipage}
    \begin{minipage}{0.119\textwidth}
            \centering
            \includegraphics[width=\textwidth]{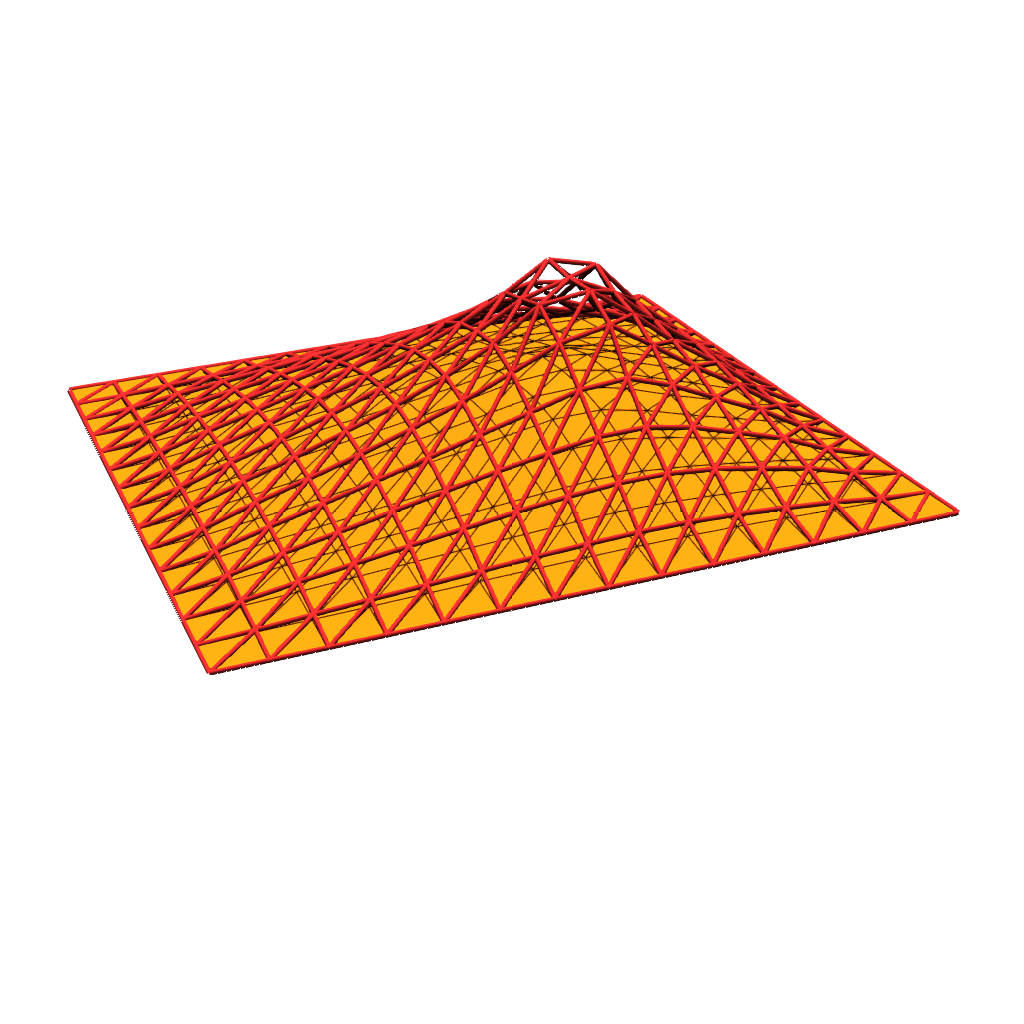}
    \end{minipage}
    \begin{minipage}{0.119\textwidth}
            \centering
            \includegraphics[width=\textwidth]{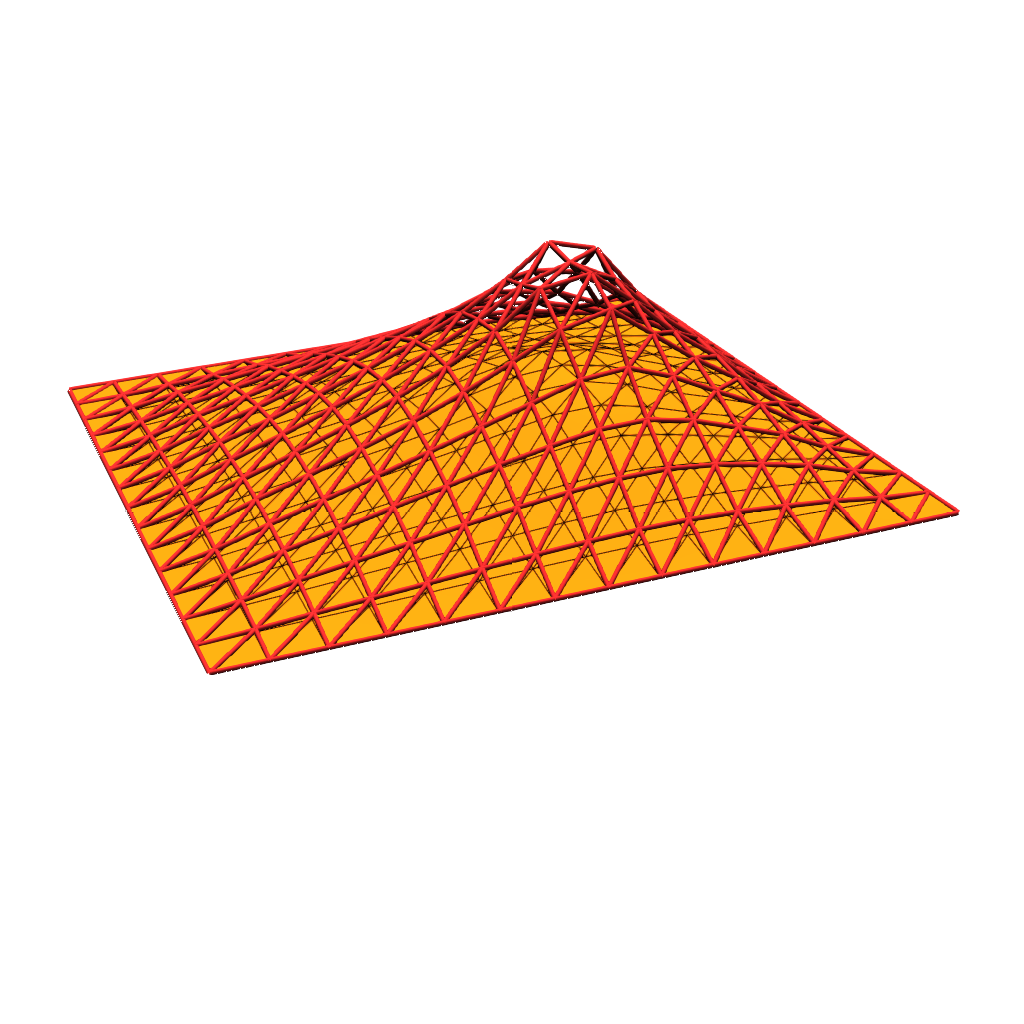}
    \end{minipage}

    \begin{minipage}{0.119\textwidth}
            \centering
            \includegraphics[width=\textwidth]{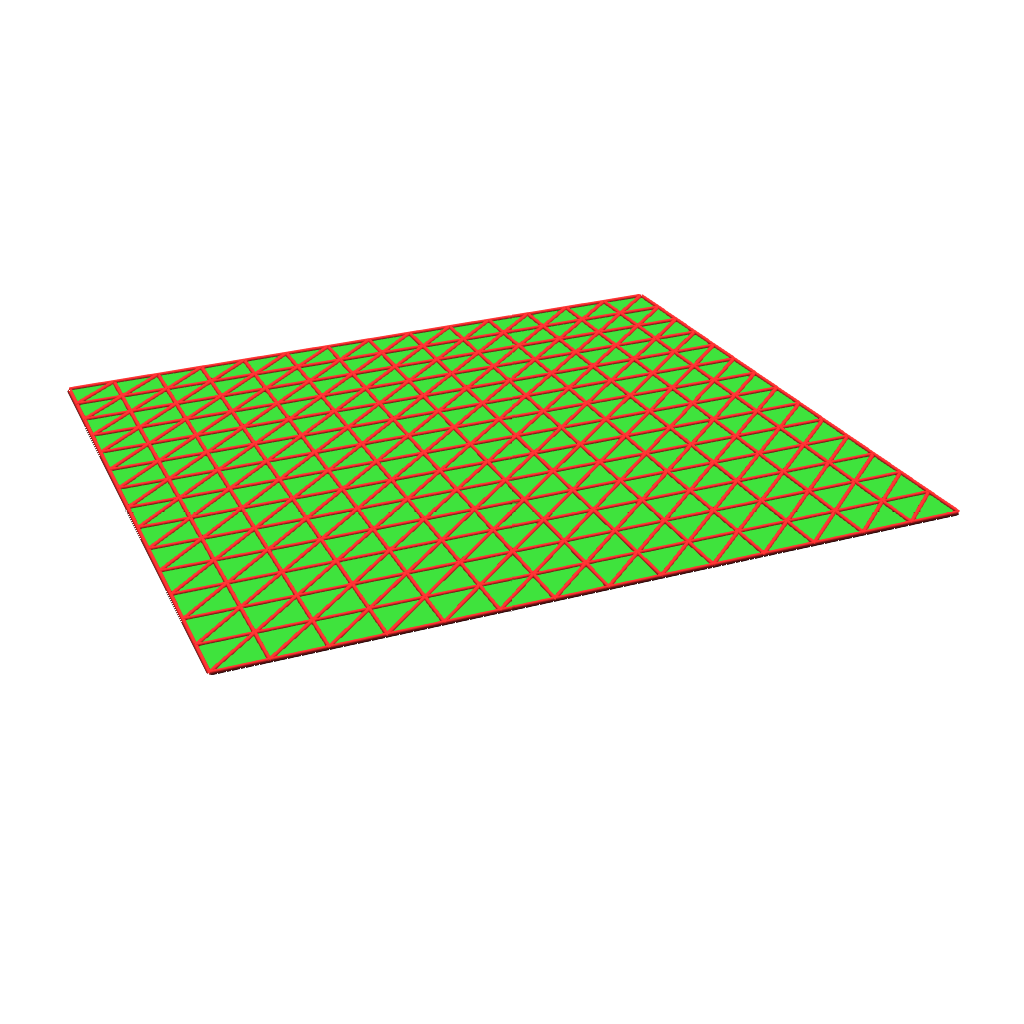}
    \end{minipage}
    \begin{minipage}{0.119\textwidth}
            \centering
            \includegraphics[width=\textwidth]{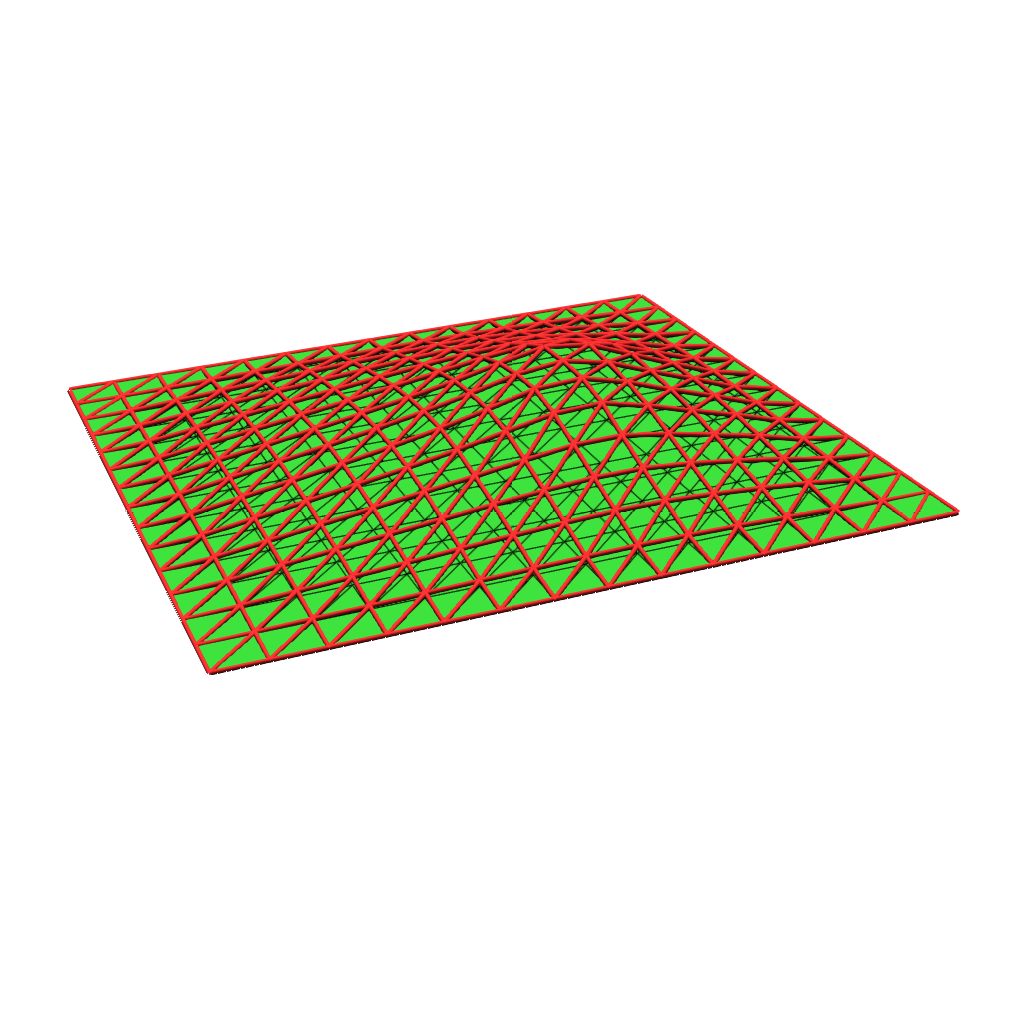}
    \end{minipage}
    \begin{minipage}{0.119\textwidth}
            \centering
            \includegraphics[width=\textwidth]{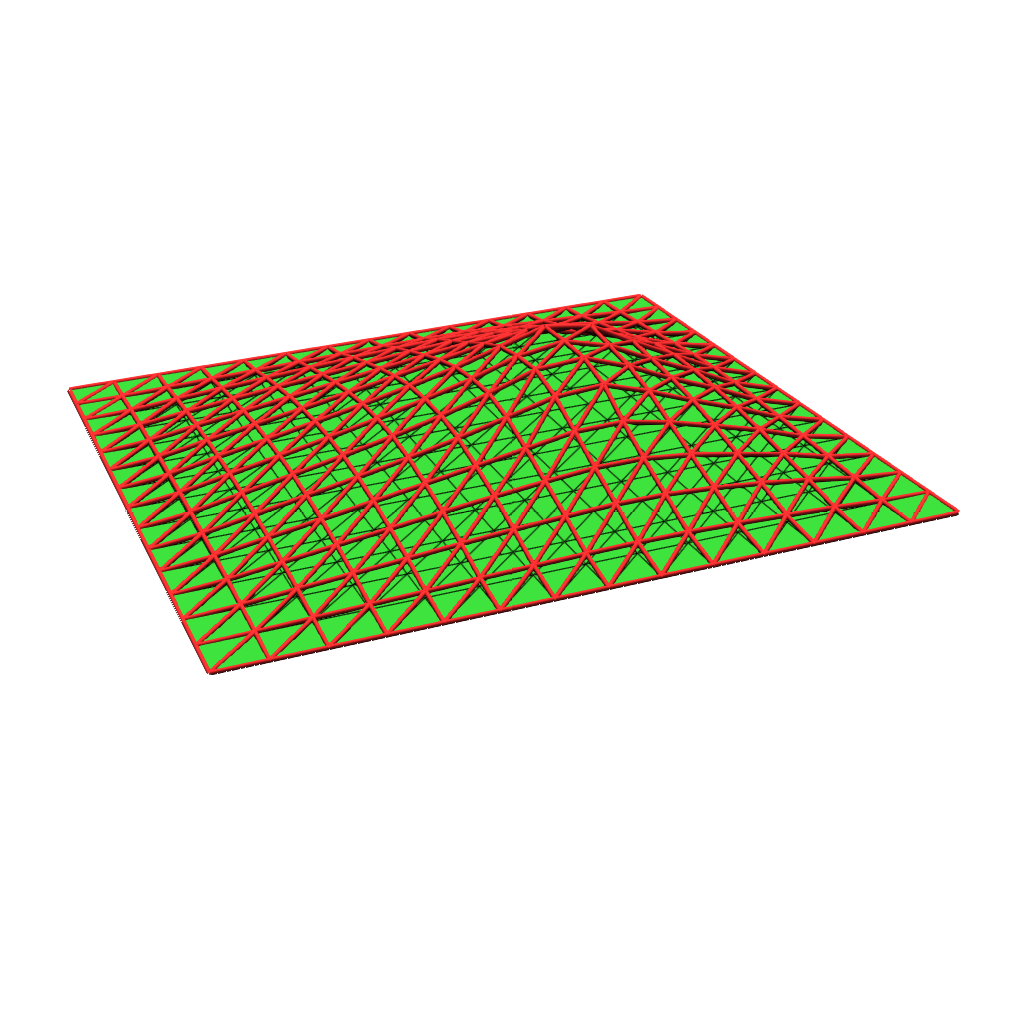}
    \end{minipage}
    \begin{minipage}{0.119\textwidth}
            \centering
            \includegraphics[width=\textwidth]{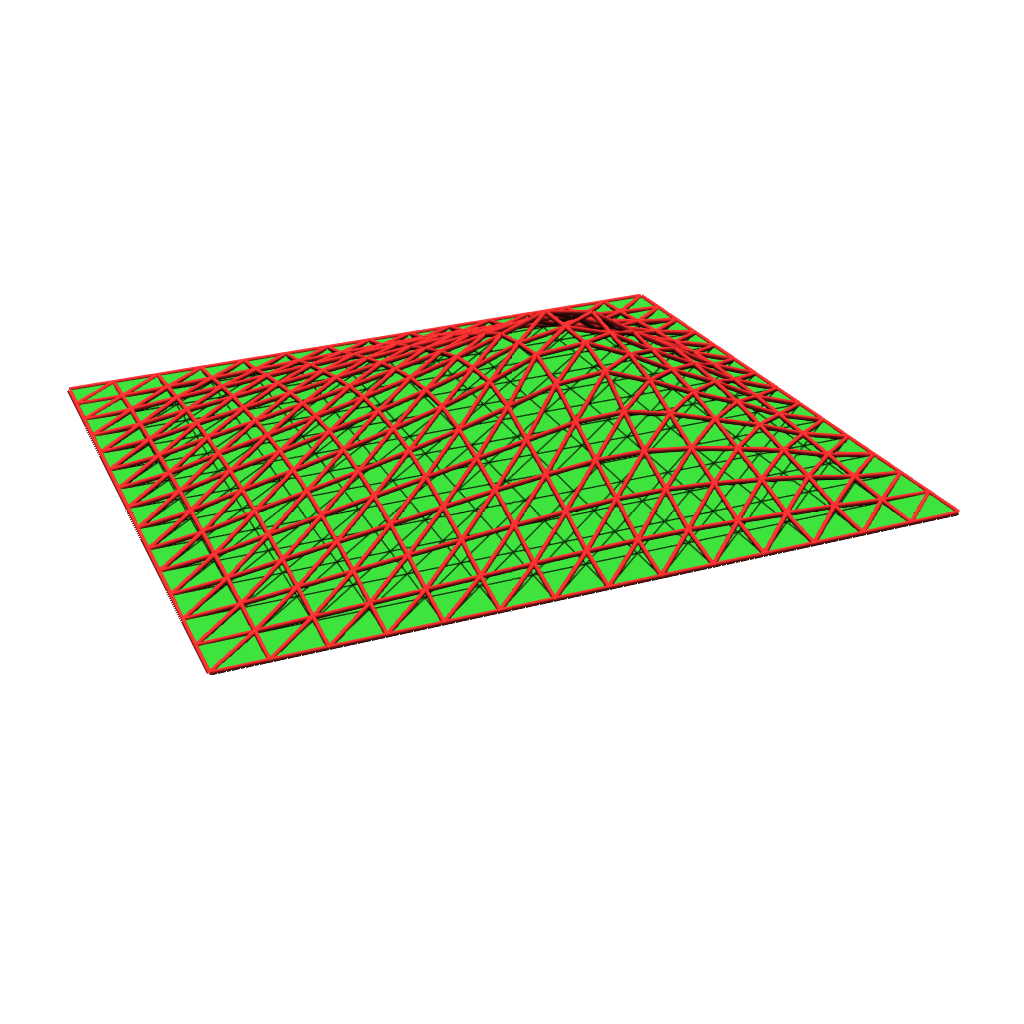}
    \end{minipage}
    \begin{minipage}{0.119\textwidth}
            \centering
            \includegraphics[width=\textwidth]{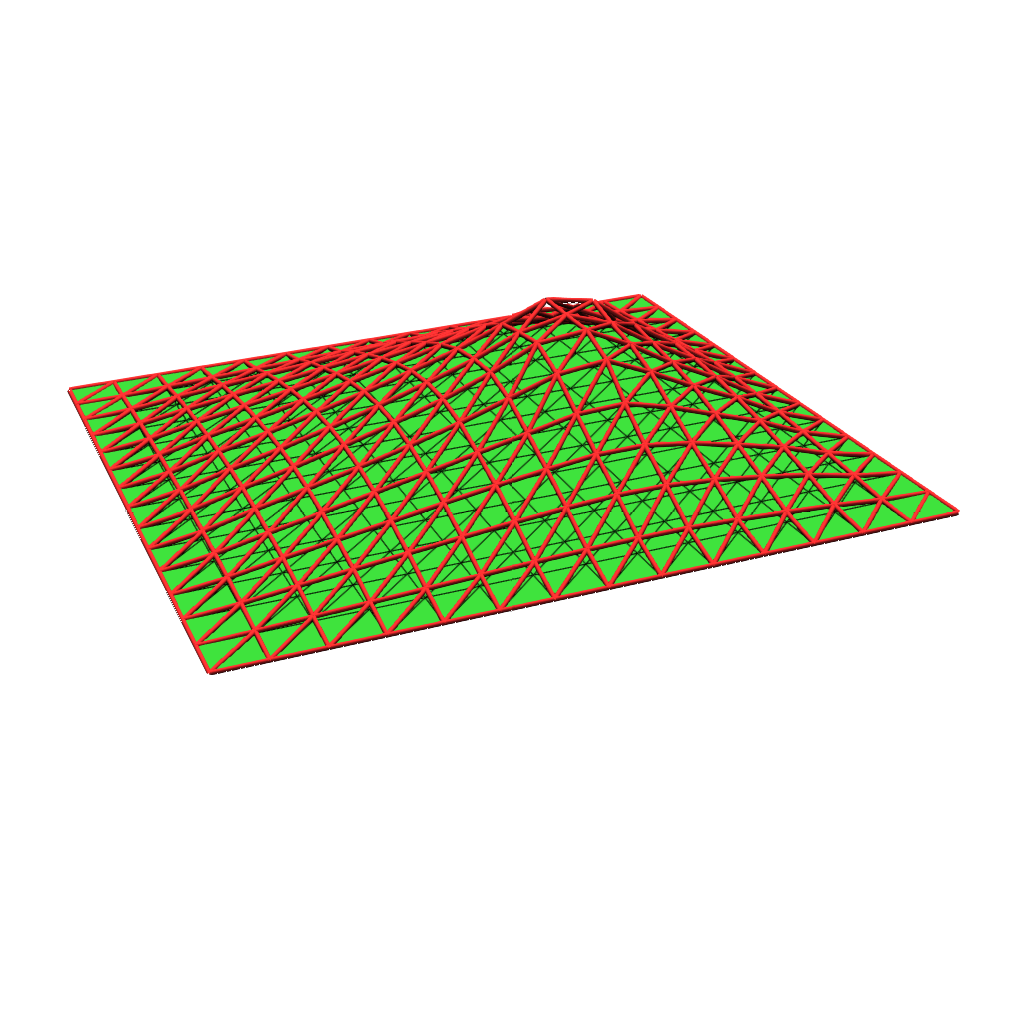}
    \end{minipage}
    \begin{minipage}{0.119\textwidth}
            \centering
            \includegraphics[width=\textwidth]{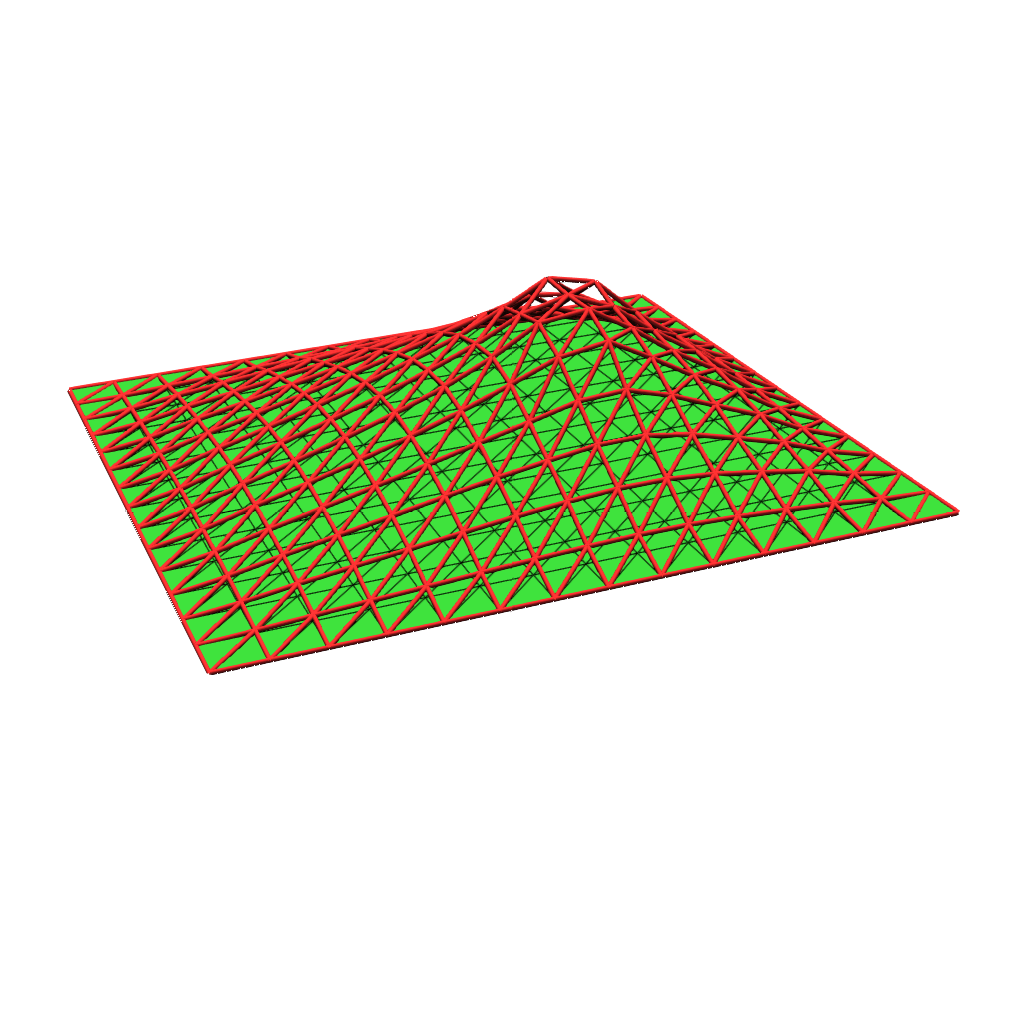}
    \end{minipage}
    \begin{minipage}{0.119\textwidth}
            \centering
            \includegraphics[width=\textwidth]{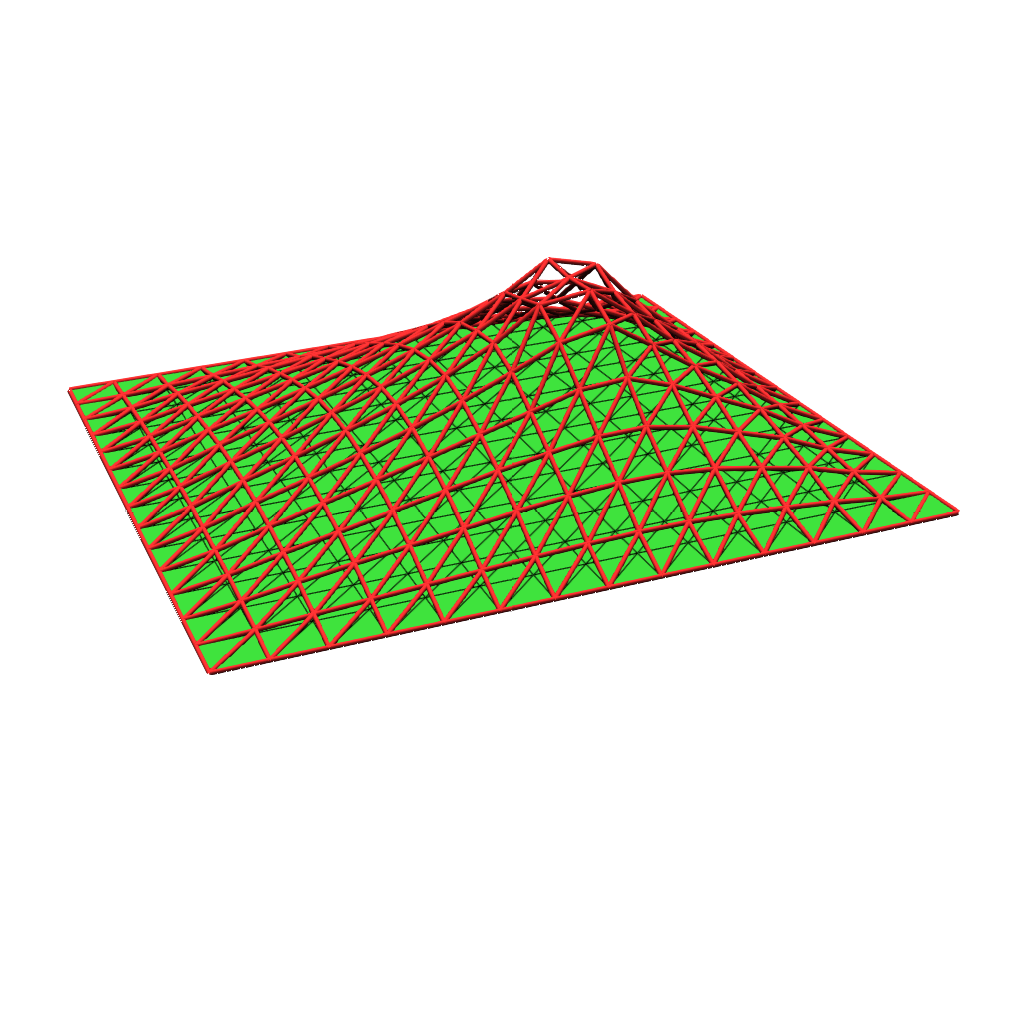}
    \end{minipage}
    \begin{minipage}{0.119\textwidth}
            \centering
            \includegraphics[width=\textwidth]{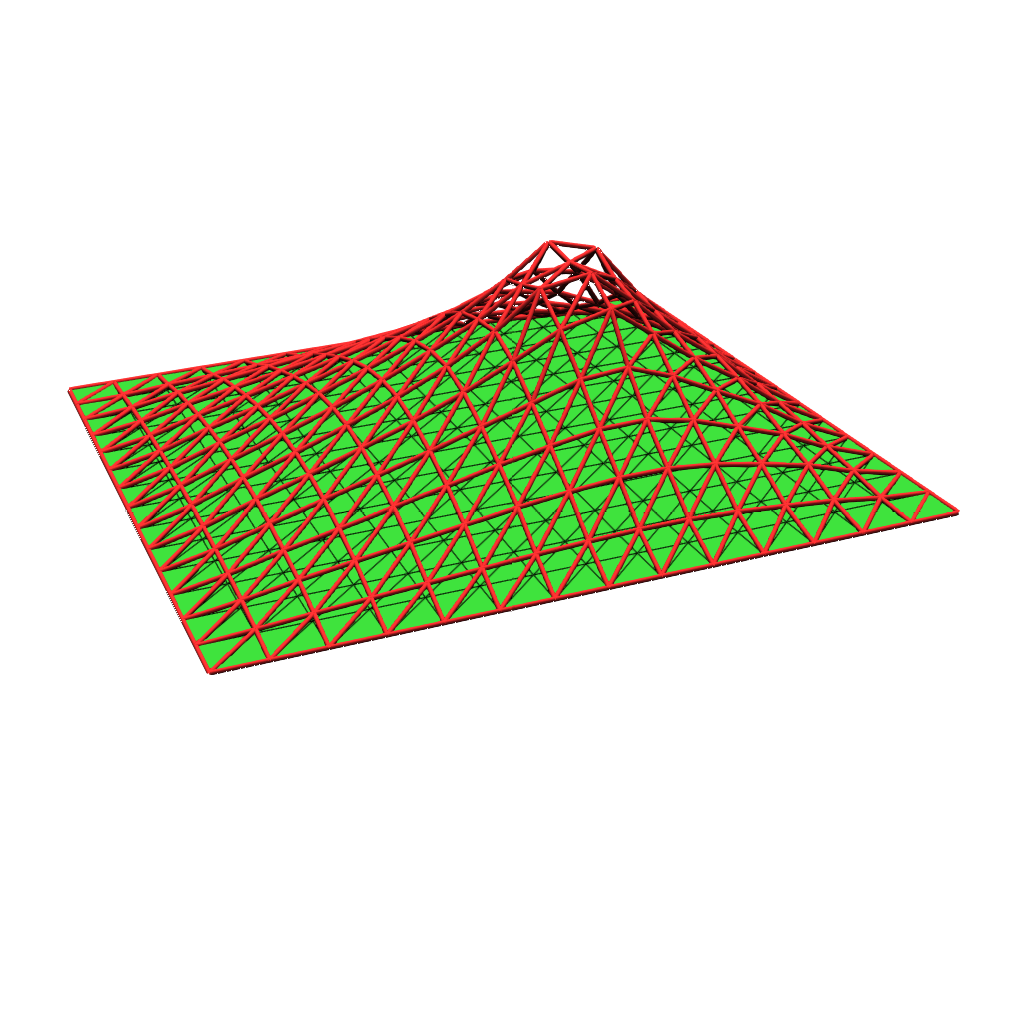}            
    \end{minipage}

    \vspace{1.0em} 
    \noindent\hrulefill 

    \vspace{-1.0em}
    \noindent\hrulefill 
    \vspace{0.5em} 

    \begin{minipage}{0.119\textwidth}
            \centering
            \includegraphics[width=\textwidth]{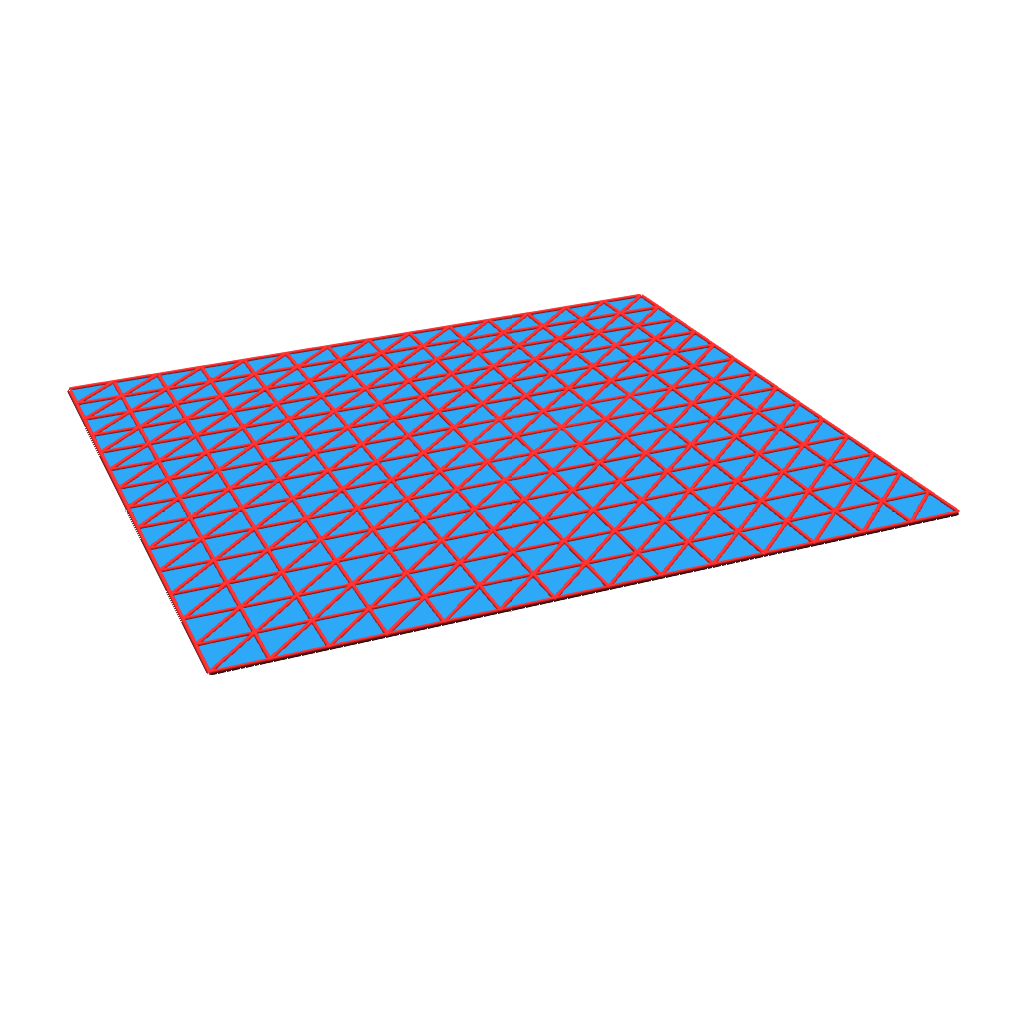}
            \subcaption*{$t=1$}
    \end{minipage}
    \begin{minipage}{0.119\textwidth}
            \centering
            \includegraphics[width=\textwidth]{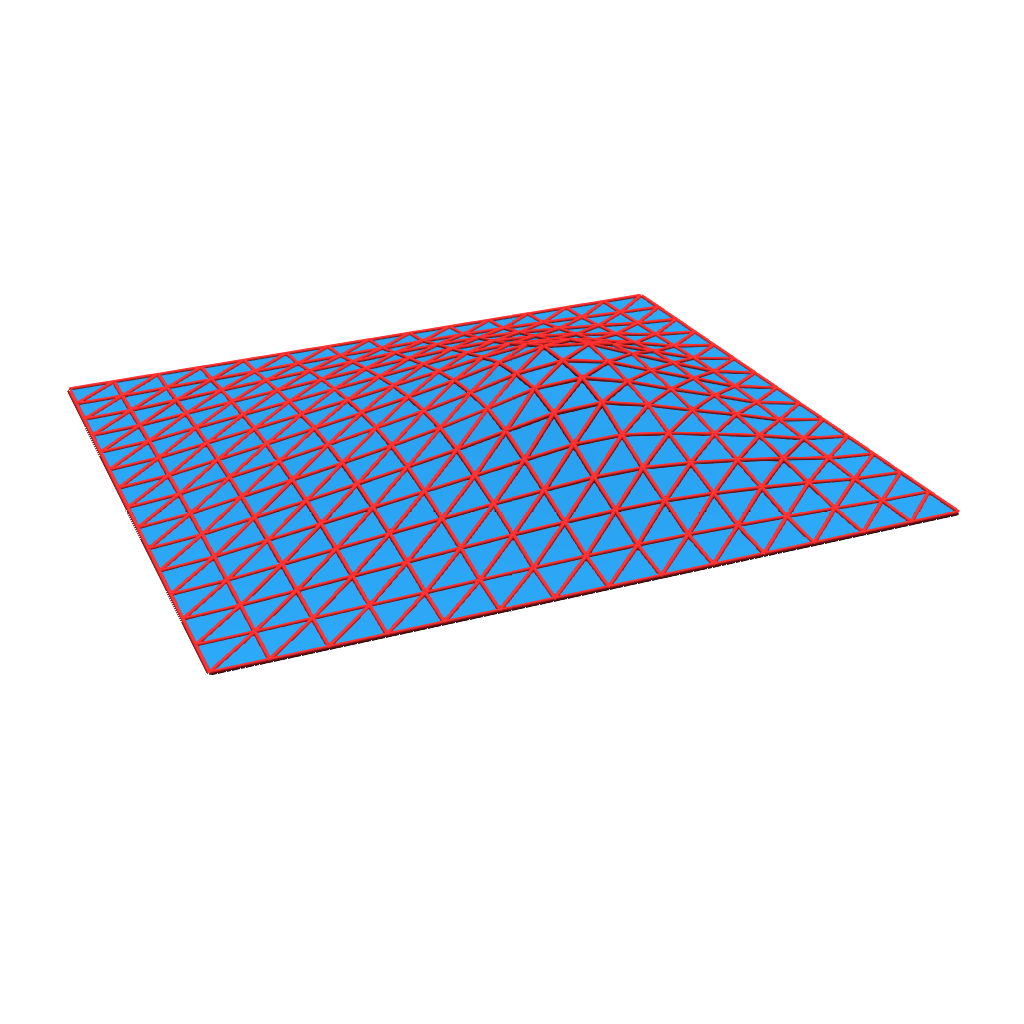}
            \subcaption*{$t=6$}
    \end{minipage}
    \begin{minipage}{0.119\textwidth}
            \centering
            \includegraphics[width=\textwidth]{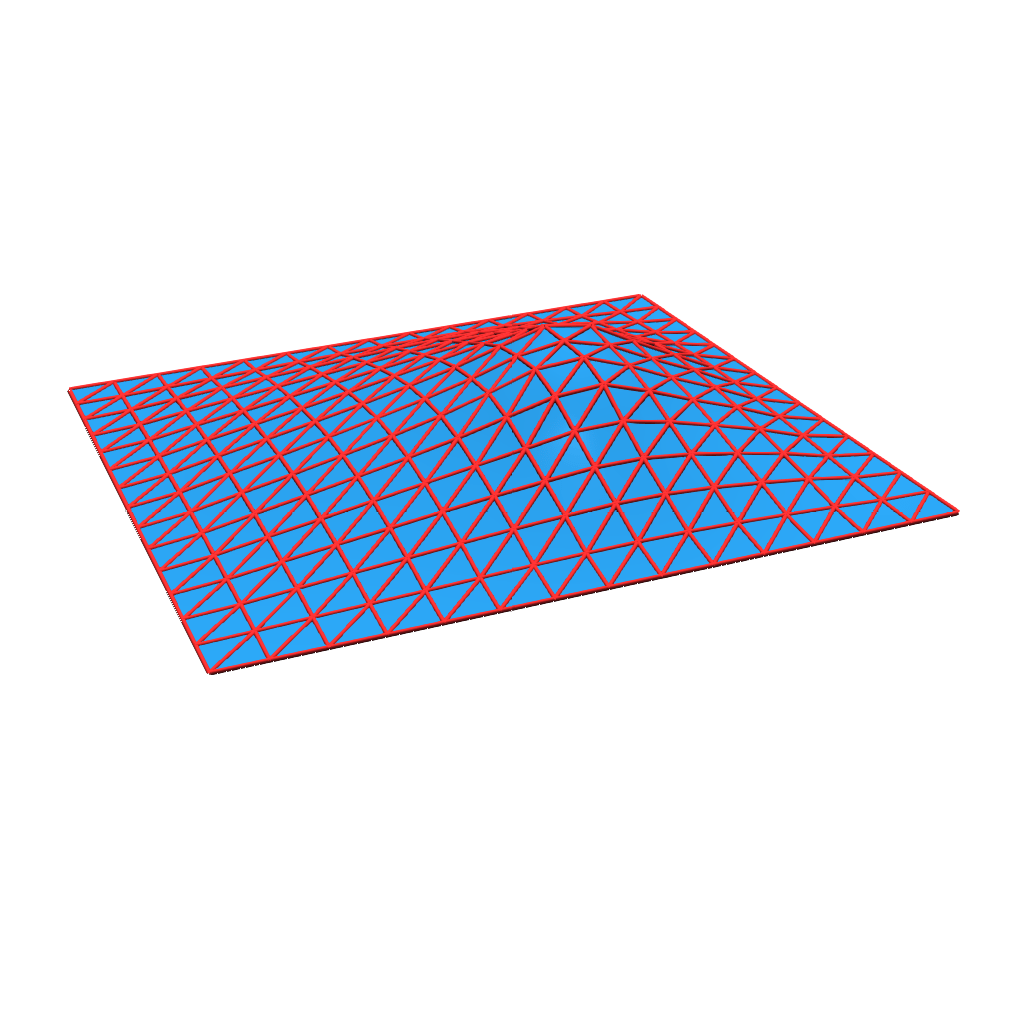}
            \subcaption*{$t=11$}
    \end{minipage}
    \begin{minipage}{0.119\textwidth}
            \centering
            \includegraphics[width=\textwidth]{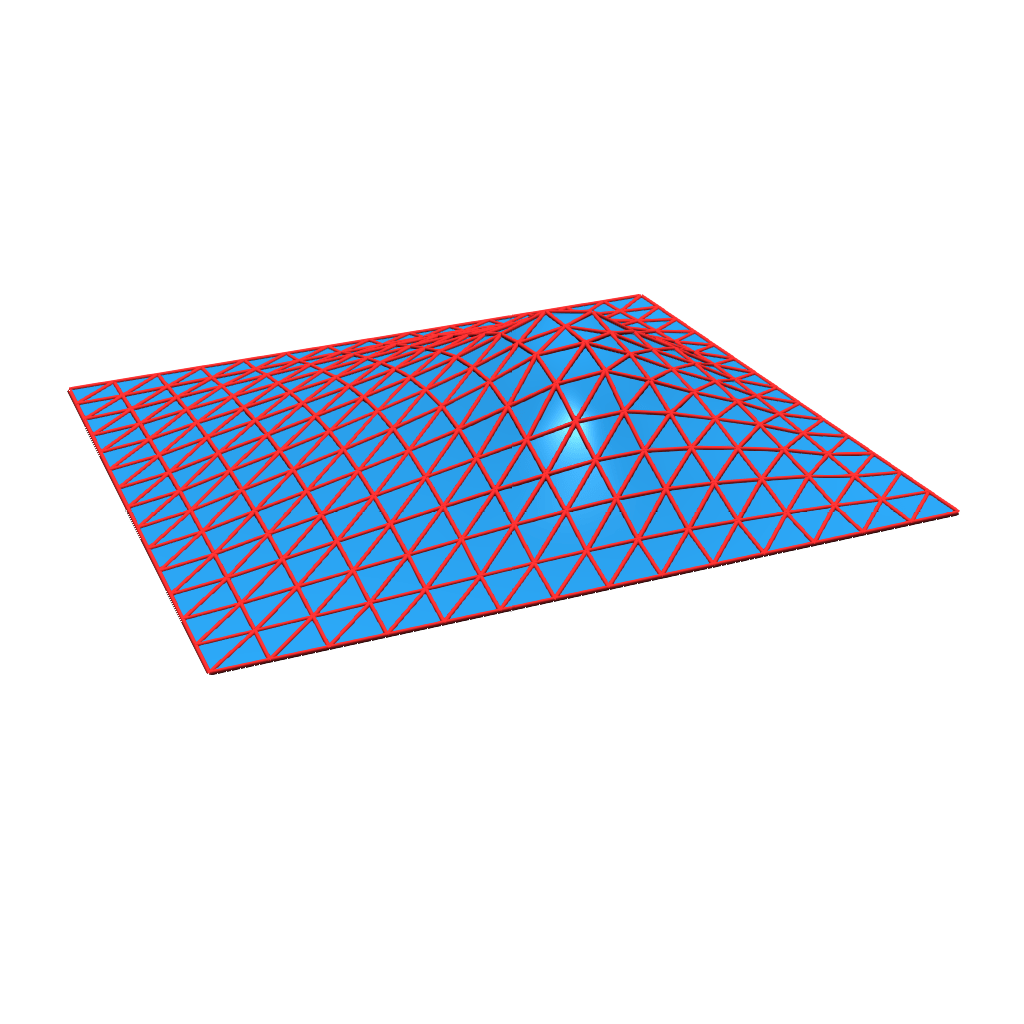}
            \subcaption*{$t=16$}
    \end{minipage}
    \begin{minipage}{0.119\textwidth}
            \centering
            \includegraphics[width=\textwidth]{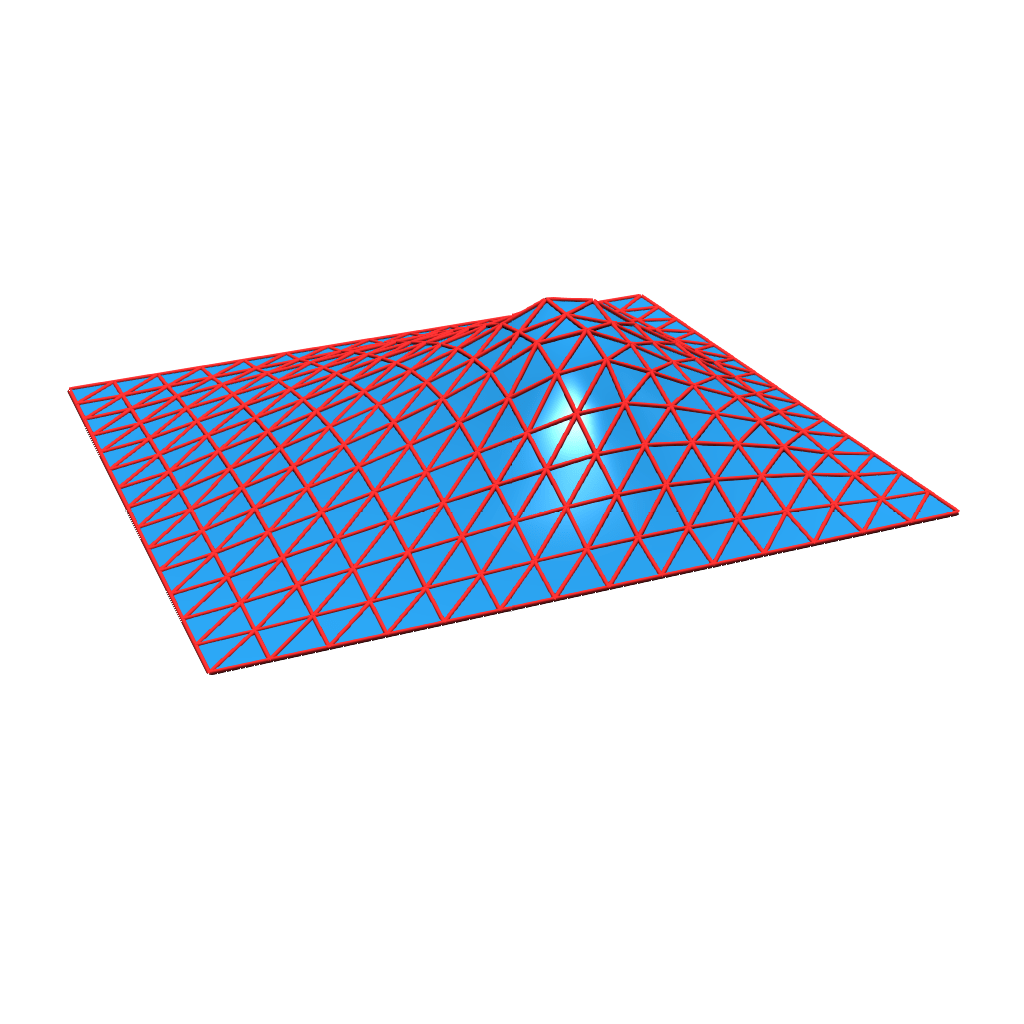}
            \subcaption*{$t=21$}
    \end{minipage}
    \begin{minipage}{0.119\textwidth}
            \centering
            \includegraphics[width=\textwidth]{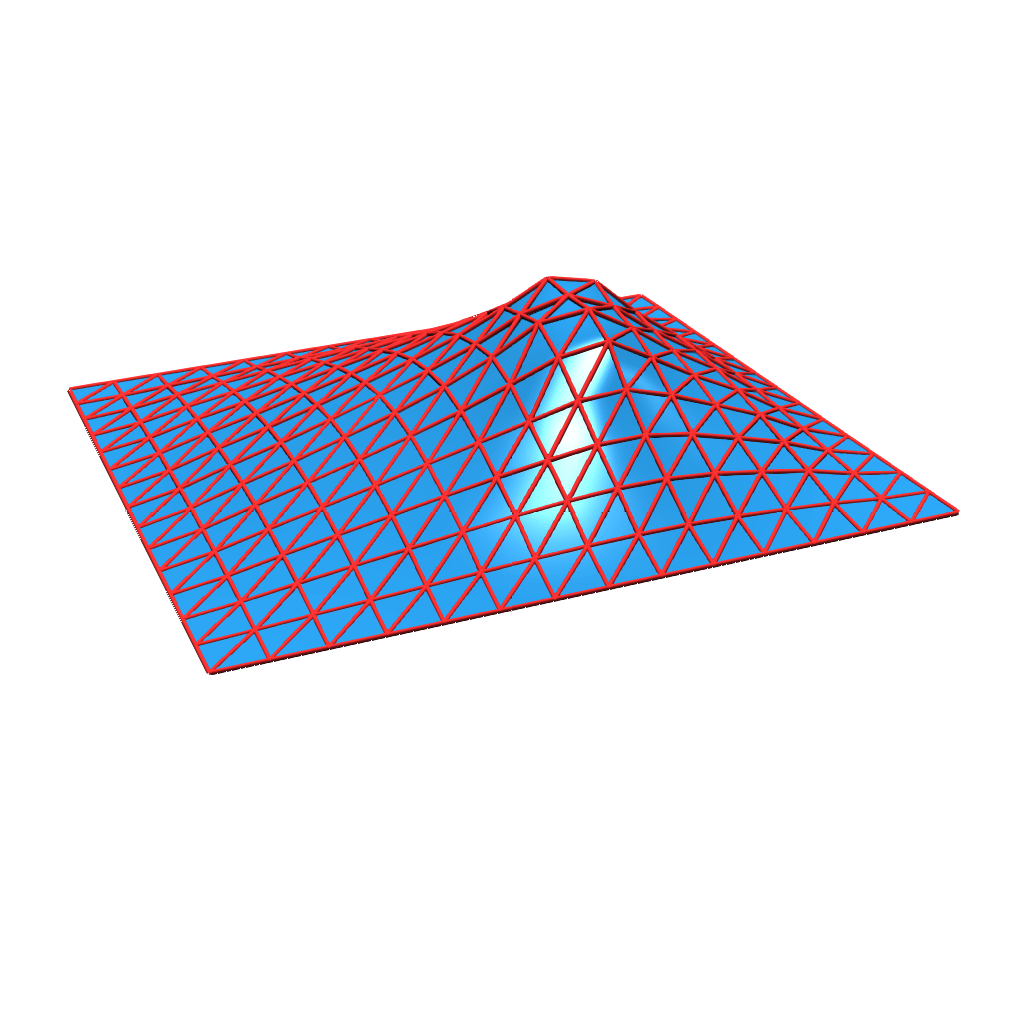}
            \subcaption*{$t=31$}
    \end{minipage}
    \begin{minipage}{0.119\textwidth}
            \centering
            \includegraphics[width=\textwidth]{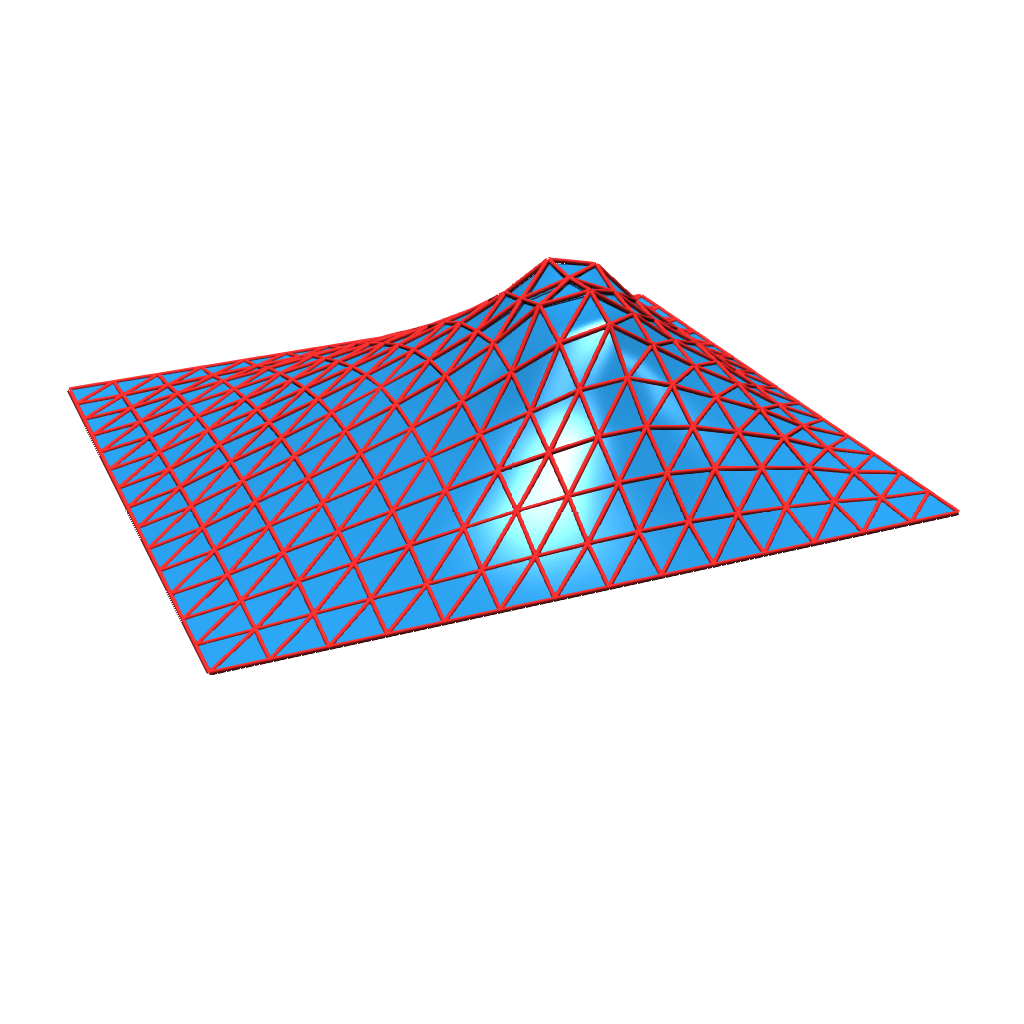}
            \subcaption*{$t=41$}
    \end{minipage}
    \begin{minipage}{0.119\textwidth}
            \centering
            \includegraphics[width=\textwidth]{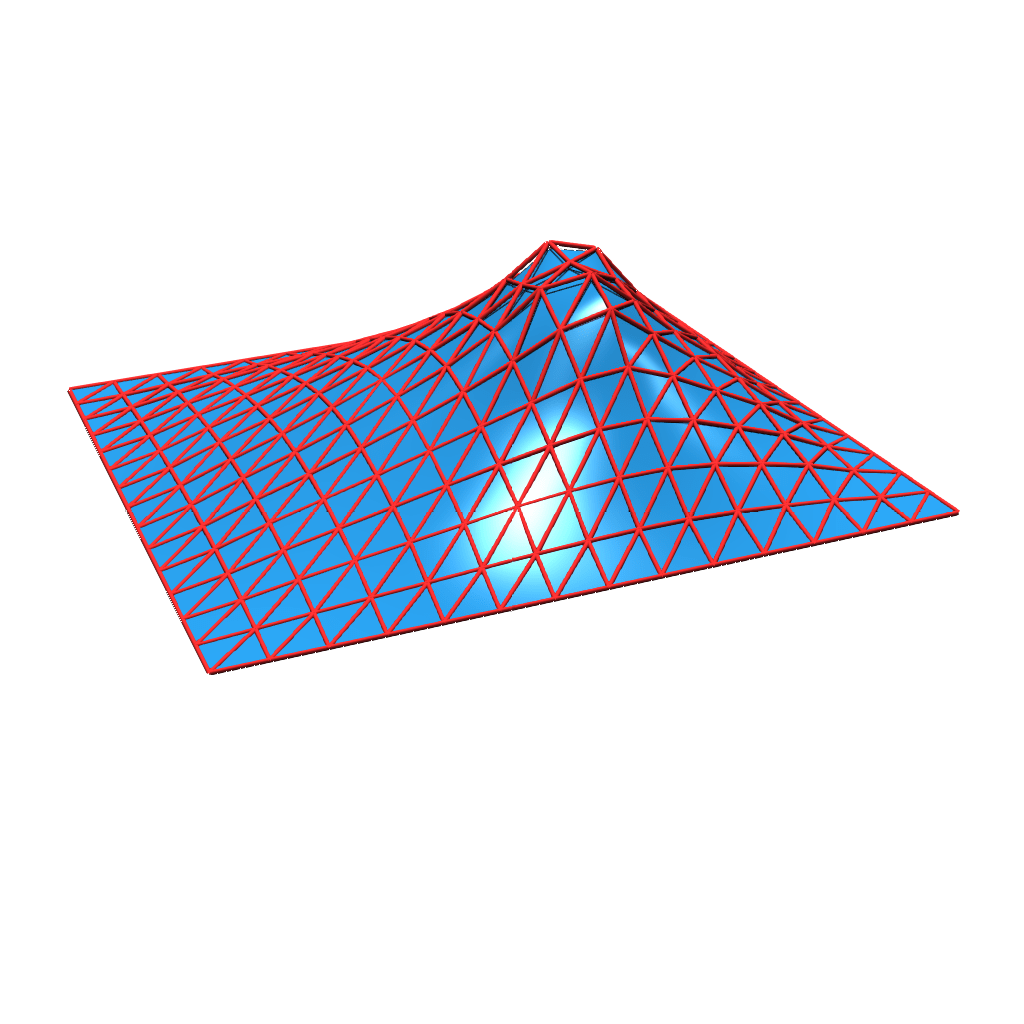}
            \subcaption*{$t=50$}
    \end{minipage}

    \vspace{0.01\textwidth}%

    \caption{
    Simulation over time of an exemplary test trajectory from the \textbf{PB-easy} task. The figure compares predictions from \textcolor{tabblue}{\textbf{MaNGO}}, \textcolor{taborange}{MGN}, and \textcolor{ForestGreen}{EGNO}. The last row, \textcolor{tabblue}{MaNGO-Oracle}, is separated by a horizontal line and represents predictions using oracle information. The \textbf{context set size} is set to $4$. All visualizations show the colored \textbf{predicted mesh}, with a \textbf{\textcolor{red}{wireframe}} representing the ground-truth simulation. \textcolor{tabblue}{\textbf{MaNGO}} accurately predicts the correct material properties, leading to a highly accurate simulation.
    }
    \label{fig:appendix_pb_easy}
\end{figure*}
\begin{figure*}[ht!]
    \centering
    \begin{minipage}{0.119\textwidth}
            \centering
            \includegraphics[width=\textwidth]{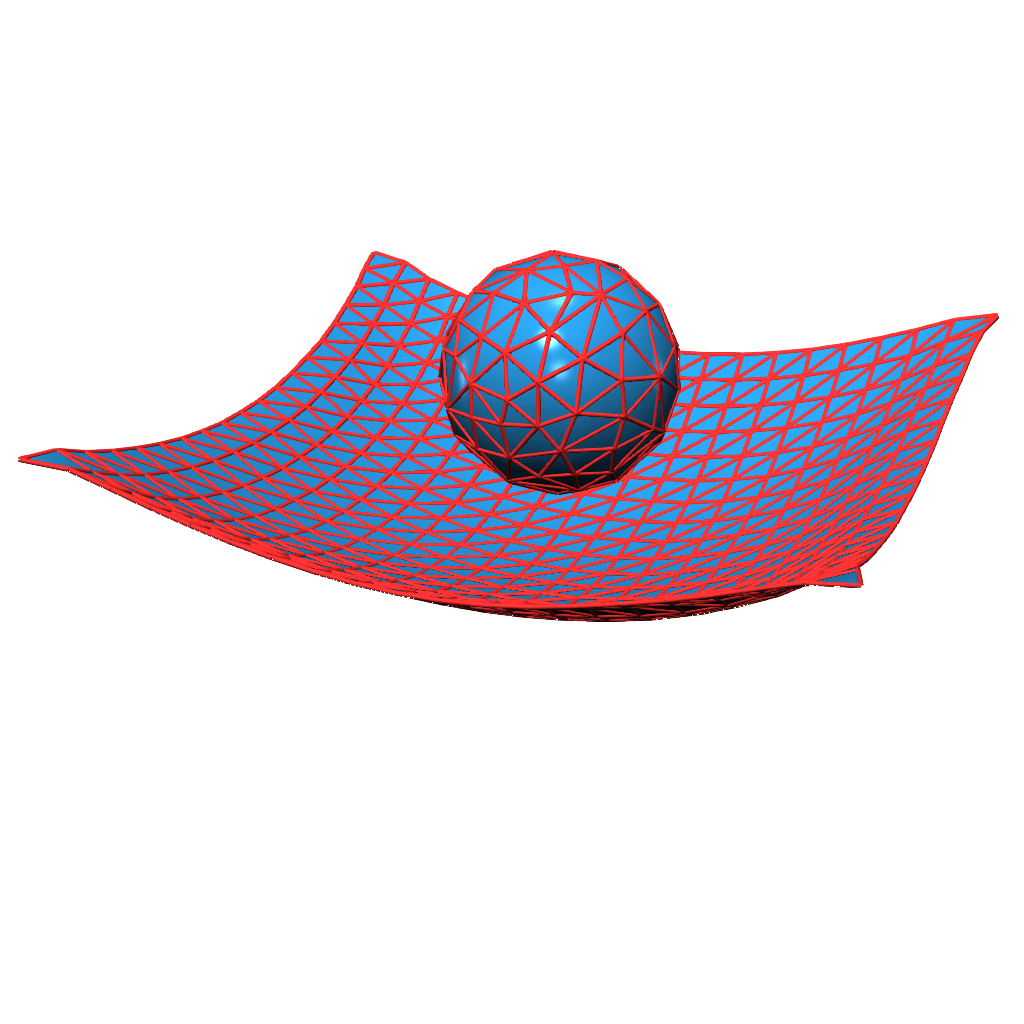}
    \end{minipage}
    \begin{minipage}{0.119\textwidth}
            \centering
            \includegraphics[width=\textwidth]{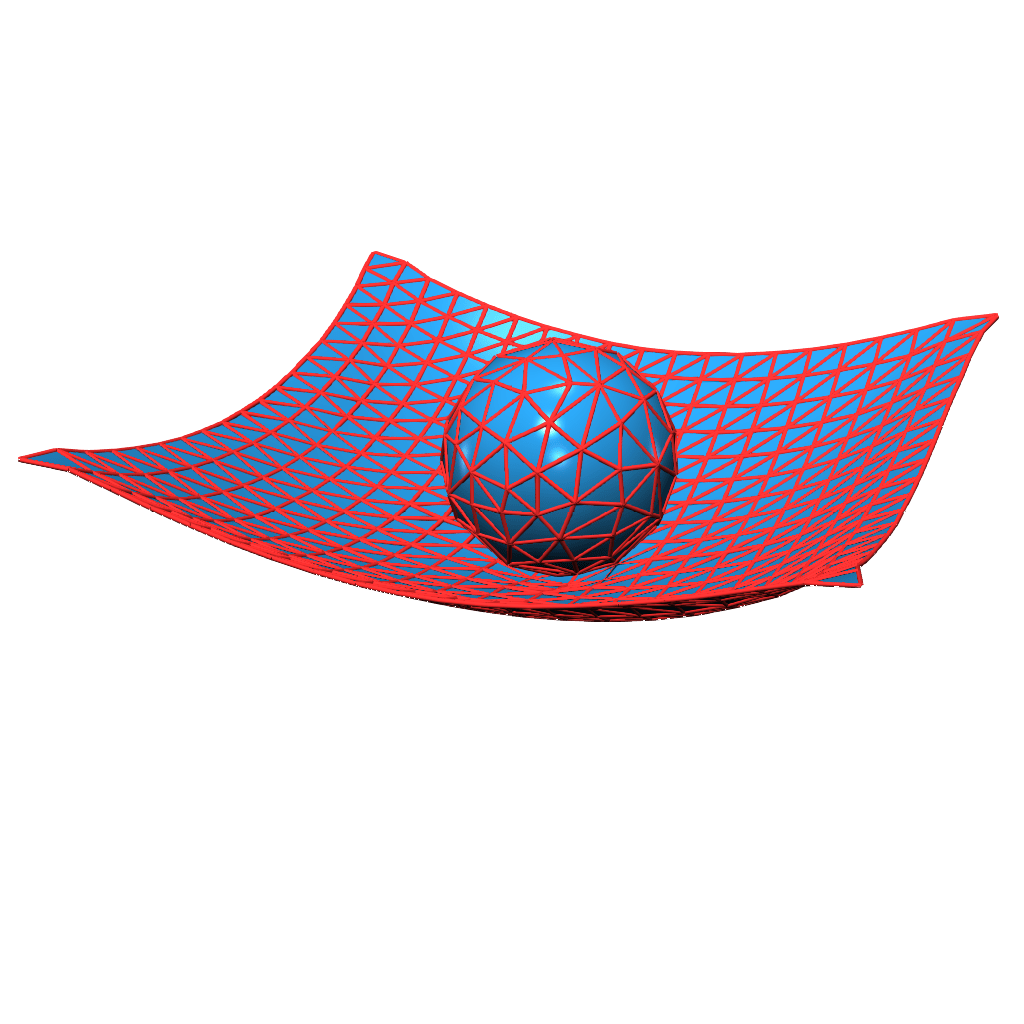}
    \end{minipage}
    \begin{minipage}{0.119\textwidth}
            \centering
            \includegraphics[width=\textwidth]{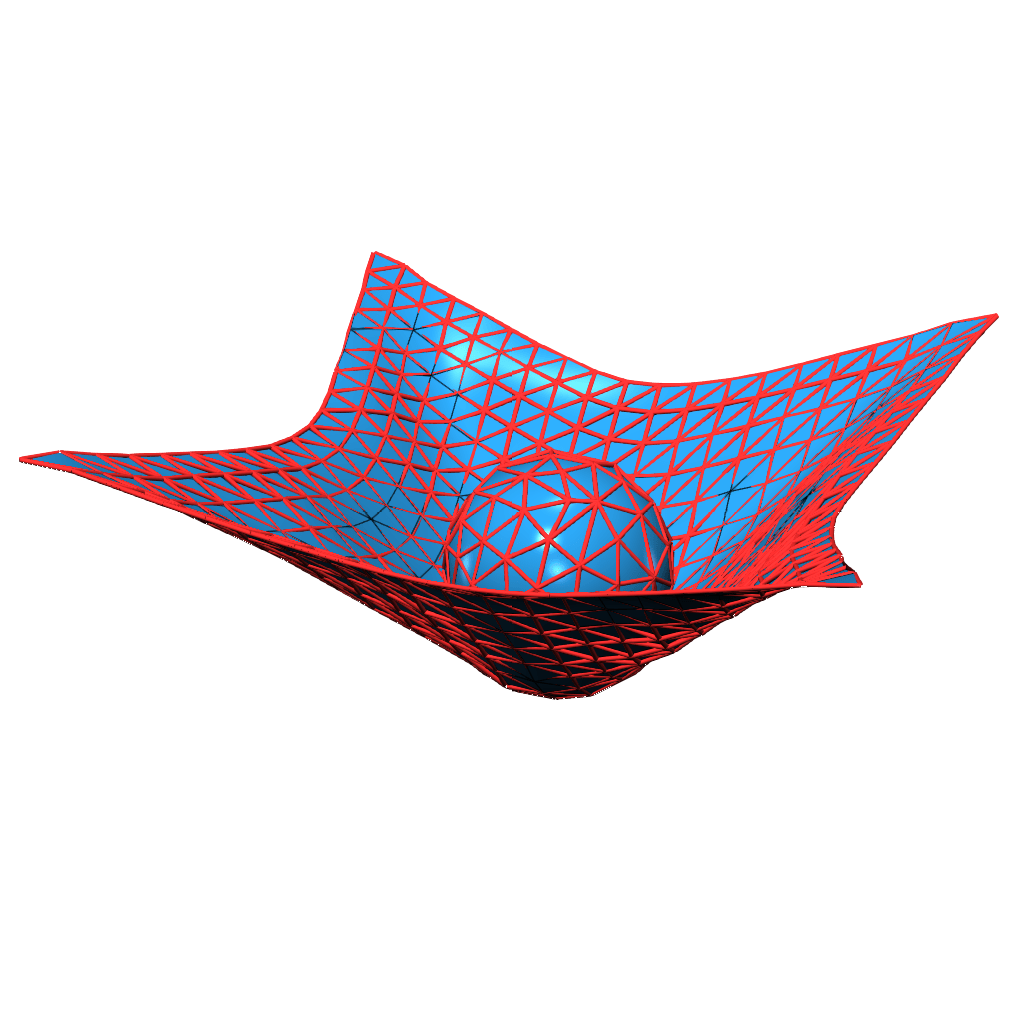}
    \end{minipage}
    \begin{minipage}{0.119\textwidth}
            \centering
            \includegraphics[width=\textwidth]{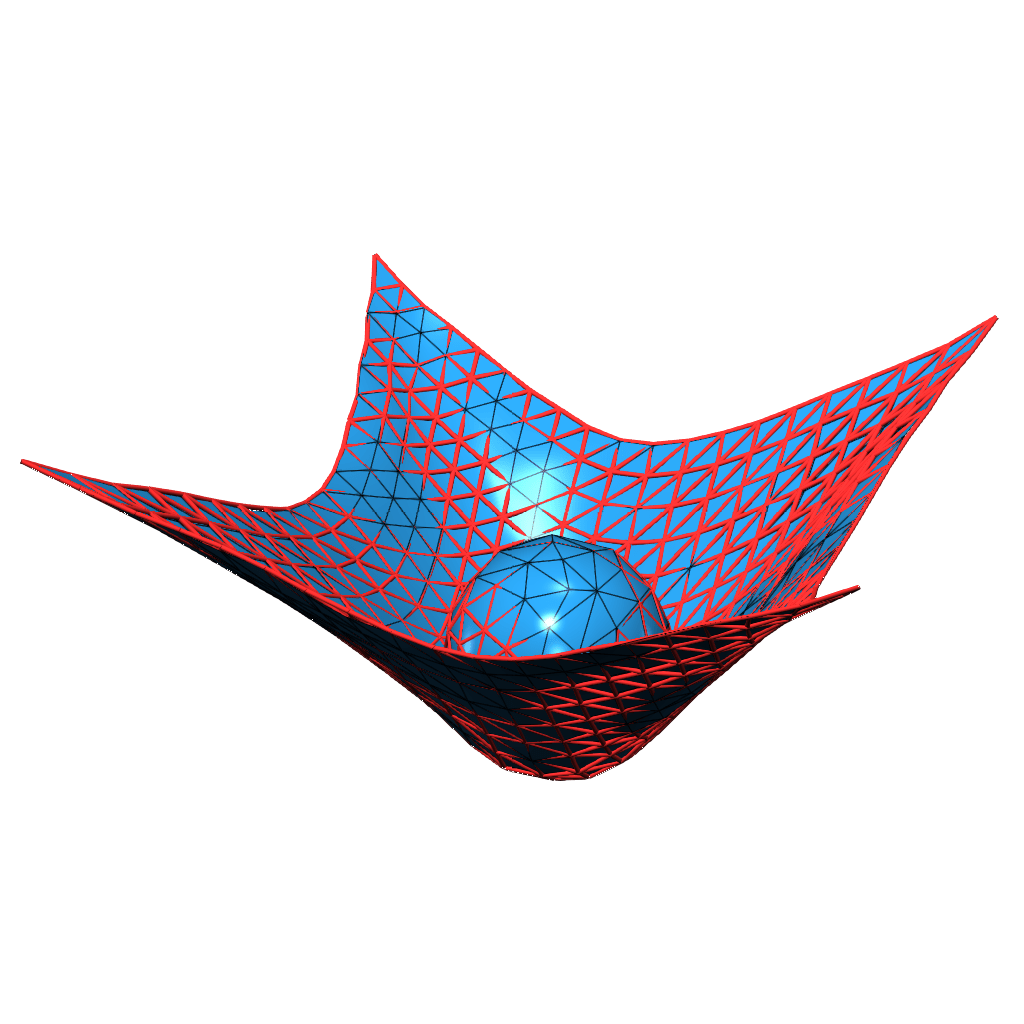}
    \end{minipage}
    \begin{minipage}{0.119\textwidth}
            \centering
            \includegraphics[width=\textwidth]{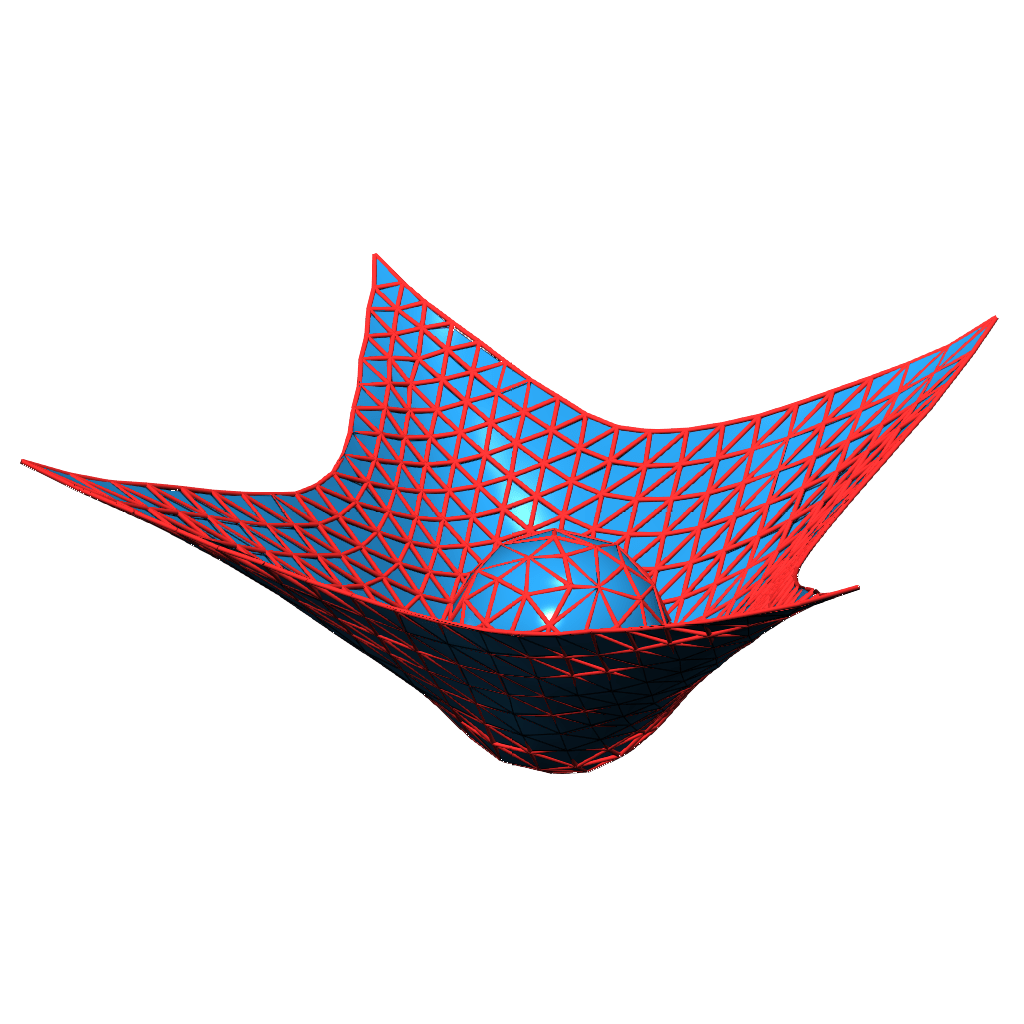}
    \end{minipage}
    \begin{minipage}{0.119\textwidth}
            \centering
            \includegraphics[width=\textwidth]{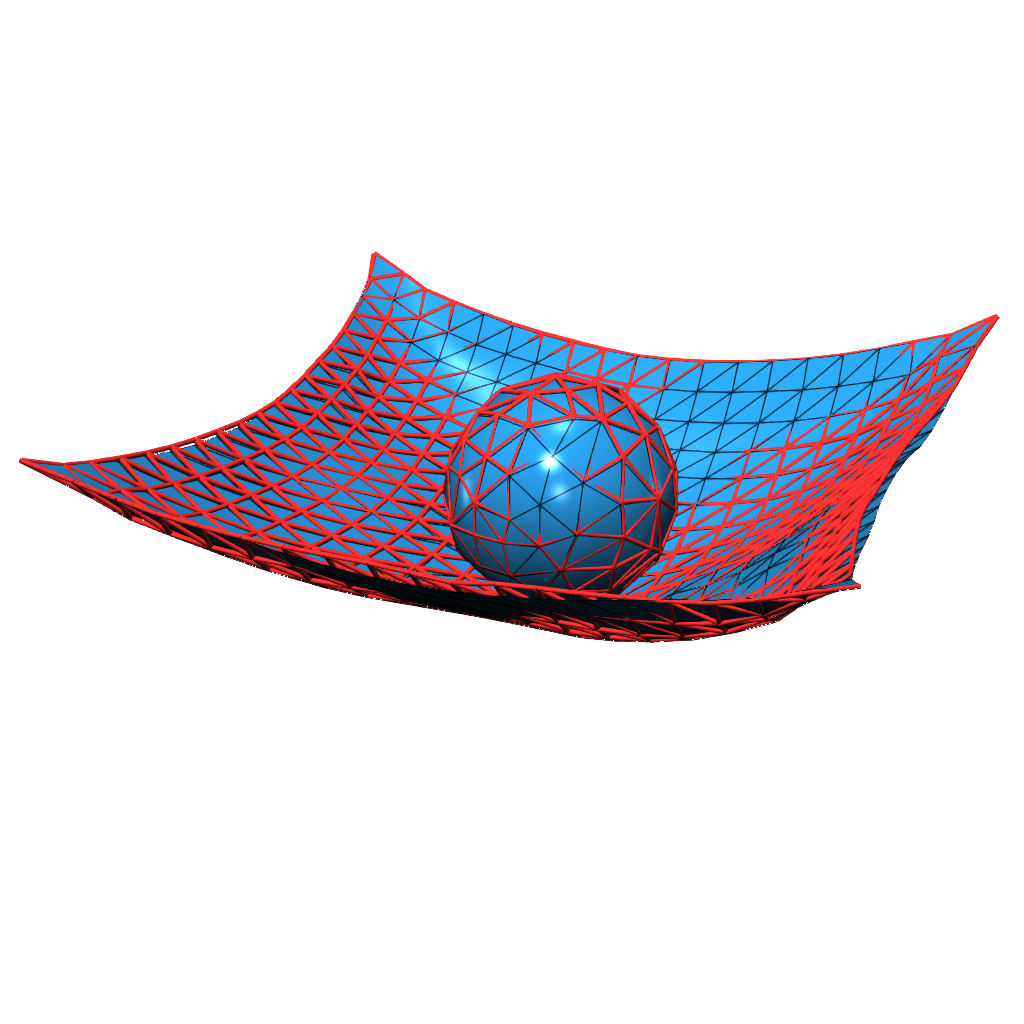}
    \end{minipage}
    \begin{minipage}{0.119\textwidth}
            \centering
            \includegraphics[width=\textwidth]{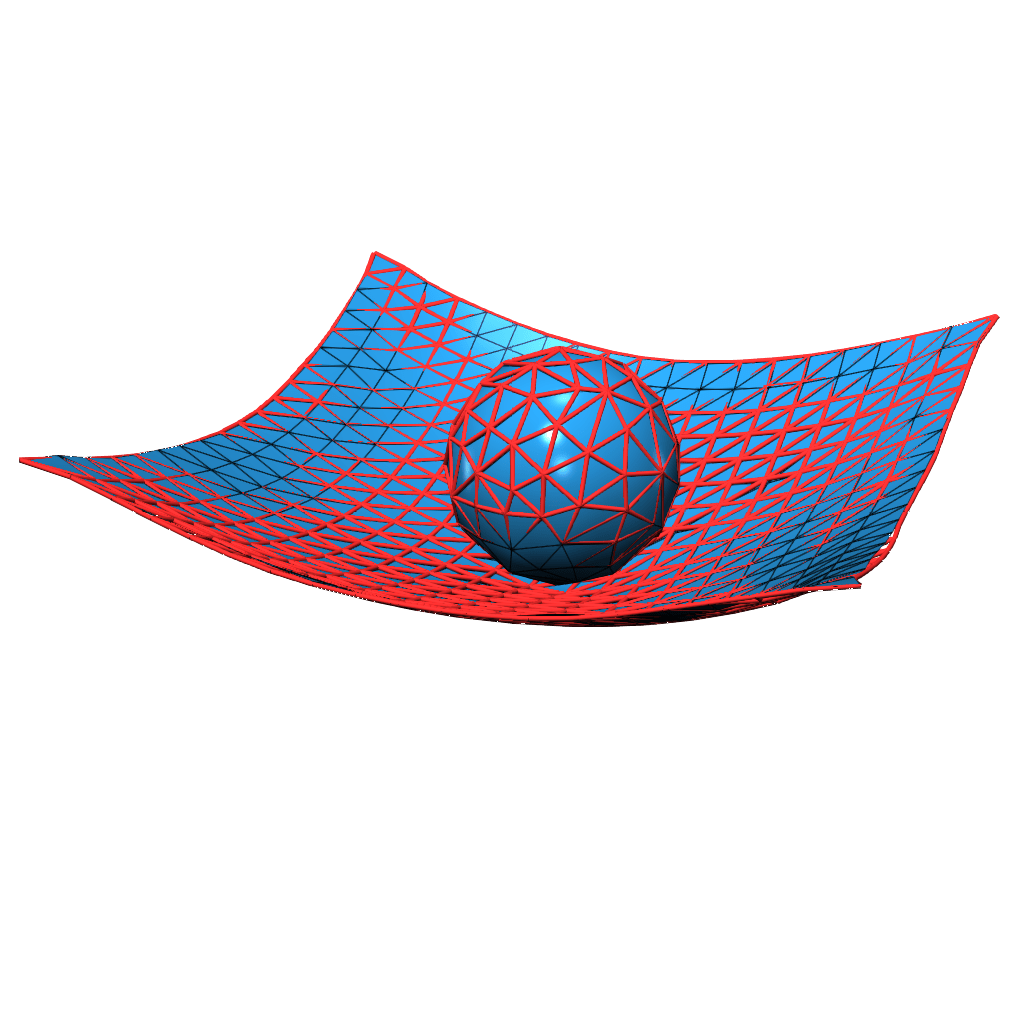}
    \end{minipage}
    \begin{minipage}{0.119\textwidth}
            \centering
            \includegraphics[width=\textwidth]{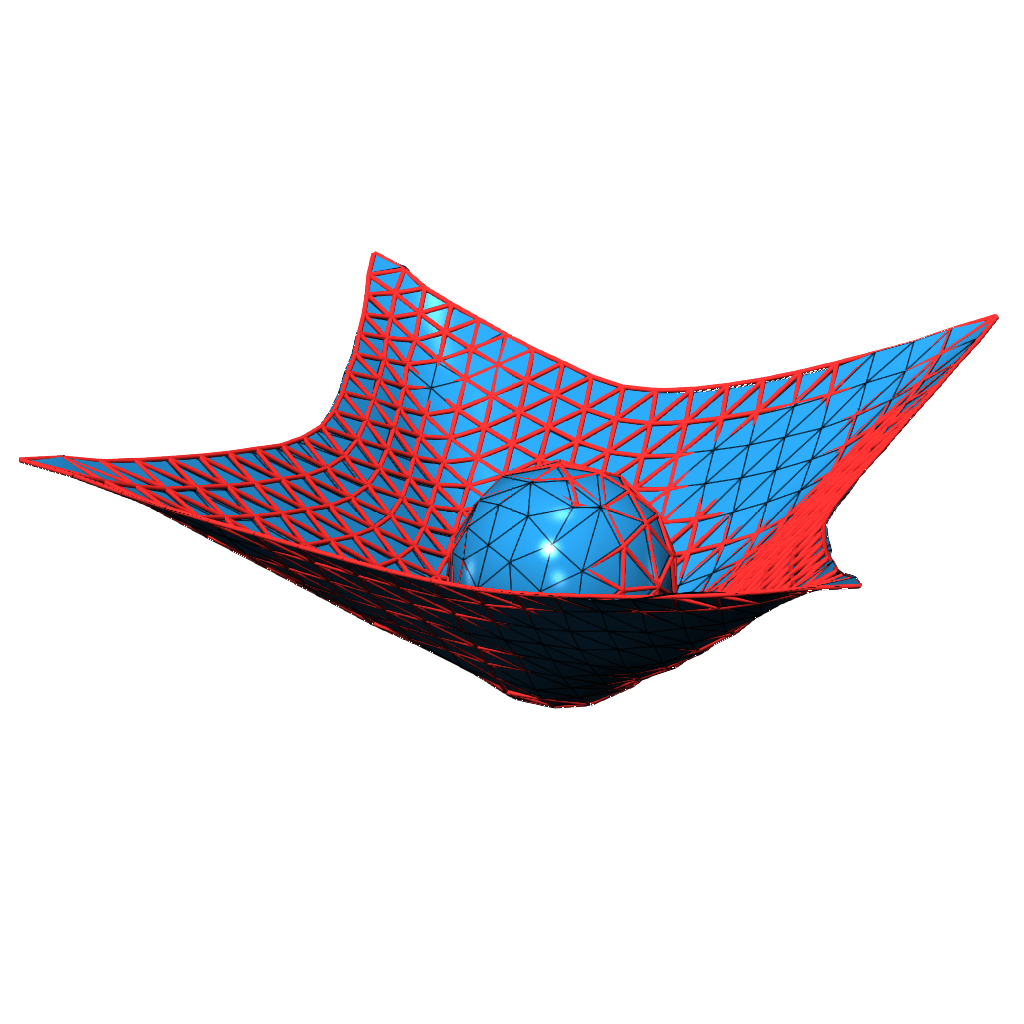}
    \end{minipage}

    \begin{minipage}{0.119\textwidth}
            \centering
            \includegraphics[width=\textwidth]{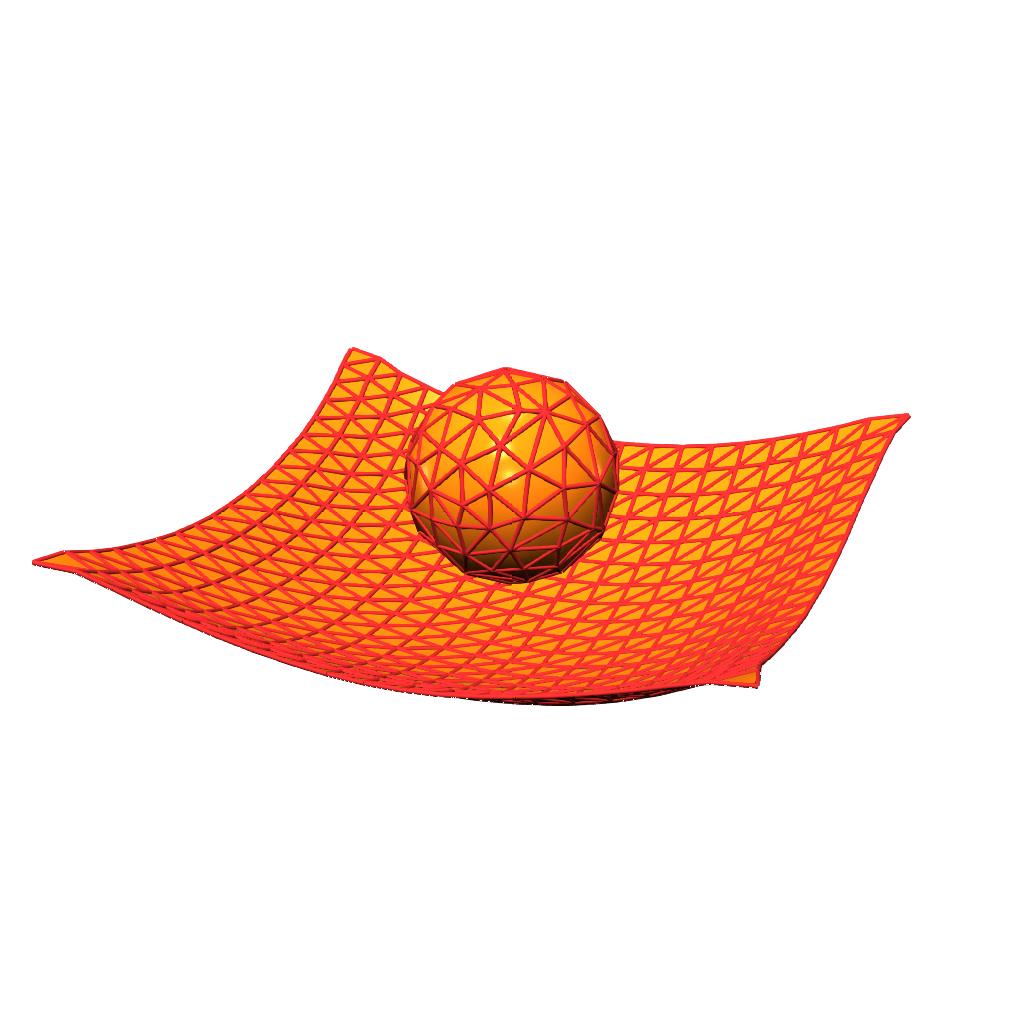}
    \end{minipage}
    \begin{minipage}{0.119\textwidth}
            \centering
            \includegraphics[width=\textwidth]{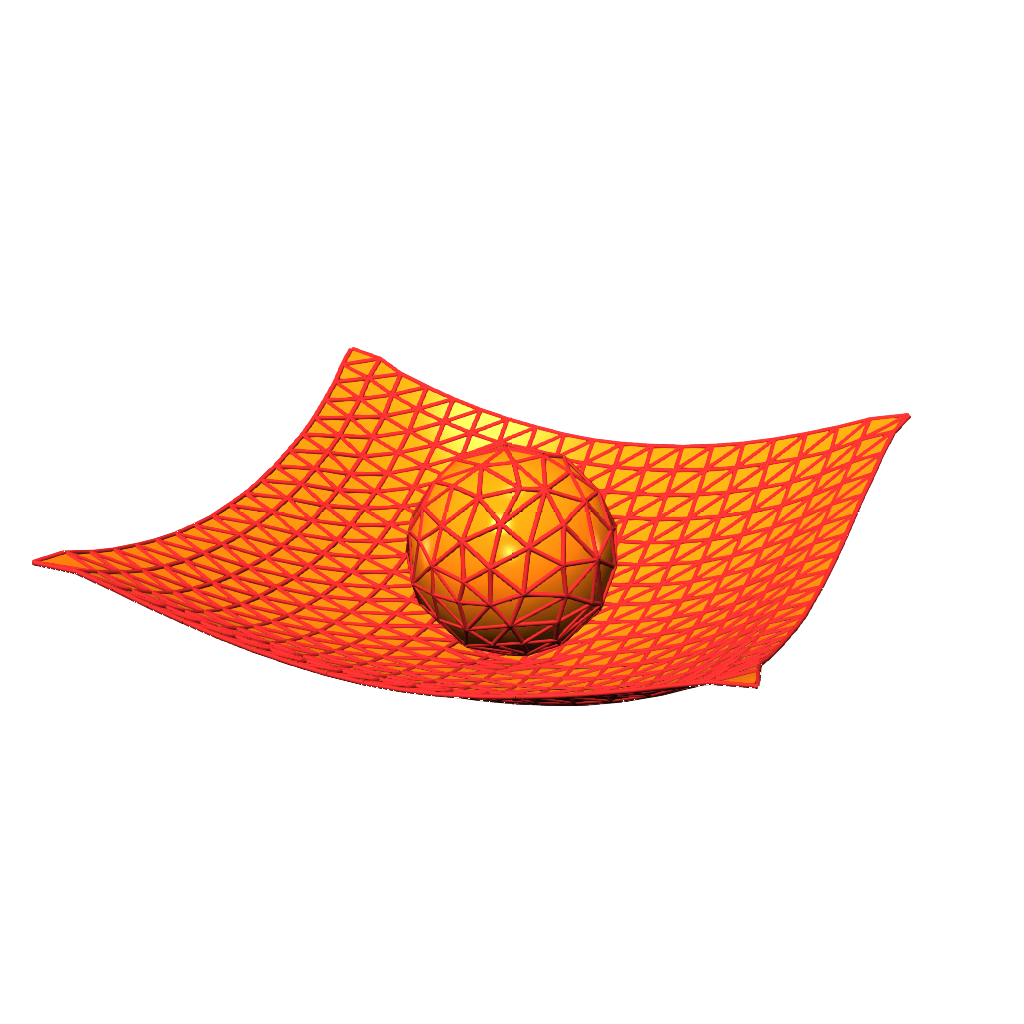}
    \end{minipage}
    \begin{minipage}{0.119\textwidth}
            \centering
            \includegraphics[width=\textwidth]{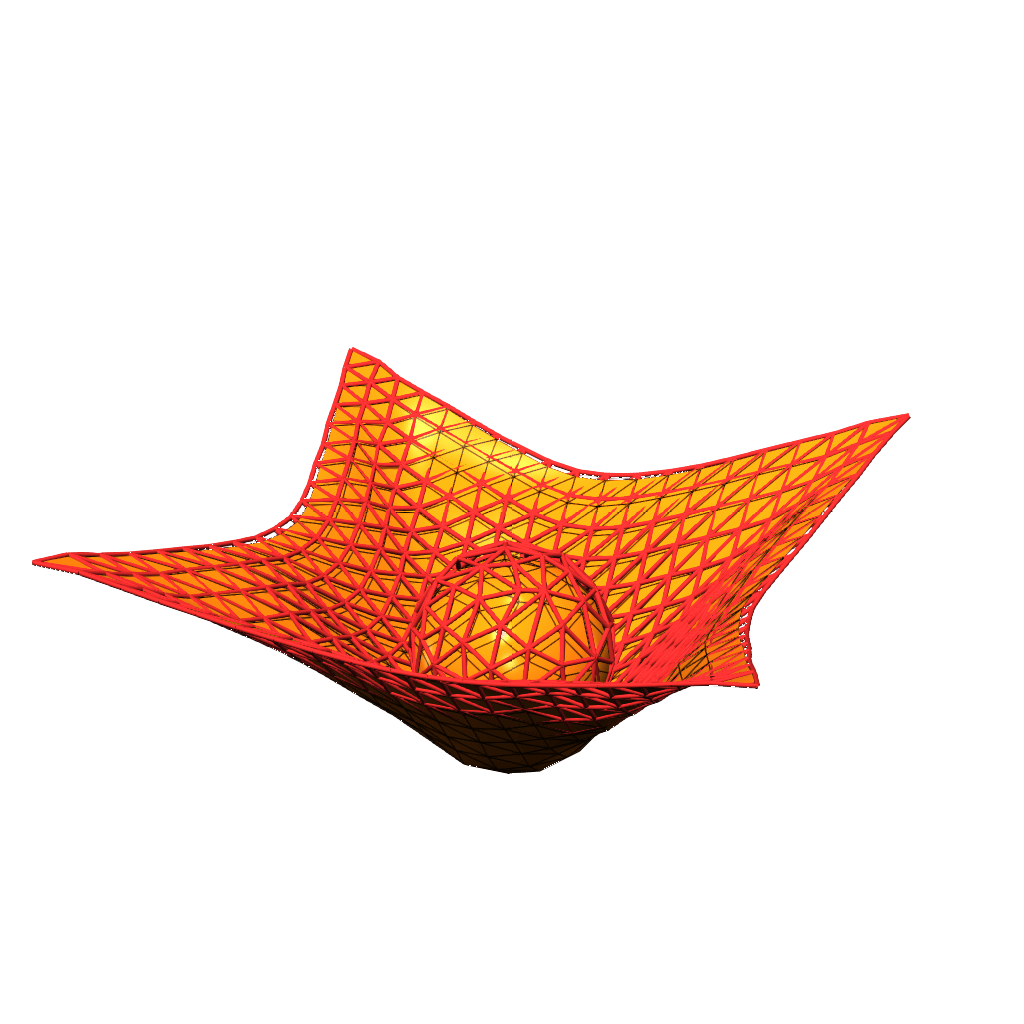}
    \end{minipage}
    \begin{minipage}{0.119\textwidth}
            \centering
            \includegraphics[width=\textwidth]{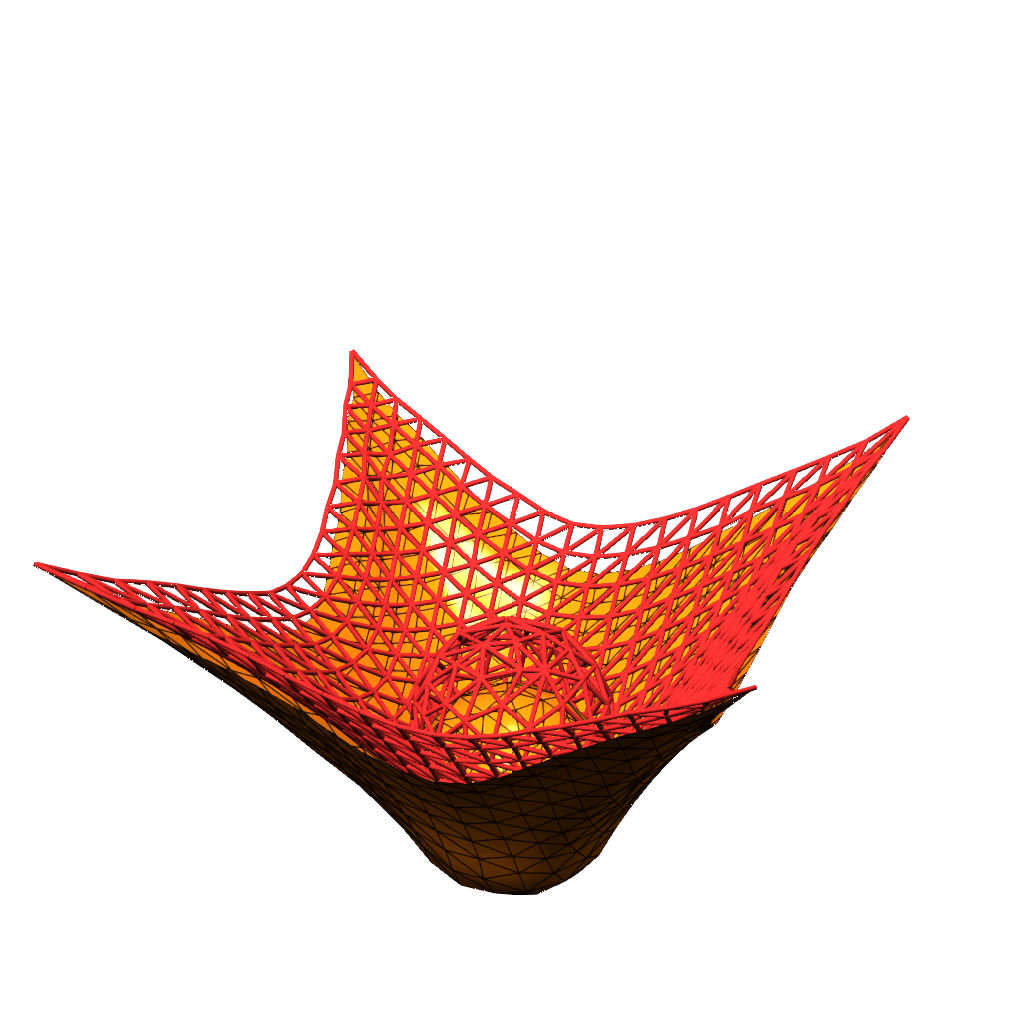}
    \end{minipage}
    \begin{minipage}{0.119\textwidth}
            \centering
            \includegraphics[width=\textwidth]{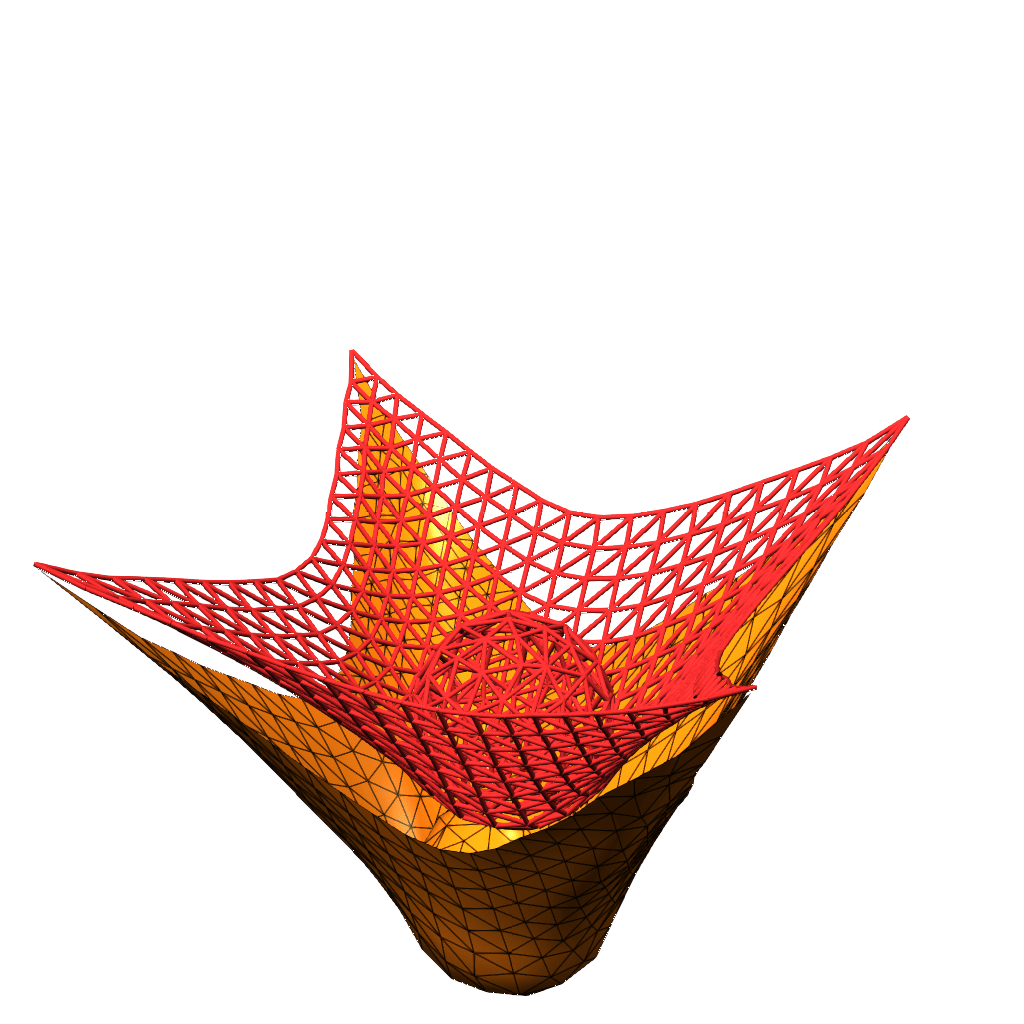}
    \end{minipage}
    \begin{minipage}{0.119\textwidth}
            \centering
            \includegraphics[width=\textwidth]{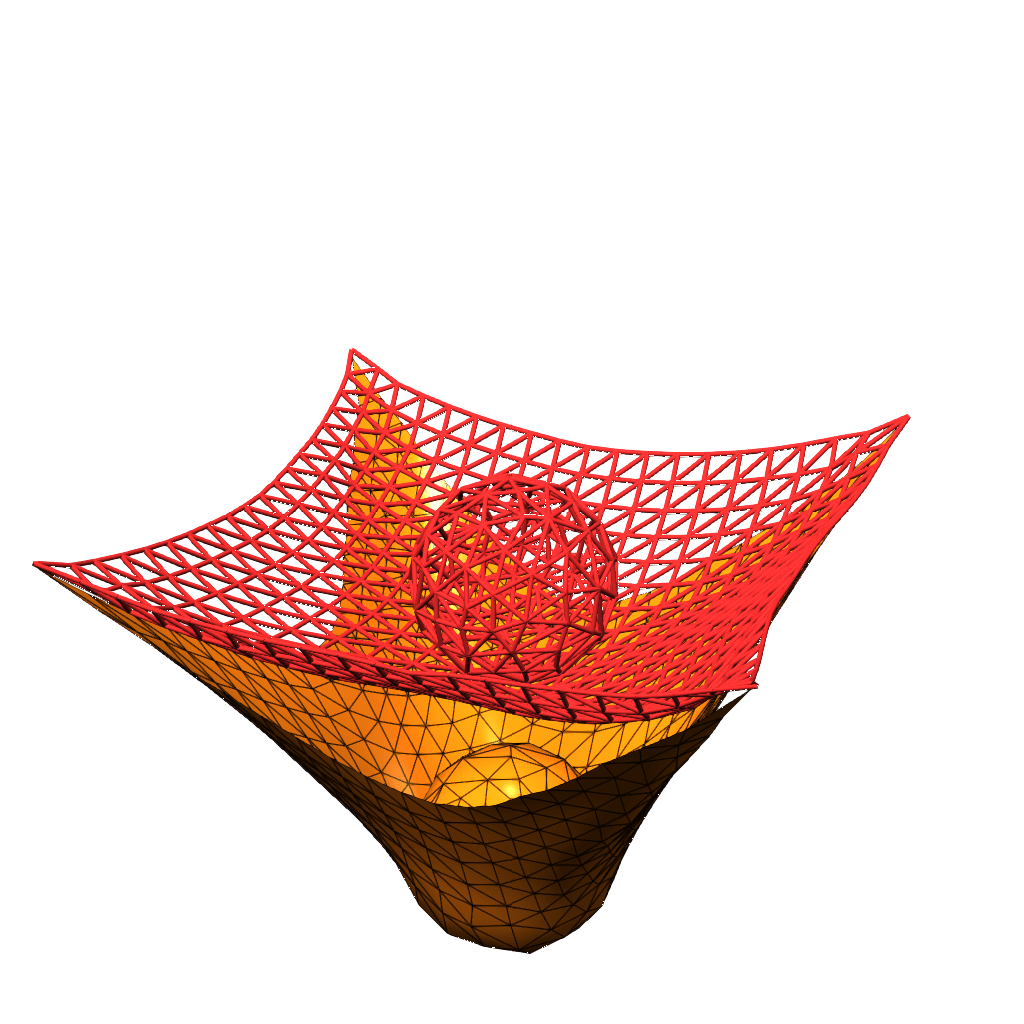}
    \end{minipage}
    \begin{minipage}{0.119\textwidth}
            \centering
            \includegraphics[width=\textwidth]{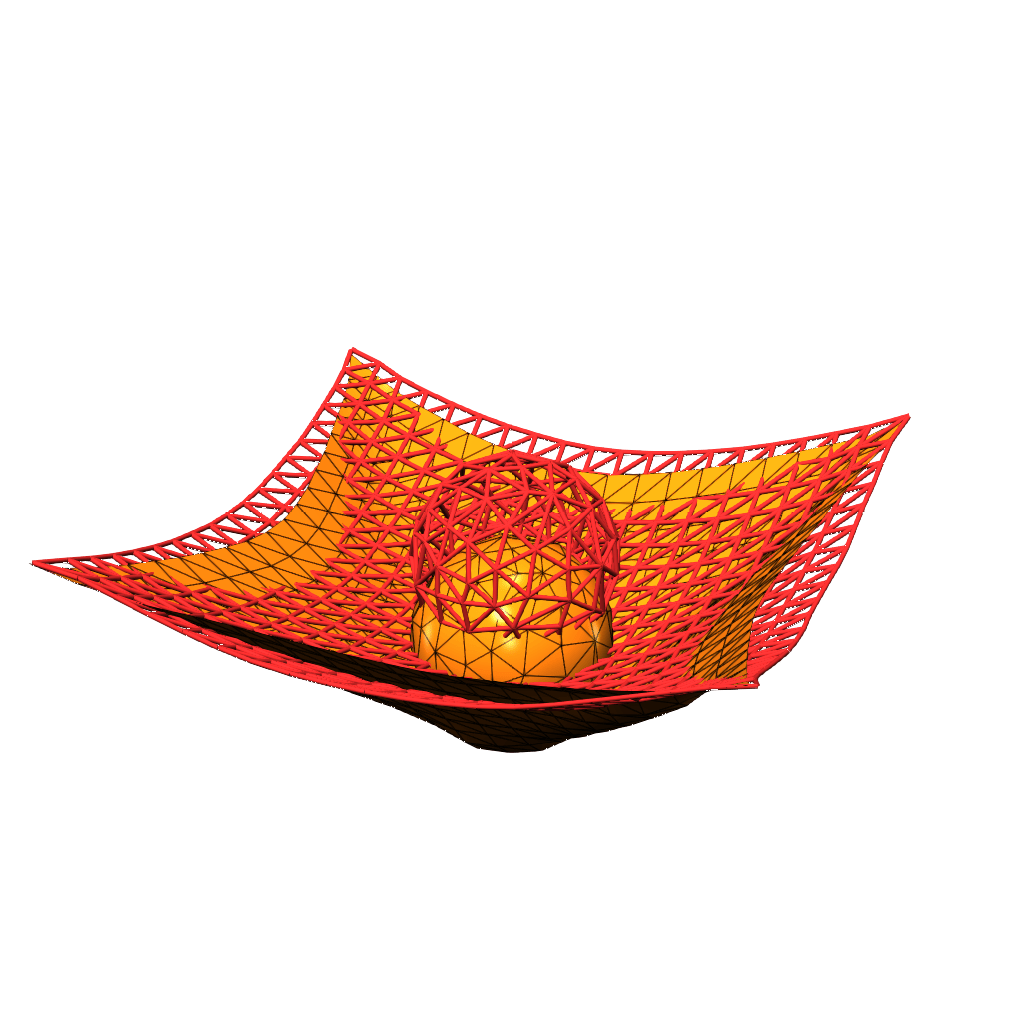}
    \end{minipage}
    \begin{minipage}{0.119\textwidth}
            \centering
            \includegraphics[width=\textwidth]{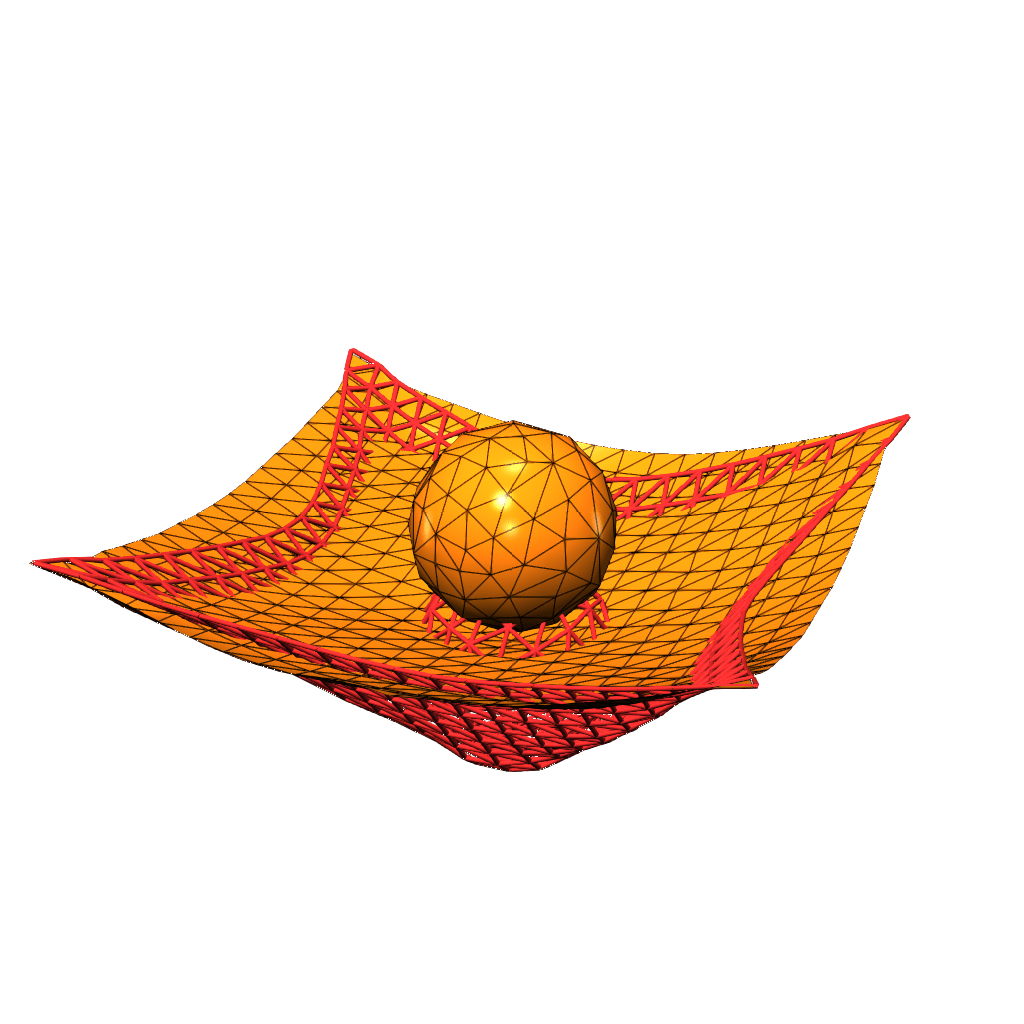}
    \end{minipage}

    \begin{minipage}{0.119\textwidth}
            \centering
            \includegraphics[width=\textwidth]{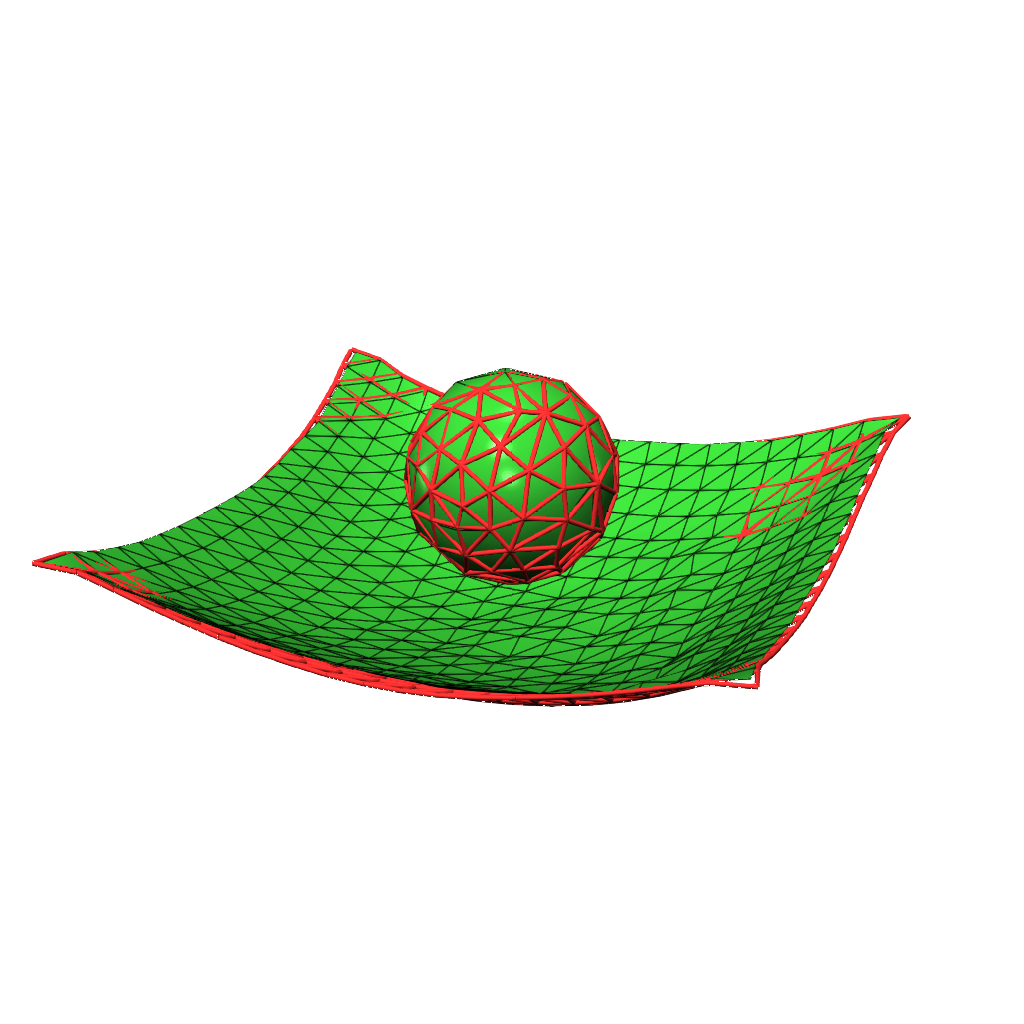}
    \end{minipage}
    \begin{minipage}{0.119\textwidth}
            \centering
            \includegraphics[width=\textwidth]{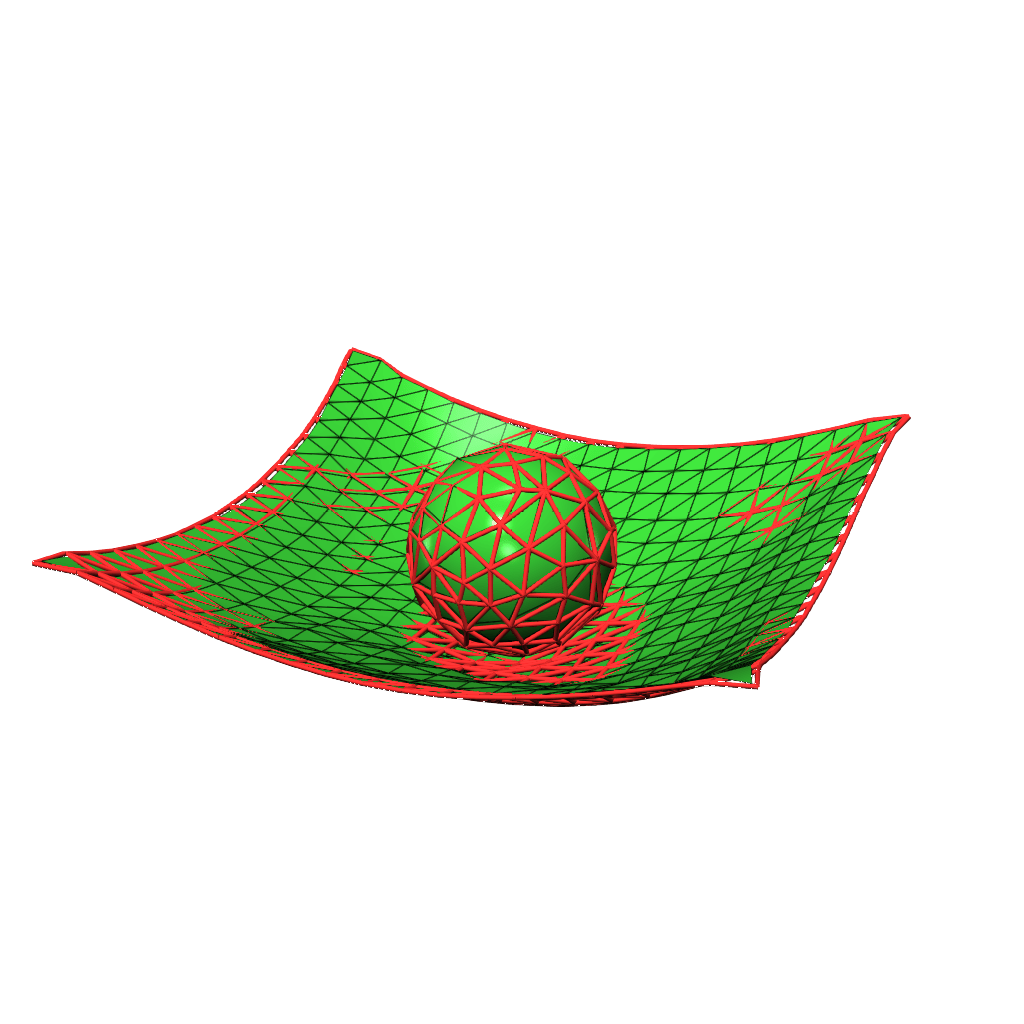}
    \end{minipage}
    \begin{minipage}{0.119\textwidth}
            \centering
            \includegraphics[width=\textwidth]{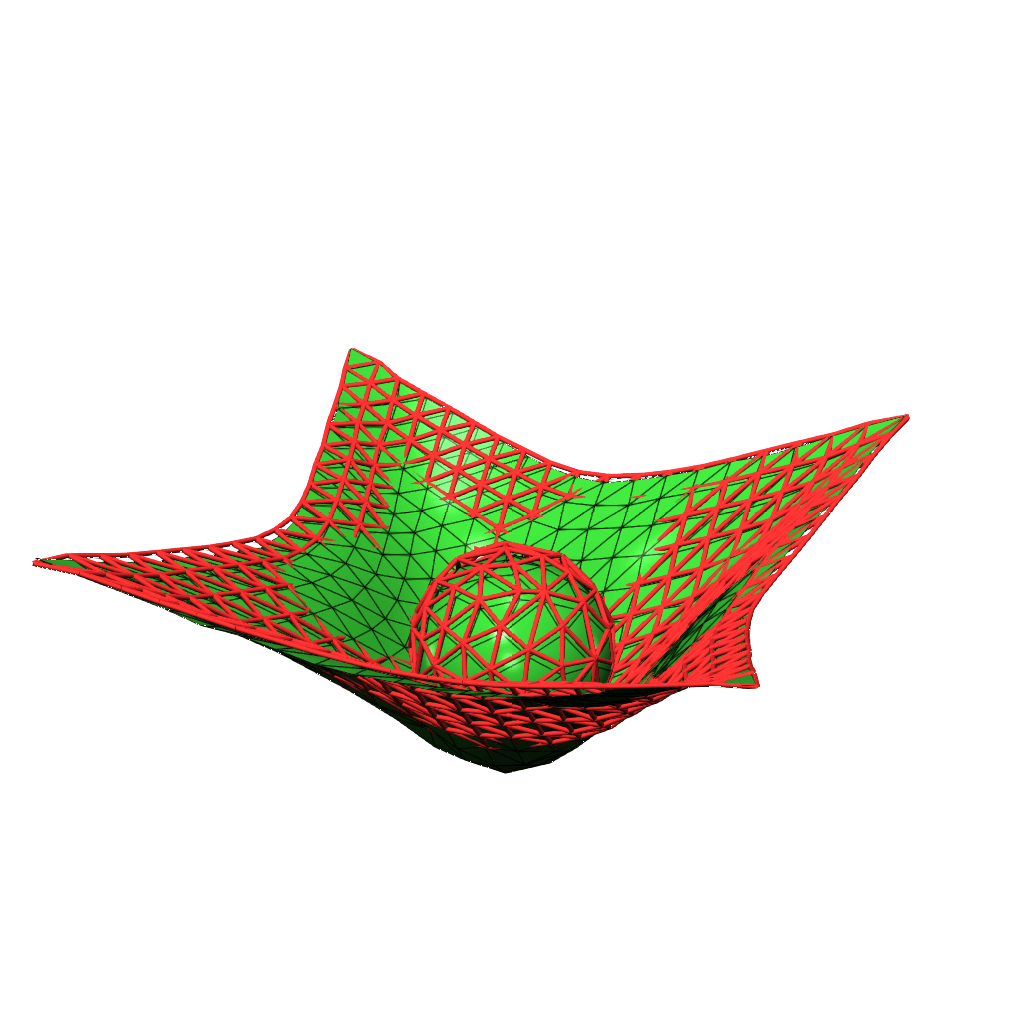}
    \end{minipage}
    \begin{minipage}{0.119\textwidth}
            \centering
            \includegraphics[width=\textwidth]{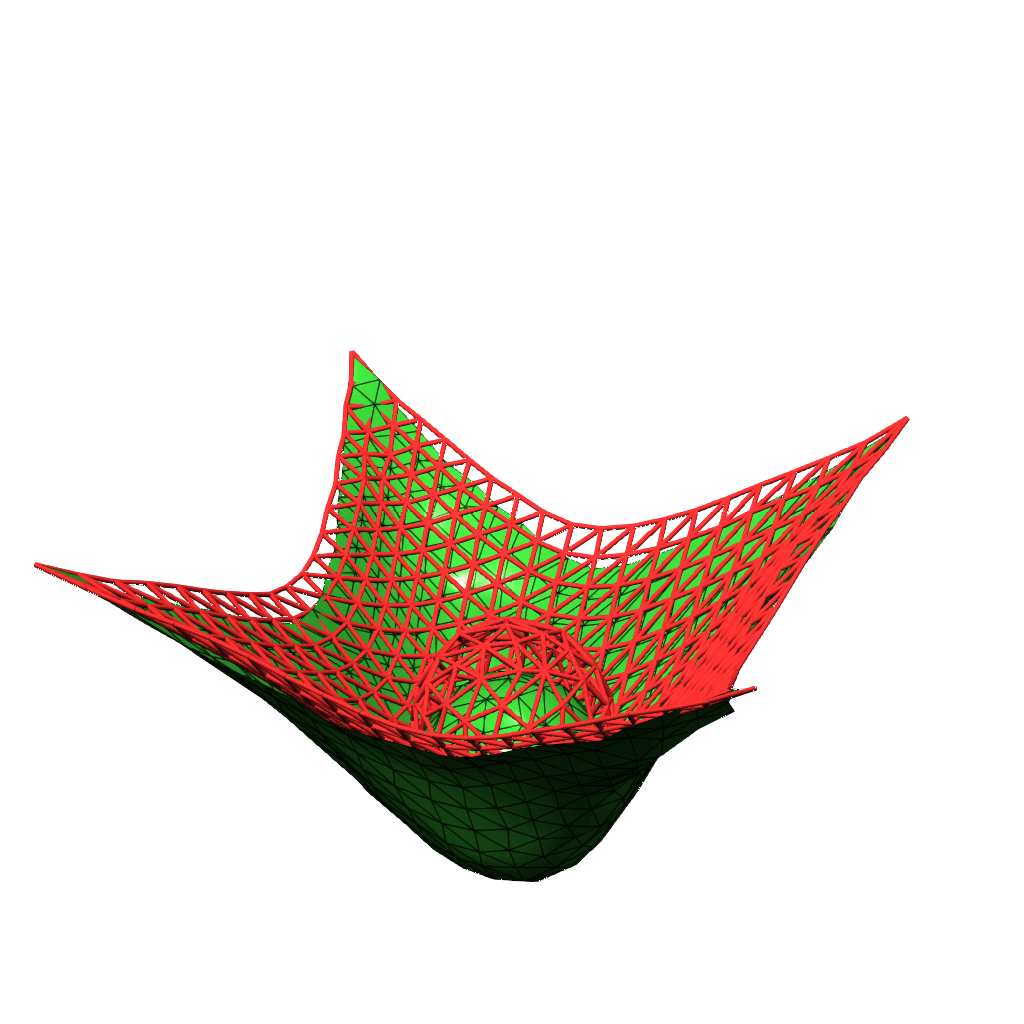}
    \end{minipage}
    \begin{minipage}{0.119\textwidth}
            \centering
            \includegraphics[width=\textwidth]{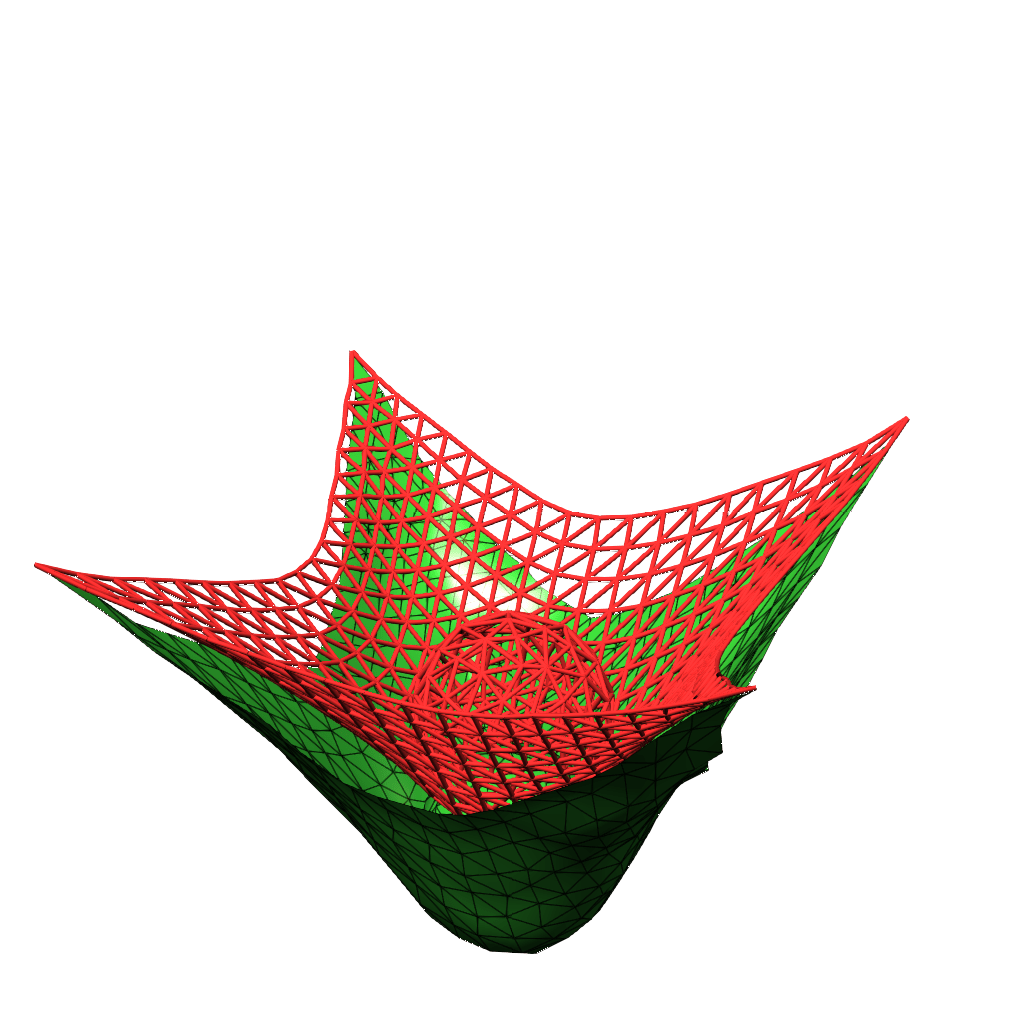}
    \end{minipage}
    \begin{minipage}{0.119\textwidth}
            \centering
            \includegraphics[width=\textwidth]{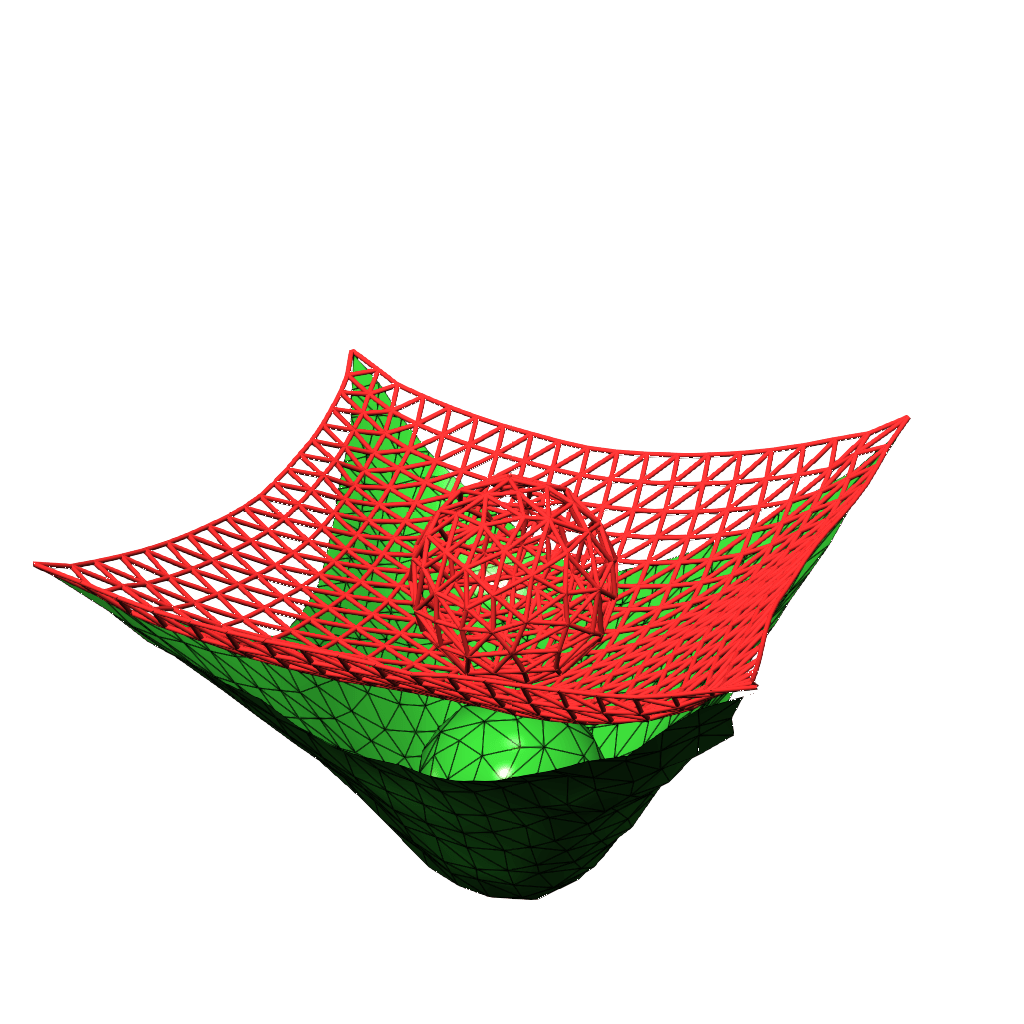}
    \end{minipage}
    \begin{minipage}{0.119\textwidth}
            \centering
            \includegraphics[width=\textwidth]{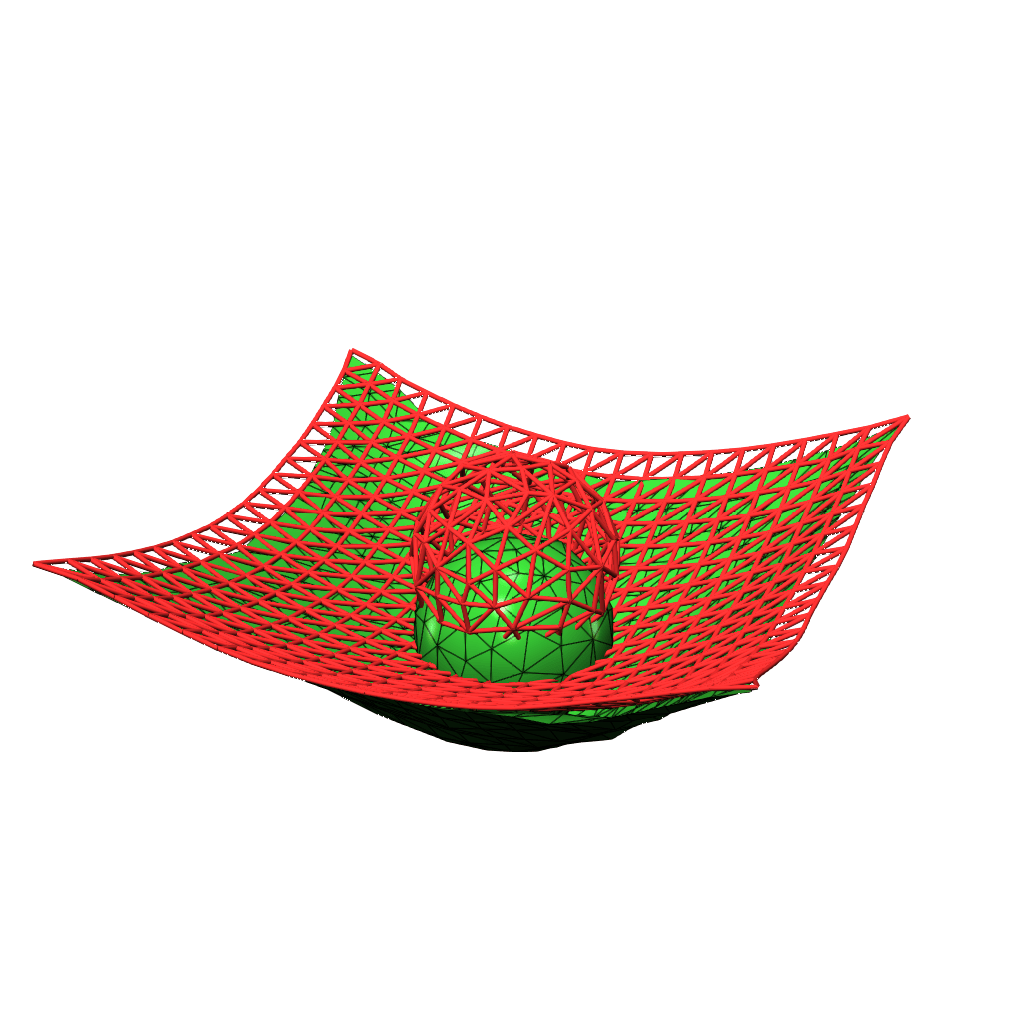}
    \end{minipage}
    \begin{minipage}{0.119\textwidth}
            \centering
            \includegraphics[width=\textwidth]{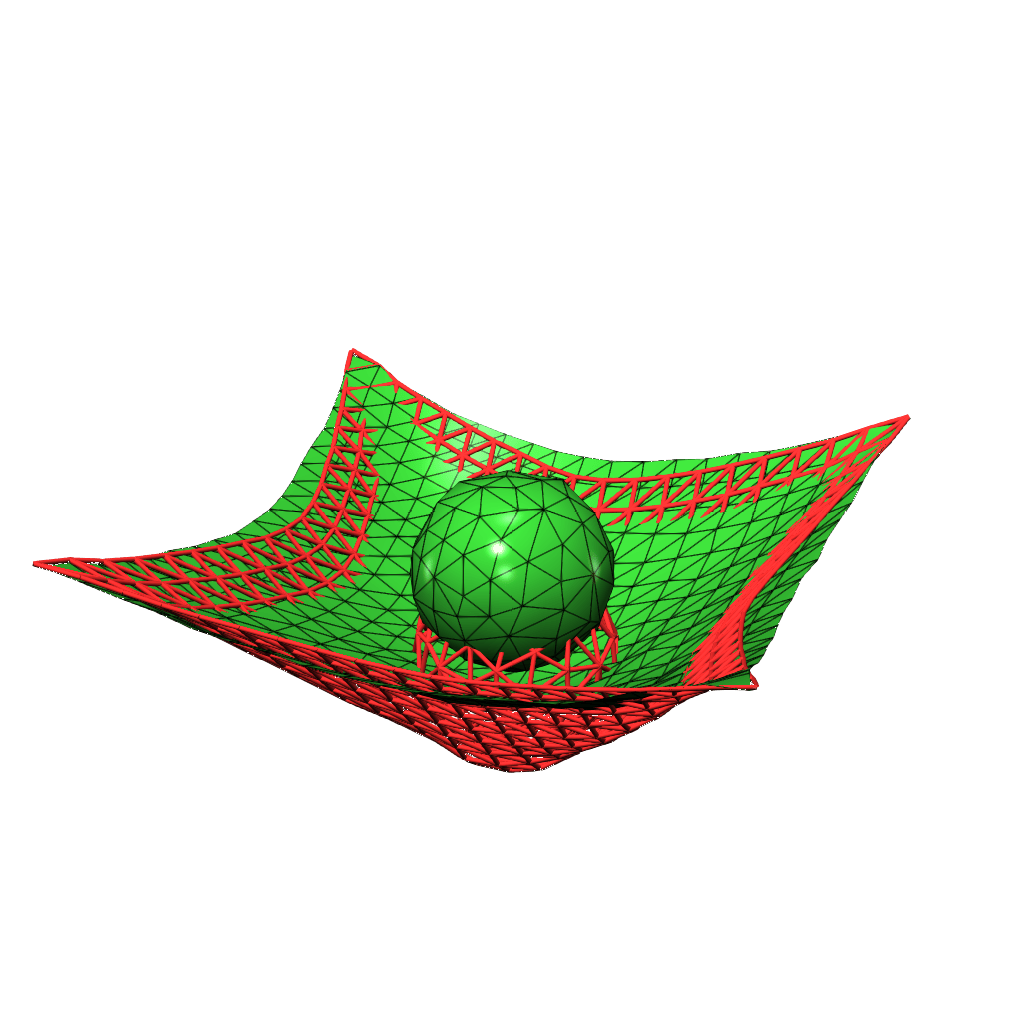}
    \end{minipage}

    \vspace{1.0em} 
    \noindent\hrulefill 

    \vspace{-1.0em}
    \noindent\hrulefill 
    \vspace{0.5em} 

    \begin{minipage}{0.119\textwidth}
            \centering
            \includegraphics[width=\textwidth]{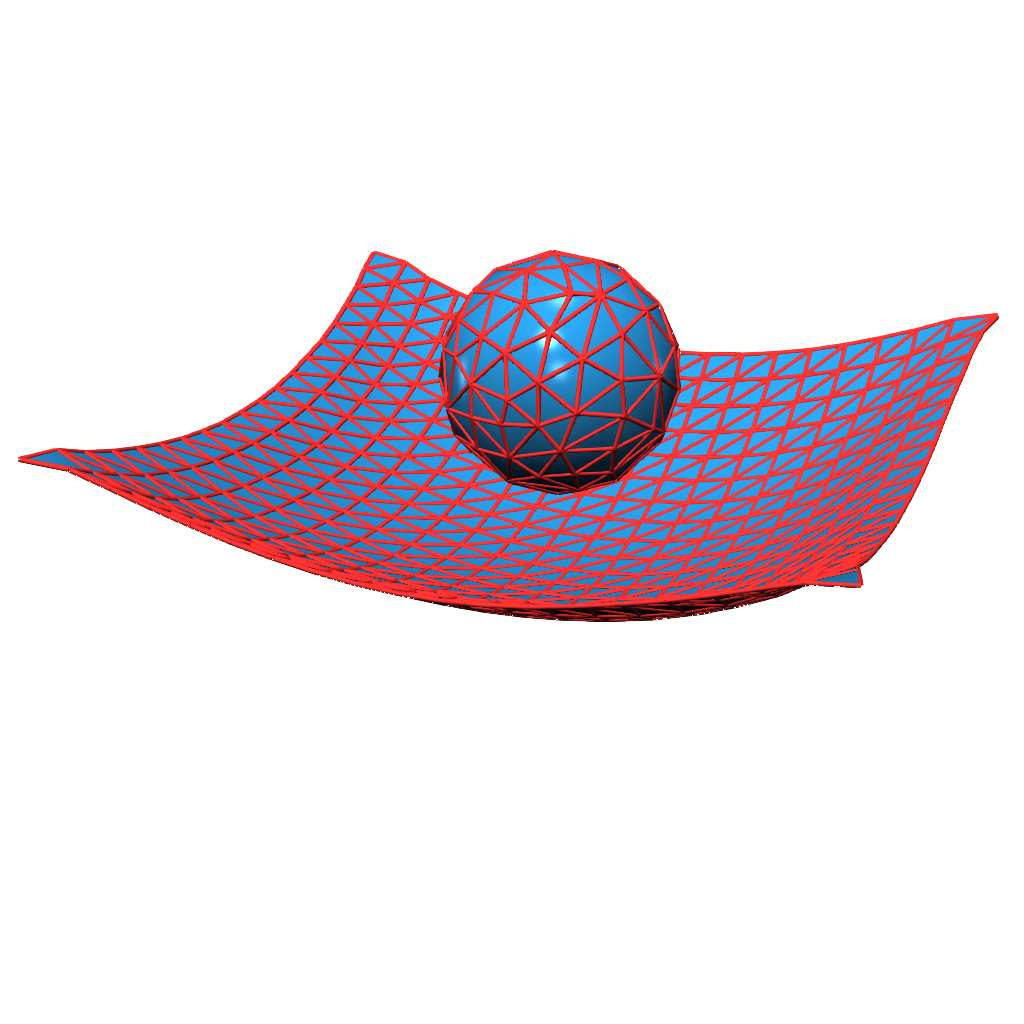}
            \subcaption*{$t=1$}
    \end{minipage}
    \begin{minipage}{0.119\textwidth}
            \centering
            \includegraphics[width=\textwidth]{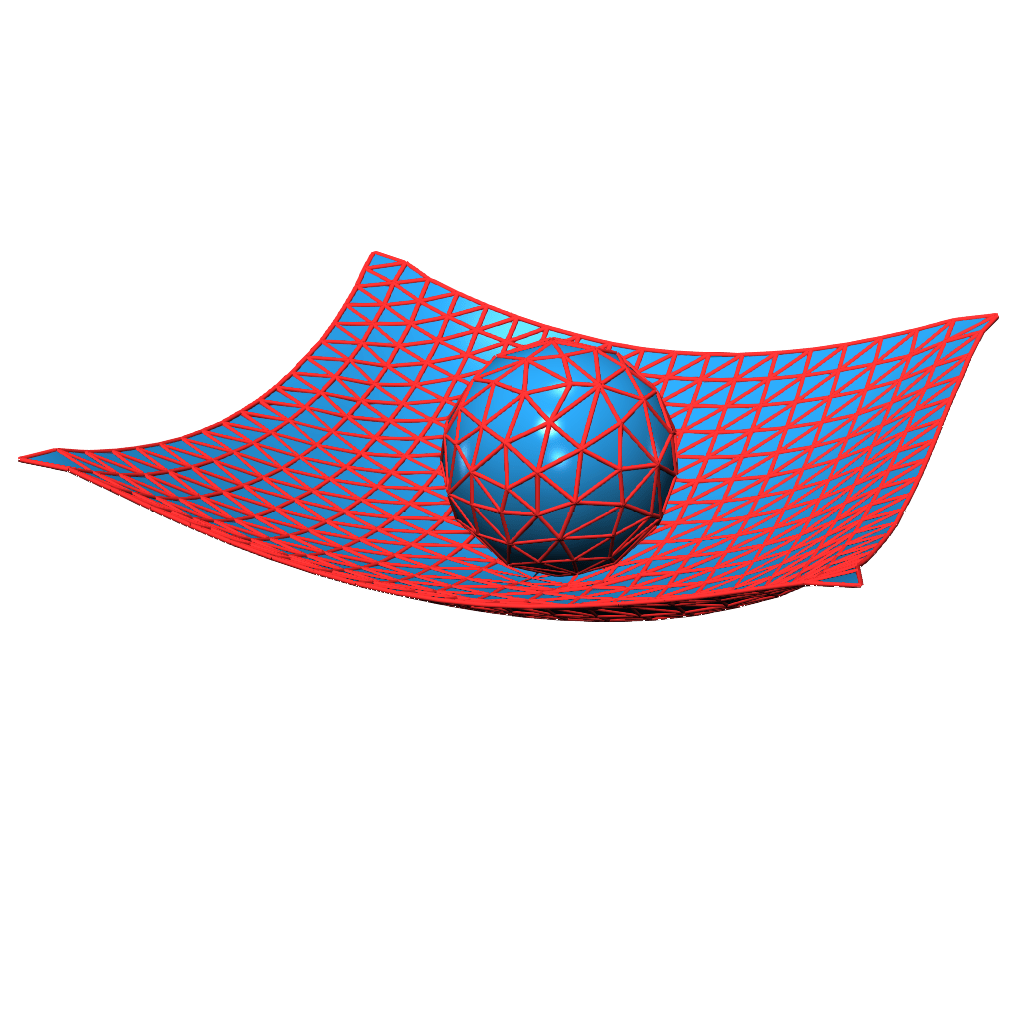}
            \subcaption*{$t=11$}
    \end{minipage}
    \begin{minipage}{0.119\textwidth}
            \centering
            \includegraphics[width=\textwidth]{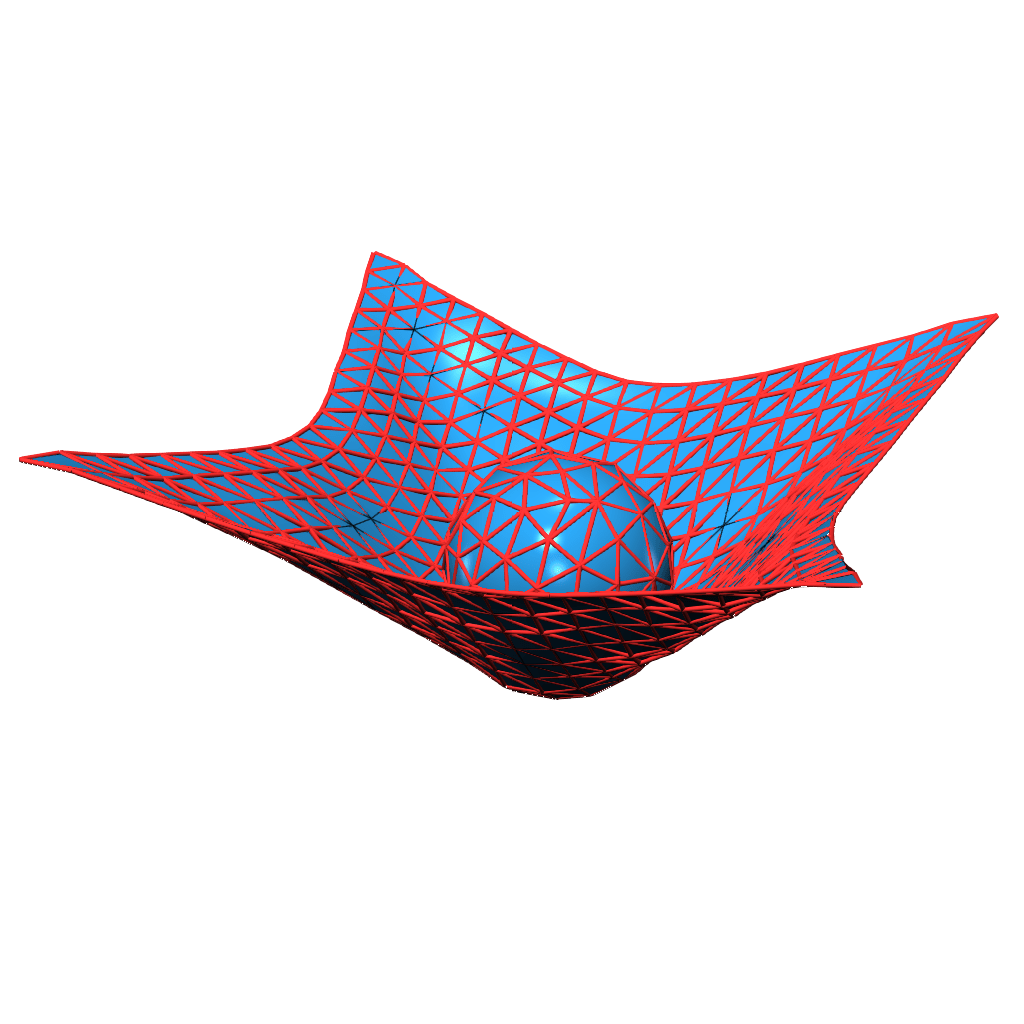}
            \subcaption*{$t=21$}
    \end{minipage}
    \begin{minipage}{0.119\textwidth}
            \centering
            \includegraphics[width=\textwidth]{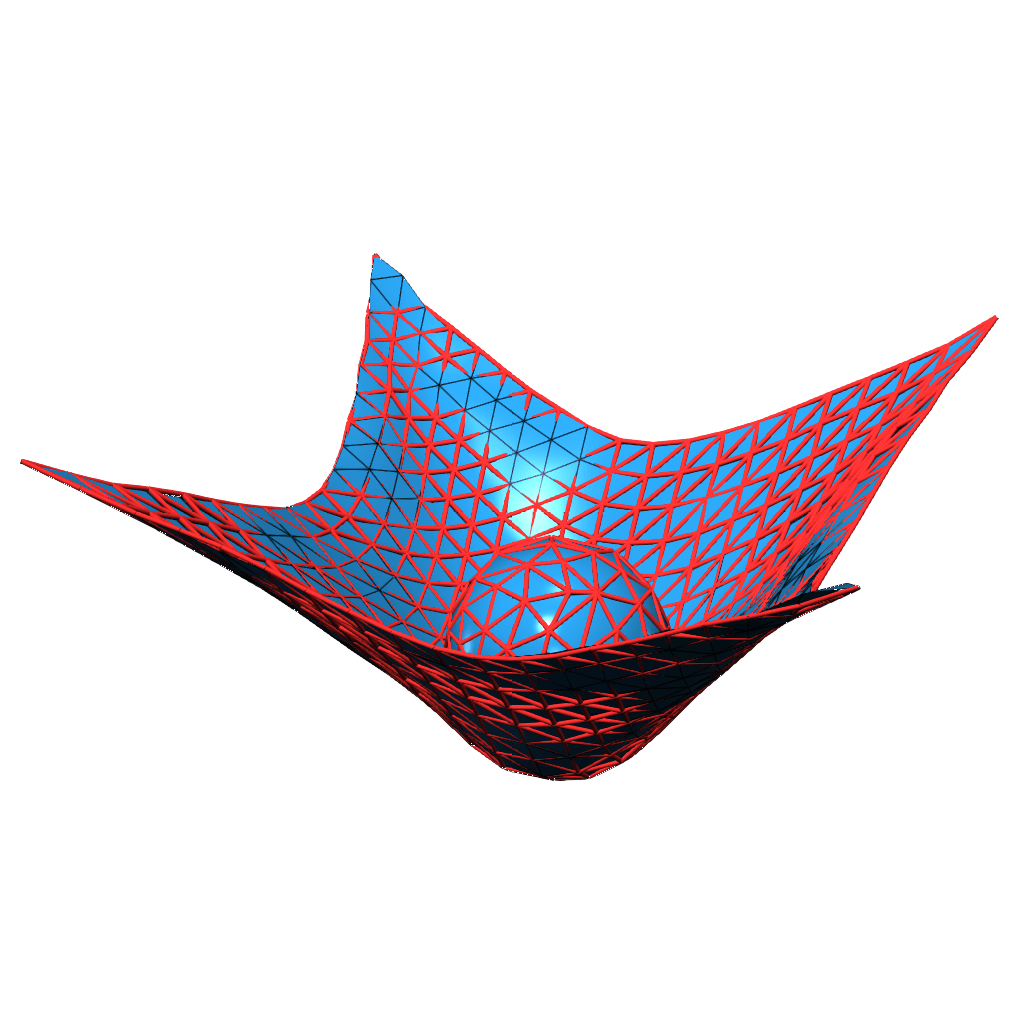}
            \subcaption*{$t=31$}
    \end{minipage}
    \begin{minipage}{0.119\textwidth}
            \centering
            \includegraphics[width=\textwidth]{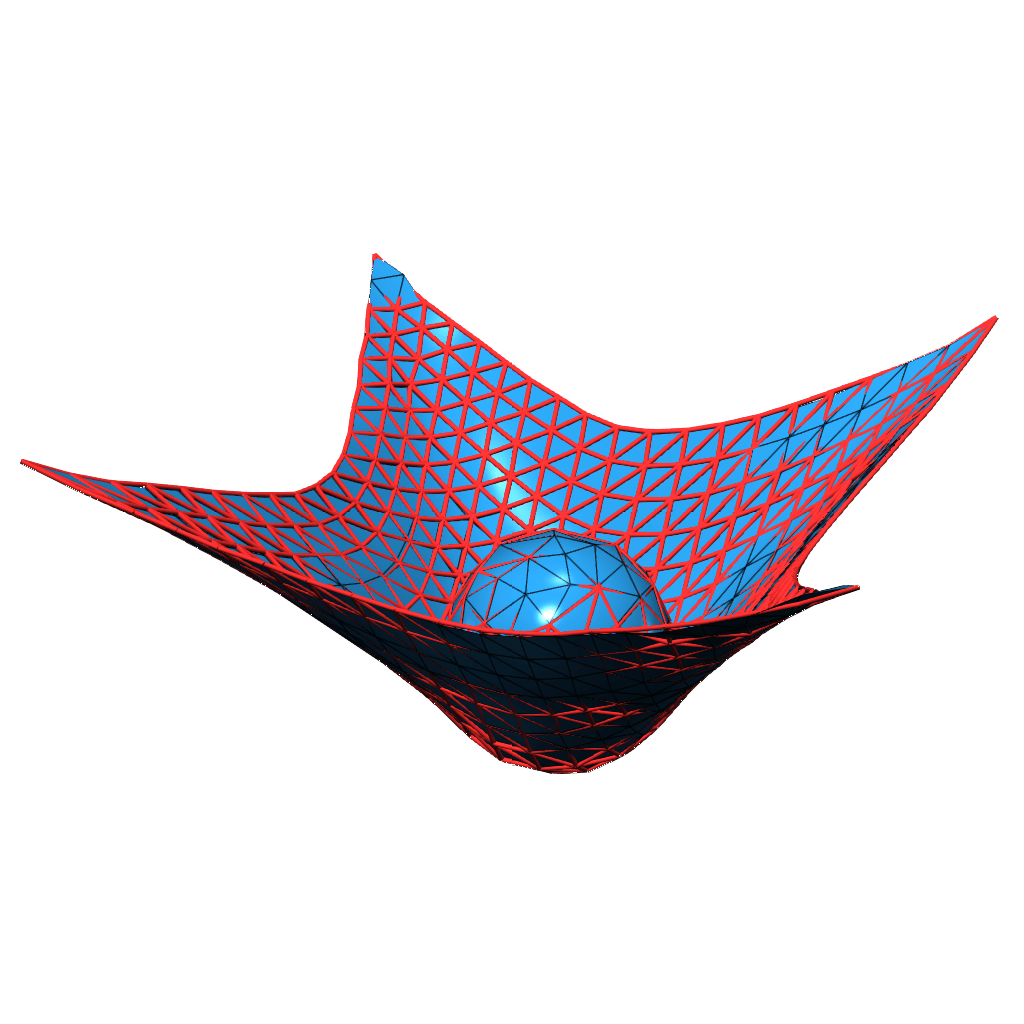}
            \subcaption*{$t=41$}
    \end{minipage}
    \begin{minipage}{0.119\textwidth}
            \centering
            \includegraphics[width=\textwidth]{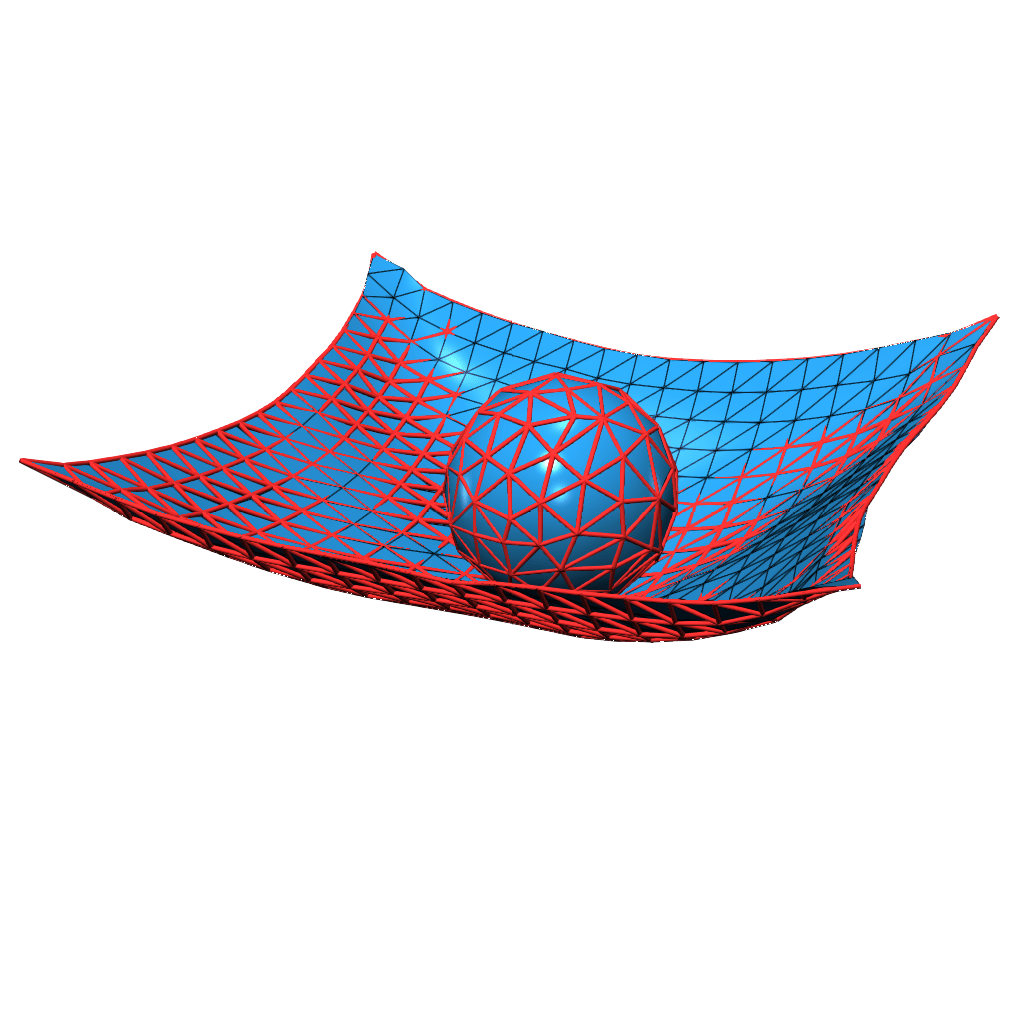}
            \subcaption*{$t=61$}
    \end{minipage}
    \begin{minipage}{0.119\textwidth}
            \centering
            \includegraphics[width=\textwidth]{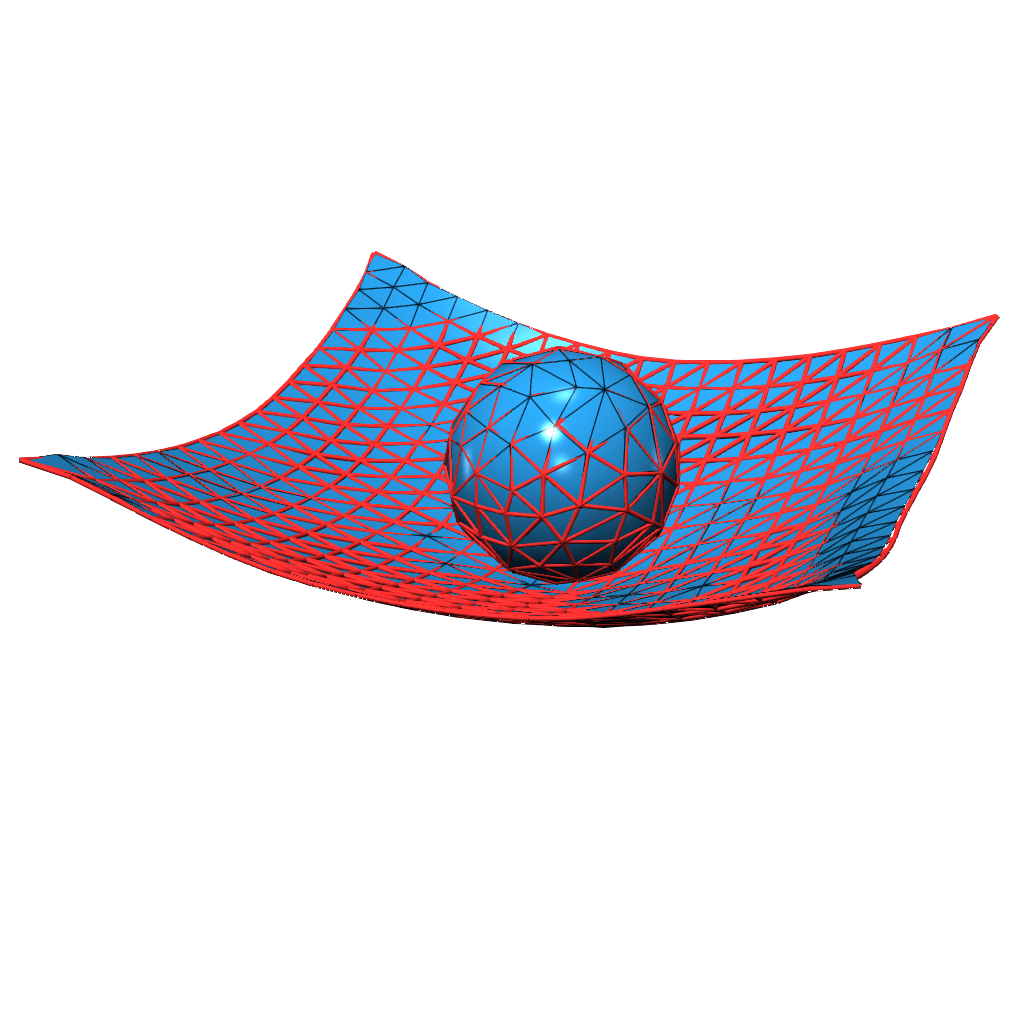}
            \subcaption*{$t=81$}
    \end{minipage}
    \begin{minipage}{0.119\textwidth}
            \centering
            \includegraphics[width=\textwidth]{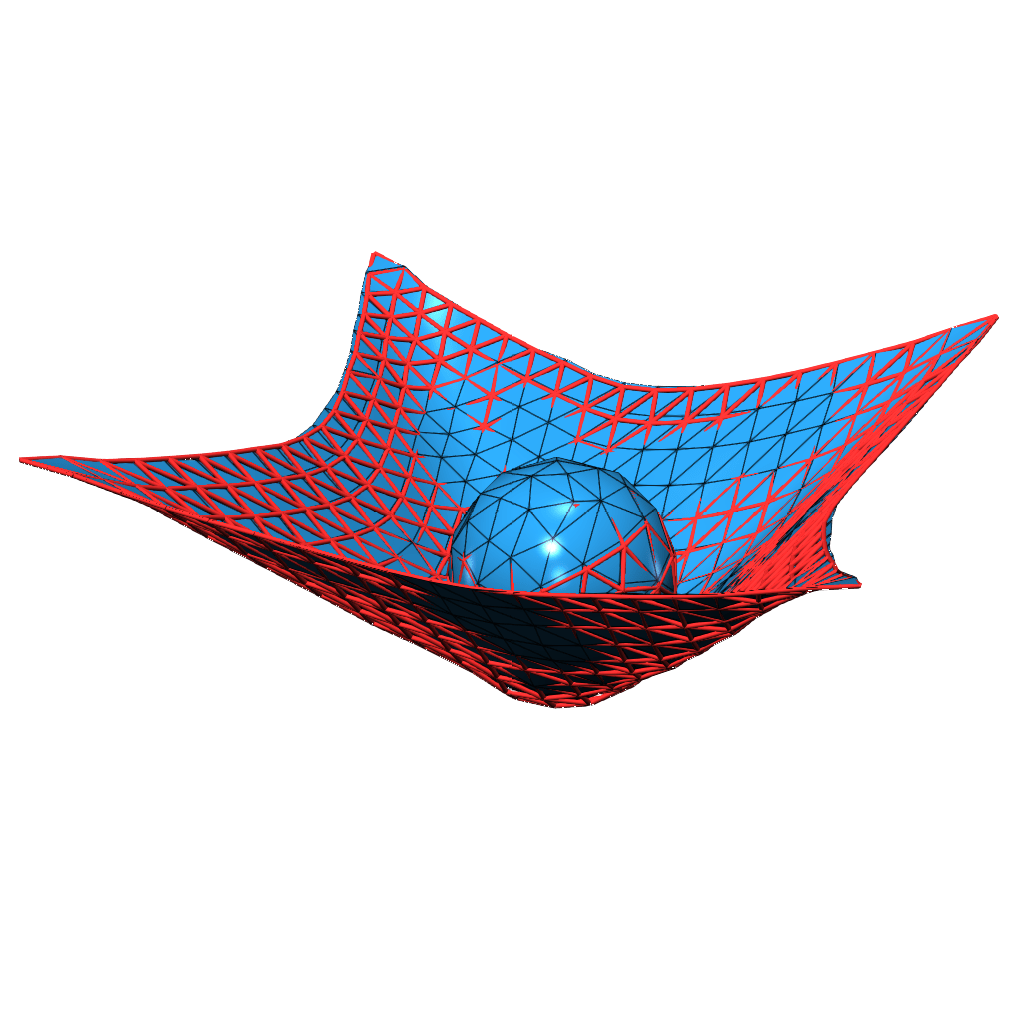}
            \subcaption*{$t=100$}
    \end{minipage}

    \vspace{0.01\textwidth}%

    \caption{
    Simulation over time of an exemplary test trajectory from the \textbf{Sphere Cloth Coupling} task. The figure compares predictions from \textcolor{tabblue}{\textbf{MaNGO}}, \textcolor{taborange}{MGN}, and \textcolor{ForestGreen}{EGNO}. The last row, \textcolor{tabblue}{MaNGO-Oracle}, is separated by a horizontal line and represents predictions using oracle information. The \textbf{context set size} is set to $4$. All visualizations show the colored \textbf{predicted mesh}, with a \textbf{\textcolor{red}{wireframe}} representing the ground-truth simulation. \textcolor{tabblue}{\textbf{MaNGO}} accurately predicts the correct material properties, leading to a highly accurate simulation.
    }
    \label{fig:appendix_sphere_cloth}
\end{figure*}


\end{document}